%% file: hdr-v02.tex
\documentclass[a4paper,11pt,twoside, openright]{book}

\usepackage[utf8]{inputenc}
\usepackage[T1]{fontenc}
 
\usepackage[francais]{babel}
\usepackage{pdfpages}
\usepackage{bm,amssymb}   

\usepackage{textcomp}

\usepackage{eurosym}

\usepackage{notations_niko}

\usepackage{geometry}

\usepackage{fancyhdr}

\usepackage{hyperref}
\hypersetup{
    bookmarks=true,         							
    unicode=false,             							
    pdftoolbar=true,          							
    pdfmenubar=true,        							
    pdffitwindow=false,     							
    pdfstartview={FitH},    							
    pdftitle={My title},         							
    pdfauthor={Author},     							
    pdfsubject={Subject},  						   	
    pdfcreator={Creator},   							
    pdfproducer={Producer}, 						
    pdfkeywords={keyword1, key2, key3},   
    pdfnewwindow=true,      						
    colorlinks=true,       								
    linkcolor=blue,         								
    citecolor=blue,        								
    filecolor=magenta,      							
    urlcolor=cyan           								
}

\usepackage[round]{natbib}   

\usepackage{enumitem}

\usepackage{amsmath,amsfonts,amssymb,amsthm,epsfig,epstopdf,titling,url,array}

\theoremstyle{plain}

\theoremstyle{definition}
\newtheorem{defn}{Définition}[section]

\theoremstyle{remark}

\usepackage{graphicx}

\usepackage{blockdiagram}
\usepackage{pgfplots}
\pgfplotsset{compat=newest}
\usepackage{forest}
\usetikzlibrary{shadows}

\usepackage{subfigure}
\usepackage{wrapfig}

\usepackage{amstext}
\usepackage{multicol}

\usepackage{titlesec}
\usepackage{tikz}\usetikzlibrary{shapes.misc}

\usepackage{amsmath}
\usepackage{amsfonts}
\usepackage{amsthm}
\usepackage{mathtools}

\newcommand{\ms}{\mbf{s}}
\newcommand{\mbf}[1]{{\mathbf #1}}
\newcommand{\msd}{{\mbf s_{\mbf d}}}
\newcommand{\me}{\mbf{e}}

\usepackage{array}

\newcolumntype{M}[1]{>{\centering}m{#1}}
\newcolumntype{C}[1]{>{\centering}p{#1}}

\usepackage{tcolorbox}

\newtcolorbox{publi}{
sharp corners=all,
boxrule=0.25mm
}

\input{macroMarchand.sty}

\usepackage{tcolorbox}

\usepackage[francais]{minitoc}
\usepackage{color,soul}

\newcommand{\cref}[1]{\textcolor{red}{#1}}
\newcommand{\todo}[1]{\textcolor{red}{#1}}

\renewcommand{\maketitle}{
\thispagestyle{empty}
\raggedbottom

\begin{center}
  \includegraphics[width = 40mm]{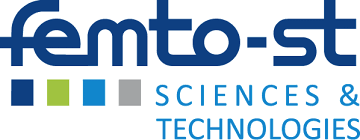} \hfill
  \includegraphics[width = 40mm]{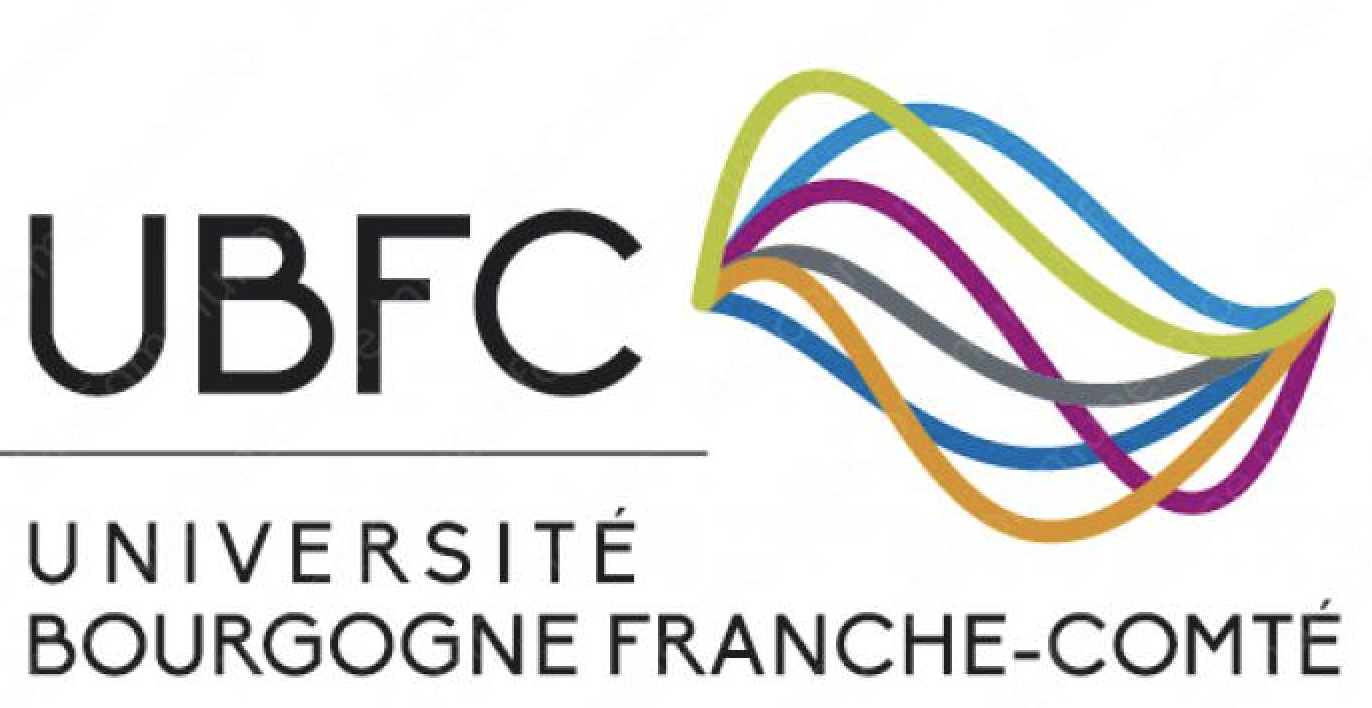} \hfill
  \includegraphics[width = 20mm]{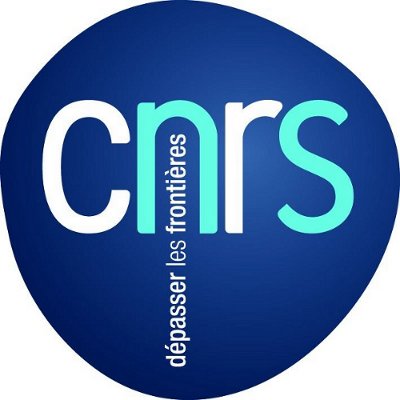}
\end{center}

\sffamily

\noindent\parbox[c][2cm]{\linewidth}{
\begin{center}
\Large \textsc{Habilitation à diriger des recherches}\\
de l'Université Bourgogne Franche-Comté \\ 
\end{center}
}

\noindent\parbox[c][3cm]{\linewidth}{
\begin{center}
\normalsize présentée par
\end{center}
}

\noindent\parbox[c][3cm]{\linewidth}{
\begin{center}
\Large \textbf{Brahim TAMADAZTE}\\
Chargé de Recherche CNRS\\
Institut FEMTO-ST - UMR 6174
\end{center}
}

\noindent\parbox[c][3.5cm]{\linewidth}{
\begin{center}
\huge \textbf{ \textcolor{blue}{Contributions à l'asservissement visuel et à l'imagerie  en médecine}}\\\
\Large \textbf{ \textcolor{black}{\emph{Contributions in visual servoing and imaging in medicine}}}
\end{center}}

\noindent\parbox[c][2cm]{\linewidth}{
\begin{center}
\Large soutenue le 12 juin 2019
\end{center}
}

\noindent
\normalsize devant le jury :
\begin{description}[labelwidth=4.9cm,labelsep=0cm,leftmargin=4.4cm]  
\item[Jocelyne \textsc{Troccaz}] Dir. de Recherche CNRS, TIMC-IMAG, Grenoble  (rapportrice)
\item[François \textsc{Chaumette}] Dir. de Recherche, Inria Bretagne-Atlantique, Rennes (rapporteur)
\item[Marie  \textsc{Chabert}] Prof. à l'Université de Toulouse, Toulouse (rapportrice)
\item[Nassir \textsc{Navab}] Prof. à Technische Universität München, Munich (examinateur)
\item[Stéphane \textsc{R\'egnier}] Prof. à Sorbonne Université, Paris (examinateur) 
\item[Nicolas  \textsc{Chaillet}] Prof. à l'Univ. Bourgogne Franche-Comté, Besançon (président)
\item[Stéphane \textsc{Chr\'etien}] Chercheur au National Physics Laboratory, Londres  (examinateur)
\item[Nicolas  \textsc{Andreff}] Prof. à l'Univ. Bourgogne Franche-Comté, Besançon (examinateur)
\end{description}

\newpage
\flushbottom
	\hbox{}\thispagestyle{empty}\newpage
}

\usepackage{titlesec}
\usepackage{tikz}\usetikzlibrary{shapes.misc}

\newcommand\sectionbar{%
\tikz[baseline,trim left=3.1cm,trim right=3cm] {
    \fill [cyan!25] (2.5cm,-1ex) rectangle (\textwidth+3.1cm,2.5ex);
    \node [
        fill=cyan!60!white,
        anchor= base east,
        rounded rectangle,
        minimum height=3.5ex] at (3cm,0) {
        \textbf{\thesection.}
    };
}%
}

\newcommand\subsectionbar{%
\tikz[baseline,trim left=2cm,trim right=3cm] {
    \fill [cyan!25] (2.5cm,-1ex) rectangle (\textwidth+2cm,2.5ex);
    \node [
        fill=cyan!60!white,
        anchor= base east,
        rounded rectangle,
        minimum height=3.5ex] at (3cm,0) {
        \textbf{\thesubsection.}
    };
}%
}

\titleformat{\section}{\Large\sffamily\bfseries}{\sectionbar}{0.1cm}{}
\titleformat{\subsection}{\large\sffamily\bfseries}{\subsectionbar}{0.1cm}{}
\titleformat{\subsubsection}{\large\sffamily\bfseries}{}{0.1cm}{}

\begin{document}

\dominitoc

\frontmatter

\maketitle

\small
\tableofcontents

\listoffigures
\normalsize

\mainmatter

\input{./fichiers/thanks.tex}

\input{./fichiers/intro_v3.tex}

\input{./fichiers/clinique_v3.tex}

\input{./fichiers/laser_v3.tex}

\input{./fichiers/wavelet_v3.tex}

\input{./fichiers/cs_v3.tex}

\input{./fichiers/conclusion_v3.tex}


\medskip
 
 
\small
\bibliographystyle{plainnat}

\bibliography{biblio_ijrr_2015_2.bib,biblio_erc.bib,biblio_cs.bib}
\normalsize

\appendix




\end{document}

%% file: fichiers/thanks.tex
\chapter*{Avant-propos et remerciements}

Ce manuscrit donne un aperçu de mes travaux de recherche réalisés au sein de l'institut FEMTO-ST à Besançon, plus précisément dans le département Automatique et Systèmes MicroMécatroniques (AS2M). Il est surtout et avant tout le fruit de mes (co)-encadrements de stagiaires, de doctorant(e)s et de post-doctorants. Je tiens à leur rendre hommage, pour leur contribution majeure à la recherche scientifique, ici et ailleurs. \\

J'ai eu la chance d'avoir un jury d'une très grande expertise à ma soutenance. Je tiens à remercier l'ensemble des membres du jury, à commencer par son président Nicolas Chaillet (Prof. Univ. Bourgogne Franche-Comté), les rapporteurs : Jocelyne Troccaz (DR-CNRS, TIMC-IMAG), Marie Chabert (Prof. Université de Toulouse) et François Chaumette (DR-Inria, Inria Bretagne-Atlantique), ainsi que les examinateurs : Nasser Navab (Prof. TUM, Munich), Stéphane Régnier (Prof. Sorbonne Université), Stéphane Chrétien (Scientifique à National Physical Lab, Londres), ainsi qu'à Nicolas Andreff (Prof. Univ. Bourgogne Franche-Comté). \\ 

Durant ces 6 années de recherche à FEMTO-ST, j'ai eu le privilège de travailler avec des collègues de grandes qualités scientifiques et humaines. J'ai tant appris. Certains ont même la double casquette de scientifique et de correcteur (ils se reconnaitront !).\\

Une équipe administrative et de gestion de grande efficacité met en musique le fonctionnement du département. J'en suis naturellement reconnaissant de ce soutien quotidien. \\

J'ai envie de clore cette séquence par une phrase de \emph{Pierre Rabhi} : "\emph{C'est dans les utopies d'aujourd’hui que sont les solutions de demain}".
 
 \vspace{8cm}
 
\begin{flushright}
\emph{A ma femme,  ma famille et ami(e)s, merci.  } 

\end{flushright}

%% file: fichiers/intro_v3.tex
\chapter*{Introduction générale}
\section{Préambule}
Ce manuscrit est le fruit de plusieurs années de recherche, d'encadrement, de collaborations et de recherches de financements. J'ai rejoint le CNRS, dans le corps des Chargés de Recherche, en 2012 sur un projet de recherche scientifique alliant la microrobotique, la commande et l'imagerie pour des applications en santé : \emph{diagnostic} et \emph{chirurgie}. Pendant cette période d'apprentissage du métier de chercheur, une question est souvent revenue : comment faire avancer la science, dans sa composante fondamentale et méthodologique, tout en gardant un \oe il avisé sur des aspects translationnels et applicatifs ? Les Sciences de l'Ingénieur offrent la possibilité de passer d'un monde à un autre, sans grandes difficultés, mais nécessitent de rester vigilant pour ne pas basculer définitivement dans l'ingénierie en délaissant la méthodologie. En robotique médicale, le travail de recherche réunit généralement trois acteurs : le patient, le chercheur et le médecin ; pour un triple bénéfice : le bien-être du patient, l'assistance au clinicien et la contribution pour l'avancée de la science. Le patient doit être naturellement placé au centre du "triplet" : médecin, patient et chercheur. 
  
\begin{wrapfigure}{l}{0.25\textwidth}
  \vspace{-20pt}
  \begin{center}
    \includegraphics[width=0.23\textwidth]{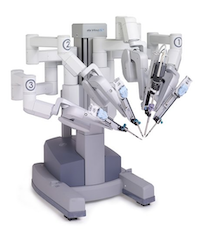}
  \end{center}
  \vspace{-0pt}
  \caption{Le robot Da Vinci \small (www.intuitive.com).\normalsize}
  \label{fig.davinci}
  \vspace{-10pt}
\end{wrapfigure}

Tout d'abord, quelques mots pour retracer l'émergence de cette nouvelle composante "médicale" de la robotique et sa progression au fur et à mesure des années. La recherche en robotique médicale est apparue au milieu des années 1980~\citep{paul1992development, lavallee1995robot}, grâce aux développements accélérés des nouvelles technologies, notamment de l'imagerie médicale et de l'informatique. Plusieurs facteurs ont permis une évolution significative de la robotique médicale : un monde médical de plus en plus ouvert aux nouvelles technologies, la miniaturisation (actionneurs, capteurs, etc.), l'évolution exponentielle des capacités informatiques, la volonté de diminuer les traumatismes post-opératoires, l'envie de mieux soigner et de soigner des pathologies jusqu'alors impossibles, etc. Les premiers robots ayant fait leur entrée dans les blocs opératoires sont issus de la robotique industrielle, sur lesquels, ont été greffés des instruments chirurgicaux et certains aspects sécuritaires ont été apportés, liés à l'environnement d'utilisation. Un des premiers robots détourné de son usage initial est le \emph{Puma 260} de la société américaine \emph{Unimation}, suivi du robot \emph{Scara} qui est le fruit d'une collaboration entre IBM et l'Université de Californie. En 1998, la société \emph{Surgical Intuitive}, grâce à son célèbre robot \emph{Da Vinci} (Fig.~\ref{fig.davinci}), a réalisé le premier pontage coronarien assisté par un robot, et ce deux ans avant sa validation par la FDA\footnote{Food and Drug Administration.} américaine. 

Le développement de la robotique médicale est motivé par les besoins :
\begin{itemize} 
\item d'améliorer l'efficacité des procédures chirurgicales grâce l'apport d'un système robotique en termes de précision, dextérité, répétabilité, ... ainsi que la capacité à coupler plusieurs sources d'informations (imagerie multimodale, capteur d'efforts/tactiles, etc.) dans la prise de décision.
\item de dépasser les limites physiques du chirurgien, notamment en apportant des moyens de visualisation au-delà de la perception humaine, d'avoir une meilleure dextérité et d'atteindre des sites opératoires inaccessibles sans un instrument/robot spécifique.
\end{itemize}

A partir de la fin des années 1990, la recherche en robotique médicale a connu un essor considérable au sein de la communauté robotique. Il ne s'agit plus d'adapter de simples robots industriels (généralement des bras manipulateurs) pour en faire des systèmes d'aide à la chirurgie, mais de concevoir entièrement des prototypes dont la finalité est médicale. Cette nouvelle approche intègre de fortes contraintes de conception, de commande, d'interface praticien-robot, ...  auxquelles s'ajoutent les questions de biocompatibilité, de stérilisation et de manière générale de réglementation. Naturellement, la robotique médicale a pris, au début des années 2000, un nouveau virage pour s'intéresser davantage à la chirurgie mini-invasive (minimally invasive surgery - MIS) voire non-invasive grâce notamment à la miniaturisation et la démocratisation des nouvelles technologies. Aujourd'hui, la communauté de la robotique médicale aborde des disciplines de recherche au moins aussi variées que celles de la robotique conventionnelle : informatique, automatique, haptique,  planification de tâches, interfaces chirurgien-robot, vision par ordinateur et commande référencée capteur (\emph{image-guided surgery}), réalité augmentée/virtuelle, sciences cognitives et neurosciences, tout en conservant de forts liens avec les applications cliniques (médecine, chirurgie, kinésithérapie, psychiatrie et social, etc.). 

\begin{wrapfigure}{r}{0.25\textwidth}
  \vspace{-20pt}
  \begin{center}
    \includegraphics[width=0.23\textwidth]{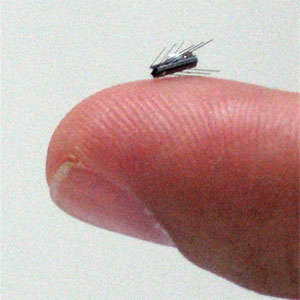}
  \end{center}
  \vspace{-20pt}
  \caption{Le microrobot Virob.}
  \label{fig.virob}
  \vspace{-10pt}
\end{wrapfigure}
Le développement de la robotique médicale est corrélé avec celui de l'imagerie associée. Dans cette discipline, certainement davantage que dans les autres de la robotique, l'imagerie (médicale) joue plusieurs rôles ; établissement d'un diagnostic, surveillance de l'évolution d'une pathologie, retour visuel pour le clinicien, ainsi que le rôle de capteur pour le guidage du robot/instrument. L'utilisation de systèmes d'imagerie médicale tels que l'IRM (imagerie à résonance magnétique), le scanner, l'échographie, la fluorescence, la tomographie à cohérence optique (OCT), ... en robotique médicale (dans le rôle de capteur) ouvrent de nouveaux challenges scientifiques, notamment en termes de commande référencée capteur (asservissement visuel), de suivi visuel d'instruments et d'organes déformables, etc. Le fait que ces imageurs sont caractérisés par des modèles de formation des images radicalement différents de ceux des capteurs (d'images) utilisés en robotique conventionnelle, tels qu’un rapport signal/bruit défavorable, une faible fréquence d'acquisition, etc. Tout ceci rajoute davantage de verrous scientifiques quant au développement d'approches de diagnostic ou de chirurgie guidées par l'image.  \\

L'autre effet direct de la capacité de miniaturisation apparue durant cette dernière décennie est l'augmentation de l'activité de recherche autour de la microrobotique médicale. Il s'agit notamment de développer des robots dédiés aux applications médicales, de plus en plus petits et qui doivent embarquer davantage de fonctionnalités (diagnostic \emph{in situ}, biopsies, largages médicamenteux, etc.). La miniaturisation engendre des contraintes de conception supplémentaires très importantes~\citep{nelson2010microrobots}. A titre d'exemple, embarquer un moyen d'actionnement et de l'énergie dans un dispositif millimétrique, voire micrométrique pour certains microrobots devient un véritable défi (Fig.~\ref{fig.virob})\footnote{Medical Robotics Laboratory at the Israel Institute of Technology (Technion).}. Par conséquent, l'actionnement adopté, par la communauté, est généralement à distance, \emph{a fortiori} magnétique voire électrique. L'actionnement magnétique en microrobotique médicale présente plusieurs avantages, comme la possibilité de contrôler la locomotion du dispositif à l'intérieur du corps humain, depuis l'extérieur et sans danger pour les organes~\citep{abbott2007robotics}. \\

Les thématiques dans lesquelles j'ai travaillé depuis que j'ai rejoint le CNRS, en 2012, s'articulent autour de l'imagerie médicale et de la commande référencée capteur pour l'aide à la réalisation d'actes chirurgicaux ou de diagnostics. J'ai également participé à la conception de certains prototypes (démonstrateurs) dans le cadre des projets de recherche dans lesquels j'émarge. Dans la suite, sont donnés, un résumé succinct de mes contributions scientifiques, ainsi que le plan du présent manuscrit. 
%
\section{Plan du manuscrit et contributions}
%
Ce manuscrit décrit, de manière chronologique, mes trois thématiques de recherche dans lesquelles j'ai travaillé depuis 2012, c'est-à-dire à partir mon recrutement au CNRS. Le socle commun de ces activités est l'imagerie et la commande référencée capteur de vision. J'ai contribué à certaines activités liées à la conception de démonstrateurs (à défaut de prototypes) précliniques de projets de recherche auxquels je participe.\\

Tout d'abord, j'ai réalisé une thèse (2007-2009) en microrobotique et en vision par ordinateur sous la direction de Sounkalo Demb\'el\'e (MCF, Univ. de Franche-Comté) et Nadine Piat (Prof., ENSMM), en forte interaction avec Éric Marchand (Prof., Univ. de Rennes)~\citep{tamadazte2009THESIS}. Durant ma thèse, j'ai travaillé sur le développement de lois de commande par asservissement visuel, d'algorithmes de suivi visuel de micro-objets sous microscopie optique, ainsi que sur des méthodes d'étalonnage de moyen de perception à forts grandissements~\citep{tamadazte2008multiscale, tamadazte2009multiscale, tamadazte2008automatic, tamadazte2009robotic, tamadazte2009real, duceux2010autofocusing, tamadazte2011TMECH, tamadazte2010cad, tamadazte2012four, agnus2013robotic, kudryavtsev2015analysis}. Après ma thèse, j'ai décidé d'orienter mes activités vers la robotique au service de la médecine en rejoignant, en janvier 2011, le laboratoire TIMC-IMAG (équipe GMCAO) à Grenoble. J'ai travaillé, sous la direction de Sandrine Voros (CR, INSERM), sur la chirurgie laparoscopique robotisée~\citep{tamadazte2013enhanced, voros2013devices, tamadazte2015multi, voros2015multi} pendant mon séjour post-doctoral.

Ce manuscrit est organisé en quatre chapitres retraçant les grands thèmes de recherche sur lesquels j'ai mené mes activités de recherche, et un cinquième consistant en un projet de recherche (perspectives). 
\subsubsection{Cadre clinique des travaux [Chapitre~\ref{chap.clinique}]}
%
Le premier chapitre résume les différents contextes cliniques dans lesquels sont réalisés les travaux de recherche discutés dans ce manuscrit. Il aborde également les différentes réalisations mécatroniques effectuées sur le volet conception de prototypes et de démonstrateurs précliniques.
%
\subsubsection{Chirurgie laser guidée par l'image [Chapitre~\ref{chap.laser}]}
%
\begin{wrapfigure}{l}{0.25\textwidth}
  \vspace{-20pt}
  \begin{center}
    \includegraphics[width=0.23\textwidth]{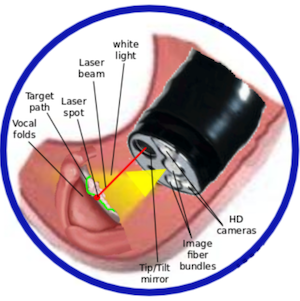}
  \end{center}
  \vspace{-20pt}
  \caption{Aperçu du concept $\mu$RALP.}
  \label{fig.uralp}
  \vspace{-10pt}
\end{wrapfigure}

Le laser de chirurgie vise à remplacer le traditionnel bistouri pour réaliser une résection ou une ablation de tissus pathologiques, ou encore une vaporisation lorsque le tissu a une teneur en eau élevée. La chirurgie laser présente plusieurs avantages par rapport à la chirurgie standard : pas de contact avec le tissu donc moins de risques d'infections, moins invasive, provoquant moins de saignements, etc. L'utilisation du laser permet également de réaliser des résections beaucoup plus précises et peut donc convenir davantage à la microchirurgie, comme en ophtalmologie, mais pas seulement. Dans le cadre du projet EU FP7 $\mu$RALP (2012-2015)\footnote{www.microralp.eu}, nous avons développé plusieurs techniques d'asservissement visuel pour contrôler le déplacement d'un spot laser sur un tissu à exciser, en l'occurrence les cordes vocales (Fig.~\ref{fig.uralp})~\citep{AndreffICINCO_2013,tamadazteIros2014,tamadazteTro2015,tamadazteIjrr2015,tamadazteTmech2016,dahroug2018some}. \\

Pour ce faire, nous avons proposé un schéma de commande qui utilise les propriétés de la géométrie trifocale~\citep{HartleyCUP06} qui régit un système de vision à trois-vues. Dans ce travail, nous avons écrit le tenseur trifocal sous une nouvelle forme, vectorielle et simplifiée, qui est plus adaptée à la commande que la traditionnelle expression tensorielle. L'autre originalité de notre approche réside dans le système de vision en lui-même. A la place d'utiliser un système à trois vues, nous n'avons utilisé que deux caméras, la troisième est remplacée par un miroir de balayage considéré comme une caméra virtuelle permettant d'acquérir un pixel à chaque instant. La commande trifocale proposée permet d'intégrer la profondeur de la scène (tissu) dans le schéma de commande de manière intuitive et sans estimation, ni au préalable, ni en ligne, de cette profondeur. Cette méthode permet de contrôler les déplacements d'un spot laser suivant des trajectoires rectilignes (point-à-point).   \\

Nous avons également développé une méthode de suivi de chemin dans laquelle l'opérateur (par exemple :  le chirurgien) peut définir n'importe quelle forme de courbe de balayage. La méthode s'inspire du suivi de chemin développé par la communauté de robotique mobile. La raison du choix du suivi de chemin plutôt que de trajectoire est pour assurer un découplage de la vitesse des autres paramètres géométriques et temporels de la courbe. Un autre avantage de cette méthode en chirurgie laser est le fait que la vitesse de progression du spot laser sur le chemin peut être définie comme un paramètre d'entrée de la commande. Par ailleurs, l'inconvénient du suivi de chemin est la non-gestion de la profondeur de la scène, comme c'est le cas de la commande trifocale. \\

Dans une troisième approche, nous avons associé les avantages de la méthode trifocale et celle du suivi de chemin dans un même schéma de commande. La vitesse du spot laser est calculée grâce à la méthode trifocale, qui est ensuite injectée dans l'expression de la vitesse obtenue par le suivi de chemin. 
\subsubsection{De la parcimonie à l'asservissement visuel [Chapitre~\ref{chap.wavelet}]}

Dans cette thématique, nous nous sommes intéressés à utiliser des informations temps-fréquence comme signal d'entrée dans une boucle de commande d'asservissement visuel. A l'image des méthodes d'asservissement visuel directes (dites également denses), comme la photométrie~\citep{collewet2011photometric}, nous avons développé plusieurs lois de commande dont l'information visuelle n'est autre que les coefficients des ondelettes (discrètes ou continues) ou encore des shearlets. Les verrous scientifiques de cette thématique résident essentiellement dans la manière de formaliser la relation entre les variations temporelles des coefficients (ondelettes ou shearlets) et le torseur vitesse de la caméra, en d'autre terme, cela revient à trouver l'expression analytique de la \emph{matrice d'interaction}, qui peut également être multi-échelle du fait que l'information visuelle considérée l'est au préalable. \\

Nous avons ainsi proposé plusieurs méthodes d'asservissement visuel direct à trois ou à six degrés de liberté (ddl) plus robustes et plus précises par rapport à l'état de l'art. Pour ce faire, nous avons choisi l'utilisation d'autres bases de projection dans lesquelles le signal-image est \emph{parcimonieux}, c'est-à-dire la grande majorité de l'information (des coefficients) est nulle ou au mieux proche de zéro. Ainsi, avec un simple seuil, il est possible de sélectionner aisément les coefficients correspondants à l'information pertinente (contours, points d'intérêts, etc.) contenue dans l'image, en d'autres termes, d'éliminer le bruit (hautes fréquences dans le domaine fréquentiel). 

\begin{wrapfigure}{r}{0.25\textwidth}
  \vspace{-20pt}
  \begin{center}
    \includegraphics[width=0.23\textwidth]{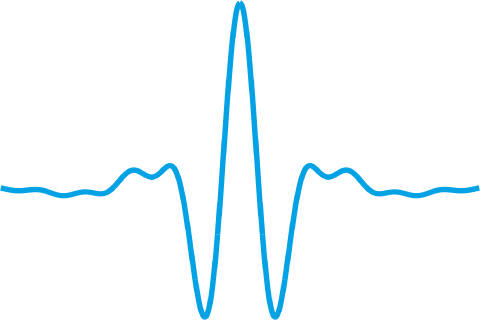}
  \end{center}
  \vspace{-20pt}
  \caption{ondelette.}
  \label{fig.wave}
  \vspace{-10pt}
\end{wrapfigure}

Les ondelettes ont été explorées (Fig.~\ref{fig.wave}), dans notre travail sur la commande par vision directe, sous deux angles : celles discrètes pour l'asservissement visuel 2D direct et celles continues pour la version 3D de l'asservissement visuel direct. Ces travaux ont été validés avec succès en simulation et expérimentalement suivant plusieurs scénarios : favorables et non-favorables, à la fois sur des plateformes expérimentales, mais dans des conditions précliniques (repositionnement automatique d'un système de biopsie optique). \\

Nous avons ensuite exploité la dernière-née des décompositions temps-fréquence multi-échelle, en l'occurrence, la transformée en shearlets dans la mise en \oe uvre de lois de commande référencées vision. La démarche est similaire à la méthode de calcul analytique de la matrice d'interaction associée aux coefficients des shearlets. Là aussi, le contrôleur développé a été validé en utilisant plusieurs scénarios expérimentaux avec différents systèmes d'imagerie (conventionnels et médicaux). 
%
\subsubsection{Acquisition comprimée [Chapitre~\ref{chap.cs}]}
Cette thématique est la plus récente de mes activités de recherche. Elle fait suite aux travaux sur la décomposition des images et de l'asservissement visuel direct décrit ci-dessus. Il s'agit de développer des méthodes d'acquisition comprimée (en anglais : compressed sensing) pour améliorer considérablement le temps d'acquisition d'une catégorie de systèmes d'imagerie médicale dite à balayage : OCT, échographie 3D, IRM, scanner, etc. 
Un problème d'acquisition comprimée peut être résumé dans la recherche d'une solution la plus parcimonieuse possible d'un système linéaire admissible sous-déterminé. Un des problèmes mathématiques sous-jacents est l'optimisation linéaire. Une des particularités des méthodes d'acquisition comprimée est la \emph{violation} du théorème d'échantillonnage de \emph{Nyquist-Shannon}. 

\begin{wrapfigure}{l}{0.30\textwidth}
  \vspace{-20pt}
  \begin{center}
    \includegraphics[width=0.29\textwidth]{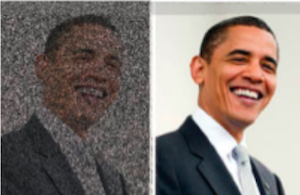}
  \end{center}
  \vspace{-20pt}
  \caption{Acquisition comprimée.}
  \label{fig.obama}
  \vspace{-10pt}
\end{wrapfigure}
Une des conditions nécessaires à la résolution d'un problème d'acquisition comprimée en imagerie, est la capacité à représenter les images dans une base parcimonieuse. On considère qu'une image est parcimonieuse (en anglais : \emph{sparse)}, lorsque la plupart de ses coefficients sont "approximativement" nuls dans une base donnée. A noter, que les décompositions en ondelettes ou shearlets offrent un moyen efficace d'obtenir cette parcimonie. 

L'imagerie médicale, de manière générale, se prête parfaitement aux méthodes d'acquisition comprimée. Les images IRM, scanner, échographie 3D, OCT, ... sont naturellement parcimonieuses et leur décomposition en ondelettes ou shearlets l'est davantage. Dans un premier temps, nous avons développé les premières approches d'acquisition comprimée dédiées aux systèmes d'imagerie OCT. Celles-ci combinent la décomposition en shearlets 2D et 3D, les méthodes de balayage optimisées et les algorithmes de rastérisation. Vis-à-vis de l'état de l'art, ces travaux constituent une grande originalité par l'implémentation d'une méthode d'acquisition comprimée directement sur un système d'imagerie (OCT) physique, tout en prenant en compte les contraintes physiques de l'OCT. Nous avons montré qu'il est possible de reconstruire une image ou un volume OCT avec une grande fidélité avec seulement 10\% des mesures. 
%
\subsubsection{Curriculum Vit\ae}
%
Mon Curriculum Vit\ae~ est présenté à la fin du document. Il renseigne le déroulement de mon parcours professionnel, l'ensemble des encadrements effectués ou en cours, les enseignements que je dispense régulièrement depuis 2012, mes activités contractuelles, les prix et les distinctions obtenus (individuelles et collectives), ainsi qu’une liste détaillée de mes publications scientifiques. 
\section{Contexte, collaborations et encadrements}
%
L'intégration dans l'institut FEMTO-ST (département AS2M) à Besançon, lors de mon recrutement au CNRS, est une opportunité et une chance pour diverses raisons (à l'exception du climat franc-comtois !) : des moyens de travail exceptionnels et un environnement humain enrichissant. En effet, j'ai eu toutes les conditions de travail nécessaires pour entamer ma tâche de chercheur CNRS. Très rapidement, j'ai intégré une jeune équipe de recherche, animée par Nicolas Andreff (Prof. à l'Univ. de Franche-Comté). \\

\begin{figure}[!h]
\centering
\includegraphics[width=\columnwidth]{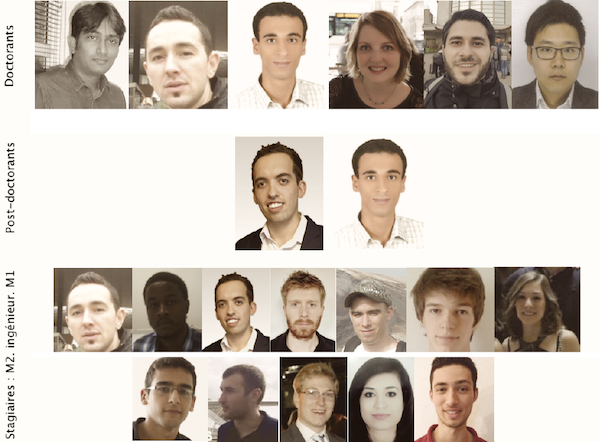}
\caption{Photographie de mes (co)-encadrements (stagiaires, doctorants et post-doctorants) réalisés depuis 2012.}
\label{fig.team}
\end{figure}

Naturellement, ces travaux sont le fruit d'un travail collectif et de nombreuses collaborations. Les travaux réalisés sur les ondelettes et les shearlets sont issus de l'encadrement de deux doctorants Mouloud Ourak (2013-2016) et Lesley-Ann Duflot (2014-2018)\footnote{En cotutelle avec l'équipe Rainbow, IRISA, Inria Bretagne Atlantique.}. Les travaux sur l'acquisition comprimée sont le fruit des travaux menés avec des doctorants : Lesley-Ann Duflot et Bassel Haydar (2017-...)\footnote{En co-tutelle avec l'équipe EnCoV, Institut Pascal, Clermont-Ferrand.}. Les tous premiers travaux, sur l'utilisation de l'information visuelle fréquentielle (transformée de Fourier) dans une boucle de commande par asservissement visuel, ont été développés dans le cadre de la thèse Naresh Marturi (2010-2013)~\citep{marturi2013visual,marturi2014visual,tamadazteTase2016,tamadazteTim2016}\footnote{J'ai participé à son encadrement sur les deux dernières années de sa thèse.}. Par ailleurs, les travaux sur le chirurgie laser guidée par l'image sont le résultat de plusieurs encadrements de stagiaires (2012-2015) (voir Fig.~\ref{fig.team}). Enfin, un travail méthodologique conséquent sur la commande d'outils chirurgicaux droits et courbés à travers des orifices, naturels ou artificiels, a été réalisé dans le cadre de la thèse de Bassem Dahroug (2014-2018)~\citep{dahroug20163d,dahrougicra2017,Dahroug2018,rosa2018online}, mais celui-ci n'a pas été abordé dans ce manuscrit pour des raisons de cohérence avec le reste des thématiques, par le fait que le projet dans lequel s'inscrivent ses travaux n'a démarré qu'en 2018. Les travaux de Bassem Dahroug peuvent être consultés dans~\citep{dahroug2018phdthesis}.

\begin{figure}[!h]
\centering
\includegraphics[width=.7\columnwidth]{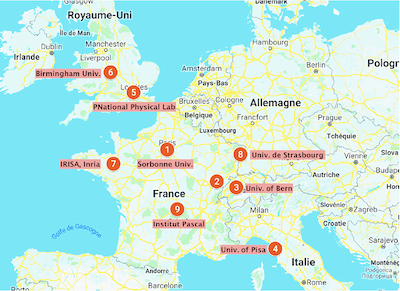}
\caption{Aperçu des collaborations initiées depuis 2012.}
\label{fig.carte_collaborations}
\end{figure}

Les travaux décrits dans ce document sont aussi le fruit de plusieurs collaborations locales, nationales et internationales initiées à partir de 2012 (Fig.~\ref{fig.carte_collaborations}). Ces collaborations sont structurées soit dans le cadre de projets de recherche, de co-encadrements ou d'échanges scientifiques : National Physical Lab (Londres, R-U), IRISA/Inria (Rennes, FR), University of Pisa (Pise, IT), Institut Pascal (Clermont-Ferrand, FR), ICube (Strasbourg, FR), Sorbonne Université/ISIR (Paris, FR), Kuka Robotics UK (Birmingham, R-U), University of Birmingham (Birmingham, R-U) et University of Bern (Berne,  CH).  

\begin{figure}[!h]
\centering
\includegraphics[width=.9\columnwidth]{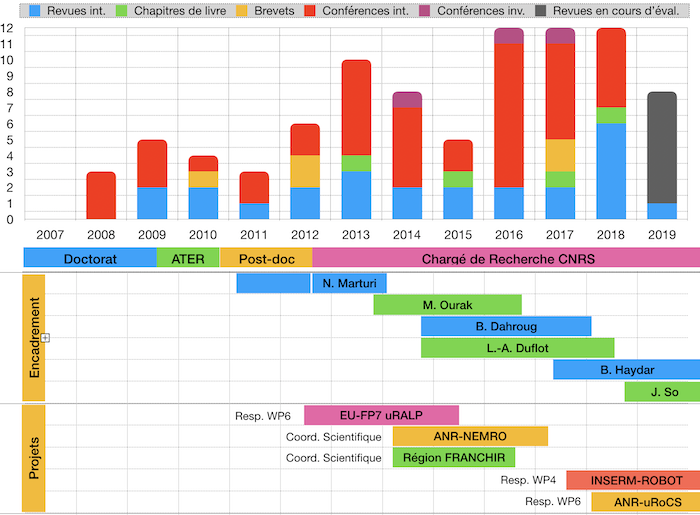}
\caption{Chronologie des thématiques, des thèses, des projets et de la production scientifique.}
\label{fig.chronologie}
\end{figure}

Par ailleurs, la Fig.~\ref{fig.chronologie} donne un aperçu sur le déroulement chronologique et scientifique des encadrements de thèses et des contrats de recherche associés. 
 
%% file: fichiers/clinique_v3.tex

\chapter{Cadre clinique des travaux}\label{chap.clinique}

\minitoc

\emph{Ce chapitre est consacré à la présentation du cadre clinique de mes travaux scientifiques décrits plus loin dans ce document. La majorité de ces travaux scientifiques traite des pathologies ORL, notamment l'oreille moyenne, les cordes vocales et la muqueuse olfactive (épithélium olfactif). A noter que le travail sur la compréhension et la caractérisation de l'olfaction a un lien direct avec les neurosciences pour le diagnostic, à un stade précoce, des maladies neurodégénératives comme l'Alzheimer ou la maladie de Parkinson. Ces travaux sont réalisés en étroites collaborations avec les cliniciens des hôpitaux de Besançon et de la Pitié-Salpêtrière à Paris. A un degré moindre, j'émarge à des travaux sur le diagnostic et la chirurgie de tissus cancéreux dans le système digestif.   \\
Nous avons toujours eu un souci de réaliser ces travaux dans un cadre translationnel, soit via la conception de démonstrateurs précliniques, soit via des tests précliniques, sur bancs de tests, soit sur cadavres. J'ai participé, à titre personnel ou par des encadrements (essentiellement des étudiants en master), à la réalisation de certains de ces démonstrateurs. 
 }
%
\section{Diagnostic des maladies neurodégénératives}
%
Ce travail est réalisé dans le cadre du projet ANR NEMRO (2014-2019)\footnote{Endoscopie nasale par OCT microrobotisée : impact du déficit olfactif sur les maladies neurodégénératives.},  dont je suis le coordinateur scientifique. L'objectif clinique de NEMRO est de démontrer cliniquement la corrélation entre la perte de l'odorat et l'apparition des maladies neurodégénératives à long terme (5 à 10 ans).  
%
\subsection{Liens entre la perte de l'odorat et la démence}
%
 Les maladies neurodégénératives, comme la maladie d'Alzheimer, sont devenues la principale priorité des autorités sanitaires de nombreux pays développés. Le "\emph{World Alzheimer Report 2014}" indique que le nombre de personnes atteintes de la maladie d'Alzheimer dans le monde est estimé à 44 millions. Ce dernier devrait doubler d'ici 2030 et plus que tripler d'ici 2050. Le coût estimé à l'échelle mondiale (2010) était de 604~Mrds$\$$ pour la maladie d'Alzheimer. La "\emph{International Federation of Alzheimer Associations}", en accord avec les spécialistes de la maladie, a souligné la nécessité d'envisager de nouvelles pistes de recherche pour le diagnostic, et développer des nouveaux moyens d'accompagnement des malades et thérapeutiques en rupture. Aujourd'hui, le diagnostic se fait généralement après l'apparition des premiers signes de perte de mémoire/motricité, ce qui est déjà bien tard et que les soins actuels sont davantage palliatifs, que thérapeutiques. 

Dans ce projet, nous étudions la possibilité de développer une méthode de diagnostic précoce de certaines maladies de démence par l'analyse du fonctionnement de l'odorat dans ses états sains et pathologiques. Il a été démontré cliniquement que le dysfonctionnement de l'olfaction sans raisons apparentes (traumatismes, pathologie nasale, ...) est le lien de cause à effet entre ce dysfonctionnement et l'apparition d'une pathologie de type démence dans le cerveau~\citep{Joussain2015ApplicationOT}.  
\begin{wrapfigure}{r}{0.25\textwidth}
  \vspace{-20pt}
  \begin{center}
    \includegraphics[width=0.24\textwidth]{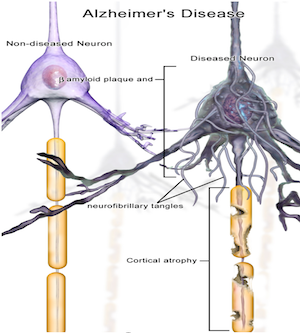}
  \end{center}
  \vspace{-20pt}
  \caption{Neurones : sain et mort.}
  \label{fig.neurone}
  \vspace{-10pt}
\end{wrapfigure}

\emph{Marcel Proust} (1871-1922), dans son livre "\emph{A la recherche du temps perdu}" (1$^{er}$ tome) décrit parfaitement ce lien existant entre l'odorat et la mémoire. Il s'agit du célèbre passage connu sous le nom de la "\emph{madeleine de Proust}" dans lequel il décrivit la façon dont l'odeur de la madeleine le ramenait à son enfance avec de nombreux détails précis (temps, lieu, contexte, ...). Cette connexion entre la mémoire et l'olfaction a suscité l'intérêt de plusieurs neurologues et scientifiques pour en comprendre le fonctionnement. 

Généralement, le vieillissement s'accompagne d'une diminution du volume du cerveau, de l'épaisseur corticale, de l'intégrité de la substance blanche et de l'activité neuronale, ... Ces variations s'opèrent particulièrement au niveau du "lobe temporal médian" et des pôles frontaux du cerveau. Il s'agit des régions cérébrales associées à la mémoire, aux fonctions émotionnelles et comportementales ainsi qu'au traitement olfactif~\citep{Kemper1984}. Dans plusieurs études scientifiques très sérieuses, il a été démontré que le dysfonctionnement de l'olfaction peut être la cause d'une maladie neurodégénérative~\citep{Alves2014, Godoy2014, Joussain2015ApplicationOT}. A titre d'exemple, une grande étude clinique réalisée sur 39 patients a montré que les troubles olfactifs affectent à la fois la détection et la reconnaissance~\citep{Rahayel2014}. 
%
%
\begin{figure}[!h]
\centering
\includegraphics[width=0.7\columnwidth]{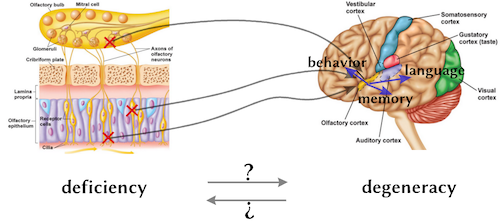}
\caption{Corrélation entre la déficience olfactive et la dégénérescence neuronale.}
\label{fig.link}
\end{figure}

Durant les cinq dernières années, des scientifiques se sont intéressés à la manière d'inverser les signes de la maladie d'Alzheimer sur des modèles animaux (souris transgéniques) en travaillant sur la réparation du dysfonctionnement de l'olfaction~\citep{Saar2015}. Les résultats prometteurs issus de ces travaux ont montré que les marqueurs relatifs à l'Alzheimer ont diminué. 

Une des hypothèses fortes sur l'apparition de la maladie d'Alzheimer est le développement et l'agrégation d'une protéine appelée \emph{Bêta-Amyloïde} (\emph{B-A}) (Fig.~\ref{fig.neurone}). La question qui se pose, les \emph{B-A} se développent t'elles d'abord sur l'épithélium olfactif puis elles migrent (à travers la lame criblée) pour créer des agrégations sur les neurones pour bloquer les transmissions entre eux ou est-ce l'inverse ? C'est-à-dire les \emph{B-A} se développent dans le cerveau avant de migrer dans l'autre sens vers l'épithélium olfactif pour créer des agrégats qui provoquent un dysfonctionnement de l'olfaction. Cependant, il n'existe pas de moyens de vérifier ces hypothèses directement sur site (épithélium olfactif). 

Le projet NEMRO vise à développer un système endoscopique flexible robotisé dont le diamètre extérieur n'excède pas 2~mm. Le robot miniature doit naviguer dans les fosses nasales sans collisions (l'examen clinique souhaité sera sans anesthésie générale) pour venir se placer en face de la muqueuse olfactive. Aussi, le robot doit être équipé, en sa partie distale, d'un moyen de caractérisation capable de faire la différence entre une muqueuse olfactive saine et une pathologique. Le mécanisme du robot flexible est inspiré des travaux de~\citep{Webster2010review} sur les tubes concentriques. Ce type de mécanisme a été choisi pour disposer d'un canal axial libre, qui permettra le passage d'un outil de caractérisation \emph{in situ}. Cette caractérisation peut être réalisée grâce à un système d'imagerie miniature (fibré) pour l'acquisition de biopsies optiques 2D ou 3D. Nous travaillons sur la microscopie confocale à fluorescence et la tomographie par cohérence optique. Les avantages de ces deux modalités d'imagerie sont leur capacité à imager les tissus en profondeur et à des résolutions axiales et spatiales micrométriques. Les premiers résultats obtenus sur des souris transgéniques sont prometteurs~\citep{Etievant2019}.
%
\subsection{Prototype NEMRO}
%

\begin{figure}[!h]
\centering
\includegraphics[width=\columnwidth]{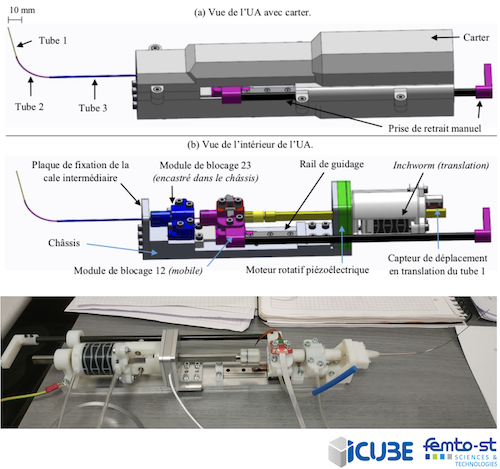}
\caption{Concept du robot NEMRO (CAO et prototype).}
\label{fig.demo_nemro}
\end{figure}

Un des objectifs scientifiques de ce travail sur le fonctionnement du système olfactif et son impact sur les maladies neurodégénératives est de concevoir, dans une version préclinique, l'ensemble du système de "diagnostic" de la démence. Ce démonstrateur est en cours de réalisation grâce à la collaboration entre FEMTO-ST (Besançon) et ICube (Strasbourg) sous les recommandations cliniques d'un chirurgien ORL de l'hôpital de Besançon~\citep{girerd18phdthesis}\footnote{Thèse encadrée par Pierre Renaud (ICube) et Kanty Rabenorosoa (FEMTO-ST) dans le cadre du projet ANR NEMRO.}.  Il doit intégrer les contributions scientifiques sur la navigation automatique intranasale et sans collisions, l'acquisition de biopsies optiques répétitives (chapitre~\ref{chap.wavelet}) et l'acquisition comprimée (chapitre\ref{chap.cs}). 

La Fig.~\ref{fig.demo_nemro} montre un aperçu (CAO et première version du dispositif ) de l'unité d'actionnement, ainsi que de la partie insérable dans les fosses nasales. Ce dispositif a été pensé de manière à être capable d'examiner environ 80\% des patients sans apporter de modification sur la partie distale. La conception a été réalisée en s'appuyant sur les résultats de l'analyse de l'intervariabilité nasale (forme, longueur, largeur, ...) en utilisant les données (scanners) d'une trentaine de patients adultes~\citep{girerd17abme}. 
%
\section{Chirurgie laser des cordes vocales}
\begin{wrapfigure}{l}{0.25\textwidth}
  \vspace{-20pt}
  \begin{center}
    \includegraphics[width=0.24\textwidth]{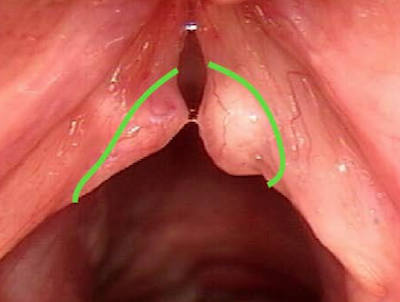}
  \end{center}
  \vspace{-20pt}
  \caption{Cancer des cordes vocales.}
  \label{fig.vocal_folds}
  \vspace{-10pt}
\end{wrapfigure}

%
Les cordes vocales, appelées également "plis vocaux", sont constituées de deux membranes fines, flexibles et fragiles situées dans le larynx, formant une structure en "V" pointée vers l'avant. Parmi les pathologies des cordes vocales les plus fréquentes, on peut citer les nodules, des petites boules de tissu qui peuvent se développer un peu partout sur la surface des cordes vocales ; les kystes appelés intracordaux ; les cancers des cordes vocales (Fig.~\ref{fig.vocal_folds}), ... Ces pathologies touchent typiquement les fumeurs et certaines catégories de métiers, qui provoquent un surmenage ou un malmenage vocal : les gens du spectacle, les enseignants, etc.
%
%
\subsection{Phonochirurgie laser}
%

Le traitement de ces pathologies est généralement réalisé par chirurgie. De fait de la fragilité des cordes vocales et de la petite taille des nodules ou tumeurs cancéreuses qui s'y développent, la chirurgie (résection ou ablation) est effectuée à l'aide d'un laser de puissance (laser de chirurgie). La phonochirurgie laser est la solution la plus préconisée pour garantir au patient de garder les fonctionnalités des cordes vocales, en particulier la parole. C'est une chirurgie délicate, réalisée systématiquement sous anesthésie générale par un chirurgien confirmé par une laryngoscopie en suspension. Lors de l'intervention, le praticien dispose d'une panoplie d'instruments de microchirurgie, d'un laser de chirurgie et d'un microscope optique placé à l'extérieur de patient.  

\begin{figure}[!h]
\centering
\subfigure[Premier concept.]{
\includegraphics[width=0.3\columnwidth,height=3.8cm]{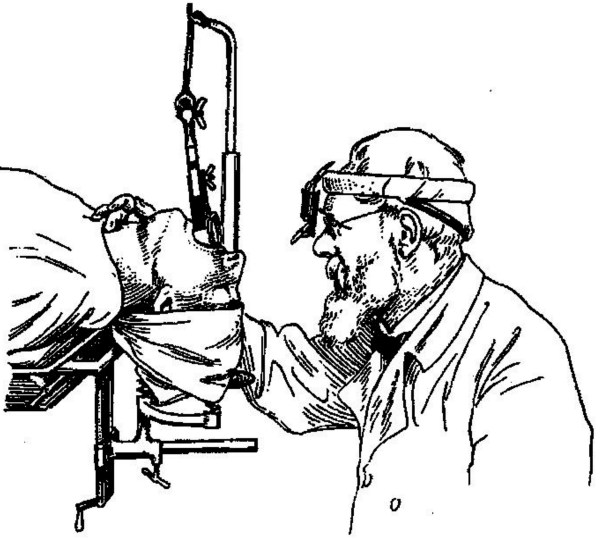}}
\subfigure[Système AcuBlade.]{
\includegraphics[width=0.3\columnwidth,height=3.8cm]{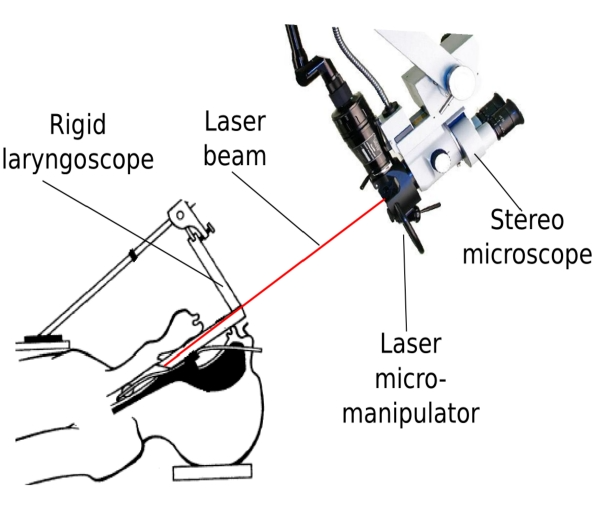}}
\subfigure[Système  $\mu$RALP.]{
\includegraphics[width=0.3\columnwidth,height=3.8cm]{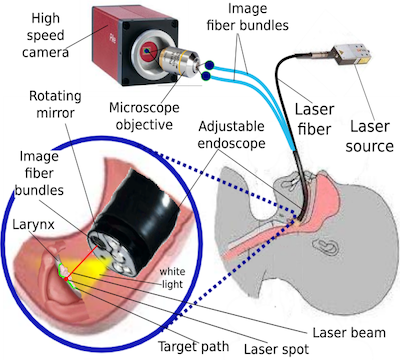}}
\caption{Laryngoscope : a) le premier concept, (b) l'actuel système de chirurgie laser des cordes vocales, et (b) celui développé dans le cadre du projet $\mu$RALP.}
\label{fig:history}
\end{figure}

Il existe plusieurs types de lasers chirurgicaux comme le CO$_2$, LBO ou encore le KTP, probablement les deux les plus utilisés en chirurgie laser~\citep{Mattos2014}. Les pratiques actuelles de l'utilisation d'un laser chirurgical en chirurgie des cordes vocales (ou en laryngoscopie, de manière générale) présentent plusieurs difficultés et limitations, même avec les systèmes (partiellement robotisés) les plus récents (Fig.~\ref{fig:history}(b)) :
\begin{itemize}
\item l'utilisation d'un système de vision indirect, c'est-à-dire un microscope optique placé à l'extérieur du patient, qui nécessite une extension maximale de la nuque du patient pour obtenir une ligne droite entre le système de vision et les cordes vocales. Cette position, très inconfortable pour le patient, peut exclure une certaine catégorie de patients, dont la nuque est moins souple, et peut engendrer des traumatismes post-opératoires, plus ou moins importants. 
\item la distance, environ 400~mm, entre la source laser et la cible (cordes vocales) peut poser un problème de précision de l'acte chirurgical, notamment à cause de l'effet levier. Les dégâts dus à ces erreurs de positionnement du spot laser sur les cordes vocales peuvent être irréversibles.
\item le danger que peut présenter le laser chirurgical, qui se retrouve sans protection entre la source et la bouche du patient. 
\item la précision et la réussite de l'acte chirurgical, dépend fortement de l'expertise du praticien du fait que les mouvements de celui-ci doivent être d'une grande précision. 
\item le balayage laser sur les cordes vocales est effectué en mode manuel ou semi-automatique à l'aide d'un joystick (qui contrôle les miroirs de balayage, de type galvanométrique).
\end{itemize}

Le projet EU-FP7 $\mu$RALP (2012-2015) \footnote{(Micro-Technologies and Systems for Robot-Assisted Laser Phonomicrosurgery) dont les partenaires sont : IIT (Gênes, Italie), FEMTO-ST (Besançon, France), LUH (Hanovre, Allemagne), l'hôpital de Besançon et l'hôpital de Gênes.}, dont j'étais responsable d'un WP, a pour objectif de développer une nouvelle génération de systèmes de chirurgie laser des cordes vocales à fort degré d'"intégrabilité". Il vise à dépasser les contraintes et limitations actuelles de ce type de chirurgie, souvent réalisées par exemple par le dispositif AcuBlade\footnote{Commercialisé par la société Lumenis (www.bernas-medical.com/fr).} illustré à la Fig.~\ref{fig:history}(b), par le développement d'un robot endoscopique qui doit intégrer les composants suivants : 
\begin{itemize}
\item un système d'imagerie stéréoscopique (en couleur) avec un minimum de 25 images par seconde,
\item un système d'imagerie stéréoscopique (monochrome ou en couleur) à très haute cadence c'est-à-dire plusieurs centaines d'images/seconde,
\item un éclairage fibré à lumière froide,
\item un laser de chirurgie et un laser de visée (visible) couplés dans la même fibre,
\item un dispositif de balayage laser miniature de type miroir actionné à 2 DDL.
\end{itemize}

Le projet impose des contraintes dimensionnelles et d'encombrements très forts, à savoir que, tous les éléments, mentionnés ci-dessus, doivent tenir dans la partie tubulaire du système endoscopique dont le diamètre extérieur ne doit dépasser 20~mm. Deux éléments constituent des challenges élevés de conception : le microrobot (miroir actionné), qui doit assurer le balayage laser sur le tissu doit être contenu dans un volume inférieur à 1~mm$^3$, et le dispositif de vision rapide stéréoscopique qui nécessite une conception optomécatronique adaptée (la dimension du système d'imagerie ne doit pas dépasser quelques millimètres). J'ai contribué, soit à titre personnel ou par l'encadrement d'étudiants, au développement de ces deux éléments : mécatronique (miroir actionné) et optomécatronique (système de vision rapide fibré).  
%
\subsection{Prototype $\mu$RALP}
%

\begin{wrapfigure}{l}{0.20\textwidth}
  \vspace{-20pt}
  \begin{center}
    \includegraphics[width=0.19\textwidth]{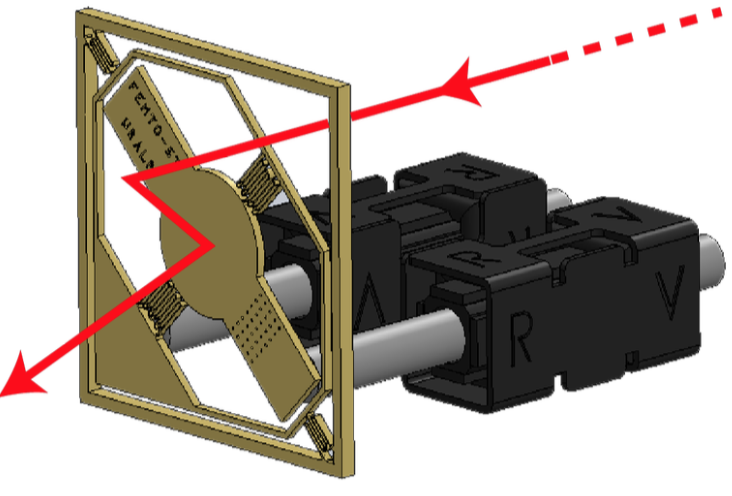}
  \end{center}
  \vspace{-20pt}
  \caption{Miroir passif et actionneurs \emph{Squiggle}.}
  \label{fig.mirror}
  \vspace{-10pt}
\end{wrapfigure}

Plusieurs dispositifs miniatures (microrobots) ayant pour objectif le contrôle de l'orientation du spot laser sur les cordes vocales, ont été développés. Un de ces microrobots, a été intégré dans le démonstrateur final du projet, consistait à combiner une partie active (mobile) obtenue avec deux actionneurs piézoélectriques linéaires de type \emph{Squiggle}\footnote{https://www.newscaletech.com.} et une partie passive (miroir en silicium mobile mais non actionné). Les actionneurs piézoélectriques viennent pousser, séparément, par des mouvements de translation les deux parties rotatives du miroir en silicium pour guider la rotation du faisceau laser réfléchi dessus. Le miroir dispose également de deux micro-ressorts, qui rappellent indépendamment les deux parties mobiles du miroir, lorsque les actionneurs \emph{Squiggle} se retirent, c'est-à-dire translatent dans l'autre sens (voir Fig.~\ref{fig.mirror}). Cette cinématique permet d'obtenir des angles de rotation allant jusqu'à 45$^\circ$ sur chacun des axes et atteignant une résolution angulaire de 0.017$^\circ$ (équivalent à 1~$\mu$m sur les cordes vocales).

\begin{figure}[!h]
\centering
\includegraphics[width=.8\columnwidth]{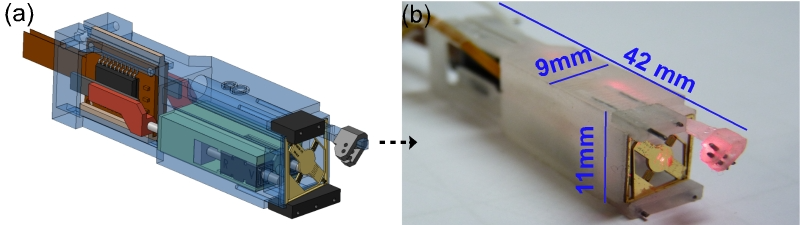}
\caption{Aperçu  du microrobot \emph{SquipaBot} intégré dans le robot de chirurgie laser.}
\label{fig.squiggle_rob}
\end{figure}

Le dispositif, appelé \emph{SquipaBot}, de guidage laser a été intégré dans un boîtier mesurant : longueur = 42~mm, hauteur = 11~mm et largeur = 9~mm. Ce boîtier doit s'insérer dans la longueur de l'endoscope robotisé $\mu$RALP respectant parfaitement les contraintes dimensionnelles de la chirurgie laser des cordes vocales. Celui-ci embarque plusieurs autres éléments : le miroir en silicium, les deux actionneurs linéaires, deux capteurs magnétiques (un pour chaque actionneur), un prisme d'un millimètre-cube pour le renvoi du laser sur le miroir mobile, un laser fibré, et des câbles de d'alimentation. Une fois tous ces éléments assemblés, nous obtenons un microrobot de guidage laser complètement emboîté sans aucun collage ni soudure, simplement grâce à un système de jeux mécaniques préalablement calculé. Le dispositif final est illustré sur la Fig.~\ref{fig.squiggle_rob}. 

\begin{figure}[!h]
\centering
\includegraphics[width=.8\columnwidth]{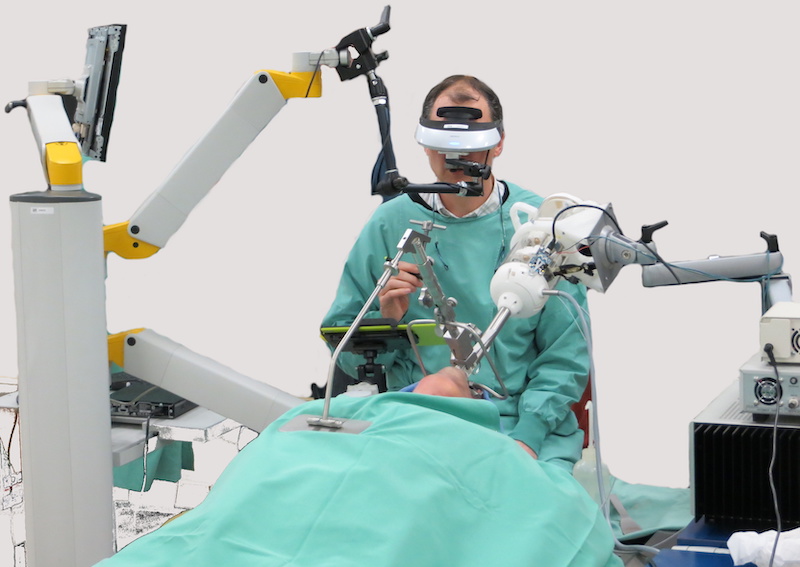}
\caption{Photographie du démonstrateur final du projet $\mu$RALP lors de tests d'évaluation sur cadavre.}
\label{fig.demo_uralp}
\end{figure}

Le démonstrateur final du projet $\mu$RALP intègre les contributions (matériels et logiciels) de tous les partenaires du projet (Fig.~\ref{fig.demo_uralp}). Le système de phonochirurgie laser, dans son ensemble, a été évalué par deux chirurgiens, à plusieurs reprises sur cadavres. Son fonctionnement est le suivant :  le chirurgien insert la partie flexible de l'endoscope robotisé dans le larynx jusqu'à ce que la partie distale soit en face-à-face avec les cordes vocales (à environ 20~mm de distance). En plus du dispositif de guidage laser, la partie distale embarque également : un système d'éclairage fibré, deux caméras miniatures couleur, et deux caméras fibrées rapides. Le chirurgien dispose d'un système de vision augmentée sur lequel est projeté le modèle 3D des cordes vocales reconstruit (en temps réel) à partir des images issues des caméras miniatures. Par ailleurs, le chirurgien dessine sur une tablette tactile la trajectoire que le laser doit suivre pour réaliser la résection des tissus pathologiques.

%
\section{Chirurgie de l'oreille moyenne}
%
La chirurgie de l'oreille moyenne est la troisième brique de nos travaux sur les pathologies ORL après ceux sur le larynx (cordes vocales) et le nez (olfaction et maladies neurodégénératives). Ce développement est réalisé en collaboration avec les hôpitaux de Besançon et de Pitié Salpêtrière Paris, ainsi que deux partenaires académiques : FEMTO-ST (Besançon) et l'ISIR (Paris), réunis dans le projet ANR $\mu$RoCS\footnote{MicroRobot-Assisted Cholesteatoma Surgery.}. Ce projet vise à développer, entre autres, deux dispositifs innovants, l'un mécatronique (microrobot à fort degré de courbure) et l'autre optique (système OCT fibré pour l'acquisition de biopsie optique) pour la différenciation des tissus sains de ceux cholestéatomeux à une résolution cellulaire. Je suis responsable du WP traitant de ce dernier point.
%
\subsection{Cholestéatome}

\begin{wrapfigure}{r}{0.20\textwidth}
  \vspace{-20pt}
  \begin{center}
    \includegraphics[width=0.19\textwidth]{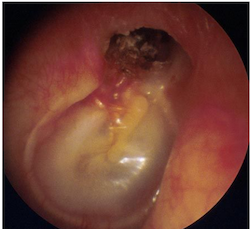}
  \end{center}
  \vspace{-20pt}
  \caption{Exemple d'un cholestéatome.}
  \label{fig.cholesteatome}
  \vspace{-10pt}
\end{wrapfigure}
L'otite chronique représente un état inflammatoire de la muqueuse de la cavité tympanique. Elle peut faire suite à un épisode infectieux aigu. Elle peut, dans d'autres cas, être le résultat d'un problème ou d'une association de plusieurs pathologies primitives de l'oreille moyenne tels que le dysfonctionnement de la trompe d'Eustache, un défaut de résorption des gaz ou une hyper-production du mucus de la muqueuse ou une perforation tympanique. Cet état peut se compliquer par l'introduction, dans l'oreille moyenne, de tissu épidermique qui va grossir progressivement, appelé le \emph{cholestéatome}~\citep{sheehy1977cholesteatoma,kemppainen1999Finland}. Il se présente par une oreille qui suinte avec, à l'examen, la présence de squames sur un tympan remanié (Fig.~\ref{fig.cholesteatome}). La consistance est comparée à celle d'un "\emph{marron glacé}", sous forme d'une tumeur assez bien limitée s'émiettant au contact. Son agressivité locale, avec érosion osseuse, cause les différentes complications suivantes : surdité, vertiges, paralysie faciale, etc. 

\begin{figure}[!h]
\centering
\includegraphics[width=.6\columnwidth]{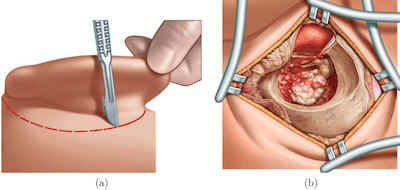}
\caption{Illustration de la chirurgie du cholestéatome telle que réalisée actuellement.}
\label{fig.chirurgie_oreille}
\end{figure}

Le cholestéatome touche environ 10 000 personnes par an en France. Le seul moyen d'y remédier est la chirurgie pour exciser méthodiquement tous les tissus cholestéatomeux. Les cellules cholestéatomeuses résiduelles seront responsables, à court terme (6 à 18 mois), d'un cholestéatome qu'il faudra réséquer à nouveau par une seconde chirurgie, celle-ci doit être plus exhaustive que la première afin d'éviter une troisième intervention. 

La chirurgie actuelle consiste à réaliser un trou dans la partie postérieure du pavillon suffisamment large pour accéder à la cavité de l'oreille moyenne où se développe le cholestéatome. Ce trou est réalisé dans la mastoïde, mesure généralement entre 10~mm et 20~mm. Il sert à la vision indirecte du chirurgien (en utilisant un microscope optique placé à l'extérieur) et à faire passer des instruments adaptés pour la résection des tissus cholestéatomeux (voir Fig~\ref{fig.chirurgie_oreille}).

Une fois le trou réalisé, le chirurgien procède à l'enlèvement des tissus pathologiques présents dans la cavité de l'oreille moyenne tout en faisant attention de ne pas abimer certains organes comme le nerf facial ou les osselets (lorsqu'ils ne sont pas abîmés). Cependant, le chirurgien ne dispose d'aucun moyen de vérification, à l'exception de son expérience, et de l'exhaustivité de l'acte. Cette technique (il existe d'autres méthodes plus ou moins proches) pose au moins deux problèmes, celui de l’invasivité de l'intervention et celui de manque de moyens de différentiation en temps-réel des tissus pathologiques de ceux sains pour éviter une seconde, voire une troisième intervention. 
%
\subsection{Prototype $\mu$RoCS}
%
Le dispositif $\mu$RoCS consiste à combiner deux approches robotiques à deux échelles : un robot médical de dimensions standards existant de placement d'osselets développé en partie par notre partenaire (l'ISIR), et commercialisé par la société "Collin Medical" sous le nom \emph{RobOtol}\footnote{http://www.collinmedical.fr/portfolio-item/robotol/}, et un microrobot flexible à fort degré de courbure dont la partie axiale est libre (en cours de conception). Le système microrobotique vient se fixer sur RobOtol comme un effecteur à plusieurs degrés de liberté. La combinaison des deux cinématiques permet de créer un système redondant dont la partie distale est un microrobot de diamètre extérieur ne dépassant pas 2~mm~\citep{Dahroug2018}. 

\begin{figure}[!h]
\centering
\includegraphics[width=.8\columnwidth]{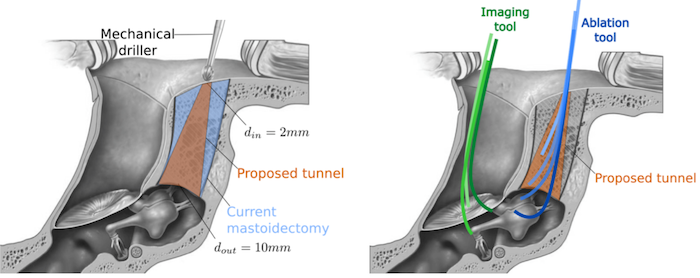}
\caption{Conceptualisation de l'approche chirurgicale visée dans le projet $\mu$RoCS.}
\label{fig.concept_urocs}
\end{figure}

Pour atteindre la cavité de l'oreille moyenne, nous envisageons deux voies d'accès ; à travers un trou de 2~mm de diamètre percé dans la mastoïde, et un autre réalisé le long du conduit auditif (en décollant le tampon légèrement sur le côté afin de le préserver) (voir Fig.~\ref{fig.concept_urocs}). Un des orifices servira à faire passer le microrobot équipé d'un instrument de résection du cholestéatome et l'autre sera utilisé pour insérer l'outil optique fibré de caractérisation des tissus en temps réel. La procédure de résection s'effectue en deux étapes ; la première consiste à enlever le maximum des tissus cholestéatomeux par un instrument de résection "mécanique" et la seconde utilisera un laser de puissance pour carboniser les cellules pathologiques résiduelles détectées par le système optique de caractérisation. 

Par conséquent, la chirurgie du cholestéatome devient moins invasive et surtout exhaustive grâce à l'outil de différenciation des tissus sains et pathologiques. 

\begin{figure}[!h]
\centering
\includegraphics[width=.9\columnwidth]{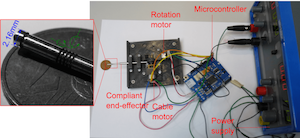}\\
\includegraphics[width=.55\columnwidth]{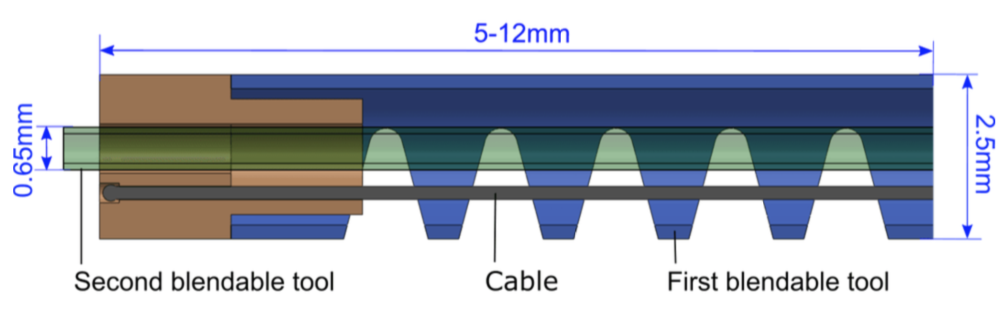}
\caption{Premier prototype réalisé pour la partie microrobotique du démonstrateur $\mu$RoCS.}
\label{fig.demo_urocs}
\end{figure}
%
%
\section{Cancer du système digestif}
%
\begin{figure}[!h]
\centering
\includegraphics[width=.45\columnwidth, height = 8cm]{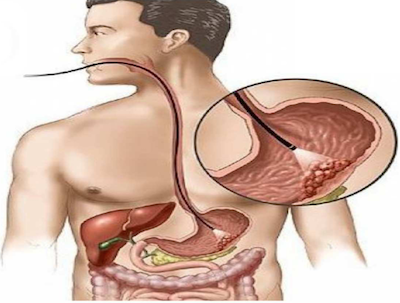}
\caption{Concept d'un fibroscope robotisé pour la chirurgie du cancer du système digestif.}
\label{fig.demo_robot}
\end{figure}

Nos travaux sur la biopsie optique robotisée du système digestif sont réalisés dans le cadre du projet INSERM-ROBOT (2017-2020) \footnote{Robotics and Optical Coherence Tomography (OCT) for Optical Biopsy in the Digestive Tract.} (Fig.~\ref{fig.demo_robot}). L'idée est d'équiper un endoscope (sous forme d'un fibroscope) robotisé\footnote{Concept développé initialement dans le cadre du Labex ACTION (http://www.labex-action.fr/fr/micro-mechatronics).} par un nouveau concept de système de tomographie par cohérence optique miniature (développé par le département MN2S de l'institut FEMTO-ST) capable de réaliser une acquisition volumique sans tâche de balayage grâce à un interféromètre intégré.

Dans ROBOT, je m'occupe de la partie relevant de la conception de lois de commande pour la réalisation de biopsies optiques répétitives capable de différencier les tissus cancéreux de ceux sains. Le chirurgien aura alors une vision temps-réel du site opératoire lui permettant d'exciser de manière exhaustive les tissus pathologiques grâce à une visualisation en profondeur (3D). Des méthodes d'acquisition comprimées seront également développé pour rendre l'acquisition des biopsies optiques les plus rapides possibles. 

\begin{figure}[!h]
\centering
\includegraphics[width=.8\columnwidth]{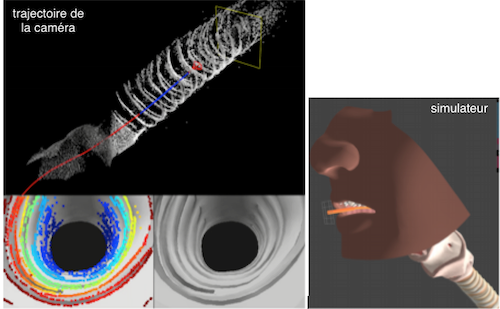}
\caption{Exemple de navigation intracorporelle autonome à l'aide d'un robot continu à tubes concentriques.}
\label{fig.slam_robot}
\end{figure}

Par ailleurs, nous investiguons également dans le développement de méthodes de navigation intracorporelle guidées par la vision pour contrôler la progression du robot fibroscopique dans le système digestif. Ces méthodes sont inspirées des travaux sur la navigation automatique des véhicules autonomes dites SLAM (\emph{simultaneous localization and mapping}). Ces méthodes permettent à la fois la navigation sans collisions dans le système digestif mais aussi la reconstruction 3D (cartographie) de ce dernier. En d'autres termes, construire au fur et à mesure de la progression, le modèle 3D du système digestif, permettra au praticien de se repérer précisément. Sur le modèle reconstruit, il sera alors possible de recaler les biopsies optiques, donnant ainsi une cartographie multimodale du système digestif. Pour ce faire, plusieurs verrous scientifiques doivent être levés : navigation dans un environnement inconnu avec de fortes variations, images peu texturées et généralement de faible résolution, cinématique complexe (robot continu), etc. La Fig.~\ref{fig.slam_robot} montre un exemple d'un premier résultat de navigation intracorporelle d'un robot continu dans un simulateur modulaire développé dans le cadre de ce projet.
%
\section{Bilan}
%
Ce chapitre a donné un aperçu des applications cliniques dans lesquelles s'inscrivent une grande partie de mes contributions scientifiques, qui seront abordées dans la suite du document. Comme chacun peut le constater, une majorité de mes actions ont pour cadre clinique, les pathologies ORL. Ce choix de se concentrer davantage sur une partie du corps permet de constituer une cohérence scientifique et clinique de mes travaux de recherche. 
Par ailleurs, ce chapitre a résumé certaines contributions relatives aux développements de certains démonstrateurs et à la réalisation d'études cliniques (validation de méthodes, vérifications d'hypothèses, ...). \\

Ci-dessous, la liste des publications scientifiques issues de ces travaux. 
\footnotesize
\subsubsection{Liste des publications scientifiques [depuis 2012]}
\begin{enumerate}
%
%
\item  [Ji]  C. Girerd, T. Lihoreau,  K. Rabenorosoa, \textbf{B. Tamadazte}, M. Benassarou, L. Tavernier, L. Pazart, and E. Haffen, N. Andreff, and P. Renaud, \emph{In Vivo Inspection of the Olfactory Epithelium: Feasibility of Robotized Optical Biopsy}, Annals of Biomedical Engineering, 2018, 1-11.
\item  [Ji]  \underline{\underline{R. Renevier}}, \textbf{B. Tamadazte}, K. Rabenorosoa, and N. Andreff, \emph{Microrobot for laser surgery: design, modeling and control}, IEEE/ASME Trans. on Mechatronics, 2018, 22 (1), 99-106.
\item  [Ji] \underline{B. Dahroug}, \textbf{B. Tamadazte}, S. Weber, L. Tavernier, and N. Andreff, \emph{Review on Otological Robotic Systems: Towards Micro-Robot Assisted System for Cholesteatoma Surgery}, IEEE Review on Biomedical Engineering, 2018, 11, 125-142.
\item  [Ji]  A. Etiévant, J. Monnin, T. Lihoreau, \textbf{B. Tamadazte}, L. Tavernier, L. Pazart, E. Haffen \emph{Comparison of non-invasive imagery methods to observe healthy and degenerated olfactory epithelium of mice for early diagnosis of neurodegenerative diseases}, J. of Neurosciences Methods,  \todo{(revised)}. 
\item [Ji] A. V. Kudryavtsev, C. Girerd, P. Rougeot, N. Andreff, P. Renaud, K. Rabenorosoa, and \textbf{B. Tamadazte}, \emph{Autonomous Intracorporeal Deployment of Concentric Tube Robots}, IEEE Trans. on Robotics, \todo{(under review)}. 
\end{enumerate}
\normalsize
\vspace{.5cm}
\noindent
\underline{P. Nom} : doctorant(e)\\
\underline{\underline{P. Nom}} : stagiaire\\

\noindent
Ji : journal international avec comité de lecture\\

%% file: fichiers/laser_v3.tex
%
%
\chapter{Chirurgie laser guidée par l'image}\label{chap.laser}
%
\vspace{1cm}

\newcommand{\pl}{\pnt{\tilde{p}}{{}_{\texttt{L}}}}
\newcommand{\epil}{\pnt{e}{{}_{\texttt{L}}}}
\newcommand{\epir}{\pnt{e}{{}_{\texttt{R}}}}
\newcommand{\epi}{\pnt{e}{{}_{ \texttt{0}}}}

\newcommand{\epin}{\pnt{e}{{}_{\texttt{n}}}}
\newcommand{\epim}{\pnt{e}{{}_{\texttt{m}}}}

\newcommand{\cenn}{\pnt{c}{{}_{\texttt{n}}}}
\newcommand{\cenm}{\pnt{c}{{}_{\texttt{m}}}}

\newcommand{\cenl}{\pnt{c}{{}_{\texttt{L}}}}
\newcommand{\cenr}{\pnt{c}{{}_{\texttt{R}}}}
\newcommand{\cen}{\pnt{c}{{}_{\texttt{0}}}}
\newcommand{\pr}{\pnt{\tilde{p}}_{{}_{\texttt{R}}}}
\newcommand{\dotpl}{\pnt{\dot{\tilde{p}}}_{{}_{\texttt{L}}}}
\newcommand{\vecteur}[1]{\overrightarrow{\point{#1}}} 
\newcommand{\dotpr}{\pnt{\dot{\tilde{p}}}_{{}_{\texttt{R}}}}
\newcommand{\Fmat}[2][]{{}^{{}^#1}\!\matx{F}_{\!{}_{#2}}}
\newcommand{\Fl}{\Fmat[0]{\texttt{L}}}
\newcommand{\Fr}{\Fmat[0]{\texttt{R}}}
\newcommand{\hr}{\vect{h}_{\texttt{R}}}
\newcommand{\hLr}{\vect{h}_{\texttt{L}}}
\newcommand{\xn}[1][]{{\,}^{#1}\unit{x}}
\newcommand{\yn}[1][]{{\,}^{#1}\unit{y}}
\newcommand{\dothr}{{\hr{\dot{h}}}{{}_{\texttt{R}}}}
\newcommand{\dothl}{{\hLr{\dot{h}}}{{}_{\texttt{L}}}}

\minitoc

\emph{Ce chapitre traite de la chirurgie laser guidée par l'image. Le fil conducteur de ces travaux est l'utilisation de l'information visuelle associée à la géométrie projective, pour la formulation de lois de commande efficaces et précises pour le contrôle des déplacements d'un spot laser (respectivement, un faisceau laser) sur une cible (tissu, organe) dont la forme est quelconque. Pour ce faire, nous avons développé plusieurs approches s'inspirant de la géométrie à deux ou à trois vues et de la robotique mobile, séparément ou combinées dans le même schéma de commande. Sous les nombreuses contraintes de la chirurgie laser (fréquence, précision, interaction laser-tissu, étalonnage caméra ou caméra/robot, ...), nous avons dû repenser certains principes de la géométrie pour nous adapter au mieux aux applications. Les méthodes développées ont été validées avec succès suivant plusieurs scénarios : en simulation, sur banc d'essais, ainsi que sur cadavre humain (sous la supervision d'un chirurgien).} 
%
\section{Contexte et positionnement}\label{sec.cahier}
%
Pour rappel, ces travaux ont été réalisés dans le cadre du projet EU FP7 $\mu$RALP, qui traitait de la microchirurgie laser des cordes vocales.  Le cahier des charges de la microchirurgie laser des cordes vocales est le suivant :  1) la précision requise lors d'une tâche de résection (erreur minimale admissible entre les positions finale et désirée du laser sur le tissu), 2) la fréquence de passage du spot laser sur une trajectoire de résection doit être supérieure à 200Hz (pour avoir une découpe nette et éviter la carbonisation des tissus adjacents), et 3) la vitesse de déplacement du laser sur le tissu doit être constante et décorrélée des aspects géométriques et temporels de la courbe de résection/ablation. \\

Pour ce faire, nous avons travaillé sur l'utilisation de commande référencée vision pour contrôler efficacement le déplacement du laser chirurgical sur le tissu. Pour rappel, deux caméras miniatures sont disposées sur la partie distale de l'endoscope robotisé (chapitre~\ref{chap.clinique}). L'idée est alors d'utiliser tout le potentiel d'un système de vision stéréoscopique dans la commande du spot laser. D'une part, il est donc tout à fait possible d'utiliser les propriétés de géométrie épipolaire dans la conception des lois de commande. D'autre part, il est également possible de considérer le miroir de balayage comme une caméra virtuelle à un pixel.  Ajoutée au système de vision stéréoscopique, la conception des lois de commande trifocale (propriétés de la géométrie à trois vues) peut être envisagée.\\

Plusieurs méthodes d'asservissement visuel ont été proposées dans la littérature où les propriétés de la géométrie épipolaire ont été intégrées dans la loi de commande~\citep{hespanha1998can, ruf1999, Rives2000, LamEspAndHor00a, Pari10, Alkhalil12}. Toutefois, en dépit de l'intérêt que ces approches présentent, certaines lacunes subsistent. Par exemple, la contrainte de géométrie épipolaire est difficile à satisfaire dans le cas d'une "baseline" courte ou encore avec des scènes planes comme le montre~\citep{Becerra2009, Yang2014, Montijano2014} qui présentent des singularités dans le contrôleur. Pour y remédier, il est intéressant d'ajouter un capteur supplémentaire (laser, caméra, etc.) pour aboutir à un système à trois vues régi par la géométrie trifocale. Cette géométrie est principalement utilisée dans la reconstruction de scènes~\citep{Liu2013}, le calcul de pose 3D~\citep{Becerra2009}, et aussi pour concevoir des lois de commande d'asservissement visuel~\citep{ShademanJ10, Lopez2010, Becerra2011, Sabatta2013, Becerra2014a, zhang2018visual}. L'apport de la géométrie trifocale permet de lever certaines ambiguïtés dans la loi de commande. Cependant, la conception de lois de commande en faisant appel à la contrainte trifocale, nécessite, inévitablement, la manipulation de plusieurs matrices et donc des inversions matricielles (souvent approchées numériquement), dont les solutions ne sont pas évidentes. \\

Dans notre approche, nous avons revisité la configuration d'un système de vision à trois vues en utilisant uniquement deux caméras et une source laser pour l'écriture d'une version simplifiée de contrainte trifocale plus adaptée au contrôle du déplacement d'un laser sur une surface inconnue. 
%
\section{Bases de la géométrie à plusieurs vues}\label{sec.3vues}
%
La vision géométrique a connu, durant les trois dernières décennies, des avancées \emph{exceptionnelles} dans plusieurs domaines : étalonnage de caméras, suivi visuel, calcul de pose, reconstruction 3D, commande référencée capteur de vision, etc. Les méthodes de commande laser développées dans ce chapitre s'inspirent à la fois de la géométrie à deux vues et de celle à trois vues. Ainsi, en guise de préambule, nous commençons par quelques rappels sur la géométrie d'un point de vue vision par ordinateur.
%
\subsection{Géométrie à deux vues : matrice fondamentale}
%
%
\begin{figure}[!h]
\centering
\includegraphics[width=0.6\columnwidth]{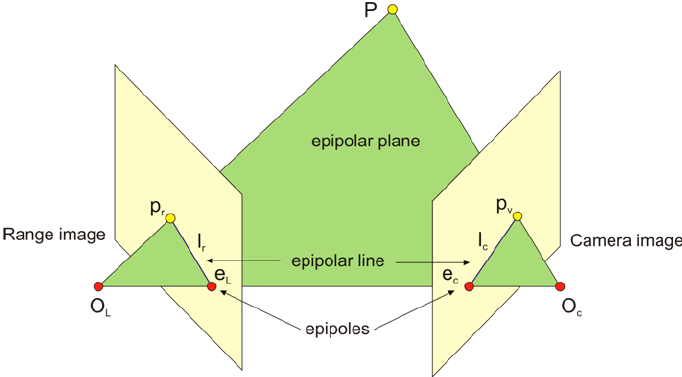}
\caption{Représentation de la géométrie à deux vues et les différentes notations associées.}
\label{fig.epipolar_geo}
\end{figure}

La géométrie à deux vues ou stéréovision consiste à utiliser deux caméras positionnées à deux endroits différents et qui regardent la même scène 3D. Ainsi, considérons un système de vision à deux caméras (stéréoscopique) dont les centres optiques sont notés $\cenl$ et $\cenr$ observant simultanément un point-monde $^0\mathbf{P} = (X, Y, Z)$. Ce dernier peut être projeté en points-image  $\pnt p_\texttt{L} = (x_\texttt{L}, y_\texttt{R})^\top$ et $\pnt p_\texttt{R} = (x_\texttt{R}, y_\texttt{R})^\top$ respectivement dans les plans-image gauche $\phi_\texttt{L}$ et droit $\phi_\texttt{R}$. Le segment ($\cenl \cenr$) définit la \emph{baseline} entre les deux caméras. L'intersection entre le segment ($\cenl \cenr$) et les plans-image $\phi_\texttt{L}$ et $\phi_\texttt{R}$ donnent, respectivement, les points $\epil$ et $\epir$ appelés épipôles. En outre, les droites ($\epil \pnt p_\texttt{L}$) et ($\epir \epir$) sont appelées lignes épipolaires (Fig.~\ref{fig.epipolar_geo}). Il existe alors un ensemble de relations géométriques entre ces différents points, droites, plans, ... communément appelées \emph{contraintes épipolaires}.  Pour tout connaître sur la géométrie à deux ou trois vues, le lecteur est invité à se référer aux livres~\citep{HartleyCUP06,Faugeras1993} où ces concepts sont traités remarquablement et de manière quasi-exhaustive. Dans ce document, nous nous contenterons de la \emph{matrice fondamentale} dont la définition est la suivante :
\begin{defn}
La matrice fondamentale $\mathbf{F}$, appelée également "\emph{tenseur bifocal}" est une matrice 3$\times$3 de rang 2, qui régit un système de vision stéréoscopique. En effet, si $\pnt p_\texttt{L}$ et $\pnt p_\texttt{R}$ forment une paire de points-image 2D issus de la projection du même point-monde $^w\mathbf{P}$, alors on peut écrire que  $\pl^\top \Fmat[\texttt{L}]{\texttt{R}} \pr = 0$ où $\pl= (\pnt p_\texttt{L}, 1)^\top$ et $\pr = (\pnt p_\texttt{R}, 1)^\top$ sont, respectivement, les formes homogènes des points-image $\pnt p_\texttt{L}$ et $\pnt p_\texttt{R}$. 
\end{defn}
La matrice fondamentale peut être déterminée sans connaissance des matrices de projection $\mathbf{P}_\texttt{L}$ et $\mathbf{P}_\texttt{R}$ des deux caméras gauche et droite, ni sur les centres de projection $\cenl$ et $\cenr$, mais à partir d'un ensemble de coordonnées de points-images $\pnt p_\texttt{L}$ et $\pnt p_\texttt{R}$ mis en correspondance. Elle exprime la dualité entre les points-image et les droites 2D dans l'image, c'est-à-dire un point $\pnt p_\texttt{L}$ de l'image de gauche peut être projeté en une droite $l_\texttt{R}$ ($l_\texttt{R}= \Fmat[\texttt{L}]{\texttt{R}} \pl$) dans l'image de droite et inversement  $l_\texttt{L}= \Fmat[L]{R}^\top \pr$  ou encore $\pl^\top l_\texttt{L} = 0$  et  $\pr^\top l_\texttt{R} = 0$  (Fig.~\ref{fig.epipolar_geo}). En d'autres termes, un point de l'image de gauche $p_\texttt{L}$ a pour correspondant un point situé sur la droite épipolaire $l_\texttt{R}$ dans l'image de droite et \emph{vice-versa}. Un cas particulier de la matrice fondamentale est la matrice essentielle $\mathbf{E}$. Elle intervient lorsque les coordonnées-image sont orthonormées, dont l'origine est la projection perpendiculaire des centres de projection $\cenl$ et $\cenr$ sur, respectivement, les plans-image $\phi_\texttt{L}$ et $\phi_\texttt{R}$. 
%
\subsection{Géométrie à trois vues : tenseur trifocal}
%
\begin{figure}[!h]
\centering
\includegraphics[width=0.6\columnwidth, height = 6.5cm]{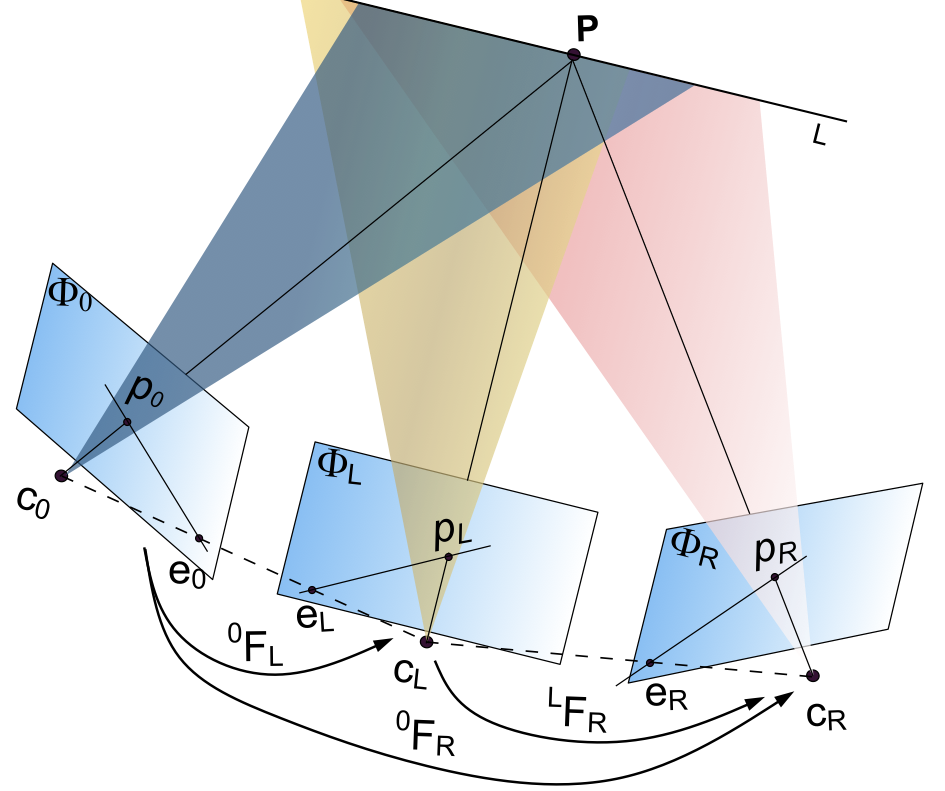}
\caption{Représentation de la géométrie à trois vues et des relations géométriques qui en découlent.}
\label{fig.trifocal_geo}
\end{figure}

La géométrie à trois vues peut être considérée comme une analogie de celle à deux vues. L'ajout d'une troisième caméra permet de décrire de manière plus complète une scène 3D inconnue par un système de relations de géométrie projective entre les trois vues. Ces relations sont encapsulées dans un formalisme qui s'appelle le \emph{tenseur trifocal} définit comme suit : 
\begin{defn}
Le tenseur trifocal $\mathcal{T}$, appelé également "\emph{tri-tenseur}" est un tenseur 3$\times$3$\times$3 (27 éléments dont 18 sont indépendants) englobant les relations géométriques entre les points et les droites dans les différentes vues.
\end{defn}

En rajoutant une troisième caméra au système stéréoscopique de la Fig.~\ref{fig.epipolar_geo}, nous obtenons ainsi un système trifocal comme illustré sur la Fig.~\ref{fig.trifocal_geo}. On note $\cen$ le centre de projection de la troisième caméra, $\pnt p_\texttt{0} = (x_\texttt{0}, y_\texttt{0})^\top$ est la projection du point 3D $^w\mathbf{P}$ dans le plan image $\phi_\texttt{0}$ et $\epi$ l'épipôle associé. En conséquence, il est possible d'écrire les contraintes épipolaires, en utilisant les matrices fondamentales $\Fmat[i]{j}$, pour chaque paire de caméras comme suit : 
\begin{eqnarray}
	\label{eq.trifocal}
   \tilde {\pnt p}_\texttt{0}^\top \Fmat[\texttt{0}]{\texttt{L}} \pl  = 0  \\
   \pl^\top \Fmat[\texttt{L}]{\texttt{R}} \pr  = 0  \\
	 \pnt{\tilde{p}_\texttt{0}} \Fmat[\texttt{0}]{\texttt{R}} \pr^\top   = 0 
\end{eqnarray}

Par ailleurs, les trois contraintes ci-dessus peuvent être regroupées dans une seule transformation point-point-point régie par le tenseur trifocal associé~\citep{HartleyCUP06}. On peut écrire alors,
\begin{equation}
[\pl]_\times  \left( {\sum\limits_{i = 1}^3 {\pnt{\tilde{p}}_i } \textbf{T}_i } \right)[\pr]_ \times   = 0_{3 \times 3} 
\label{eq.torseur}
\end{equation}
où $\as{\vect{v}}$ est la matrice asymétrique associée au produit vectoriel croisé ($\times$) par $\vect{v}$ et l'ensemble des matrices $\textbf{T}_1$, $\textbf{T}_2$, and $\textbf{T}_3$ représentent le tenseur trifocal $\mathcal{T}$$_{3\times 3\times 3}$. De même, il est possible d'écrire la transformation point-point-point comme suit :
\begin{equation}
[\pl]_\times  (\pnt{\tilde{p}}_i \mathcal{T}_i ^{jk} )[\pr]_ \times= 0_{3 \times 3} 
\label{eq.torseur1}
\end{equation}
avec $i$,$j$,$k$ $\in$ [1;3].
%
\section{Asservissement visuel trifocal}\label{sec.trifocal}
%
\subsection{Redéfinition du tenseur trifocal}
\begin{figure}[!h]
  \centering
  \includegraphics[width=0.65\columnwidth]{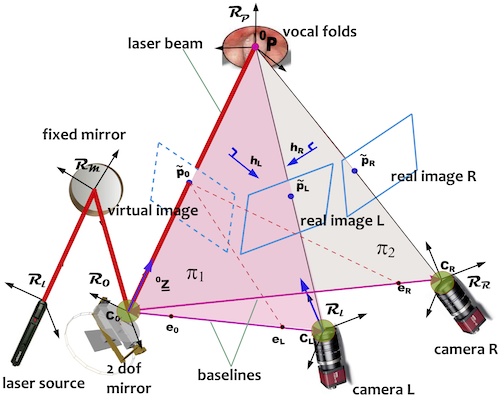}
  \caption{Nouvelle représentation de la géométrie à trois vues utilisant deux caméras et un miroir actionné.}
  \label{fig.new_trifocal_geo}
\end{figure}

Dériver une loi de commande d'asservissement visuel en prenant en compte la contrainte trifocale introduite dans (\ref{eq.torseur}) n'est pas trivial. Ceci nécessite beaucoup de manipulations mathématiques souvent complexes (inversions de matrices, approximations, etc.). Pour simplifier, deux idées originales ont été mises en commun dans ce travail : 1) utiliser le miroir actionné qui contrôle le déplacement du laser comme une \emph{caméra virtuelle} permettant l'acquisition d'un pixel à chaque itération et 2) reformuler le tenseur trifocal sous forme vectorielle simplifiée. Pour ce faire, la Fig.~\ref{fig.trifocal_geo} peut être représentée par la Fig.~\ref{fig.new_trifocal_geo}.

Par ailleurs, notons que $\zn[0]$ le vecteur unitaire représentant la direction du faisceau laser qui part du miroir rotatif vers la cible (scène) et ${}^0\bf{P}$ la position 3D du spot laser dans la scène et  $\pnt{\tilde{p}}$ sa projection dans l'image virtuelle. Il est alors possible d'écrire
\begin{equation}
{}^0 \bf{P} =   \textit{d}\zn[0]  = \textit{Z} \pnt{\tilde{p}_\texttt{0}} 
\label{eq.P=dz}
\end{equation}
où, $d$ la distance entre le miroir actionné et la scène, et $Z$ la profondeur (distance parallèle à l'axe optique de la caméra virtuelle). 

A noter que les distances $d$ et $z$ sont difficiles à déterminer contrairement à $\zn[0]$ qui peut obtenu par triangulation. En conséquence, dans la suite, nous nous efforcerons de les supprimer dans la formulation de la loi de commande. 

A partir de~(\ref{eq.P=dz}), il est aisé de remarquer que $\zn[0]$ and $\tilde{\pnt p}_\texttt{0}$ sont "projectivement" similaires. Ainsi,  
\begin{equation}
\zn[0] \equiv \pnt{\tilde{p}_\texttt{0}}
\label{eq:z0}
\end{equation}

En fait, $\pnt {\tilde {p}_\texttt{0}} $ est la projection perspective sur un plan, tandis que $\zn [0] $ est la projection perspective sur une sphère. En accord avec la schéma montré Fig.~\ref{fig.new_trifocal_geo}, il est possible de réécrire la contrainte trifocale (\ref{eq.trifocal}) et (\ref{eq:z0}) sous la forme suivante :
\begin{eqnarray}
\hr=\Fmat[\texttt{0}]{\texttt{R}} \pr \quad \quad \text{et}  \quad  \quad
\hLr=\Fmat[\texttt{0}]{\texttt{L}} \pl 
\label{eq.hrhl}
\end{eqnarray}
 \begin{eqnarray}
  \label{eq:3epipolar}
   \pl {}^{\top}\, \Fmat[\texttt{L}]{\texttt{R}}\, \pr&=&0\\
  \label{eq:zFpr}
   \zn[0]^{\top}\, \Fmat[\texttt{0}]{\texttt{R}}\, \pr&=&0\\
  \label{eq:zFpl}
   \zn[0]^{\top}\, \Fmat[\texttt{0}]{\texttt{L}}\, \pl&=&0
\end{eqnarray}

Laissons de côté les formulations (\ref{eq:3epipolar}), (\ref{eq:zFpr}), et (\ref{eq:zFpl}) pour introduire les notations suivantes $\hr$ et $\hLr$ qui représentent respectivement les vecteurs normaux aux plans épipolaires (Fig.~\ref{fig.new_trifocal_geo}). Ils sont exprimés comme suit :
\begin{eqnarray}
\hr=\Fmat[\texttt{0}]{\texttt{R}} \pr  \quad \quad \text{et}  \quad  \quad
\hLr=\Fmat[\texttt{0}]{\texttt{L}} \pl 
\label{eq.hrhl}
\end{eqnarray}

Par conséquent, de (\ref{eq:zFpr}) et (\ref{eq:zFpl}), il est trivial de constater que le produit croisé $\hr$~$\times$~$\hLr$ est parallèle à $\zn[0]$ (c'est-à-dire $\zn[0] \sim \hr$~$\times$~$\hLr$). Pour éviter les problèmes d'échelle, cela peut être formulé par le produit croisé (\ref{eq.z000}) :
\begin{equation}
  \zn[0]\times \left( \hr \times \hLr \right) = 0
	\label{eq.z000}
\end{equation}

Ainsi, la formulation de transformation point-point-point (tenseur trifocal) exprimé (\ref{eq.torseur1}) peut être récrite sous cette forme :
\begin{equation}
  \zn[0] \times \left(\Fmat[\texttt{0}]{\texttt{R}}\pr \times \Fmat[\texttt{0}]{\texttt{L}} \pl \right) = 0
	\label{eq.ntensor}
\end{equation}

Cette nouvelle formule représente la contrainte trifocale exprimée sous une forme algébrique à l'aide de produits vectoriels $zn[0]$, $\pl$ et $\pr$, contrairement à la formulation matricielle (\ref{eq.torseur}). Nous pouvons noter que la manipulation dans la nouvelle expression de la contrainte trifocale ne nécessite plus la manipulation de matrices (inversion, dérivée, etc.). Néanmoins, elle présente au moins un inconvénient par rapport à l'expression originale : l'impossibilité de calculer $\pr$ lorsque $\zn[0]$ est coplanaire avec $\epil$ et $\epir$ (lorsque $\zn[0]$ franchit la baseline).
%
%
Les quantités $\hLr$ et $\hr$ peuvent être calculées, soit à l'aide de la position désirée du spot laser dans les deux images gauche et droite et en utilisant les matrices fondamentales $\Fmat[\texttt{0}]{\texttt{L}}$ et $\Fmat[\texttt{0}]{\texttt{R}}$ (\ref{eq.hrhl}), soit grâce aux épipôles (Fig.~\ref{fig.new_trifocal_geo}). Sur ce dernier point, il est possible d'utiliser les expressions suivantes :
\begin{equation}
{\hLr}  =  \epil \times {\zn[0]} \quad \quad \text{et}  \quad  \quad
{\hr}  =  \epir \times {\zn[0]}
\label{eq.epip0}
\end{equation}
où $\Fl \epil = 0$ et $\Fr \epir = 0$. Le signe des matrices fondamentales et celui des épipôles doivent être cohérents entre eux, c'est-à-dire exprimés comme suit : 
\begin{equation}
(\Fl {\pl})^\top (\epil \times {\zn[0]}) >0 \quad \quad \text{et}  \quad  \quad  (\Fr {\pr})^\top(\epir \times {\zn[0]})>0
\end{equation}
de manière à ce que (\ref{eq.hrhl}) and (\ref{eq.epip0}) soient équivalents à un scalaire strictement positif.

Par conséquent, à partir de~(\ref{eq.epip0}), il est trivial d'écrire : 
\begin{equation}
   \hr \times \hLr = (\epir \times \zn[0]) \times (\epil \times \zn[0])
\end{equation}

En s'inspirant de la formule du triple produit vectoriel suivante : 
\begin{equation}
  \label{eq:axbxc}
	\vect{a} \times (\vect{b}  \times \vect{c}) = \vect{b}\left(\vect{a}^\top\vect{c}\right) -  \vect{c} \left(\vect{a}^\top\vect{b}\right)
\end{equation}

Après quelques arrangements et simplifications, nous arrivons au résultat suivant : 
\begin{equation}
  \label{eq:hrxhl=epsilon.z}
   \hr \times \hLr = - \zn[0]^\top (\epil\times\epir) \zn[0]
\end{equation}

\begin{figure}[!h]
  \centering
  \includegraphics[width=0.6\columnwidth]{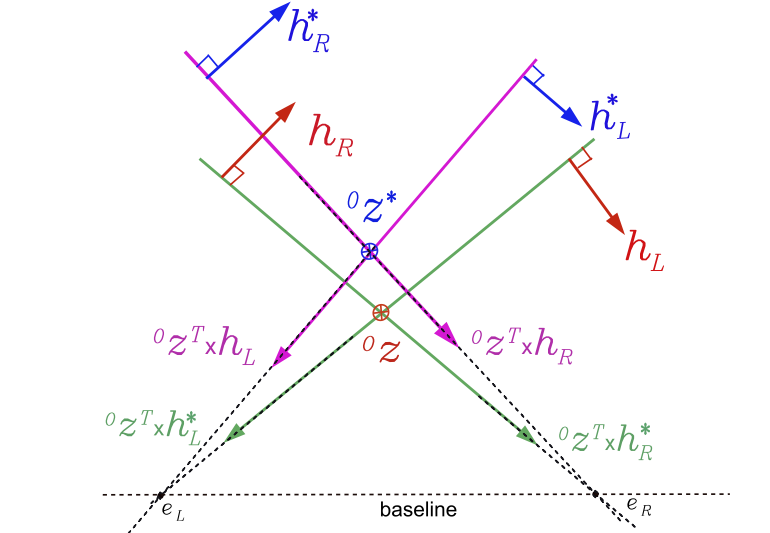}
  \caption{Problèmes de signes liés à la position de la ligne reliant les deux épipôles.}
  \label{fig.product_vect}
\end{figure}

Dans cette expression (\ref{eq:hrxhl=epsilon.z}), il apparaît une distance orientée, c'est-à-dire un signe (-) devant  "$\zn[0]^\top (\epil\times\epir) \zn[0]$",  entre $\zn[0]$ et la droite passant entre les deux épipôles (Fig.~\ref{fig.product_vect}). Ainsi, deux cas de figures se présentent : 
\begin{itemize} 
\item la présence d'une singularité de la contrainte trifocale se produisant, lorsque le spot laser franchit la ligne formée entre les centres de projection des deux caméras dont le résultat est que $ \hr \times \hLr$ est égal à $\vect{0}$.
\item en dehors de cette singularité, les relations suivantes sont valables 
\begin{eqnarray}
    \label{eq:norm_hrxhl}
    \| \hr \times \hLr \| &=& | \zn[0]^\top (\epil\times\epir) |\\
    \label{eq.zo_cross}
    \zn[0]&=& \epsilon \frac{ \hr \times \hLr}{\| \hr \times \hLr \|}
  \end{eqnarray}
où $\epsilon$ = - $\mathrm{sign}(\zn[0]^\top (\epil\times\epir))$.
\end{itemize}

\begin{figure}[!h]
  \centering
  \includegraphics[width=0.6\columnwidth]{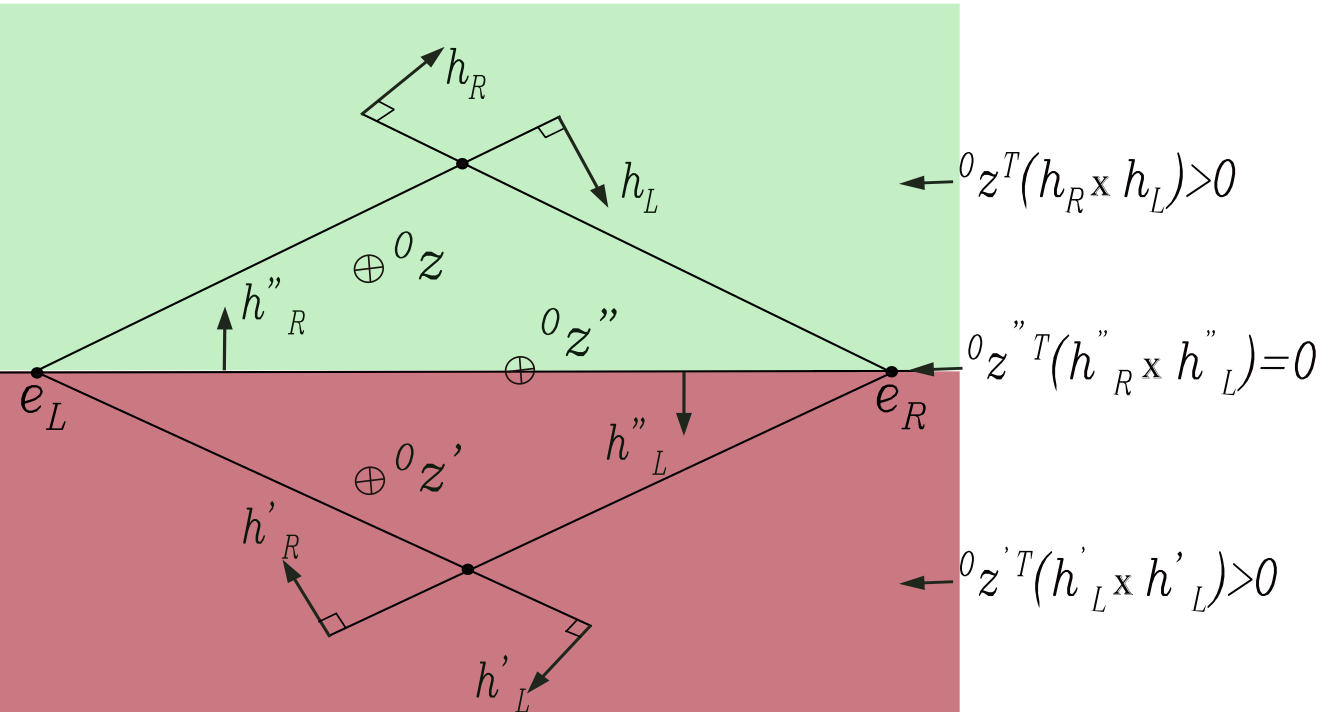}
  \caption{Différents cas sur la position du spot laser par rapport à la droite ($\epil \epir$).}
  \label{fig.cas_figure_el_er}
\end{figure}

Dans la suite de cette section, nous allons aborder la méthodologie de la mise en \oe uvre de la loi de commande pour le contrôle du spot laser dans l'image dérivée de la réécriture de la contrainte trifocale. 
%
\subsection{Loi de commande}
%
A présent, tous les \emph{ingrédients} nécessaires pour la mise en \oe uvre de la loi de commande par asservissement visuel sont réunis. Commençons par les contraintes cinématiques qui en découlent de la nouvelle formule du tenseur trifocal. La dérivée temporelle de la contrainte trifocale (\ref{eq.z000}) permet d'obtenir :
\begin{equation}
\dotzn[0] \times \left( \hr \times \hLr \right) + \zn[0]  \times \left( {\vect{\dot{h}}}_{R} \times \hLr \right) +
\zn[0]  \times \left( \hr \times {\vect{\dot{h}}}_{L} \right) = 0
\label{eq.dev.tensor}
\end{equation}

Par ailleurs, il est possible d'écrire que ${\vect{\dot{h}}}_{R} = \Fr \dotpr$ et  ${\vect{\dot{h}}}_{L} = \Fl \dotpl$. Par conséquent, (\ref{eq.dev.tensor}) peut être réécrite comme suit :
\begin{equation}
\dotzn[0] \times  \left( \hr \times \hLr \right) + \zn[0]  \times \left( (\Fr \dotpr) \times \hLr \right) +
\zn[0]  \times \left( \hr \times ( \Fl \dotpl) \right) = 0
\end{equation}
ou encore, 
\begin{equation}
\dotzn[0] \times  \left( \hr \times \hLr \right) = - \zn[0] \times \left( (\Fr \dotpr) \times \hLr \right) -
\zn[0]  \times \left( \hr \times ( \Fl \dotpl) \right) 
\end{equation}

Après quelques décompositions et réorganisations, nous arrivons à la formulation suivante 
\begin{equation}
  \dotzn[0] \times  \left( \hr \times \hLr \right) = \zn[0]  \times \left( \hLr \times (\Fr \dotpr )
-  \hr \times (\Fl \dotpl) \right) 
	\label{eq.eq_trif}
\end{equation}

Dans un premier temps, si nous effectuons une division de chaque côté de ($\ref{eq.eq_trif}$) par le terme $\| \hr \times \hLr \|$, il en résulte :
\begin{equation}
  \dotzn[0] \times  \frac{\hr \times \hLr}{\| \hr \times \hLr \|} = \frac{\zn[0]}{\| \hr \times \hLr \|} \times \left( \hLr \times (\Fr \dotpr ) -  \hr \times (\Fl \dotpl) \right) 
\label{eq:dotzn0}
\end{equation}

Dans un second temps, en utilisant (\ref{eq.zo_cross}) et en inversant le produit vectoriel $\dotzn[0] \times \zn[0]$, nous obtenons finalement une relation cinématique, très intéressante, décrivant le lien  entre les vitesses $\dotpl$ et $\dotpr$ du spot laser dans respectivement chacune des images $\mathbf{I}_\texttt{L}$ et $\mathbf{I}_\texttt{R}$, et les vitesses angulaires du faisceau laser, c'est-à-dire $\dotzn[0]$.  La formule résultante est la suivante :
\begin{equation}
 \zn[0] \times  \dotzn[0] =  - \frac{\hr \times \hLr}{\| \hr \times \hLr \|^2} \times \left( \hLr \times (\Fr \dotpr ) -  \hr \times (\Fl \dotpl) \right) 
\label{eq:dotzn}
\end{equation}

A présent, il est nécessaire de lier les vitesses du faisceau laser $\dotzn[0]$ à la vitesse de rotation du miroir actionné, noté $\Vrot$. Sachant que $\dotzn[0]$ est un vecteur unitaire (c'est-à-dire $\zn [0]^\top \zn[0] = 1$ et $\dotzn[0]^\top \zn[0] = 0$), on peut donc écrire
\begin{equation}
  \dotzn[0] = \Vrot \times \zn[0]
\label{eq.dotzn00}
\end{equation}

La solution analytique est définie par 
\begin{equation}
\label{eq: Vrot}
\Vrot = \zn [0] \times \dotzn[0] + k \zn[0], ~~\texttt{avec} ~~ k \in \Re
\end{equation}

Cependant, une rotation du faisceau laser autour de son axe n'est pas significative. Par conséquent, $k$ est choisi égal à 0 et de fait, nous pouvons écrire 
\begin{equation}
 \Vrot = \zn[0] \times \dotzn [0]
\label{eq.dotzn1}
\end{equation}

Par conséquent, en identifiant (\ref{eq:dotzn}) et (\ref{eq.dotzn1}), nous pouvons facilement déduire l'expression de la vitesse du miroir sous forme de fonction des vitesses du point laser :
\begin {equation} 
  \Vrot = - \frac{\hr \times \hLr}{\| \hr \times \hLr \|^2}  \times \left( \hLr \! \times \!  (\Fr \dotpr )
-  \hr \! \times \! ( \Fl \dotpl) \right) 
\label {eq.cmd2}
\end {equation}

Pour une décroissance exponentielle de la fonction de coût $\mathbf{e} = \left( {\begin{array}{*{20}c}{\tilde{\pnt p}_\texttt{L} - \consigne{\pnt{\tilde{p}}}_\texttt{L}}  \\
   {\tilde{\pnt p}_\texttt{R} - \consigne{\pnt{\tilde{p}}}_\texttt{R}}  \\
\end{array}} \right)$, nous introduisons, dans les deux images, un comportement du premier ordre de l'erreur, entre la position actuelle $\pnt {\tilde {p}} $ et la position désirée $\consigne {\pnt {\tilde {p}}} $ du spot laser
\begin{equation}
\label{eq:ctrl_pnt}
  \pnt{\dot{\tilde{p}}_i}= - \lambda (\pnt{\tilde{p}_i} - \consigne{\pnt{\tilde{p}}}_i) + \consigne{\pnt{\dot{\tilde{p}}}}_i,~~\texttt{avec} ~~ i \in \{ L,R\}
\end{equation}
où $ \lambda $ est un gain positif et $ \consigne {\pnt {\dot {\tilde {p}}}}_ i $ est le terme de "\emph{feed-forward}" en cas de suivi de trajectoire. Cette dernière peut être définie directement par le clinicien à travers une Interface Robot-Chirurgien (tablette).

En s'appuyant sur (\ref{eq:ctrl_pnt}) et (\ref{eq.cmd2}), il devient possible d'exprimer le vecteur des vitesses angulaires $\Vrot$ sous la forme suivante :
\begin{eqnarray}
\nonumber 
  \Vrot &=&  \lambda \frac{ \hr \times \hLr}{\| \hr \times \hLr \|^2}  \times \Big( \hLr \! \times \!  \big(\Fr   (\pnt{\tilde{p}_\texttt{R}} - \consigne{\pnt{\tilde{p}}}_\texttt{R})\big)
-  \hr \! \times \! \big( \Fl (\pnt{\tilde{p}_\texttt{L}} - \consigne{\pnt{\tilde{p}}}_\texttt{L}) \big) \Big)\\ 
&& -  \frac{  \hr \times \hLr}{\| \hr \times \hLr \|^2} \times \left(\hLr \! \times \! ( \Fr \consigne{\pnt{\dot{\tilde{p}}}}_\texttt{R}) - \hr \! \times \! (\Fl \consigne{\pnt{\dot{\tilde{p}}}}_\texttt{L}) \right)
\label{eq.omega2}
\end{eqnarray}

Rappelons que nous pouvons écrire $\vect{h}_{i}=\Fmat[\texttt{0}]{i} \pnt{\tilde{p}_i}$ et avec  l'introduction de  notations $\consigne{\vect{h}_{i}}=\Fmat[\texttt{0}]{i} \consigne{\pnt{\tilde{p}}}_i$ et $\consigne{{\vect{\dot{h}}}_{i}}= \Fmat[\texttt{0}]{i} \consigne{\pnt{\dot{\tilde{p}}}_i}$, l'équation (\ref{eq.omega2}) peut être simplifiée en l'expression du contrôleur final, comme suit :
\begin{equation}
   \Vrot =  \underbrace{- \lambda \frac{ \hr \times \hLr}{\| \hr \times \hLr \|^2}  \times \Big( \hLr \! \times \!  \consigne{\hr} 
-  \hr \! \times \! \consigne{\hLr} \Big)}_{comportement~1^{er}~ordre }  -  \underbrace{\frac{ \hr \times \hLr}{\| \hr \times \hLr \|^2}  \times \Big( \hLr \! \times \! \consigne{{\vect{\dot{h}}}_{R}} - \hr \! \times \! \consigne{{\vect{\dot{h}}}_{L}} \Big)}_{terme~defeed-forward}
\label{eq:cmd_final}
\end{equation} 

\begin{figure}[!h]
  \centering
  \includegraphics[width=0.9\columnwidth]{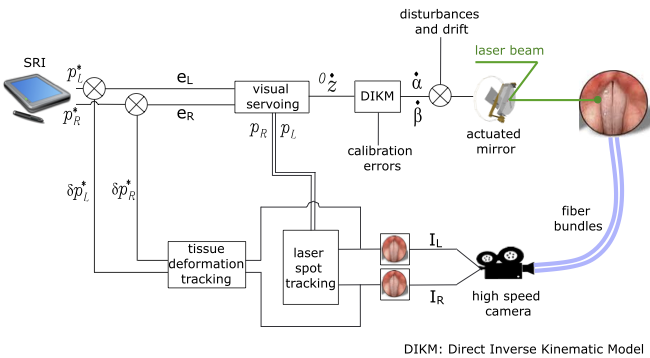}
  \caption{Schéma de commande d'asservissement visuel trifocal d'un faisceau laser.}
  \label{fig.loop_laser}
\end{figure}

A noter que l'expression de la loi de commande définie ci-dessus (schématisée dans la Fig.~\ref{fig.loop_laser}) est entièrement fondée sur les mesures-image. Ainsi, elle devrait présenter une convergence découplée (avec un comportement similaire) de chaque composante de la position du spot laser dans chaque image. La décroissance de l'erreur doit avoir également une forme exponentielle, qui est un comportement typique d'un asservissement visuel classique grâce à la (pseudo) inversion de la matrice d'interaction associée. Cependant, cette particularité est moins évidente ici, du fait, que nous inversons l'équation du mouvement (inversion exacte du mouvement) sans inversion de la matrice d'interaction, en tant que telle, et qu'aucune simplification ou approximation n'est introduite dans cette opération.   

%
\subsection{Validation}
%
\begin{figure}[!h]
  \centering
  \includegraphics[width=\columnwidth]{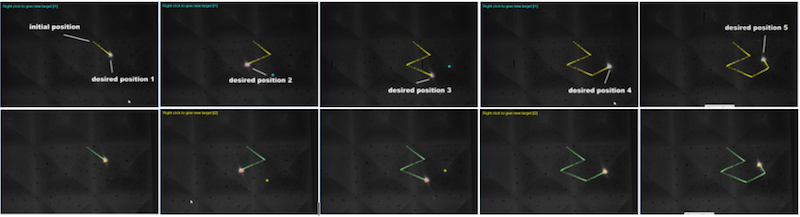}
  \caption{Séquence d'images capturées lors de la réalisation d'une tâche de positionnement (sans planification) sur une surface 3D inconnue.}
  \label{fig.seq_trifocal}
\end{figure}

La loi de commande, décrite ci-dessus, a été validée suivant plusieurs scénarios : numériques et expérimentaux. Le fonctionnement du contrôleur est le suivant : l'opérateur dispose des images gauche et droite de la scène (et du spot laser). Il clique sur l'image de gauche (ou de droite) pour définir la ou les position(s) désirée(s) du spot laser, la position correspondante géométriquement dans l'autre image est définie par à la contrainte épipolaire et par une méthode de recherche de mises en correspondance fondée sur l'autocorrélation. Un exemple de ce fonctionnement est montré sur la Fig.~\ref{fig.seq_trifocal} où plusieurs positions désirées successives ont été définies.
%

A souligner que la trajectoire du spot laser, entre deux points successifs dans l'image du spot laser, est quasi-rectiligne. Ceci est effectué sans aucune connaissance préalable ni estimation tridimensionnelle (par exemple, reconstruction 3D ou calcul de profondeur) de la surface sur laquelle le spot laser se déplace. 

%
\begin{figure}[!h]
    \centering
    \subfigure[caméra gauche.]{\includegraphics[width=0.45\columnwidth, height = 4.2cm]{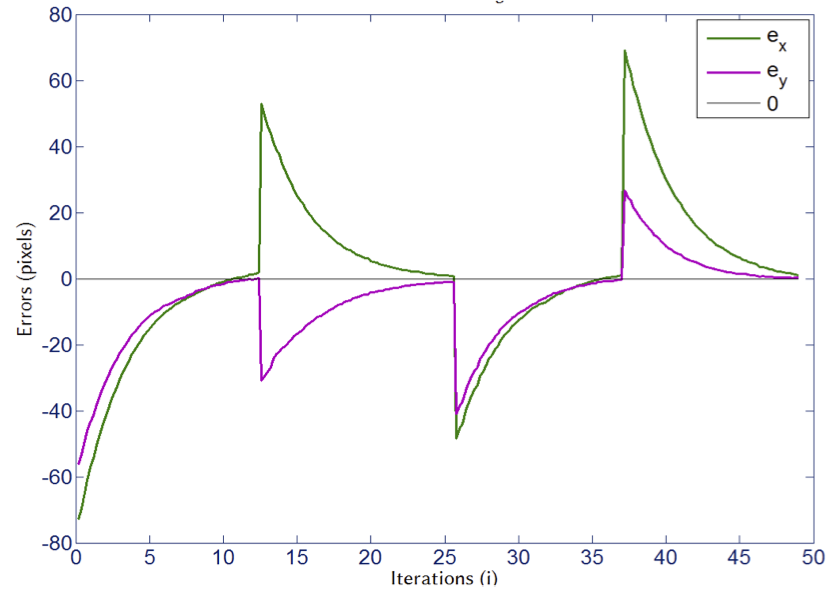}}
    \subfigure[caméra droite.]{\includegraphics[width=0.45\columnwidth, height = 4.2cm]{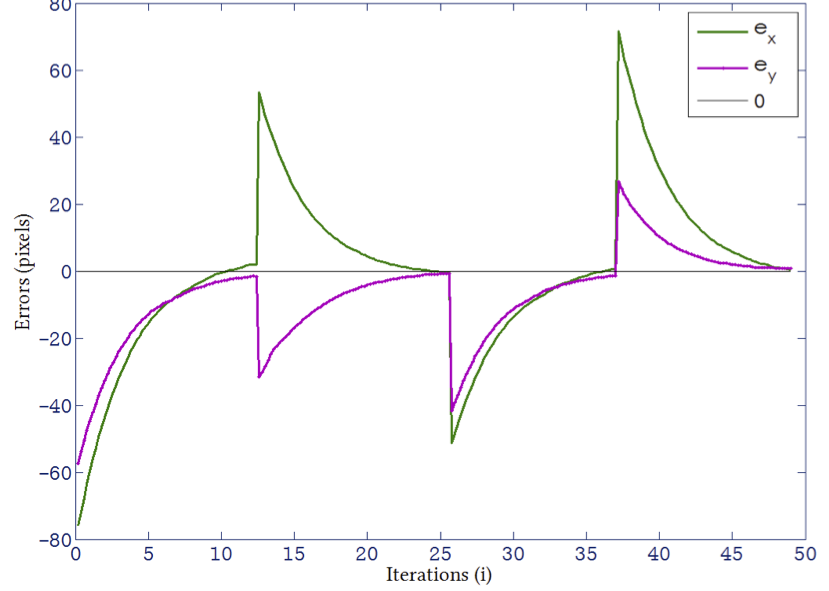}}
    \caption{Régression des erreurs dans l'image (en pixels) \emph{versus} nombre d'itérations $i$: (a) caméra de gauche, et (b) caméra de droite.}
    \label{fig.trajs_tr}
\end{figure}

\begin{figure}[!h]
  \centering
  \includegraphics[width=0.8\columnwidth, height = 6cm]{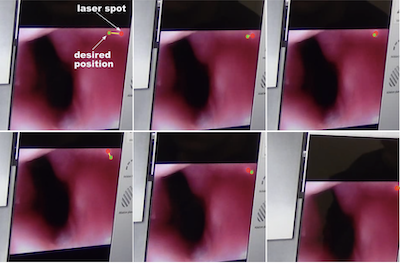}
  \caption{Validation sur les cordes vocales au laboratoire d'anatomie de l'hôpital de Besançon.}
  \label{fig.seq_cadavre}
\end{figure}

Comme le montre la Fig.~\ref{fig.trajs_tr}, la décroissance de l'erreur dans les images (gauche et droite) est sous forme exponentielle comme attendue. Le contrôleur a été également validé avec succès lors de tests sur cadavre humain avec dans le rôle de l'opérateur, un chirurgien ORL. Même s'il fallait adapter l'algorithme de suivi permettant la détection du spot laser, la loi de commande trifocale a parfaitement fonctionner sur les cordes vocales (Fig.~\ref{fig.seq_cadavre}). 
  
L'implémentation de la loi de commande proposée se fait avec seulement une vingtaine de lignes de code (en C++)~\citep{tamadazteIros2014, tamadazteIjrr2015}. A cela s'ajoutent plusieurs caractéristiques fort intéressantes en commande référencée capteur. En effet, cette implémentation : 
\begin{itemize}
\item ne nécessite pas d'inversion de matrice ni de manipulation de formules mathématiques complexes (tenseurs) ;
\item fonctionne de manière similaire sur les scènes 2D et 3D sans connaissance a priori de la géométrie de la cible 
\item ne nécessite qu'un faible étalonnage (robot/caméra, caméra, etc.) ;
\item est précise, c'est-à-dire une erreur de positionnement, dans l'image, inférieur à 50$\mu$m, même lorsque l'étalonnage est approximatif ;
\item est capable de fonctionner à très haute cadence, c'est-à-dire plusieurs centaines de Hz (limitée uniquement par la cadence de l'algorithme de suivi visuel du spot laser dans les images) ;
\item présente une stabilité prouvée de manière explicite (voir~\citep{tamadazteIjrr2015}).
\end{itemize}
%
\section{Suivi de chemin dans l'image}
%
Dans cette section, il est question de présenter succinctement les différentes lois de commande développées pour contrôler par retour visuel un spot laser. Les méthodes développées, ici, sont inspirées de la robotique mobile et des travaux, qui en découlent, sur les problématiques de suivi de chemin et de trajectoire. Dans un premier temps, nous avons cherché à trouver une analogie entre le suivi de chemin en robotique mobile et la commande d'un spot laser le long d'un chemin défini dans l'image. Ensuite, nous avons étendu ces travaux en utilisant deux caméras (commande stéréoscopique) pour une prise en compte de la profondeur de la scène dans la commande, notamment en associant les résultats de la méthode trifocale discutée précédemment. 
%
\subsection{Robotique mobile \emph{vs.} chirurgie laser}
%
\begin{figure}[!h]
    \centering
    \subfigure{\includegraphics[width=0.4\columnwidth, height=4.5cm]{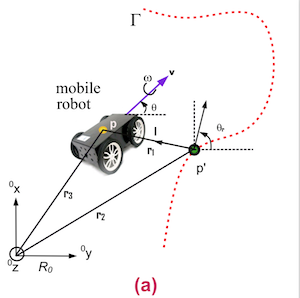}}
    \subfigure{\includegraphics[width=0.4\columnwidth, height=4.5cm]{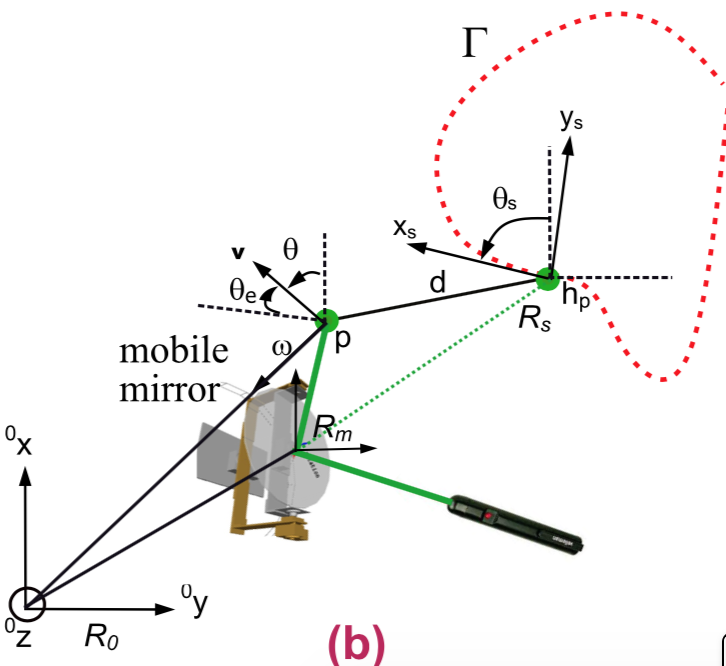}}
    \caption{Similitudes entre le modèle cinématique d'un véhicule de type unicycle (a) et un modèle cinématique d'un miroir actionné pour le guidage laser (b).}
    \label{fig.trajs_paths}
\end{figure}

Tout d'abord, montrons que le suivi de chemin par un spot laser dans l'image est très similaire à celui du suivi d'un véhicule unicycle au sol comme le montre le schéma fonctionnel de la Fig.~\ref{fig.trajs_paths}. Nous pouvons citer à titre d'exemples les similitudes suivantes entre les deux cinématiques :
\begin{itemize}
\item dans les deux cas, la direction et l'amplitude de la vitesse sont asservies indépendamment ;
\item dans le cas d'un unicycle, on s'attend généralement à ce que la vitesse soit telle que la vitesse curviligne soit constante, tandis qu'en cas de chirurgie laser, on s'attend à ce que la vitesse corresponde à une vitesse constante du spot laser sur la surface du tissu ;
\item dans les deux cas, la précision de suivi du chemin devrait être indépendante de l'amplitude de la vitesse d'avancement ;
\item les vitesses du point laser et du robot unicycle doivent être lisses (sans dépassements).
\end{itemize}

Dans ce travail, nous avons opté pour le suivi de chemin plus que celui de trajectoire pour plusieurs raisons. En fait, dans le cas de la chirurgie laser, la commande du laser consiste en deux tâches. La première a pour objectif d'assurer la compatibilité de la vitesse de déplacement du laser avec l'interaction laser-tissu (contrôle longitudinal), afin d'éviter la carbonisation du tissu tout en permettant une incision ou une ablation précise et efficace. La seconde tâche (contrôle latéral) est destinée à garantir un suivi géométrique du chemin désiré défini à main-levée par un opérateur/clinicien sur une interface Robot-Chirurgien (par exemple, une tablette)~\citep{Mattos2014}. L'ensemble de ces deux tâches définissent finalement la trajectoire du laser, c'est-à-dire la courbe géométrique + le profil de vitesse le long de cette courbe. Toutefois, il n'est pas conseillé d'utiliser le suivi de trajectoire standard, car les deux tâches doivent pouvoir être modifiées de manière intuitive et indépendante par le chirurgien. En fait, la vitesse du spot laser doit être la même quels que soient la forme, la taille ou la courbure du chemin d'incision/ablation \citep{tamadazteTro2015}. De plus, l'amplitude de la vitesse doit être adaptable par le chirurgien en fonction du type de tissu à réséquer (mince, épais, fragile, etc.) ou de la source laser (laser CO2, laser Hélium-Néon, etc.), voire même l'adapter en ligne durant la tâche de résection/ablation.  
%
\subsection{Modélisation cinématique du suivi de chemin par un laser}
%
\begin{figure}[!h]
  \centering
  \includegraphics[width=0.55\columnwidth]{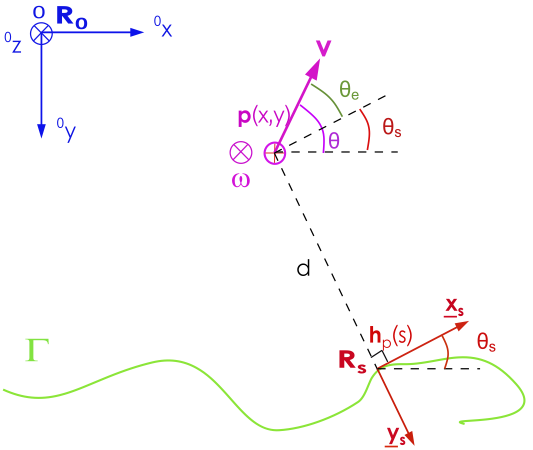}
  \caption{Représentation des différents repères associés à la méthode de suivi de chemin.}
  \label{fig.frenet_frame}
\end{figure}

Le formalisme cinématique du problème du suivi de chemin dans l'image par un laser est inspiré par les travaux de Pascal Morin et Claude Samson \citep{Morin2008}. Pour ce faire, considérons un spot laser projeté dans l'image en un point 2D $\pnt p = (x, y)^\top$ exprimé dans le repère de référence $\mathbf{R_o}$. La cinématique du spot laser exprimée dans le repère $\mathbf{R_o}$  est régie par :
\begin{eqnarray}
\label{eq.syst1}
\dot{x}& = &v \cos \theta \\
\label{eq.syst2}
\dot{y}& = &v \sin \theta \\
\label{eq.syst3}
\dot{\theta}& = &\omega
\end{eqnarray}
où, $\theta$ est l'angle entre la direction de la vitesse du laser $\mbf{\underline{v}} $ et l'axe $ {}^0 x$ du repère de référence $\mathbf{R_0}$, $v$ représente l'amplitude de la vitesse de translation du spot laser en $\mbf{R_s}$ ($\mbf{v}$ = $v$ $\underline{\mbf{v}}_k$), $\mbf{h}_{p} (s)$ est la projection orthogonale de $\mbf{p}$ = $(x, y)^{\top} $ sur la courbe $\Gamma$ et $\omega$ sa vitesse de rotation portée par l'axe ${}^0 z$ (Fig.~\ref {fig.frenet_frame}).

Ces équations cinématiques peuvent être généralisées et exprimées dans le repère de \emph{Frenet}  $\mathbf{R_s}$ attaché à la courbe à réaliser (Fig.~\ref{fig.frenet_frame}), comme démontré dans~\citep{Samson93, Morin2008}. Ainsi, les équations (\ref{eq.syst1}), (\ref{eq.syst2})  et (\ref{eq.syst3})  s'expriment dans $\mathbf{R_s}$ comme suit :
\begin{eqnarray}
\label{eq.syst2.1}
\dot{s}& = &\frac{v}{1-dC(s)}\cos\theta_e \\
\label{eq.syst2.2}
\dot{d}& = &v\sin\theta_e \\
\label{eq.syst2.3}
\dot{\theta}_e& = &\omega - \dot{\emph{s}}\emph{C(s)}
\end{eqnarray}
avec $s$ et $C(s)$ sont respectivement l'abscisse curviligne et la courbure, $\theta_e$ est la différence entre l'orientation du laser $\mathbf{\underline{v}}$ et le vecteur tangentiel de  $\mathbf{R_s}$ et $d$ la plus petite distance entre $\mathbf{p}$ = $ (x,y)^{\top}$ à la courbe $\Gamma$.

Afin de concevoir une loi de commande "implémentable", il est nécessaire d'effectuer une transformation de représentation de la manière suivante : $\left\{ {s, d, \theta_e, v, \omega} \right\}~\Longleftrightarrow~\left\{ {z_1, z_2, z_3, u_1, u_2} \right\}$ défini dans $\mathbb{R}^2\times\Big(-\frac{\pi}{2}, +\frac{\pi}{2}\Big)\times\mathbb{R}^2$. Ceci permet une transformation locale des équations (\ref{eq.syst2.1}), (\ref{eq.syst2.2}) and (\ref{eq.syst2.3}) vers le système d'équations suivant :
\begin{eqnarray}
\label{eq.syst3.1}
\dot{z}_{1} & = & u_1 \\
\label{eq.syst3.2}
\dot{z}_{2} & = & u_1z_{3} \\
\label{eq.syst3.3}
\dot{z}_{3} & = & u_2 
\end{eqnarray}
où $u_1$ et $u_2$ sont des entrées intermédiaires de commande. Au fait, si $z_1$ = $s$, alors (\ref{eq.syst3.1}) peut s'écrire comme suit :
\begin{eqnarray}
u_1 &= &\frac{v}{1-dC(s)}\cos\theta_e
\label{eq.z1dot}
\end{eqnarray}

De la même manière, si $z_2$ = $d$,  (\ref{eq.syst3.2}) devient :
\begin{equation}
z_{3}= \Big(1-dC(s)\Big)\tan\theta_{e}
\label{eq.z3}
\end{equation}

Par conséquent et en utilisant (\ref{eq.syst3.3}), $u_2$ peut être redéfini par :
\begin{eqnarray}
\nonumber
u_2 = \Big({-\dot{d}C(s)-d\frac{\partial C(s)}{\partial s}\dot{s}}\Big)\tan\theta_e 
+ \Big({1-dC(s)}\Big) \Big({1+\tan^2\theta_e}\Big)\Big(\omega - \dot{s}C(s)\Big)
\label{eq.v2.2}
\end{eqnarray}

Enfin, pour que l'erreur de distance $d$ et l'erreur d'orientation $\theta_e$ soient asservies (décroissance des erreurs vers 0), une solution de type retour d'état proportionnel peut être envisagé :
\begin{equation}
u_2= -u_1 \gamma_1 z_2-\left | u_1 \right |\gamma_2 z_3
\label{eq.v2.c}
\end{equation}
où $\gamma_1$ et $\gamma_2$ sont des gains de commande positifs  et $u_1 \ne 0$ la vitesse initiale qui peut être choisie en fonction de l'application visée. 

Jusqu'à présent, les paramètres de la courbe $\Gamma$ sont considérés connus comme dans le cas de la planification de trajectoire pour la robotique mobile. A noter que pour les applications médicales (chirurgie laser), nous ne disposons que des coordonnées discrétisées et normalisées du chemin de résection/ablation $\Gamma$ qui sont reçues au fur et à mesure que le chirurgien dessine la courbe sur une tablette. Par conséquent, le chemin à suivre est une approximation, la plus proche possible, de la courbe définie par le clinicien.
Par ailleurs, de fait que la courbure $C(s)$ est utilisée dans l'expression de la loi de commande, $\Gamma$ doit être au moins défini $C^2$ ($C(s)$ est la dérivée temporelle du vecteur tangent). 
Aussi, étant donné que la courbe $\Gamma$ est non-paramétrée, la courbure $C(s)$ et le rayon en chaque point $\pnt p$ sont estimés en utilisant trois points successifs. Néanmoins, la qualité du suivi de chemin dépend de la précision de ces approximations. En d'autres termes, plus il y a de points sur le chemin, plus la loi de commande est précise. 
%
\subsection{Détails de l'implémentation}
%
Considérons un chemin $\Gamma$ décrit par un ensemble de points $\mathcal{C}$~$=$ $\{\mathbf{c}_i$, $i$ $=$ $1$... $n\}$ avec la relation $\mathbf{c}_1$ $<$ $\mathbf{c}_2 $ $<$... $<$ $\mathbf{c}_n $. Chaque $\mathbf{c}_i$ est un quadruplet contenant deux coordonnées cartésiennes ($x$, $y$) et deux coordonnées curvilignes (courbure $C(s)$ et abscisse curviligne $s$).

\begin{figure}[!h]
  \centering
  \includegraphics[width=0.45\columnwidth]{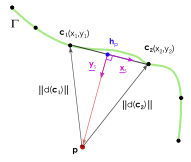}
  \caption{Méthode de calcul de la distance $d$ à chaque itération $k$ le long de la courbe $\Gamma$.}
  \label{fig.calcul_d}
\end{figure}

La position du point laser $\pnt p_k $ = $ (x_k, y_k)^{\top}$ est déterminée par un simple algorithme de suivi visuel, à chaque itération $k$. Ainsi, il est possible d'identifier les deux points successifs les plus proches dans $\mathcal{C}$, c'est-à-dire, $\mathbf {c}_{j}$ et $\mathbf{c}_{j \pm 1}$ \Big (avec $d(\mathbf{c}_{j})$ $<$ $d (\mathbf{c}_{j \pm 1}$) \Big) comme indiqués dans la Fig.~\ref {fig.calcul_d}. Pour ce faire, nous effectuons une recherche locale autour de l'abscisse curviligne prédite $s_k$ définie par :
\begin{equation}
\widehat{s_{k+T_e}} =\dot{s}T_e+ s_k
\label{eq.pred_s}
\end{equation}
avec $T_e$ la période d'échantillonnage choisie sur $\Gamma$. Cela permet d'éviter une recherche sur l'ensemble de la courbe $\Gamma$ et également d'éviter les ambiguïtés notamment aux points de croisement du chemin.

Par ailleurs, en utilisant $\mathbf{c}_{j}$ et $\mathbf{c}_{j \pm 1}$, il est possible de déterminer la projection de  $\mathbf{p}$  sur la courbe approximée $\mathcal{C}$ donnant  $\mathbf{h}_p$ = $(x_{\mathbf{h}_p}, y_{\mathbf{h}_p})^\top$. 

A partir de la Fig.~\ref {fig.calcul_d}, il est également possible de déterminer le rayon de  courbure $\mathbf{r}$ de la courbe approximée en chaque point $\mathbf{c}_{j}$. Grâce à la connaissance de $\mathbf{r}$, il devient trivial de calculer les coordonnées curvilignes du point $\mathbf{h}_p$. Ainsi, l’abscisse curviligne $s(\mathbf{h}_p) $ est déterminée par la somme des abscisses curvilignes de $\min (\mathbf {c}_j, \mathbf{c}_ {j \pm 1})$ et la distance entre ce point et $\mathbf{h}_p $ et la courbure $C(s)$ dans $\mathbf{h}_p$, c'est-à-dire, $C(\mathbf{h}_p) $ est calculée comme suit :
\begin{equation}
C(\mathbf{h}_p)   = C\Big(s(\min(\mathbf{c}_{j},\mathbf{c}_{j \pm 1}))\Big)\Big({1-\mathbf{r}}\Big) 
 + C\Big(s(\max(\mathbf{c}_{j},\mathbf{c}_{j \pm 1}))\Big)\mathbf{r}
\end{equation}

De plus, les vecteurs-unitaires $\underline{\mathbf{x}}_s $ et $\underline{\mathbf{y}}_s$ définissant le repère de \emph{Frenet} $\mathbf{R_s}$ sont calculés avec la ligne droite formée entre $\mathbf{c}_j $ et $\mathbf{c}_{j \pm 1}$ comme le montre la Fig.~\ref {fig.calcul_d}. Ainsi, la distance $d$ est recalculée par :
\begin{equation}
d = (\mathbf{p}_{k}-\mathbf{h}_{p}) \cdot \underline{\mathbf{y}}_s
\label{eq.d}
\end{equation}
où, $(\mathbf{p}_{k} - \mathbf{h}_{p})$ est le vecteur normal le long de $\underline{\mathbf{y}}_s$ du point le plus proche de la position du spot laser position (Fig.~\ref{fig.calcul_d}) et le produit scalaire ($\cdot$) permet de connaître le signe exact de la position du point laser par rapport à la courbe $\Gamma$. Comme le repère de \emph{Frenet} $\mathbf{R_s}$ est directement défini sur le point $\mathbf {h}_{p}$, il garantit la colinéarité des deux vecteurs $(\mathbf{p}_{k}-\mathbf{h}_{p})$ et $\underline{\mathbf{y}}_s$. 

Par ailleurs, l'erreur d'orientation $\theta_e$ peut être définie à chaque itération $k$ par la méthode suivante :
\begin{equation}
\theta_{e} = \arctan2 \Big(\underline{\mathbf{v}}_{k} \cdot \underline{\mathbf{x}}_s, \underline{\mathbf{v}}_{k} \cdot \underline{\mathbf{y}}_s \Big)
\label{eq.theta}
\end{equation}
où, $\mathbf{v}_{k}$ est la direction du vecteur vitesse à l'instant $k$ exprimée dans le repère  $\mathbf{R_\texttt{0}}$.

A présent, nous disposons de tous les éléments nécessaires pour dériver une loi de commande permettant au spot laser de suivre, dans l'image, les points de la courbe $\Gamma$. 
 
\subsection{Loi de commande}
%
En connaissant la position du spot laser exprimée dans le repère de \emph{Frenet} et en inversant l'équation~(\ref{eq.v2.2}), il est possible d'exprimer la vitesse angulaire $\omega$ sous une nouvelle forme :
\begin{equation}
\omega = \frac{u_2 +\Big(\dot{d}C(s) + d\frac{\partial C(s)}{\partial s}\dot{s}\Big)\tan\theta_e}{\Big(1-dC(s)\Big)\Big(1+\tan^{2}\theta _{e}\Big)} + \dot{s} C(s)
\label{eq.u2c}
\end{equation}

Par conséquent, la vitesse d'avancée du spot laser sur la courbe $\Gamma$ peut être obtenue par :
\begin{equation}
\underline{\mathbf{v}}_{k+1}=\frac{\underline{\mathbf{v}}_{k}+ \omega {}^0\underline{\mathbf{z}} \times \underline{\mathbf{v}}_{k}}{ \left \| \underline{\mathbf{v}}_{k}+ \omega {}^0\underline{\mathbf{z}} \times \underline{\mathbf{v}}_{k} \right \|}
\label{eq.v}
\end{equation}
où, $\times$ définit le produit vectoriel et $\underline{\mathbf{v}}_{k}$ représente la direction de la vitesse courante du spot laser. 

La commande du miroir, qui contrôle la direction du faisceau laser (position du spot laser dans l'image), est réalisée dans l'espace articulaire du miroir. De fait, les vitesses articulaires sont obtenues en utilisant le modèle cinématique inverse du miroir : $\mathbf{\dot q}_i$ = $\mathbf{D}_{inv}\underline{\mathbf{v}}_{k+1}$.

%
\subsection{Validation}
%
\begin{figure}[!h]
\centering
  \includegraphics[width=0.7\columnwidth]{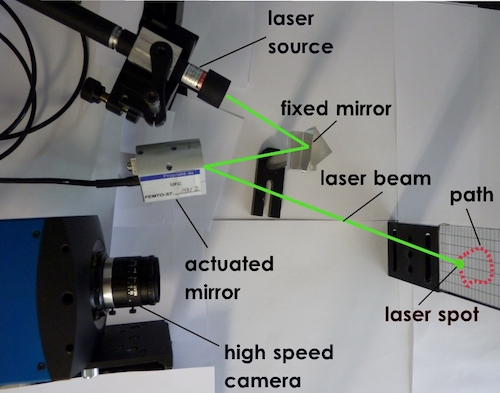}
\caption{Photographie de la station expérimentale utilisée pour la validation des différentes méthodes proposées.}
\label{fig.setup_path}
\end{figure}

La loi de commande exprimée dans (\ref{eq.v}) a été validée en simulation et dans des conditions expérimentales en utilisant le dispositif illustré dans la Fig.~\ref{fig.setup_path}. Le système d'imagerie utilisé consiste en une caméra à haute cadence (jusqu'à 10 000 images/seconde à la résolution 800$\times$600 pixels). Il s'agit de la caméra EoSens $^{\textregistered}$ CXP de chez Mikrotron. Le dispositif d'orientation du faisceau laser (ici un simple pointeur laser rouge) est un miroir à deux degrés de liberté, le S-334 de chez Physical Instruments Inc, caractérisé par une fréquence d'actionnement de 1kHz. 
\subsubsection{Simulation}
\begin{figure}[!h]
\centering
  \includegraphics[width=0.9\columnwidth, height = 5.5cm]{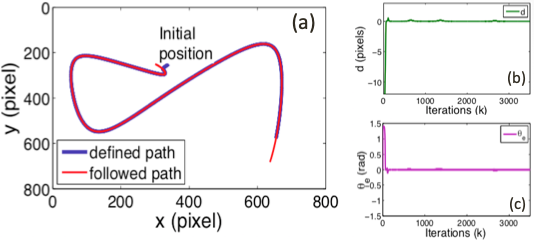}
\caption{Résultats de simulation en suivi de chemin : (a) superposition de chemin de référence et de celui effectué par le spot laser, (b) régulation de l'erreur de distance $d$, et (c) celle de l'orientation $\theta_e$.}
\label{fig.simu_path}
\end{figure}

Préalablement à la validation de la méthode sur un dispositif expérimental, nous avons réalisé plusieurs tests de simulation pour quantifier le comportement de la loi de commande, en termes de stabilité, de convergence, de précision et surtout déterminer, de manière optimale, les gains de commande $\gamma_1$ et $\gamma_2$. La Fig.~\ref{fig.simu_path} montre un exemple d'un suivi de chemin dessiné à main levée. A noter que la courbe de référence et celle réalisée par le contrôleur se superposent parfaitement (Fig.~\ref{fig.simu_path}(a)). Quant à la régulation des erreurs, respectivement, de distance $d$ et d'orientation $\theta_e$, elles convergent rapidement vers $0$ et sont maintenues à cette valeur pendant tout le processus de suivi de chemin. 

\begin{figure}[!h]
\centering
  \includegraphics[width=0.7\columnwidth, height = 5.5cm]{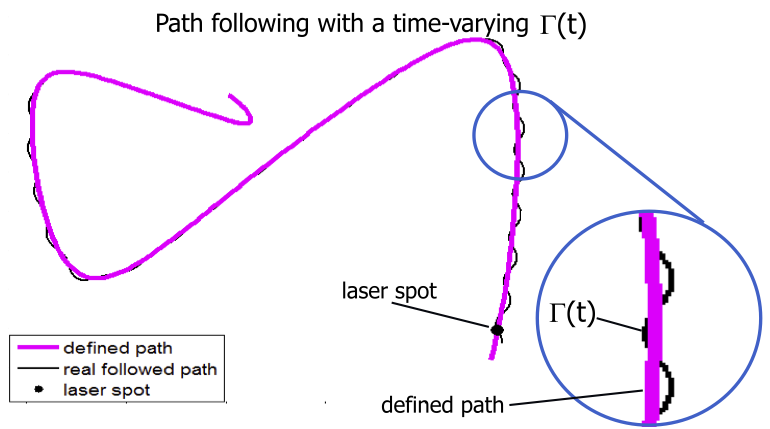}
\caption{Image capturée pendant la simulation du suivi de chemin avec une courbe variable dans le temps.}
\label{fig.path_simu_t}
\end{figure}

Grâce au simulateur que nous avons développé, il a été possible de simuler des mouvements physiologiques sur la courbe $\Gamma$. En d'autres termes, nous avons fait varier la forme et la taille de la courbe pendant le processus de suivi de chemin. Sans faire varier les paramètres de commande : la vitesse initiale $v$, les gains de la commande $\gamma_1$ et $\gamma_2$, la méthode proposée permet de suivre efficacement et avec le même ordre de précision, que le test précédent, le chemin variable $\Gamma(t)$ comme montré dans la Fig.~\ref{fig.path_simu_t}. 

Nous avons également analysé le comportement de la loi de commande en fonction de l'amplitude et du profil de vitesse d'avancée définis par l'opérateur indépendamment de la forme de la courbe à suivre. Nous avons utilisé trois profils différents : échelon, sinusoïde et succession d'échelons avec différentes amplitudes. Le Tableau~\ref{tab.vvar} montre que les erreurs de suivi $d$ et $\theta_e$ restent très similaires d'un test à un autre avec un faible écart-type. En effet, comme énoncée initialement, la qualité du suivi de chemin est complètement décorrélée de l'aspect temporel et de la forme géométrique de la courbe, même si la vitesse d'avancée varie au cours du temps. 
\begin{table}[!h]
	\caption{Erreur quadratique moyenne (RMS) et écart-type (STD) : $d$ (pixels) et $\theta_e$ ($rad$).}
	\centering
\begin{tabular}{|l||c|c||c|c|}
  \hline
  profil de $\emph{v}$ & RMS ($d$) & STD ($d$) & RMS ($\theta_e$) & STD ($\theta_e$) \\
  \hline \hline
 constant & 0.0086 & 0.0080 & 0.0195   & 0.0195\\ 
	sinusoidal  &  0.0073& 0.0065 & 0.0174 & 0.0174   \\
	succession d'échelons & 0.0086 & 0.0078 &0.0198 & 0.0198   \\
	\hline
\end{tabular}
\label{tab.vvar}
\end{table}
\subsubsection{Experimental}
\begin{figure}[!h]
\centering
  \includegraphics[width=0.8\columnwidth, height = 6cm]{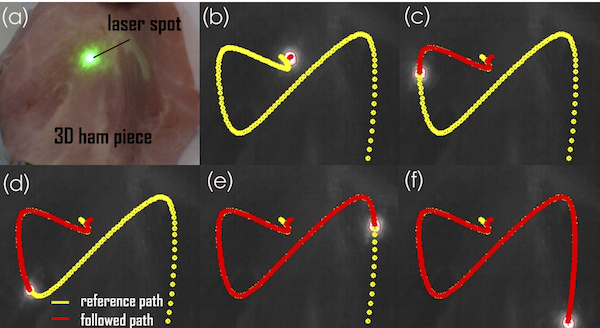}
\caption{Validation expérimentale du suivi de chemin : (a) représente une image couleur de la "pièce du boucher" utilisée pour simuler les cordes vocales, et (b) à (f) illustrent le déplacement du spot laser le long de la courbe $\Gamma$.}
\label{fig.seq_path_exp}
\end{figure}

Les tests de simulations ont été reproduits sur la plateforme expérimentale présentée à la Fig.~\ref{fig.setup_path}. La scène, sur laquelle le spot laser est projeté, est une "pièce du boucher" pour simuler la texture et la forme ondulée (3D) des cordes vocales. Nous avons choisi la même courbe (forme géométrique et taille) que dans les tests de simulation. Comme le montre la Fig.~\ref{fig.seq_path_exp}, la loi de commande permet au spot laser de suivre parfaitement la courbe $\Gamma$ (superposition de la courbe de référence avec le chemin effectué par le spot laser).  

\begin{figure}[!h]
\centering
\subfigure[$d$]{\includegraphics [width=0.4\columnwidth]{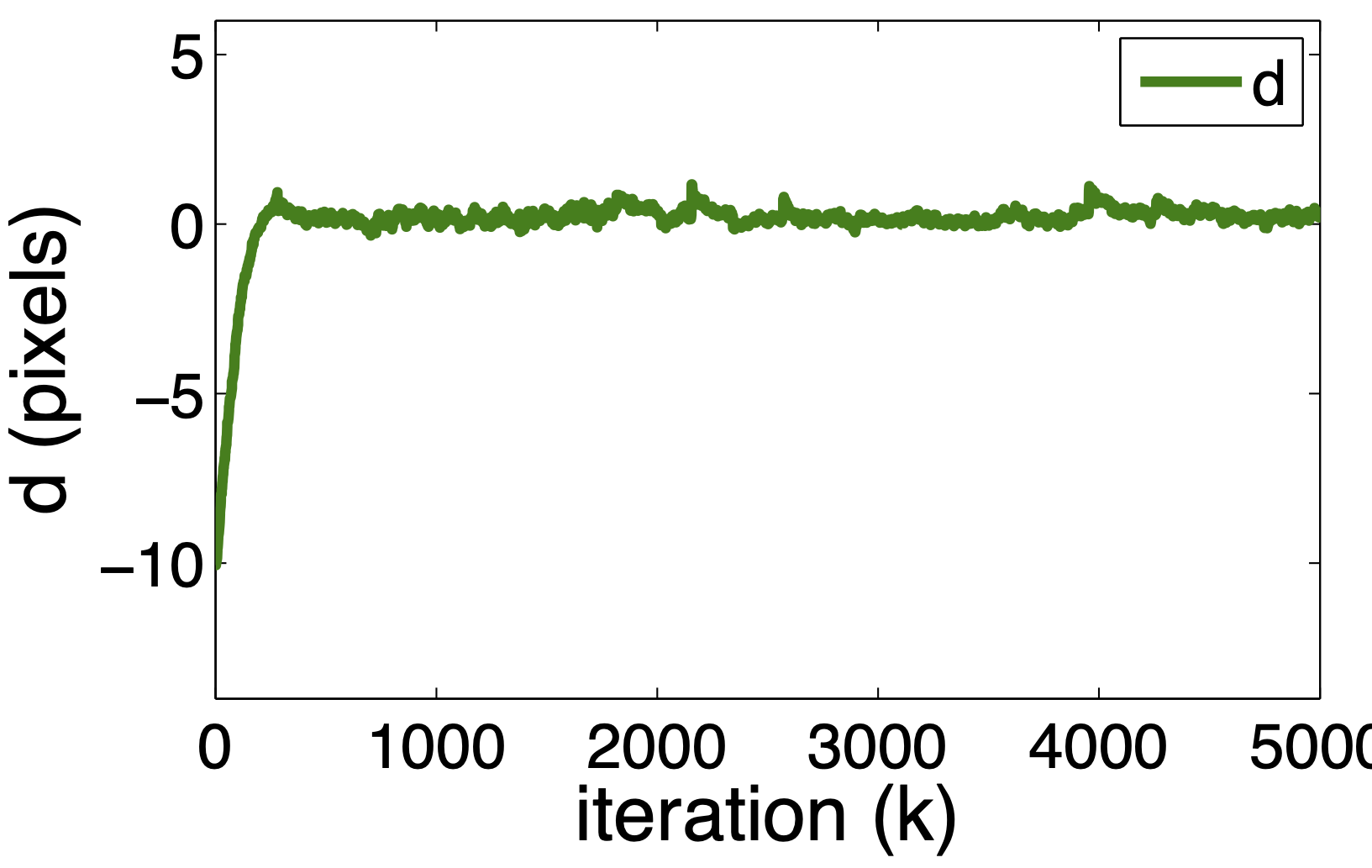}}\quad
\subfigure[$\theta_e$]{\includegraphics [width=0.4\columnwidth]{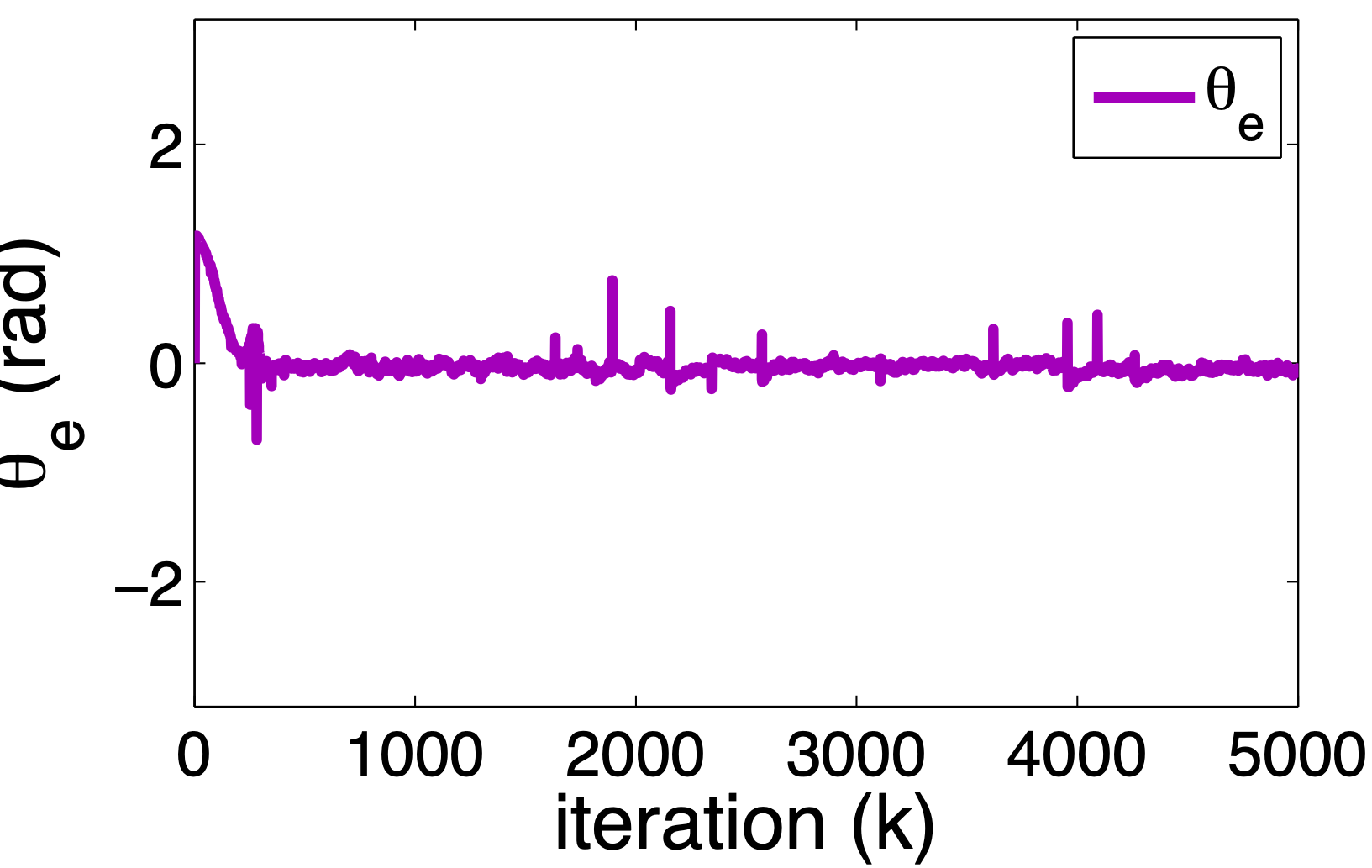}}
\caption{Evolution de $d$ et $\theta_e$ lors du suivi de chemin illustré sur Fig.~\ref{fig.seq_path_exp}.}
\label{fig.d_theta_e}
\end{figure}

Les erreurs d'avancement $d$ et d'orientation $\theta_e$ du spot laser ont été enregistrées et montrées à la Fig.~\ref{fig.d_theta_e}. Il est à noter que les erreurs, sur $d$ et $\theta_e$, diminuent rapidement vers 0 et sont maintenues à cette valeur tout le long du processus de suivi. Les pics aperçus sur la courbe de $\theta_e$ correspondent aux changements de signe et/ou la valeur de la courbure $C(s)$ en certains points de la courbe.  

La répétabilité de la méthode dans les conditions expérimentales a été également étudiée. Il s'agissait de reproduire le même test expérimental que celui dans la Fig.~\ref{fig.seq_path_exp}. Le Tableau~\ref{tab.rep} résume les erreurs moyennes (RMS) et les écarts-types (STD) de chaque série de trois tests individuels effectués sur la même courbe et dans les mêmes conditions expérimentales (gains de commande $\gamma_1$, $\gamma_2$ et vitesse d'avancée $v$). D'après ce même tableau, il est facile de constater que les erreurs (respectivement les écarts-types) ne varient d'un test à un autre que très faiblement, ce qui démontre que la méthode proposée est répétable. 
\begin{table}
	\caption{RMS et STD : $d$ (pixels) et $\theta_e$ ($rad$). Chaque ligne représente la moyenne de 3 tests individuels.}
	\centering
\begin{tabular}{|l||c|c||c|c|}
  \hline
  N$^\circ$ & RMS ($d$) & STD ($d$) & RMS ($\theta_e$) & STD ($\theta_e$) \\
\hline \hline
 	Test 1 & 0.230 & 0.556 & 0.217   & 0.556  \\ 
	Test 2 &  0.223& 0.554 & 0.209 & 0.555   \\
	Test 3 & 0.229 & 0.555 &0.216  & 0.555   \\
	Test 4 & 0.224 &  0.555& 0.210 & 0.555   \\
	Test 5 & 0.231 &  0.554& 0.217 &  0.555  \\
\hline
\end{tabular}
\label{tab.rep}
\end{table}

\begin{figure}[!h]
\centering
  \includegraphics[width=0.8\columnwidth, height = 6cm]{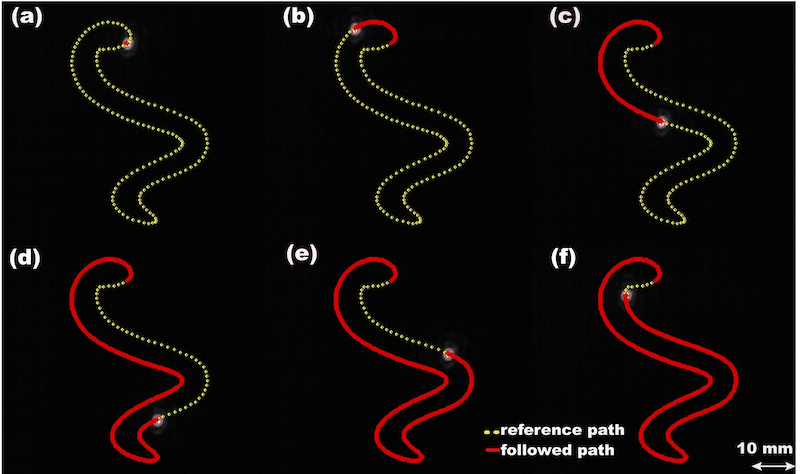}
\caption{Validation expérimentale du contrôleur sur une courbe sous forme de "serpent".}
\label{fig.seq_path_serp}
\end{figure}

D'autres tests expérimentaux ont été réalisés en utilisant des courbes quelconques et plus complexes, par exemple, avec une plus grande courbure. La Fig.~\ref{fig.seq_path_serp} illustre la réalisation d'un suivi de chemin sur une courbe sous forme d'un "serpent" qui peut représenter une délimitation d'un tissu cancéreux à ablater. La précision du suivi de chemin reste du même ordre que les tests évoqués ci-dessus. 

\begin{figure}[!h]
\centering
  \includegraphics[width=0.8\columnwidth, height = 6.5cm]{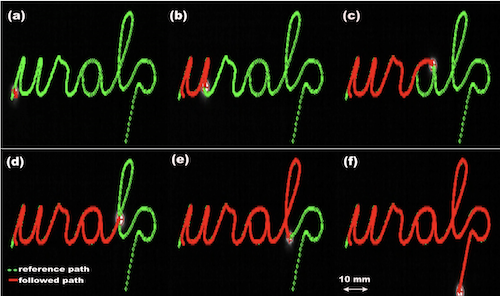}
\caption{Validation expérimentale de la loi de commande sur la courbe constituée des lettres manuscrites "uralp".}
\label{fig.seq_path_uralp}
\end{figure}

\begin{figure}[!h]
\centering
  \includegraphics[width=0.7\columnwidth]{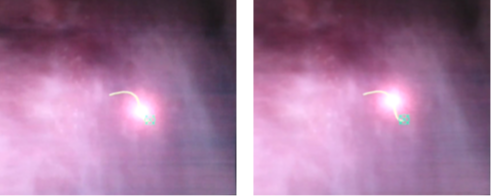}
\caption{Exemple du suivi de chemin réalisé sur les cordes vocales d'un cadavre humain.}
\label{fig.path_cadavre}
\end{figure}

Nous avons également étendu la validation expérimentale plus loin en réalisant un suivi de chemin sur une courbe encore plus complexe. L'idée est de réaliser un suivi de chemin sur les lettres manuscrites "\emph{uralp}" qui présentent de fortes courbures, ainsi que des intersections de la courbe. Comme le montre la Fig.~\ref{fig.seq_path_uralp}, notre loi reste fonctionnelle et précise (superposition des courbes de référence et effectuée). Par ailleurs, la Fig.~\ref{fig.path_cadavre} montre un exemple réalisé par un chirurgien sur les cordes vocales d'un cadavre humain, ce qui démontre le fonctionnement de la méthode même dans les conditions cliniques. 

\subsubsection{Analyse des performances de la méthode} 
Il est à noter que dans le cas de la microchirurgie laser, le laser doit passer à plusieurs reprises et à haute cadence (plusieurs centaines de Hz) sur le chemin de résection afin d'avoir une découpe précise du tissu et ainsi d'éviter la carbonisation des tissus sains avoisinants. Comme le montre le Tableau~\ref{tab.freq}, il convient de noter que le contrôleur proposé ne nécessite que 1,702 millisecondes ($ms$), ce qui est  équivaut à une fréquence de $f$ = 587Hz (temps d'échantillonnage), malgré la lenteur de la communication USB utilisée entre l'ordinateur de commande et le contrôleur du robot (miroir actionné). 
\begin{table}[!h]
	\caption{Temps d'exécution des tâches principales du contrôleur développé.}
	\centering
\begin{tabular}{|l|c|}
  \hline
  Tâche & Temps ($ms$) \\
  \hline \hline
  acquisition d'une image & 0.064 \\
  suivi visuel du spot laser & 0.471 \\
	calcul de la commande &  0.002\\
	communication (USB 2.0) &  0.989\\
	autres & 0.176\\
	\hline
	\textbf{ensemble du processus} & \textbf{1.702 ($f$ = 587 Hz)}\\
	\hline
\end{tabular}
\label{tab.freq}
\end{table}

La précision du contrôleur est très intéressante, en simulation mais aussi dans des conditions expérimentales proches de celles d'un bloc opératoire. La précision est quantifiée dans divers scénarios de validation ; elle est subpixellique en simulation (quelques centièmes de pixel en translation et moins d'un degré en rotation) et de quelques dixièmes de pixel expérimentalement (équivalent à quelques centièmes de millimètres en translation et moins d'un degré en rotation). La méthode proposée montre également de très bonnes performances en termes de robustesse (grandes courbures, variation géométrique de la courbe, vitesse d'avancée variable, etc.) et de répétabilité (plusieurs tests réalisés montrant les mêmes performances). 

Néanmoins, notre méthode utilise uniquement les informations de l'image et une modélisation bidimensionnelle du chemin à suivre, ce qui limite son efficacité sur les surfaces 3D où la profondeur de la scène doit être prise en compte. Cette limitation ainsi que les propositions d'amélioration seront discutées dans la section suivante. 

%
\section{Contrainte trifocale dans le suivi de chemin}\label{sec.hybrid_laser}
\begin{figure}[!h]
\centering
  \includegraphics[width=0.9\columnwidth, height = 7.5cm]{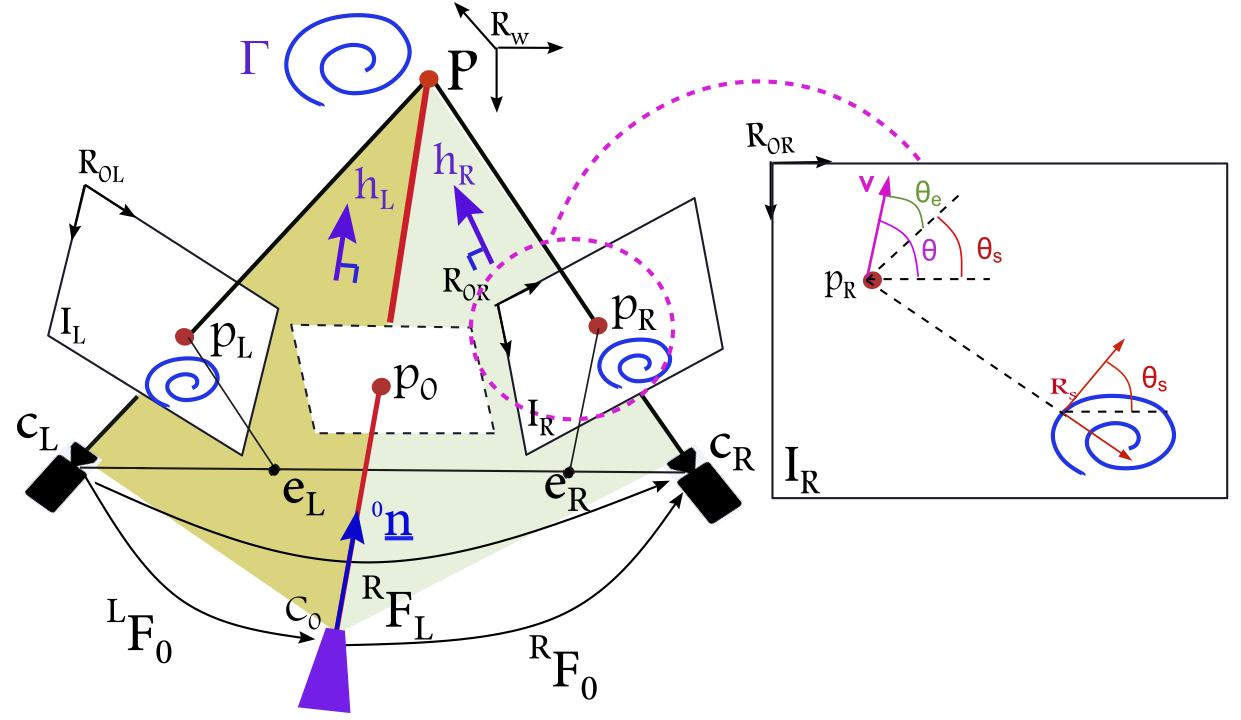}
\caption{Vue d'ensemble d'un système à trois vues comprenant deux caméras réelles et une source laser (caméra virtuelle) dans le cas d'un suivi de chemin 3D.}
\label{fig.concept_path_trifocal}
\end{figure}

%
Cette section est consacrée à la présentation d'une méthode de suivi de chemin associant la contrainte trifocale décrite dans la section~\ref{sec.trifocal}. Cette combinaison de deux concepts géométrique et cinématique dans le même schéma de commande constitue une forte originalité.  Ce concept est schématisé sur la Fig.~\ref{fig.concept_path_trifocal} où nous pouvons   reconnaitre la modélisation trifocale présentée précédemment et la méthode de suivi de chemin à trois vues : deux caméras réelles et une caméra virtuelle (miroir actionné + laser). 

%
\subsection{Loi de commande}
%
Dans la méthode trifocale, nous avons montré que, grâce à la contrainte trifocale, il est possible de lier la vitesse angulaire d'un miroir actionné (ou la vitesse du faisceau laser) à celle du point laser $\dotpl$ et $\dotpr$ définies dans les repères des images gauche et droite $\Repere{i} $($i \in[L, R]$). Par conséquent, le problème du suivi de chemin 3D peut être vu comme la combinaison \emph{cohérente} de deux lois de commande de suivi de chemin 2D. La contrainte trifocale (\ref{eq.z000}) permet de donner un sens géométrique et cinématique à cette association, même si une simple moyenne des vitesses $\dotpl$ et $\dotpr$ peut être envisagée. 
 
Ainsi, les vitesses d'avancées du spot laser dans chacune des images peuvent être exprimées de la manière suivante :
 \begin{equation}
\mathbf{v}_{\texttt{L}} = \frac{ \dotpl + \omega  \times \dotpl} {\left \| \dotpl + \omega \times \dotpl \right \|}, ~~\text{et}~~\mathbf{v}_{\texttt{R}} = \frac{\dotpr + \omega  \times \dotpr} {\left \| \dotpr + \omega  \times \dotpr \right \|}
\label{eq.pl}
\end{equation}   
où, les vitesses dans l'image $\dotpl$ et $\dotpr$ sont obtenues, respectivement, par la différenciation numérique de $\pl$ et $\pr$ calculées à l'aide de l'algorithme de suivi visuel du spot laser.

Ainsi, en injectant les expressions $\mathbf{v}_{\texttt{L}}$ et $\mathbf{v}_{\texttt{R}}$ (\ref{eq.pl}) dans (\ref{eq.cmd2}), nous obtenons la vitesse angulaire du miroir en fonction des vitesses 2D $\mathbf{v}_{\texttt{L}}$ et $\mathbf{v}_{\texttt{R}}$, et de $\dotpl$ et $\dotpr$. Ainsi, à chaque itération, il devient possible de calculer la nouvelle vitesse angulaire notée $\omega$ en utilisant l'expression suivante :
\begin{eqnarray}
\omega &=& - \eta \times \left(\hLr \! \times \! (\Fr \mathbf{v}_{\texttt{R}})
-  \hr \! \times \! (\Fl \mathbf{v}_{\texttt{L}}) \right) 
\label{eq.lb22}
\end{eqnarray}
ainsi, 
\begin{equation}
\omega  =  - \eta \times \Bigg( \hLr \! \times \! \bigg (\Fr \frac{\dotpr + \omega  \times \dotpr} {\left \| \dotpr + \omega  \times \dotpr \right \|} \bigg) - \hr \! \times \! \bigg( \Fl \frac{ \dotpl + \omega  \times \dotpl} {\left \| \dotpl + \omega \times \dotpl \right \|}\bigg) \Bigg) 
\label{eq.lb2}
\end{equation}
où, $\eta = \frac{\hr \times \hLr}{\| \hr \times \hLr \|^2}$ a le même effet qu'un gain de commande constant.
%
\subsection{Validation}
%
Ce nouveau contrôleur a été validé d'abord numériquement sur un simulateur, puis sur une station microrobotique que nous avons développée à cette occasion. 
%
\subsubsection{Simulation}
%
\begin{figure}[!h]
\centering
  \includegraphics[width=0.5\columnwidth, height = 8cm]{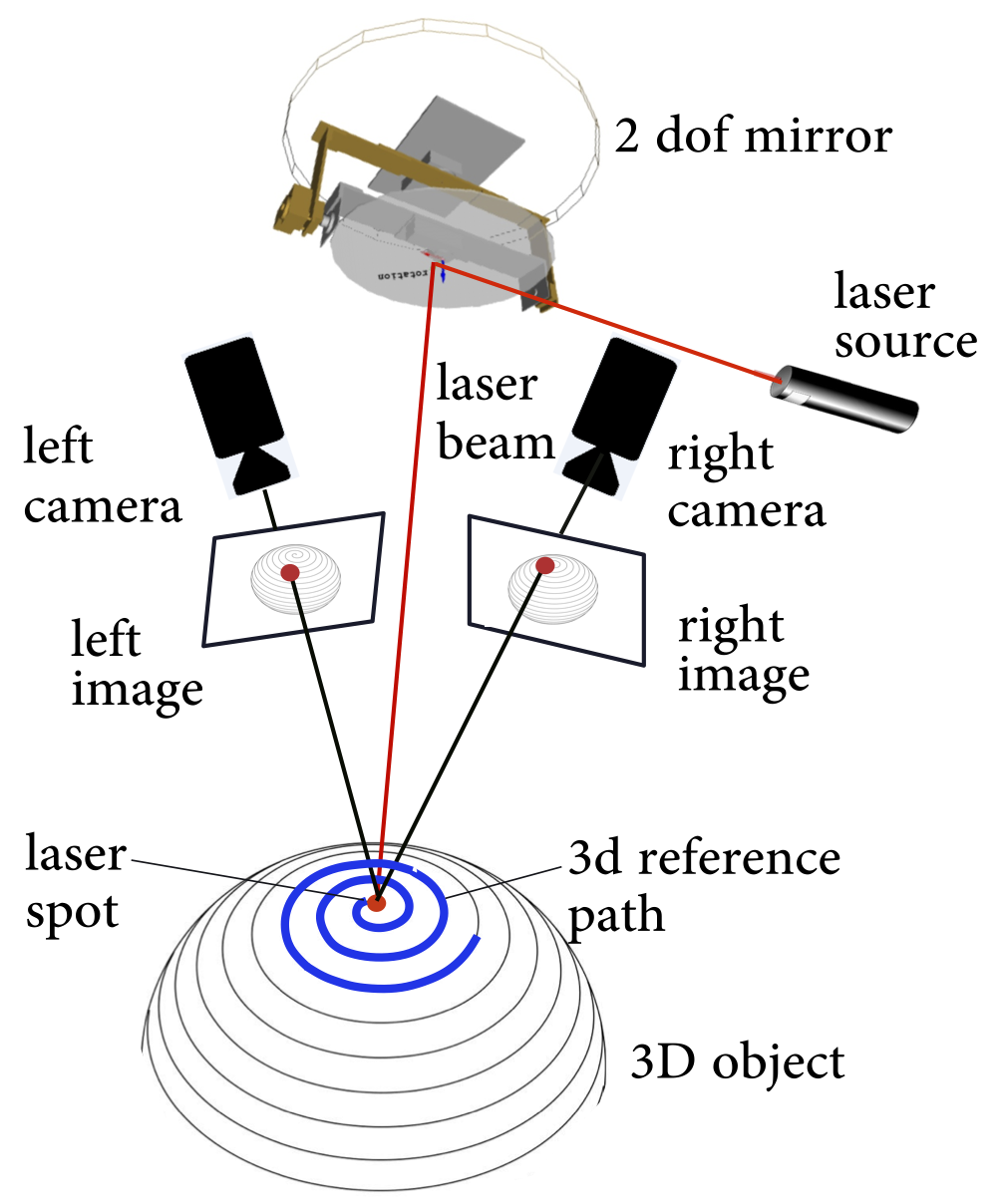}
\caption{Illustration du simulateur de suivi de chemin 3D.}
\label{fig.setup_simu_stereo}
\end{figure}

Préalablement à la validation expérimentale, la loi de commande proposée a été validée à l'aide d'un simulateur développé (dans une version en script \textsc{MATLAB}). Afin d'être le plus proche possible des conditions expérimentales, le simulateur développé inclut les composants principaux mis en \oe uvre dans la configuration expérimentale. Par conséquent, deux caméras (un système de vision stéréoscopique), une source laser, un miroir actionné et un objet 3D sont reproduits dans le simulateur (Fig.~\ref{fig.setup_simu_stereo}). En outre, les différents composants sont positionnés de manière à imiter la configuration réelle du système chirurgical au laser (par exemple, la partie distale d'un endoscope~\citep{rabenorosoa2015}. Les tests de validation ont été réalisés dans des conditions de travail favorables et défavorables (erreurs de mesure et d'étalonnage, cibles déformables, etc.) afin d'évaluer les performances de la loi de commande (précision et robustesse).

Concernant les matrices fondamentales $\Fl$ (entre le miroir et la caméra gauche) et $\Fr$ (entre le miroir et la caméra droite) qui interviennent dans l'expression de la loi de commande, sont "grossièrement" estimées comme suit :
\begin{itemize}
  \item $\mathbf{K}_\texttt{L}  = \mathbf{K}_\texttt{R}  =  \left({\begin{array}{*{20}c}
   {900} & 0 & {320}  \\
   0 & {900} & {240}  \\
   0 & 0 & 1  \\
   \end{array}} \right)$, est la matrice des paramètres intrinsèques des caméras simulées
	\item $\Fmat[0]{\texttt{L}}$ = $\as{{}^0\bf{t}_{\texttt{L}}}{}^0 \bf{R}_{\texttt{L}} \bf K^{-1}$, la matrice fondamentale entre la vue virtuelle et la caméra de gauche, est obtenue (par estimation grossière) grâce au vecteur de translation ${}^0 \bf{t}_{\texttt{L}}$ = (-40, 35, -20)$^\top$~(mm) et la matrice de rotation ${}^0 \bf{R}_{\texttt{L}}= \bf{I}_{{\texttt{3}} \times {\texttt{3}}}$
	\item De même, $\Fmat[0]{\texttt{R}}$ = $\as{{}^0\bf{t}_{\texttt{R}}} {}^0 \bf{R}_{\texttt{R}} \bf K^{-1}$ est obtenue (également par estimation grossière) en utilisant le vecteur de translation ${}^0 \bf{t}_{\texttt{L}}$ = (40, 35, -20)$^\top$~(mm) et la matrice de rotation ${}^0 \bf{R}_{\texttt{R}} = \bf{I}_{{\texttt{3}} \times {\texttt{3}}}$
\end{itemize}

Le seul étalonnage précis nécessaire dans cette méthode est la matrice fondamentale $\Fmat[\texttt{R}]{\texttt{L}}$ entre l'image de gauche et celle de droite. Cette matrice fondamentale permet de définir la courbe $\Gamma_{\texttt{R}}$ dans l'image de droite, tout en gardant une cohérence géométrique,  à partir d'une courbe définie par l'opérateur dans l'image de gauche. A noter que la matrice $\Fmat[\texttt{R}]{\texttt{L}}$  est calculée une seule fois (hors ligne) et n'intervient pas dans le fonctionnement de la loi de commande. 

%
\begin{figure}[!h]
\centering
\includegraphics[width=\columnwidth]{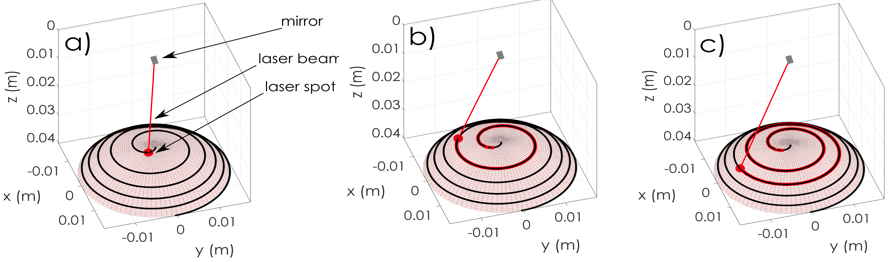}
\caption{Fonctionnement du suivi de chemin 3D sur une sphère : (a) position initiale  (itération $i$ = 0), (b) position intermédiaire ($i$ = 200), et (c) position finale ($i$ = 350).}
\label{fig.seq_simu_3d}
\end{figure}

\begin{figure}[!h]
\centering
\includegraphics[width=\columnwidth]{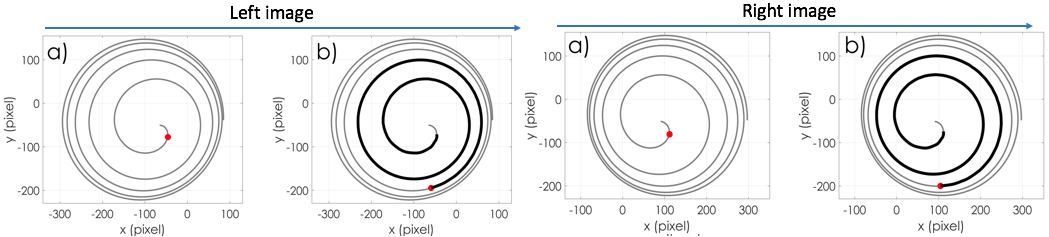}
\caption{Projection dans les images gauche et droite du suivi de chemin 3D sur une sphère.}
\label{fig.seq_simu_3d_proj}
\end{figure}

Comme le montre la Fig.~\ref{fig.seq_simu_3d}, la méthode de suivi de chemin 3D avec l'association de la contrainte trifocale dans la loi de commande fonctionne parfaitement avec un étalonnage grossier et sans connaissance a priori de la scène. La qualité du suivi peut être aussi observée dans les superpositions entre les courbes de référence et projetées dans les images de gauche et de droite (Fig.~\ref{fig.seq_simu_3d_proj}). Les valeurs numériques de la précision du contrôleur obtenues sur plusieurs tests sont de l'ordre de 0.03~mm de moyenne (RMS) avec un écart-type de 0.35~mm.  

\begin{figure}[!h]
\centering
\includegraphics[width=.7\columnwidth]{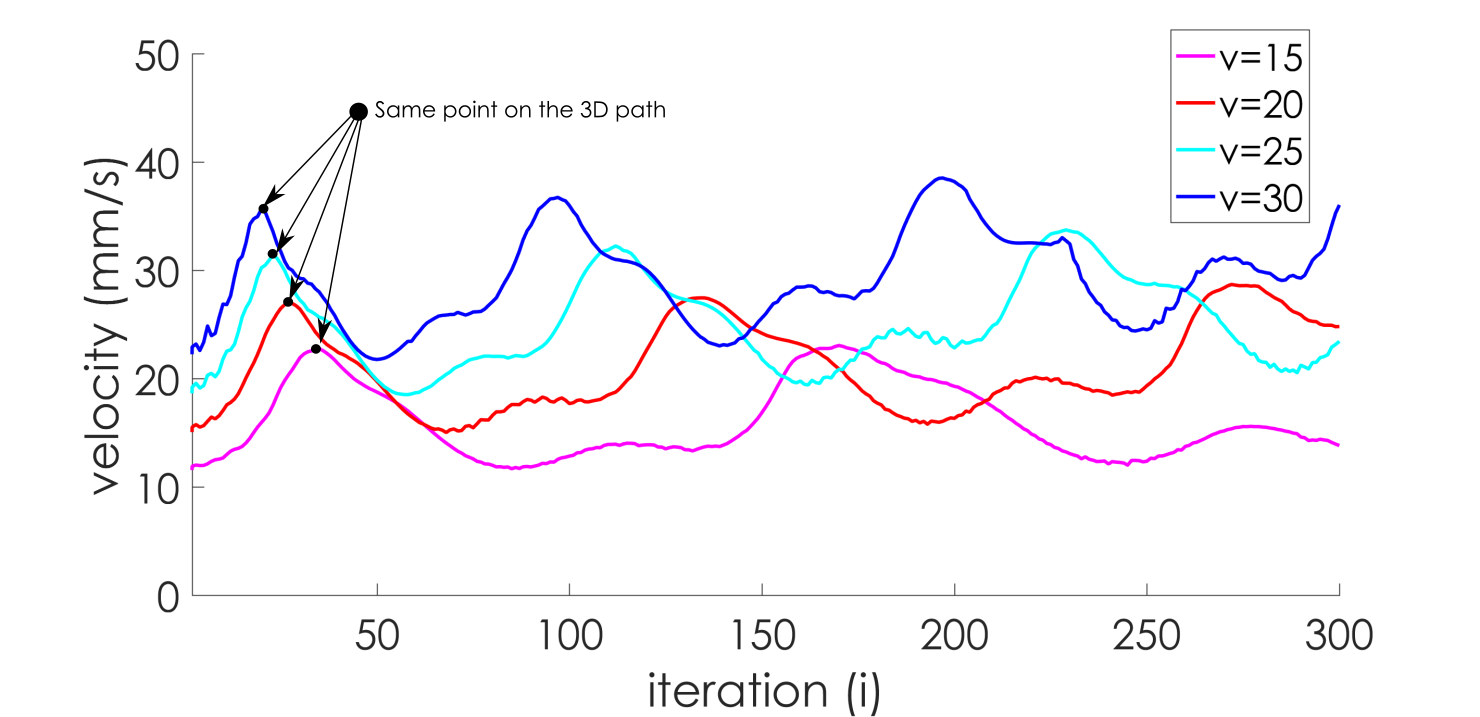}
\caption{Illustration du profil de la vitesse d'avancée le long d'un chemin 3D.}
\label{fig.prof_v_3d}
\end{figure}

Par ailleurs, comme énoncée préalablement, l'autre caractéristique de cette méthode est la capacité à garder la vitesse d'avancée constante sur le long du chemin indépendamment du temps et de la forme géométrique (ici 3D) de la courbe à suivre. Ce découplage est visible sur la Fig.~\ref{fig.prof_v_3d}. 

\begin{figure}[!h]
\centering
\includegraphics[width=\columnwidth]{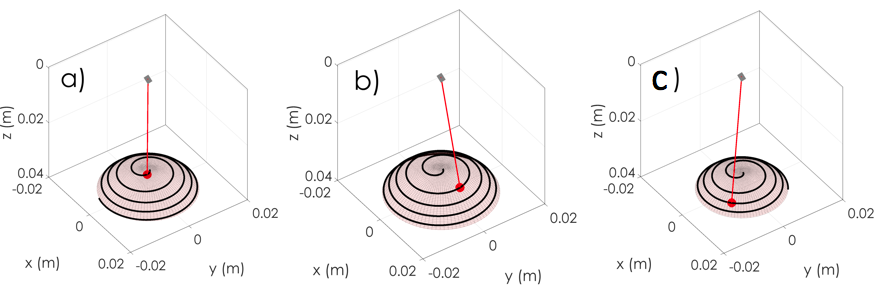}
\caption{Quelques illustrations du suivi de chemin sur une courbe 3D variable $\Gamma_{3d}(t)$.}
\label{fig.seq_path_3d_var}
\end{figure}

Des tests de validation supplémentaires ont également été réalisés sur des courbes 3D variables dans le temps pour simuler une scène en mouvement ou encore les mouvements physiologiques (battements du c\oe ur et respiration) lorsqu'il s'agit d'un cadre applicatif clinique. La Fig.~\ref{fig.seq_path_3d_var} montre le fonctionnement de la méthode sur une portion de sphère qui change sinusoïdalement de taille (largeur et hauteur) d'environ 800\% (entre la plus petite et la plus grande taille). Il est à noter que la loi de commande reste fonctionnelle et précise malgré les grandes variations de la scène. 
%
\subsubsection{Expérimentalation}
%
\begin{figure}[!h]
\centering
\includegraphics[width=.6\columnwidth]{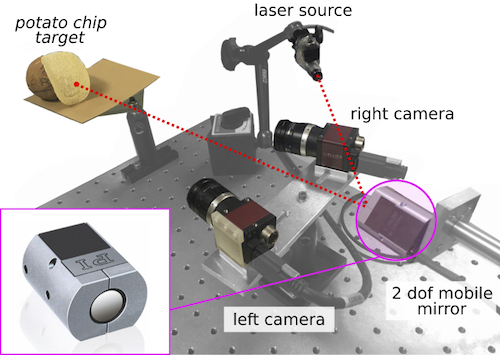}
\caption{Photographie de la nouvelle station expérimentale utilisée pour la validation du suivi de chemin 3D.}
\label{fig.setup_exp_3d}
\end{figure}

Finalement, cette nouvelle loi de commande a été validée expérimentalement en utilisant la plateforme montrée à la Fig.~\ref{fig.setup_exp_3d}.  Contrairement au cas de la simulation, la définition de la courbe 3D est plus complexe dans la réalité. En fait, il nécessaire de reconstruire la courbe 3D à partir d'uniquement la courbe 2D dessinée sur une image (respectivement sur une interface de type tablette). 
\begin{figure}[!h]
\centering
\includegraphics[width=.7\columnwidth, height = 5cm]{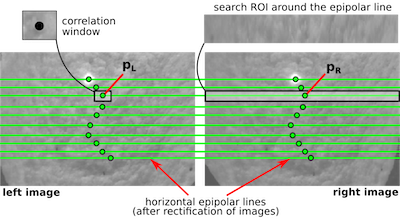}
\caption{Exemple montrant la méthode de détermination d'une courbe 3D en utilisant la géométrie épipolaire. (a) ensemble de points 2D ($\Gamma_{\texttt{L}}$) définis par un opérateur dans l'image de gauche, et (b) la courbe 2D ($\Gamma_{\texttt{R}}$)  correspondante calculée automatiquement dans l'image de droite.}
\label{fig.def_path_3d}
\end{figure}

Pour ce faire, nous utilisons les principes de la géométrie épipolaire et la triangulation. D'abord, il faut calculer la matrice fondamentale $\Fmat[\texttt{R}]{\texttt{L}}$  la plus précise possible, entre les images de gauche et de droite.  Une fois que $\Fmat[\texttt{R}]{\texttt{L}}$  est déterminée, nous opérons une rectification d'images pour avoir les lignes épipolaires parallèles et horizontales dans les deux images. Finalement, l'opérateur défini la courbe échantillonnée $\Gamma_{\texttt {L}}$ (sous forme d'un vecteur de points 2D) dans l'image de gauche, par exemple et grâce à la matrice $\Fmat[\texttt{R}]{\texttt{L}}$, à la rectification des images et à la  recherche de mise en correspondance (auto-corrélation)  le long des lignes épipolaires,  nous pouvons retrouver la courbe $\Gamma_{\texttt{R}}$ dans l'image droite (Fig.~\ref{fig.def_path_3d})~\citep{tamadazteTmech2016}. 
\begin{figure}[!h]
\centering
\includegraphics[width=\columnwidth, height = 3cm]{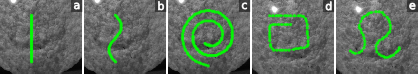}
\caption{Echantillons de courbes utilisées pour la validation du suivi de chemin 3D.}
\label{fig.exple_path_3d}
\end{figure}

\begin{figure}[!h]
\centering
\includegraphics[width=\columnwidth, height = 3cm]{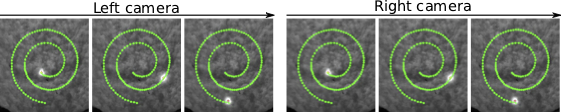}
\caption{Exemple du suivi de chemin 3D sur une courbe en spirale.}
\label{fig.seq_exp_path_3d}
\end{figure}

Nous avons testé la méthode de suivi de chemin 3D sur différentes courbes générées automatiquement ou dessinées à la main. La Fig.~\ref{fig.exple_path_3d} montre un échantillon de ces courbes. Par ailleurs, la Fig.~\ref{fig.seq_exp_path_3d} montre l'exécution du chemin correspondant à la spirale. Nous pouvons noter que le laser suit précisément la courbe à la fois dans les images gauche et droite, mais également dans le repère monde, une fois projetée sur la scène 3D.  Pour davantage de détails sur ces travaux, le lecteur est invité à consulter~\citep{tamadazteTmech2016}.

Toutes les autres courbes illustrées à la Fig.~\ref{fig.exple_path_3d} ont été réalisées et les résultats obtenus, en terme de précision, sont résumés dans le Tableau~\ref{tab.exp_3d}. Nous pouvons constater que la méthode de suivi de chemin 3D fonctionne sur l'ensemble des courbes choisies et que la précision reste très intéressante de l'ordre du millimètre malgré un étalonnage très grossier des matrices fondamentales $\Fl$ et $\Fr$. 
\begin{table}[!h]
\centering
\caption{Analyse de la précision sur différents chemins 3D.}
\label{tab.exp_3d}
\begin{tabular}{|l||c||c||c||c||c||c|}
\hline
                  & \multicolumn{3}{c||}{$e_\texttt{L/R}$ (px)} & \multicolumn{3}{c|}{$e_\texttt{L/R}$ (mm)} \\ 
\hline
                  & moyenne  &  STD  & RMS   & moyenne  & STD  & RMS   \\
\hline \hline
ligne             & 2.45  & 1.12  & 2.70  & 0.87   & 0.39  & 0.96  \\
sinusoïdale             & 2.23  & 1.29  & 2.58  & 0.79   & 0.46  & 0.91  \\
spirale           & 2.80   & 1.96  & 3.44  & 0.99   & 0.69  & 1.22  \\
sigma             & 2.35   & 1.37  & 2.72  & 0.83   & 0.48  & 0.97  \\
leo              & 2.02   & 1.48  & 2.51  & 0.72  & 0.52  & 0.89  \\
\hline
\end{tabular}
\end{table}
%
\section{Bilan}\label{sec.bilan_laser}
%
Les travaux décrits dans ce chapitre traite de la chirurgie laser guidée par l'image. Nous nous sommes appuyés sur des concepts de la vision par ordinateur (géométrie), et de la robotique mobile (suivi de chemin) pour proposer trois lois de commande originales et génériques pour le contrôle des déplacements d'un spot laser (respectivement, un faisceau laser) sur une surface quelconque. Les applications qui en découlent sont nombreuses, non seulement en chirurgie (cordes vocales, prostate, ophtalmologie, etc.), mais également en industrie (micro-usinage laser, soudure, découpe, etc.).  

La méthode est fondée sur l'insertion de la nouvelle contrainte trifocale \emph{revisitée} pour une meilleure intégration dans une loi de commande de type asservissement visuel. Dans ce travail, nous avons exprimé la nouvelle contrainte trifocale en utilisant uniquement deux caméras réelles et une caméra virtuelle (la source laser) pour construire une géométrie à trois vues. Cette méthode présente plusieurs avantages : 
\begin{itemize}
\item pas d'inversion de matrice, ni de manipulation de formules mathématiques complexes ;
\item pas de connaissance a priori de la configuration du robot (miroir actionné) ;
\item sans aucune connaissance de la structure géométrique (exemple, la profondeur) de la scène ; 
\item ne nécessite qu'un étalonnage faible (paramètres intrinsèques des caméras, poses du miroir vis-à-vis des caméras, etc.) ;
\item ne nécessite qu'une vingtaine de lignes de code C++ pour l'implémentation du contrôleur ;
\item capacité à exprimer explicitement les conditions de stabilité du contrôleur.
\end{itemize}

Nous avons également travaillé sur la commande du laser le long d'un chemin quelconque et non paramétré, méthode inspirée de la robotique mobile. Ce travail a pour but de proposer une méthode simple et efficace pour la chirurgie laser par retour visuel. Notre but était de proposer un moyen de contrôler un laser le long d'une courbe tout en garantissant une vitesse d'avancée du laser complètement décorrélée de la géométrie et de l'aspect temporel de la courbe à suivre. Cette méthode présente plusieurs avantages tels que : 
\begin{itemize}
\item une très bonne précision (<0.5~mm dans les conditions expérimentales), 
\item une robustesse par rapport à des perturbations (courbe variable dans le temps),
\item une haute fréquence de commande, environ 600~Hz pour l'ensemble de la boucle de commande (sans optimisation),
\item une robustesse par rapport aux erreurs d'étalonnage (caméra, miroir, robot/miroir, etc.).
\end{itemize}

Pour une meilleure prise en compte des surfaces 3D, nous avons développé une troisième famille de contrôleurs intégrant dans le même schéma de commande, le suivi de chemin et la contrainte trifocale, ce qui constitue une originalité complémentaire. Les performances de cette nouvelle famille de contrôleurs respectent parfaitement le cahier des charges de la micro/chirurgie laser. 

Enfin, ces travaux sur la commande laser par retour visuel ont donné lieu à plusieurs publications scientifiques. Ci-après une liste non-exhaustive de ces publications.

\footnotesize
\subsubsection{Liste des publications scientifiques [depuis 2012]}
\begin{enumerate}

\item [Ji] \textbf{B. Tamadazte}, \underline{\underline{R. Renevier}}, \underline{\underline{J.-A. Séon}}, and N. Andreff, \emph{Laser beam steering along 3-dimensional paths}, IEEE/ASME Trans. on Mechatronics, 2018, 23(3), 1148-1158.
\item [Ji]\underline{\underline{R. Renevier}}, \textbf{B. Tamadazte}, K. Rabenorosoa, and N. Andreff, \emph{Microrobot for laser surgery: design, modeling and control}, IEEE/ASME Trans. on Mechatronics, 2018, 22 (1), 99-106.
\item [Ji] N. Andreff,  and \textbf{B. Tamadazte}, \emph{Laser steering using virtual trifocal visual servoing}, Int. J. of Robotics Research, DOI: 0278364915585585, 2016.
\item [Ji] \underline{\underline{J.-A. Séon}}, \textbf{B. Tamadazte}, and N. Andreff, \emph{Path tracking and visual servoing for laser surgery}, IEEE Trans. on Robotics, DOI: 10.1109/TRO.2015.2400660, 2015.
\item [ ----- ]
\item [CL] \underline{B. Dahroug}, \textbf{B. Tamadazte}, and N. Andreff, \emph{3D path following with remote center of motion constraints}, selected paper from ICINCO 2015, LNCS Springer, 2016. 
\item [ ----- ]
\item [Ci]  \underline{B. Dahroug}, \underline{\underline{J.-A. Séon}}, \textbf{B. Tamadazte}, N. Andreff,  A. Oulmas, T. Xu, and S. Régnier, \emph{Some examples of path following in microrobotics}, MARSS, Nagoya, Japan, (in press).
\item [Ci] \underline{B. Dahroug}, \textbf{B. Tamadazte}, and N. Andreff, \emph{Visual servoing controller for performing time-invariant 3D path following with remote center of motion constraints}, IEEE ICRA, Singapore, 2017, pp. 3612-3618.
\item [Ci] \underline{B. Dahroug}, \textbf{B. Tamadazte}, and N. Andreff, \emph{3D Path following with remote center of motion constraints}, ICINCO, Lisbonne, pp.~84-91, 2016.
\item [Ci]  \textbf{B. Tamadazte}, and N. Andreff, \emph{Weakly Calibrated Stereoscopic Visual Servoing for Laser Steering: Application to Microphonosurgery}, IEEE/RSJ IROS,  Chicago, IL, pp. 743-748, 2014.
\item [Ci]  N. Andreff, S. Dembélé, \textbf{B. Tamadazte}, and Z. Hussnain, \emph{Epipolar geometry for vision-guided laser surgery}, ICINCO, 2013. 
\item [Ci]  S. Dembélé, \textbf{B. Tamadazte}, Z. Hussnain, and N. Andreff, \emph{Preliminary variation on vision-guided laser phonomicrosurgery using multiview geometry}, 1st Russian 
German Conf. on Biomedical Eng., 2013.
\item [ ----- ]
\item [Wi]  \underline{\underline{J.-A. Séon}}, \textbf{B. Tamadazte}, and N. Andreff, \emph{Path following: from mobile robotics to laser surgery}, ViCoMor workshop, IEEE/RSJ IROS, Chicago (USA), pp. 1-6, 2014.
\end{enumerate}
\normalsize

\vspace{.5cm}

\noindent
\underline{P. Nom} : doctorant(e)\\
\underline{\underline{P. Nom}} : stagiaire\\

\noindent
Ji : journal international avec comité de lecture\\
CL :  chapitre de livre\\
Ci : conférence internationale avec actes et comité de lecture\\
Wi : workshop international avec actes et comité de lecture

%% file: fichiers/wavelet_v3.tex
%
\chapter{De la parcimonie à l'asservissement visuel}\label{chap.wavelet}

\minitoc

\emph{Ce chapitre décrit les contributions scientifiques relatives à l'asservissement visuel dit direct. Par direct, nous sous-entendons, toutes les méthodes, qui ne nécessitent pas d'extraction et de sélection d'informations visuelles, de mise en correspondance et de suivi visuel de ces informations. Autrement dit, les informations globales de l'image sont modélisées et utilisées dans la boucle de commande sans segmentation proprement dit. Nous avons travaillé sur une nouvelle famille d'asservissements visuels directs où le signal d'entrée de la boucle de commande est un vecteur d'informations fréquentielles ou spatio-fréquentielles parcimonieuses obtenues par la décomposition multi-échelle de l'image dans une autre base. Ces décompositions peuvent être la transformée de Fourier, les ondelettes ou encore les shearlets. Ces méthodes montrent une plus grande précision, mais sont aussi plus robustes par rapport aux méthodes existantes dans la littérature. Elles s’adaptent davantage à certaines applications cliniques, dont les images utilisées sont caractérisées par un rapport signal/bruit défavorable, une faible texture, et l'absence de formes géométriques. \\
Les lois de commande proposées, dans ce chapitre, ont été validées avec succès en considérant les images endoscopiques conventionnelles, l'échographie ou encore la tomographie à cohérence optique.} \\
%
%
\section{Contexte et positionnement}
\begin{figure}[!h]
  \centering
  \includegraphics[width=\columnwidth]{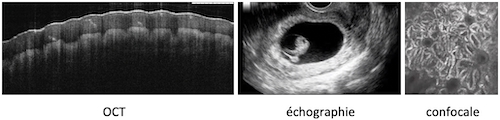}
  \caption{Illustration de la particularité de certaines images médicales.}
  \label{fig.images_medicales}
\end{figure}

%
A l'instar du chapitre précédent, celui-ci décrit les travaux, qui ont une visée clinique. L'objectif est de mettre en \oe uvre des lois de commande par asservissement visuel capables de fonctionner même dans des conditions défavorables. Les applications cliniques de ces travaux sont multiples : réalisation de tâches de positionnement automatiques et très précises comme le diagnostic (par exemple, échographie), le suivi d'un organe en mouvement (compensation des variations physiologiques), suivi de l'évolution d'une pathologie (biopsie optique répétitive), etc. Ces méthodes doivent opérer avec une grande robustesse vis-à-vis de la qualité des images utilisées (faible texture, rapport signal/bruit défavorable, ... Fig.~\ref{fig.images_medicales}), des paramètres d'étalonnage (caméra, robot, caméra/robot, etc.) et d'autres types de perturbations, comme les variations de texture des tissus pendant le temps, occultations (informations tronquées), etc. \\

Les tâches de positionnement sont réalisées à l'aide de lois de commande d'asservissement visuel. Ces approches utilisent généralement des informations visuelles de nature 2D~\citep{espiau1992new}, 2D 1/2~\citep{malis19992} ou 3D \citep{martinet1996vision, wilson1996relative} extraites de l'image. Le lecteur peut se référer aux tutoriels \citep{chaumette2006visual, chaumette2007visual} pour tout savoir sur les bases du formalisme théorique des grandes familles d'asservissement visuel. Plus récemment, d'autres techniques ont vu le jour. Ces techniques dites "\emph{directes}" ou "\emph{denses}" permettent de s'affranchir des étapes de détection, de mise en correspondance, et de suivi visuel (et de reconstruction) des informations visuelles utilisées dans le schéma de commande. Ces nouvelles approches considèrent toute l'information contenue dans l'image, nous parlons alors d'informations globales de l'image. Les premières approches directes sont apparues au milieu des années 90 avec des méthodes reposant sur l'analyse en composantes principales \citep{nayar1996subspace, deguchi1996visual} permettant de contrôler 2 ddl en réduisant le dimensionnement des données constituant l'information visuelle. En 2007, d'autres méthodes sont proposées avec, d'une part une méthode utilisant uniquement les homographies~\citep{benhimane2007homography}, et d'autre part une méthode fondée sur les noyaux de convolution~\citep{kallem2007kernel}. \\

A partir de 2008, une nouvelle étape est franchie avec l'utilisation de l'information visuelle globale dans l'asservissement visuel, à l'exemple, de la modélisation de l'intensité des pixels de toute l'image associée à la contrainte du flux optique, pour la mise en \oe uvre d'une loi de commande dite photométrique~\citep{collewet2011photometric,tamadazte2011direct}. Il faut noter, que cette contrainte du flux optique stipule que l'intensité d'un point image est indépendante du temps, ce qui n'est pas toujours le cas lors du changement des conditions d'illumination de la scène, par exemple. D'autres approches plus robustes, à ce type de conditions défavorables, sont apparues utilisant les gradients de l'image~\citep{marchand2010using}, l'information mutuelle~\citep{dame2011mutual}, la somme de variances conditionnelles d'une image~\citep{richa2011variance, delabarre2012visual}, les modèles de mélanges de Gaussiennes~\citep{crombez2018visual}, ou encore, plus récemment, les histogrammes~\citep{bateux2017histograms}. Une technique utilisant les moments photométriques a également été mise en \oe uvre dans~\citep{bakthavatchalam2013photometric}. En effet, l'utilisation récente de méthodes d'apprentissage de type réseaux de neurones a été étudiée dans~\citep{bateux2018training}.\\

D'autres types d'informations visuelles globales peuvent être également prises en compte dans la conception d'une loi de commande d'asservissement visuel direct. Elles peuvent être de type fréquentielles (utilisant la transformée de Fourier)~\citep{marturi2014visual, tamadazteTase2016}, ou encore spatio-fréquentielles obtenues grâce aux méthodes de décomposition multi-échelle comme les ondelettes, les shearlets, etc. Ces informations visuelles peuvent apporter davantage de robustesse et de précision, notamment lorsque les images utilisées sont de mauvaise qualité, comme dans le cas de l'imagerie médicale.\\

Dans la suite de ce chapitre, il sera question de ces nouvelles méthodes, dont nous détaillerons au mieux la méthodologie de mise en \oe uvre.   
%
\section{Transformée de Fourier}
\begin{figure}[!h]
  \centeringÒ
  \includegraphics[width=0.8\columnwidth]{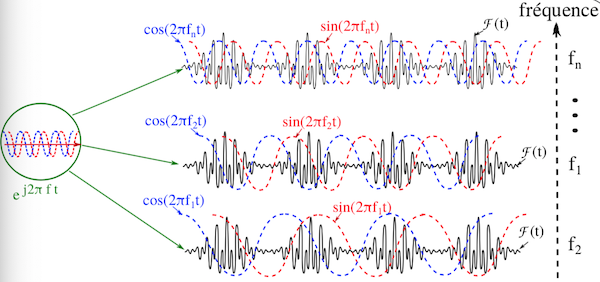}
  \caption{Transformée de Fourier sur plusieurs fréquences \citep{Gao2011}.}
  \label{fig.fourier}
\end{figure}

%
Dans cette section, nous allons rappeler les principes de base de la transformée de Fourier et ses différentes variantes, notamment celle dite à "fenêtre glissante". Cette dernière est considérée comme un premier pas vers la transformée en ondelettes, en curvelets, ou encore les shearlets. Ces brefs rappels seront nécessaires pour comprendre notre démarche dans la conception de lois de commande par asservissement visuel, dont les informations visuelles, prises en compte les matrices d'interactions associées, sont de type fréquentielles ou spatio-fréquentielles. \\
%
%
Introduite en 1822, la transformée de Fourier est sans doute l'outil le plus utilisé pour le traitement des signaux stationnaires~\citep{fourier_2009}. Elle peut être vue comme un prisme, qui décompose le signal sur une base de fonctions trigonométriques (sinus et cosinus). Elle permet de représenter un signal, défini initialement dans l'espace-temps, dans un nouvel espace fréquentiel (Fig.~\ref{fig.fourier}). Plusieurs variantes de cette transformée existent : continue, discrète, rapide, etc.

A noter que la transformée de Fourier repose sur le constat que toute fonction périodique peut être décomposée en plusieurs séries trigonométriques (séries de Fourier) exprimées comme suit :
\begin{equation}
f(x) = \sum a_n \cos(nx) + b_n \sin(nx),
\label{eq.sfourier}
\end{equation}
où, $a_0$ est la composante continue, $a_1$ et $b_1$ sont les composantes fondamentales et $a_n$ et $b_n$ ($n>1$) constituent les harmoniques.

Cette représentation a été par la suite étendue à toutes fonctions intégrables dans $\mathbb{R}$ donnant naissance à la transformée de Fourier. Elle est définie comme suit :
\begin{equation}
\mathcal{F}(f): \omega ~\rightarrow~ \hat{f}(\omega)= \int_{-\infty }^{-\infty } f(t) e^{- i \omega t}dt,
\label{eq.tfourier}
\end{equation}
avec $f$ une fonction intégrale dans $\mathbb{R}$, $i$ est un nombre imaginaire, $t$ représente le temps, et $\omega$ les fréquences. 

La transformée de Fourier admet également une transformée inverse (fréquence $\rightarrow$ temps) notée $\mathcal{F}^{-1}$ définie par :
\begin{equation}
f(t) = \mathcal{F}^{-1}(\hat{f})(t) = \frac{1}{2\pi} \int_{-\infty }^{-\infty } \hat{f}(\omega) e^{+i \omega t}d\omega.
\label{eq.invfourier}
\end{equation}

En traitement d'images, la transformée de Fourier correspondante est bi-dimensionnelle. Ainsi, l'aspect temporel devient spatial et ainsi pour toute fonction $f$ définie dans $\mathbb{R}^2$, sa transformée de Fourier 2D s'écrit :
\begin{equation}
\nonumber
\hat{f}(\omega_1, \omega_2)= \int_{-\infty}^{-\infty} \int_{-\infty}^{-\infty} f(x,y) e^{- i (\omega_1 x + \omega_2 y}) dx dy
\label{eq.tfourier}
\end{equation}

\begin{figure}[!h]
  \centering
  \includegraphics[width=0.8\columnwidth]{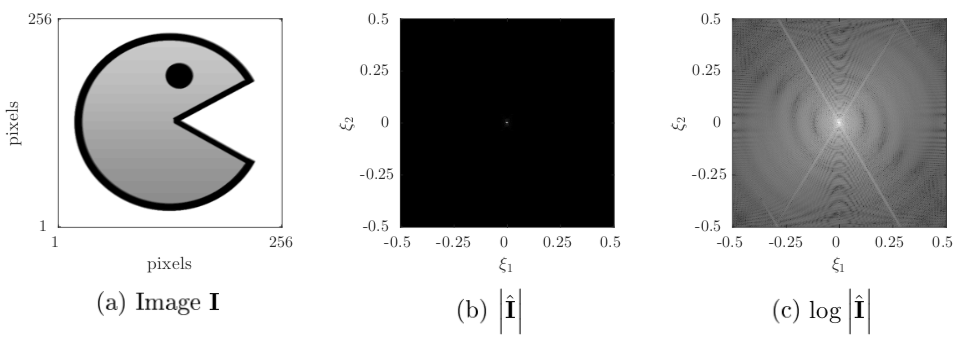}
  \caption{Transformée de Fourier 2D appliquée à l'image.}
  \label{fig.tf_image}
\end{figure}

La visualisation de l'information visuelle dans le domaine fréquentiel est généralement réalisée en échelle logarithmique pour à la fois représenter les hautes et les basses fréquences (Fig.~\ref{fig.tf_image}). 

\begin{figure}[!h]
  \centering
  \includegraphics[width=0.8\columnwidth]{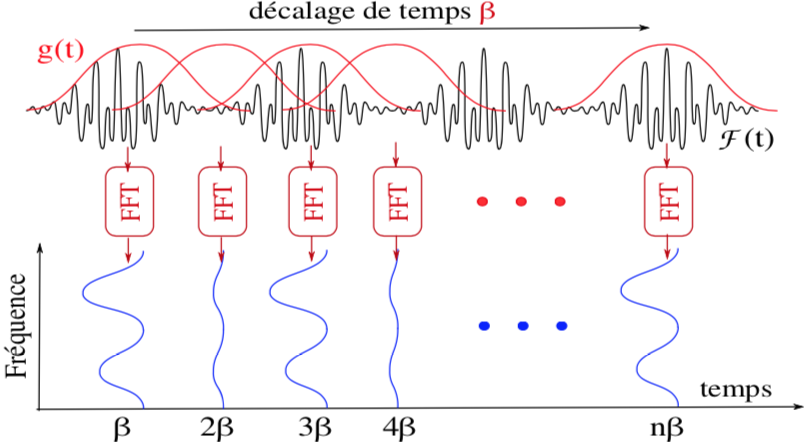}
  \caption{Transformée de Fourier à fenêtre glissante~\citep{Gao2011}.}
  \label{fig.tfGabor}
\end{figure}

Par ailleurs, la transformée de Fourier se révèle inadaptée, quant à la représentation des signaux/images faisant apparaître des évènements transitoires (informations non-stationnaires)~\citep{fourier_2009}. Les variations locales de l'image (contours, points d'intérêt) deviennent des caractéristiques globales dans le domaine fréquentiel qui peuvent rendre difficile une modélisation précise. Il est alors préférable de décomposer l'image en plusieurs parties (segments) et ensuite d'appliquer la transformée de Fourier indépendamment sur chacun des segments. Pour ce faire,~\citep{Gabor1946} a proposé la transformée de Fourier à fenêtre glissante comme illustrée dans la Fig.~\ref{fig.tfGabor}. 

Grâce à la transformée de Fourier à fenêtre glissante, il devient possible d'avoir une représentation spatio-fréquentielle de l'information visuelle. Néanmoins, la fenêtre de Gabor est fixée au préalable, limitant ainsi l'efficacité de la décomposition. 
\subsubsection{Lois de commande}
Nous avons développé deux méthodes d'asservissement visuel direct dont l'information visuelle utilisée est spectrale. La première méthode consiste à utiliser la représentation \emph{log-polaire} de l'image 2D dans le domaine fréquentiel, pour calculer une matrice d'interaction analytique. Cette approche peut être classée dans la famille des asservissements visuels 2D directs (la régulation de l'erreur s'effectue dans l'image).  Nous avons réalisé des validations en simulation et en conditions expérimentales, sur une station de microrobotique à 6 degrés de liberté, dont les performances sont très intéressantes, ceci en termes de précision et robustesse par rapport au bruit de l'image. Pour davantage de détails sur cette méthode, le lecteur peut se référer aux travaux publiés dans~\citep{tamadazteTase2016, tamadazteTim2016}. La seconde méthode peut être considérée comme un asservissement visuel 3D direct. Grâce à l'information spectrale, nous avons reconstruit la pose partielle (4 degrés de liberté) de la caméra. La loi de commande développée permet de réguler à zéro l'erreur relative 3D entre deux positions. La méthode a été validée expérimentalement avec succès en utilisant des images médicales à faible texture et un rapport signal sur bruit défavorable. Les détails de l'implémentation de cette approche peuvent être consultés dans~\citep{tamadazteIcra2016a, tamadazteIros2016a}.\\

Nous avons choisi de ne pas détailler ces méthodes dans ce document pour davantage se consacrer aux récentes approches utilisant les ondelettes et les shearlets dans la boucle de commande. 
%
\section{Transformée en ondelettes}
\begin{figure}[!h]
  \centering
  \includegraphics[width=0.8\columnwidth]{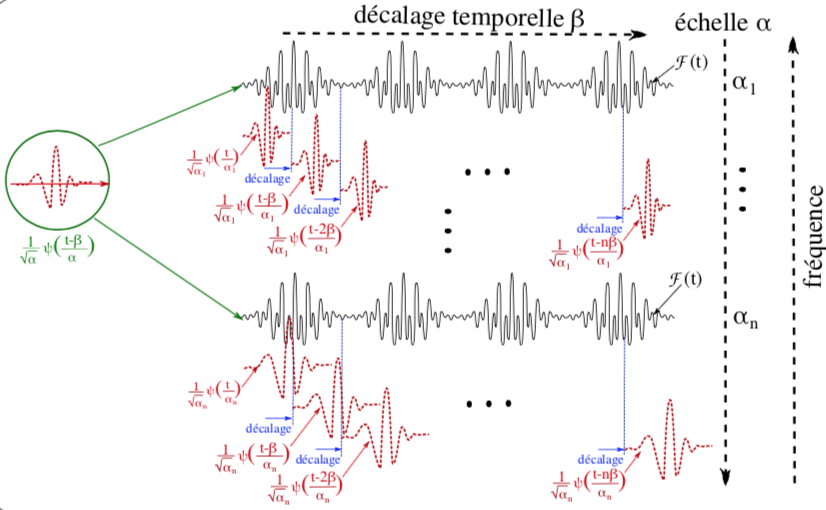}
  \caption{Transformée en ondelettes \citep{Gao2011}.}
  \label{fig.wavelet}
\end{figure}

%
Les ondelettes sont considérées comme l'extension naturelle de la transformée de Fourier standard et à fenêtre glissante. Elles utilisent des fonctions analysantes à la place des fonctions trigonométriques complexes propres à la transformée de Fourier. La particularité des ondelettes est leurs aptitudes à se comprimer et à se dilater pour s'adapter automatiquement aux différentes composantes du signal. En effet, les fenêtres étroites analysent les composantes transitoires à hautes fréquences et celles plus larges analysent les basses fréquences (Fig.~\ref{fig.wavelet}). Ces propriétés font des ondelettes un outil d'analyse multi-échelle efficace considéré comme un \emph{microscope mathématique à zoom variable}.

Par exemple, en imagerie, une transformée en ondelettes peut donc localiser l'information visuelle, de manière simultanée, dans les deux domaines spatial et fréquentiel, ce qui constitue un réel avantage par rapport à la transformée de Fourier. La première définition d'une "\emph{ondelette}" fut donnée par Alfred Haar, en 1910~\citep{haar1910theorie}. Cependant, c'est grâce à un travail novateur, vers la fin des années 1980 et début des années 1990, de pionniers tels que Stéphane Mallat~\citep{Mallat}, Ingrid Daubechies~\citep{daubechies1988orthonormal} et Yves Meyer~\citep{meyer92} qui ont travaillé sur de nouvelles familles d'ondelettes discrètes. Les applications des ondelettes sont ensuite devenues prédominantes dans divers domaines de l'imagerie numérique (compression, filtrage, suivi visuel, etc.) et des mathématiques appliquées, de manière générale. Néanmoins, bien que les ondelettes soient bien adaptées à la description de caractéristiques transitoires, telles que les singularités dans les signaux 1D, elles ne sont pas nécessairement adaptées à la détection des caractéristiques curvilignes que nous trouvons dans les signaux 2D (images)~\citep{donoho2001sparse}. Par conséquent, d'autres méthodes de décomposition multi-échelle utilisant des bases de fonctions projectives anisotropes ont vu le jour comme les curvelets~\citep{candes2000curvelets}, les contourlets~\citep{DoV03} et plus récemment les shearlets~\citep{LabateShearlets}. \\

Malgré l'efficacité démontrée de ces méthodes de décomposition multi-échelle et une double représentation temps-fréquence de l'information visuelle, peu de tentatives existent sur leur utilisation en asservissement visuel. En fait, la littérature rapporte un seul article qui traite de l'utilisation des ondelettes dans une boucle de commande d'asservissement visuel pour le contrôle, en position (sans calcul de la matrice interaction associée), d'un seul degré de liberté. \\

Dans la suite de ce chapitre, nous allons discuter de la conception de plusieurs lois de commande d'asservissement visuel direct à 6 degrés de liberté, dont le signal d'entrée est soit les coefficients d'ondelettes (multi-échelle, sous-échantillonnés ou non), soit ceux de shearlets (sous-échantillonnés ou non). Pour chacune des méthodes, la matrice d'interaction associée est déterminée de manière analytique. Au-delà du travail méthodologique original réalisé pour ces nouvelles méthodes d'asservissement visuel, plusieurs validations numériques et expérimentales (avec différents types d'imageurs : conventionnels et médicaux) ont été mené. 
%
\subsection{Ondelettes : les bases}
%
Tout d'abord, commençons par rappeler quelques bases des ondelettes, nécessaires à comprendre notre démarche pour obtenir l'expression des matrices d'interaction associées et des lois de commande qui en découlent. \\

Une transformée en ondelettes, dans l'espace des fonctions bi-dimensionnelles de carrées sommables $L^2 (\mathbb{R}^2)$, peut être construite en décalant et en redimensionnant des fonctions dites "génératrices". Ces dernières sont définies comme des produits de tenseurs d'une fonction de mise à l'échelle à 1-dimension $\phi \in L^2 (\mathbb{R})$ et d'une ondelette à 1-dimension $\psi \in L^2 (\mathbb{R})$. La fonction de mise à l'échelle $\phi$ est utilisée pour définir une "\emph{approximation grossière}" d'un signal, tandis que la fonction $\psi $, souvent appelée \emph{ondelette mère}, peut être utilisée pour encoder les "\emph{détails}" d'un signal à différentes étapes de sa décomposition. \\

Soit deux opérateurs définis dans $L^2 (\mathbb{R}^2) $, l’opérateur de \emph{dilatation} par :
\begin{equation}
	\DO{2}f(x,y)=2f(2x,2y),
	\label{equ.1Ddilatation}
	\end{equation}
 et celui de la \emph{translation} par :
	\begin{equation}
	\TO{\mbf{m}}f(x,y) = f(x-m_1,y-m_2),
	\label{equ.1Dtranslation}
\end{equation}
où $(x, y)$ représentent les coordonnées métriques d'un point image, $\mathbf{m} = (m_1, m_2)$ un vecteur de translation, et $f$ une fonction définie dans $L^2 (\mathbb{R}^2)$. Le facteur d'échelle dyadique $2$ associé à la décomposition multi-échelle (il sera brièvement décrit plus loin dans cette section), permet une implémentation plus facile des ondelettes discrètes sous-échantillonnées rapides~\citep{beylkin1991fast}.

\begin{figure}[!h]
\centering
\subfigure[ondelette mère de Haar.]{
\includegraphics[width=0.22\columnwidth,height=3cm]{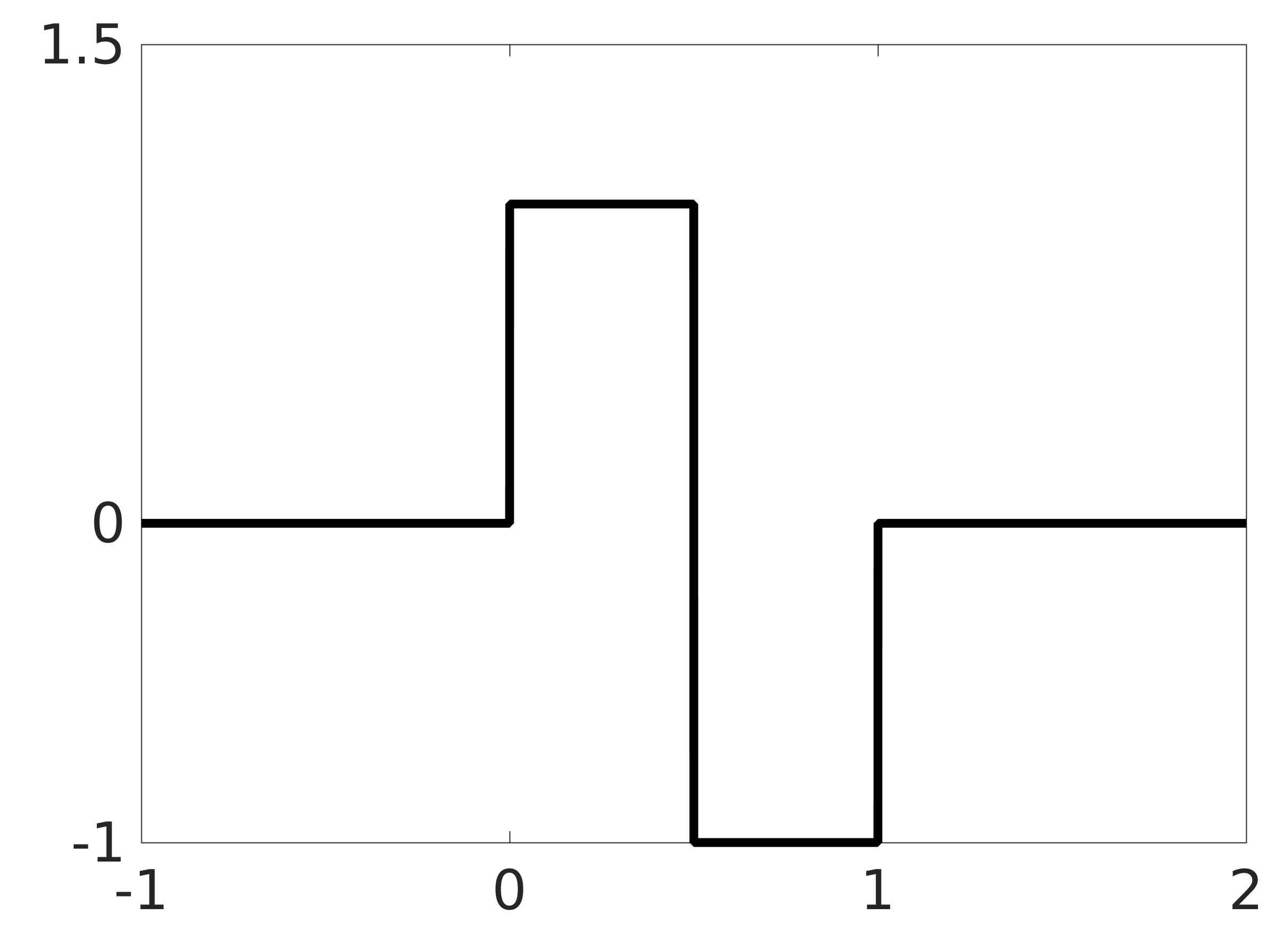}}
\subfigure[ondelette de mise à l'échelle de Haar.]{
\includegraphics[width=0.22\columnwidth,height=3cm]{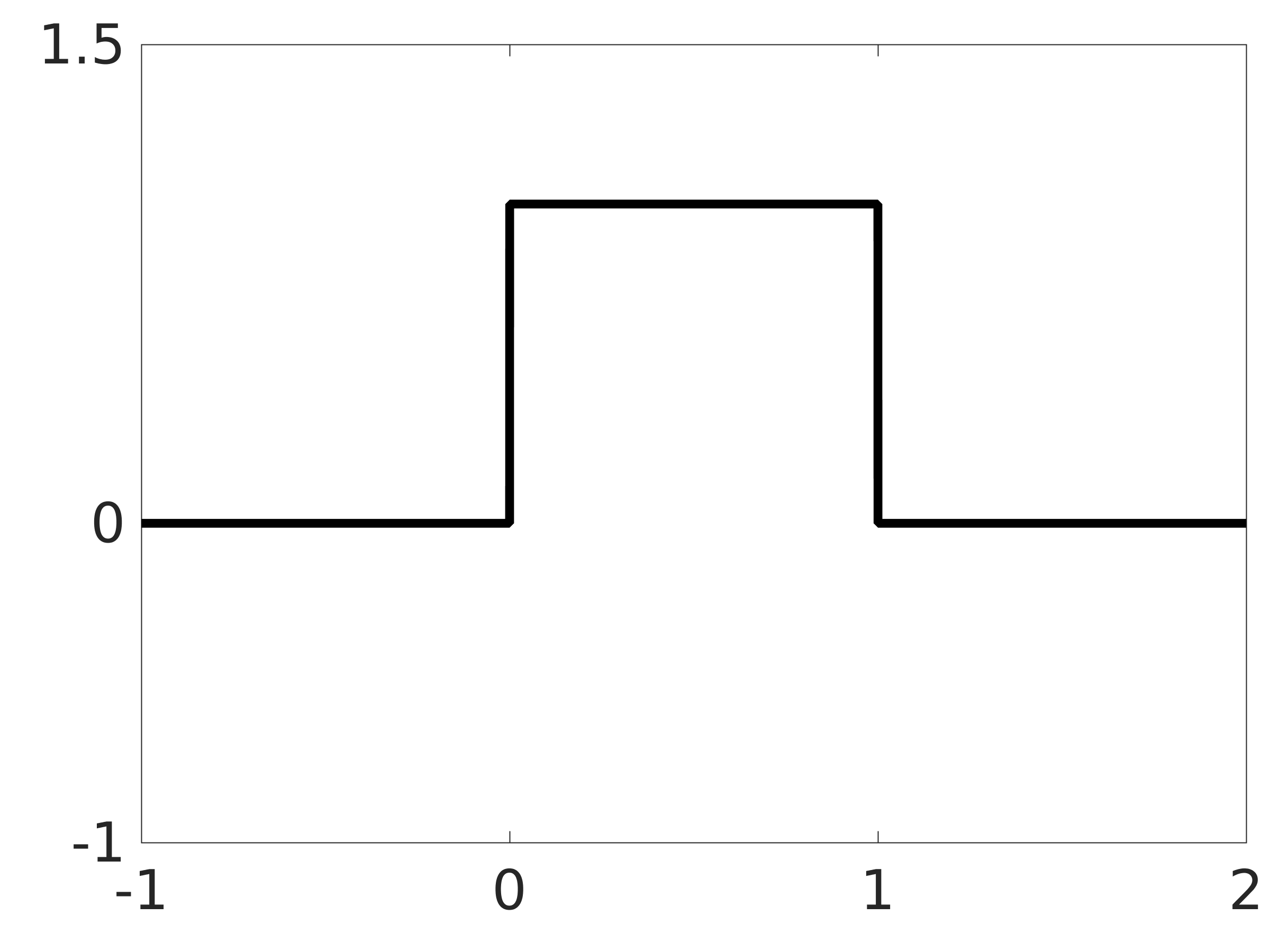}}
\subfigure[ondelette mère de Daubechies.]{
\includegraphics[width=0.22\columnwidth,height=3cm]{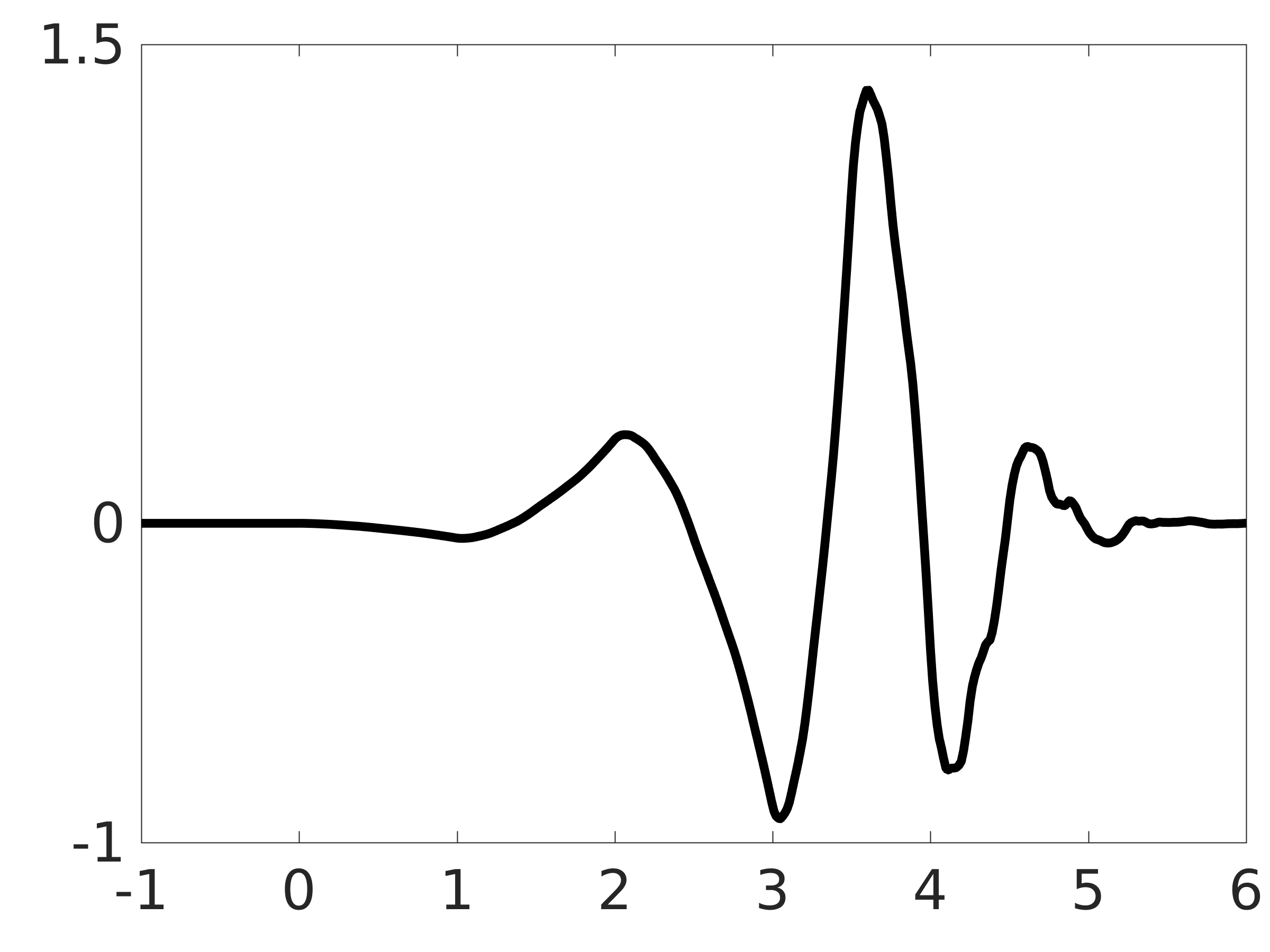}}
\subfigure[ondelette de mise à l'échelle de Daubechies.]{
\includegraphics[width=0.22\columnwidth,height=3cm]{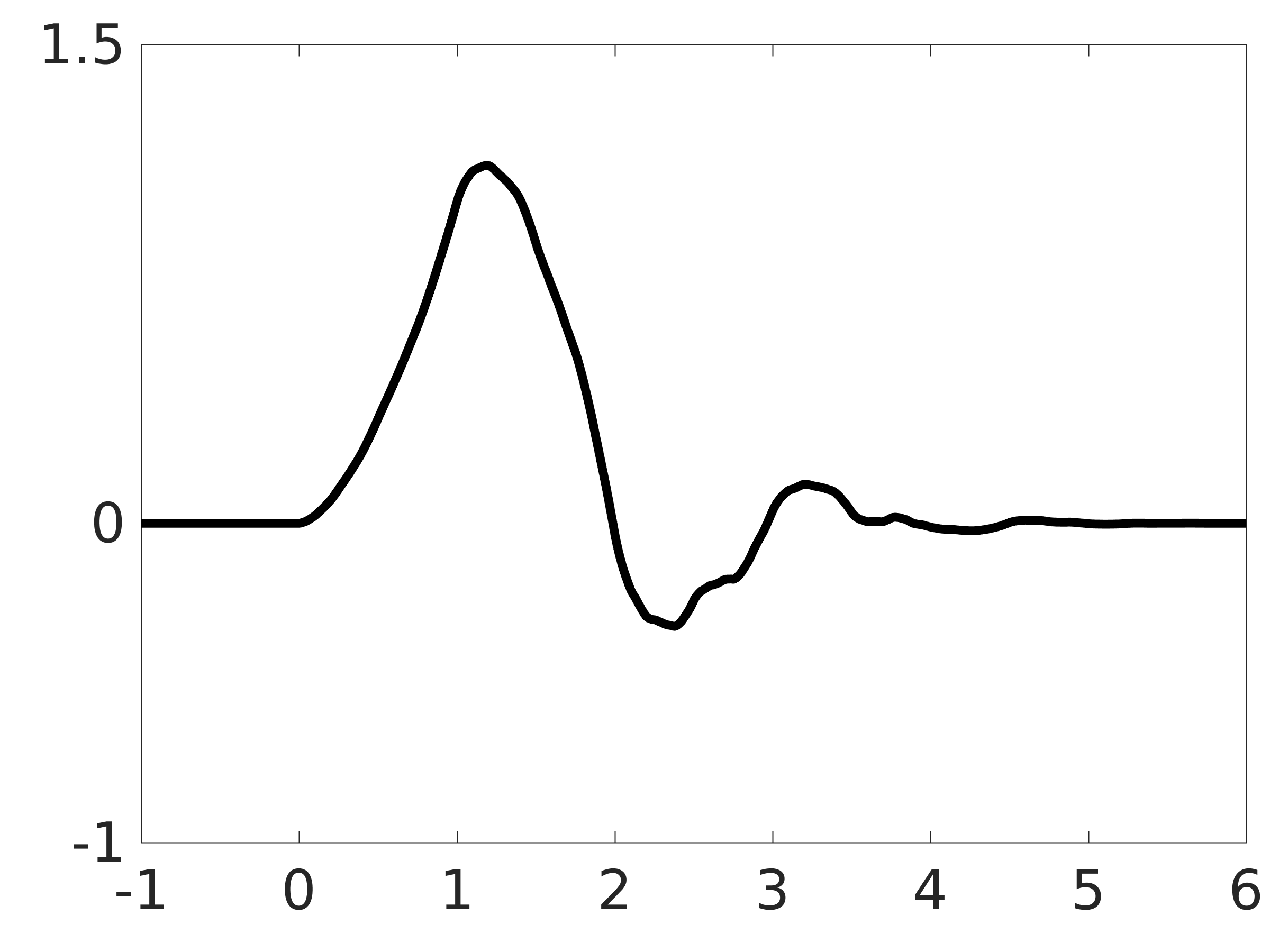}}
\caption{Différentes ondelettes mères (Haar et  Daubechies) et leurs fonctions de mise à l'échelle correspondantes.}
\label{fig.waveletplot}
\end{figure}

Ainsi, un système d'ondelettes 2D peut être construit comme suit :
\begin{equation}
\label{equ.waveletsystem}
\begin{split}
&\left\{\phi^{(1)}_{\mbf{m}} = \TO{\mbf{m}}\phi^{(1)} \colon \mbf{m} \in \mathbb{Z}^2\right\}\;  \cup \\ 
&\left\{\psi^{(l)}_{j,\mbf{m}} = \DO{2}^j\TO{\mbf{m}}\psi^{(l)}\colon j \in \bN_0, \mbf{m} \in \mathbb{Z}^2, l\in\{1,2,3\}\right\},
\end{split}
\end{equation}
où $\phi^{(1)}$ désigne la fonction génératrice d'ondelettes "passe-bas", $\psi^{(1)}$, $\psi^{(2)}$, et $\psi^{(3)}$ représentent, respectivement, les fonctions génératrices de détails verticaux, horizontaux et diagonaux, $j$ est un paramètre d'échelle et $\mbf{m}$ de translation. A noter que $j$ détermine la fréquence à laquelle l'opérateur de dilatation $\DO{2}$ est appliqué à l'ondelette génératrice. D'une manière générale, une grande valeur de $j$ donnera une ondelette extrêmement comprimée, adaptée pour encoder les composantes haute-fréquences d'une image et inversement. L'effet de l'application répétée de l'opérateur $\DO{2}$, dans le cas de filtres à base d'ondelettes, est illustré à la Fig.~\ref{fig.waveletplot}.

\begin{figure}[!h]
  \centering
  \includegraphics[width=0.9\columnwidth]{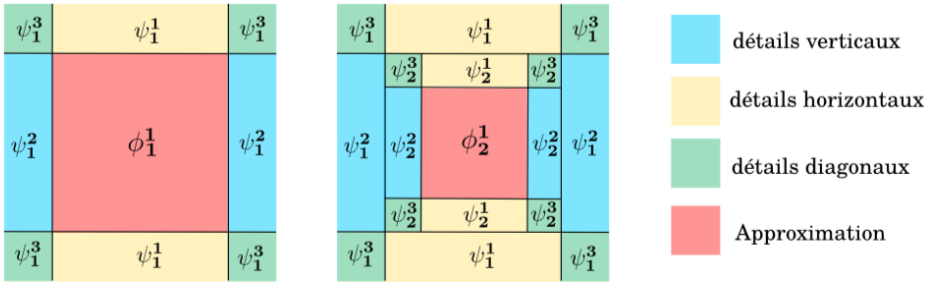}
  \caption{Découpage du domaine fréquentiel en ondelettes 2D.}
  \label{fig.dec_wavelet}
\end{figure}

Les fonctions génératrices 2D s'expriment comme suit :
\begin{equation}
\begin{split}
\phi^{(1)}(x,y) &= \phi(x)\phi(y), \\
\psi^{(1)}(x,y) &= \phi(x)\psi(y), \\
\psi^{(2)}(x,y) &= \psi(x)\phi(y),\\
\psi^{(3)}(x,y) &= \psi(x)\psi(y),
\label{equ.waveletgenerators}
\end{split}
\end{equation}
où $\phi \in L^2 (\mathbb{R})$ est une fonction de mise à l'échelle à 1-dimension et $\psi \in L^2 (\mathbb{R})$ une ondelette mère à 1-dimension. En effet, pour générer des ondelettes $\psi^{(1)}$, $\psi^{(2)}$ et $\psi^{(3)}$ illustrées graphiquement dans (Fig.~\ref{fig.dec_wavelet}), la famille suivante :
\begin{equation}
	\left\{\psi^{(l)}_{j,\mbf{m}} = \DO{2}^{j}\TO{\mbf{m}}\psi^{(l)} \colon j \in \bZ, \mbf{m} \in \mathbb{Z}^2, l\in\{1,2,3\}\right\}
\end{equation}
forme une base orthonormée dans $L^2 (\mathbb{R}^2)$.

Soit une fonction carrée intégrable $f\in L^2(\bR^2)$, sa transformée en ondelettes $\WT{\psi}{f}$, par rapport à une fonction génératrice $\psi \in L^2(\bR^2)$, s'obtient grâce à un produit interne dans $L^2$. Ce produit est donné par :
\begin{equation}
(\WT{\psi}{f})(j,\mbf{m})= \langle f , \psi_{j,\mbf{m}} \rangle = \iint\limits_{\mathbb{R}^2} f(x,y) \DO{2}^{j}\TO{\mbf{m}}\psi(x,y) \, \mathrm{d}x\mathrm{d}y,
\label{eq.ondelettes}
\end{equation}
où $j \in \mathbb{Z}$ définit l'échelle et $\mbf{m} \in \mathbb{Z}^2$ la localisation de l'ondelette $\psi_{j, m}$. Noter que compresser et dilater la génératrice $\psi$ en appliquant l'opérateur de dimensionnement dyadique $\DO{2}^{j}$ encode simultanément la fréquence et le niveau de localisation de l'information visuelle : une caractéristique très intéressante.\\

Les bases des ondelettes rappelées ci-dessus appartiennent à la famille des ondelettes continues. Néanmoins, pour une implémentation réelle, notamment en asservissement visuel, où les entrées de la boucle de commande sont discrètes et finies, nous préférons utiliser les ondelettes discrètes. Ce sont Stéphane Mallat~\citep{Mallat} et Yves Meyer~\citep{meyer92} qui ont introduit la théorie de la décomposition multi-échelle par ondelettes discrètes. Cette décomposition est fondée sur l'utilisation d'une série de filtres discrets. Ainsi, dans le cas des ondelettes discrètes, la fonction de mise à l’échelle $\phi \in L^2 (\mathbb{R}) $ et l'ondelette mère $\psi \in L^2 (\mathbb{R})$ sont choisies pour satisfaire les relations suivantes :
\begin{eqnarray}
\label{equ.scalingrelations1}
\phi(x) & = & \sqrt{2}\sum_{n\in\bZ}{h}_n\phi(2x - n),\\ 
\psi(x) & = & \sqrt{2}\sum_{n\in\bZ}{g}_n\phi(2x - n),
\label{equ.scalingrelations}
\end{eqnarray}
avec $\mbf{h}\in\ell^2(\mathbb{Z})$ un filtre passe-bas discret et $\mbf{g}\in\ell^2(\mathbb{Z})$ un filtre passe-haut discret.

En considérant une image discrète $\mathbf{I} \in \ell^2 (\mathbb{Z}^2)$ comme approximation grossière d'une fonction carrée intégrable $I$ en fonction $\phi^{(1)}$ à une échelle donnée $j \in(\mathbb{Z})$, c'est-à-dire :
\begin{equation}
I(x,y) \approx \sum_{\mbf{n}\in\bZ^2} {I}_{n_1,n_2} 2^j\phi^{(1)}(2^jx - n_1,2^jy - n_2).
\label{equ.mraassumption}
\end{equation}

Au final, les coefficients des ondelettes multi-échelles sont obtenues en réalisant successivement la convolution entre l'image discrète $\mathbf{I}$ et les filtres $\mbf{h}$ et $\mbf{g}$.\\

Nous avons développé deux approches d'asservissement visuel direct, dont les informations visuelles considérées sont les coefficients d'ondelettes.  
\subsection{Matrice d'interaction analytique : ondelettes (première approche)}
%
\def\la{\left\langle\rule{0pt}{1em}}
\def\ra{\right\rangle}

Nous nous sommes, d'abord, intéressés à la manière de relier la dérivée temporelle des coefficients des ondelettes au mouvement d'une caméra. Les coefficients d'ondelettes considérés, dans cette section, sont obtenus par la transformée en ondelettes discrète multi-échelle introduite dans~\citep{Mallat}. Nous avons alors tenté la dérivation d'une matrice d'interaction multi-échelle associée aux coefficients multi-échelles. Autrement dit, la matrice d'interaction, et de fait la loi de commande sont également multi-échelles. \\

Considérons une décomposition en ondelettes multi-résolution définie par (\ref{eq.ondelettes}). En discret, la double intégrale, peut être remplacée par un produit interne entre l'ondelette mère $\psi \in L^2 (\mathbb{R})$ et une fonction de mise à l'échelle $\phi \in L^2 (\mathbb{R})$. Aussi, dans~\citep{Mallat}, cette discrétisation peut s'effectuer par les convolutions successives entre l'image et des filtres passe-bas $\mbf{h}$ et passe-haut $\mbf{g}$.  \\

Le vecteur des informations visuelles, que nous considérons dans ces travaux, est constitué de l'image initiale décomposée à plusieurs échelles $j$, ainsi que les coefficients d'ondelettes verticaux, horizontaux et diagonaux accessibles également à plusieurs échelles $j$. Ce vecteur noté $\mbf{s}_\texttt{wm}$ est donné par :
\begin{equation}
\mbf{s}_\texttt{wm} =   \Big \langle \mbf I_{2^{-j}}(x,y), \tau^k(x,y)  \Big \rangle, \label{eq.innerProIma}
\end{equation}
où $\mbf I_{2^{-j}}$ représente les différentes sous-résolutions de l'image en fonction de $j \in \mathbb{Z}$ et $\tau^k$  ($k = [v, h, d]$, regroupe  les coefficients des ondelettes verticaux ($v$), horizontaux ($h$) et diagonaux ($d$) pour chaque résolution $j$ et $k = 0$ représente l'image à la résolution inférieure. \\

Pour estimer la matrice d'interaction, il est nécessaire d'introduire le concept de la contrainte du flux optique multi-échelle formalisé dans le cas des ondelettes dans~\citep{bernard1999wavelets}. Toujours en se basant sur les travaux sur la contrainte du flux optique appliquée aux ondelettes, nous pouvons écrire, pour chaque point $\pnt p = (x,y)^\top$, la relation suivante  :

\begin{equation}
\Bigg \langle  \mathbf{I}_{2^{-(j+1)}}(x,y) , \frac{\partial {\tau^h }(x,y)}{\partial x} \Bigg \rangle  \dot{x}+ \Bigg \langle
\mathbf{I}_{2^{-(j+1)}}(x,y),\frac{\partial \tau^v (x,y) }{\partial y}  \Bigg \rangle  \dot{y}  + \frac{\partial }{\partial t} \Bigg \langle \mathbf{I}_{2^{-(j+1)}}(x,y) ,\tau^0 (x,y) \Bigg \rangle  =  0.
\label{eq.OFCE_inner}
\end{equation}

Pour simplifier cette modélisation, nous avons délibérément omis les coefficients diagonaux qui n'apportent pas davantage à la loi de commande à ce stade. 

En introduisant les notations suivantes :
\begin{eqnarray}
{\mathbf{c}^h_{2^{-(j+1)}}(x,y)} &{\triangleq} & {\Bigg \langle  \mathbf{I}_{2^{-j}} (x,y), \frac{\partial {\tau^h }}{\partial x} (x,y)\Bigg \rangle, \label{eq.gradH}}\\
{  \mathbf{c}^v_{2^{-(j+1)}}(x,y)} &{\triangleq} & {\Bigg \langle  \mathbf{I}_{2^{-j}} (x,y), \frac{\partial {\tau^v}}{\partial y}(x,y) \Bigg \rangle, \label{eq.gradV}}
\end{eqnarray}
et en reformulant (\ref{eq.OFCE_inner}), on obtient :
\begin{eqnarray}
{\mathbf{c}^h_{2^{-(j+1)}}(x,y) ~\dot{x} +
\mathbf{c}^v_{2^{-(j+1)}}(x,y) ~\dot{y} + 
\frac{\partial }{\partial t} \mathbf{I}_{2^{-(j+1)}}(x,y) =  0.}
\label{eq.OFCE_simp}
\end{eqnarray}

Équivalent, en écriture matricielle, à :
\begin{eqnarray}
\frac{d }{d t} \mathbf{I}_{2^{-(j+1)}}(x,y) =
 \Big( \mathbf{c}^h_{2^{-(j+1)}}(x,y) ~~~ \mathbf{c}^v_{2^{-(j+1)}}(x,y) \Big)\binom{\dot{x}}{\dot{y}}.
\label{eq.mat_form_OFCE}
\end{eqnarray}

Par ailleurs, il est possible d'exprimer $\dot{x}$ et $\dot{y}$ en fonction du tenseur de vitesse de la caméra $\tc$ grâce à la matrice d'interaction $\mLp(x,y)$ associée à un point image $\pnt p$ \citep{chaumette92}. Par conséquent, 
\begin{eqnarray}
\binom{\dot{x}}{\dot{y}}=  \mLp(x,y) \tc
 \label{eq.dotxdoty}
\end{eqnarray}

Après quelques manipulations mathématiques (le détail des calculs peut être consulté dans \citep{tamadazteIcra2016b}) et en introduisant (\ref{eq.dotxdoty}) dans~(\ref{eq.mat_form_OFCE}), nous obtenons finalement :
\begin{eqnarray}
{\frac{\partial }{\partial t} \mathbf{I}_{2^{-(j+1)}} (x,y) =
- \Big( \mathbf{c}^h_{2^{-(j+1)}}(x,y) ~~ \mathbf{c}^v_{2^{-(j+1)}}(x,y) \Big) \mLp(x,y)~\tc }
\label{eq.mat_form_OFCE_2}
\end{eqnarray}

La matrice d'interaction $\mathbf{L}_{w_{2^{-(j+1)}}}$, en un point $\pnt p(x,y)^\top$, définie dans le support des ondelettes incluant à une échelle donnée $j$ : l'intensité des pixels de l'image, les coefficients verticaux et horizontaux des ondelettes peut s'écrire :
\begin{eqnarray}
\mathbf{L}_{w_{2^{-(j+1)}}}(x,y)  =
- \Bigg(\mathbf{c}^h_{2^{-(j+1)}}(x,y) ~~ \mathbf{c}^v_{2^{-(j+1)}} (x,y) \Bigg) \mLp(x,y).
\label{eq.mat_form_OFCE_22}
\end{eqnarray}

Pour obtenir la matrice d'interaction globale (pour tous les coefficients et pixels considérés), il suffit de concaténer verticalement toutes les matrices d'interactions individuelles de la manière suivante : 
\begin{eqnarray}
\mbf L_{\texttt{swm}}(x,y) = \begin{bmatrix}
\mathbf{L}_{w_{2^{-(j+1)}}(1,1)}\\ 
\mathbf{L}_{w_{2^{-(j+1)}}(1,2)}\\ 
\vdots \\ 
\mathbf{L}_{w_{2^{-(j+1)}}(M,N)}
\end{bmatrix},
\label{eq.feature_Intensities}
\end{eqnarray}
où $M \times N$ est la taille des coefficients des ondelettes à l'échelle $j$. 
%
\subsection{Loi de commande : ondelettes (première approche)}
%
A présent nous disposons d'une matrice d'interaction multi-échelle, c'est-à-dire il suffit de changer le paramètre d'échelle $j$ pour obtenir une nouvelle matrice sans calcul supplémentaire. Le paramètre $j$ peut être changé en fonction de la norme du vecteur de l'erreur cartésienne entre les positions courante et désirée.  Par exemple, si l'erreur est grande, il est conseillé d'utiliser une échelle faible ($j$ = 3 ou 4) pour une meilleure dynamique du contrôleur et utiliser $j = 1$ à l'approche de la position désirée pour une meilleure précision. 

La loi de commande, qui en découle, doit annuler le vecteur erreur $\mbf e_{\texttt{wm}}(t)$ pour converger vers zéro :

\begin{equation}
\mbf e_{\texttt{wm}}(t)= \mbf s_{\texttt{wm}}(t)  - \mbf s_{\texttt{wm}}^*. 
\label{equ.errorwv}
\end{equation}

Enfin, le tenseur de vitesse $\tc$ de la caméra à l'instant $t$ est obtenu par 
\begin{equation}
  \tc(t) = -\lambda \Big(\mathbf{L}_\texttt{swm}\Big)^+ \mbf{e}_\texttt{sw}(t),
 \label{eq.gauss}
\end{equation}
où, $\lambda > 0$ est un gain de commande et $\Big(\mathbf{L}_\texttt{swm}\Big)^+$ représente la pseudo-inverse \emph{Moore-Penrose} de $\mathbf{L}_\texttt{swm}$.

Par ailleurs, pour augmenter le domaine de convergence de la loi de commande, il est conseillé~\citep{collewet2011photometric} de remplacer la méthode d'optimisation \emph{Gauss-Newton} utilisée ci-dessus par une méthode de \emph{Levenberg-Marquardt}. Ainsi, l'expression du contrôleur \eqref{eq.gauss} devient 
\begin{equation}
\tc(t) = -\lambda \Big(\mbf{H}(t) + \mu \operatorname{diag}\big(\mbf{H}(t)\big)\Big)^{-1}\transp{\Big(\mathbf{L}_\texttt{swm}(t)\Big)} \mbf{e}_\texttt{wm}(t),
\label{equ.controller.levenberg}
\end{equation}
avec $\mathbf{H}(t)=\transp{\big(\mathbf{L}_\texttt{swm}(t)\big)}\mathbf{L}_\texttt{swm}(t)$ est la matrice Hessienne associée à  $\msWv$. Ainsi, une grande valeur de $\mu$ ($\mu = 1$) correspond à une optimisation de type descente du gradient, tandis qu'une valeur faible ($\mu = 10^{-3}$) revient à utiliser la méthode de \emph{Gauss-Newton}.
%
\subsection{Validations}
\begin{figure}[!h]
\centering
\includegraphics[scale=0.7]{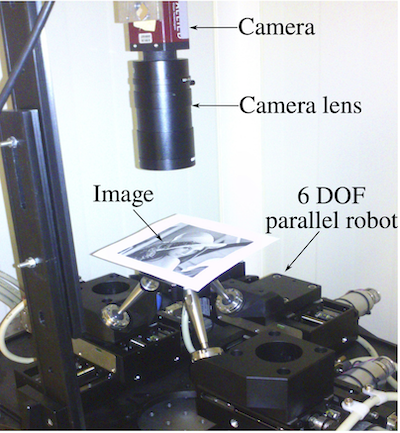} \quad
\includegraphics[scale=0.35]{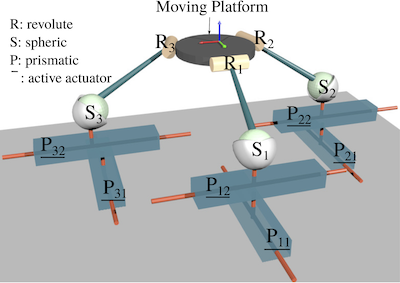}
\caption{Illustration de la plateforme microrobotique parallèle à 6 degrés de liberté.}
\label{fig.setup_as2m}
\end{figure}

Cette méthode a été validée en simulation mais aussi dans des conditions expérimentales en utilisant une plateforme microrobotique constituée d'un robot parallèle à 6 degrés de liberté et d'une caméra CCD placée dans une configuration \emph{eye-to-hand} (Fig.~\ref{fig.setup_as2m}). Dans la suite, nous allons discuter certains des résultats obtenus. 

\subsubsection{Validation numérique}
\begin{figure}[!h]
  \centering
  \includegraphics[width=\columnwidth]{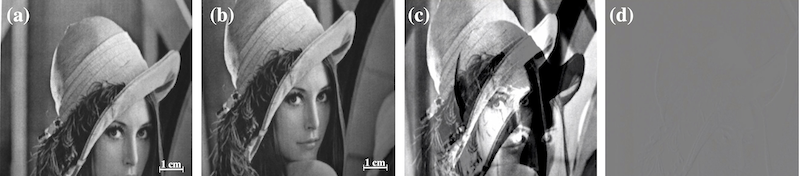}
  \caption{Tâche de positionnement par asservissement visuel par ondelettes.  (a) image initiale $\mbf I$ , (b) image désirée $\mbf I^*$  , (c) image de différence ($\mbf I - \mbf I^*$) initiale, (d) ($\mbf I - \mbf I^*$) finale obtenue à convergence.}
  \label{fig.seq_simu_wav_nom}
\end{figure}

%
Nous avons développé un simulateur en C++ en s'appuyant sur la plateforme logicielle ViSP\footnote{Visual Servoing Platform (https://visp.inria.fr).}~\citep{visp2005} dédiée à la vision par ordinateur et à l'asservissement visuel. La méthode des ondelettes a été validée dans des conditions à la fois favorables et défavorables. La Fig.~\ref{fig.seq_simu_wav_nom} montre la réalisation d'une tâche de positionnement définie dans $SE(3)$. Il est possible de constater que l'image de différence $\mbf I - \mbf I^*$ (Fig.~\ref{fig.seq_simu_wav_nom} d), la méthode converge avec précision vers la position désirée. 

%
\subsubsection{Validation expérimentale}
\begin{figure}[!h]
  \centering
  \includegraphics[width=\columnwidth]{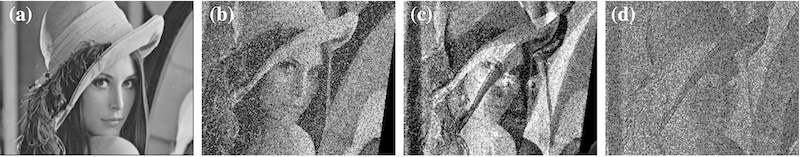}
  \caption{Tâche de positionnement par asservissement visuel par ondelettes dans le cas d'images bruitées.  (a) image initiale $\mbf I$, (b) image désirée bruitée $\mbf I^*$, (c) image de différence ($\mbf I - \mbf I^*$) initiale, (d) ($\mbf I - \mbf I^*$) finale obtenue à convergence.}
  \label{fig.seq_exp_wav_noise}
\end{figure}

%
Dans \citep{tamadazteIcra2016b}, nous avons testé les performances du contrôleur dans diverses conditions de travail défavorables (occultations partielles, présence de bruit de type "\emph{sel et poivre}", éclairage instable, erreurs d'étalonnage de la caméra ou de la position de celle-ci vis-à-vis du robot, etc.). Des tests de comparaison avec la méthode photométrique et une méthode de type d'asservissement visuel géométrique ont été également réalisés.  La Fig.~\ref{fig.seq_exp_wav_noise} montre une séquence d'images réalisée sur la plateforme microrobotique (Fig.~\ref{fig.setup_as2m}). Sur l'image désirée, nous avons ajouté un bruit pour tester la robustesse de la commande et la comparer à la méthode photométrique. 
 
\begin{figure}[!h]
  \centering
  \includegraphics[width=.8\columnwidth]{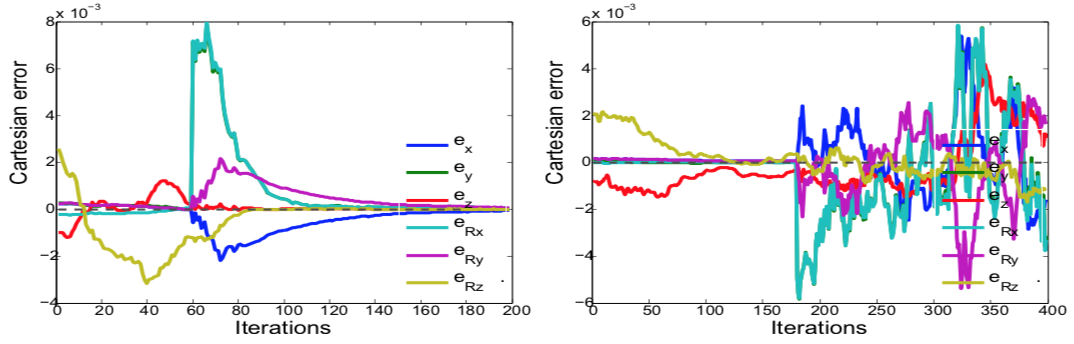}
  \caption{Régulation de l'erreur à zéro, à gauche la méthode par ondelettes, à droite celle avec la photométrie.}
  \label{fig.seq_exp_wav_noise_err}
\end{figure}

Comme le montre la décroissance de l'erreur (sur chaque axe) entre les images courante et désirée, la méthode par ondelettes converge vers la position finale (Fig.~\ref{fig.seq_exp_wav_noise_err}) (droite), tandis que la méthode photométrique échoue à cause de la présence du bruit (Fig.~\ref{fig.seq_exp_wav_noise_err}) (gauche). 

Comme le montre les résultats des validations numérique et expérimentale, la loi de commande fondée sur l'utilisation des coefficients d'ondelettes présente des performances en précision et en robustesse intéressants (pour davantage de résultats expérimentaux, le lecteur est invité à consulter nos travaux dans \citep{OurakTMECH2019}. \\

Dans la suite, nous allons discuter d'une autre approche pour la représentation analytique de la matrice d'interaction qui peut être généralisée à d'autres types d'informations visuelles issues de différentes décompositions mutli-échelles comme les shearlets. 
%
\subsection{Matrice d'interaction analytique : ondelettes (deuxième approche)}
%
La conception d'une loi de commande d'asservissement visuel direct, dont les informations visuelles sont les coefficients d'ondelettes, passe d'abord par le calcul de la matrice d'interaction, notée $\mathbf{L}_\texttt{sw}$, associée à ces coefficients. Autrement dit, il nous faut trouver la relation qui relie les mouvements de la caméra aux variations temporelles des coefficients des ondelettes.  \\

Pour ce faire, commençons par considérer un ensemble fini dans $L \in \mathbb{N}$ générant des ondelettes $(\psi^{(l)})_ {l\leq L} \subset L^2 (\mathbb{R}^2)$, qui consiste typiquement en un ensemble de quatre fonctions génératrices séparables construites à partir de fonctions d'ondelettes 1-dimension et de fonctions de  mise à l'échelle (\ref{equ.waveletgenerators}). Par conséquent, le vecteur des caractéristiques visuelles (coefficients d'ondelettes) pour un ensemble de $N \in \mathbb{N}$ triplets, c'est-à-dire $(l_n, j_n, \mbf{m}_n)_{n \leq N} \subset \{1, \ldots, L \} \times \mathbb{N} \times \mathbb{Z}^ 2$, où $l_n$ désigne une génératrice d'ondelettes, $j_n$ représente l'échelle et $\mbf{m}_n$ la translation évaluée à un instant $t$. Ce vecteur de coefficients d'ondelettes, noté $\msWv$, est donné par 
\begin{equation}
\msWv(t) = \transp{\left((\WT{\psi^{(l_1)}}{I_t})(j_1,\mbf{m}_1), (\WT{\psi^{(l_2)}}{I_t})(j_2,\mbf{m}_2), \ldots,\, (\WT{\psi^{(l_N)}}{I_t})(j_N,\mbf{m}_N)\right)}.
\label{equ.mswv}
\end{equation}

Le but, à présent, est de calculer une matrice d'interaction $\mathbf{L}_\texttt{sw} \in \mathbb{R}^{N \times 6}$ reliant le mouvement de la caméra défini par le vecteur de vitesse $\tc = (v_x, v_y, v_z, \omega_x, \omega_y, \omega_z)^\top$ à la dérivée temporelle du vecteur des coefficients d'ondelettes $\msWv(t)$ :
\begin{equation}
\frac{\mathrm{d}\msWv(t)}{\mathrm{d}t} = \mathbf{L}_\texttt{sw}(t) \tc(t).
\label{equ.dmswv}
\end{equation}

Supposons que toutes les ondelettes génératrices ainsi que les intensités élémentaires de l'image $I$ sont continuellement différentiables. Grâce à la généralisation en 2-dimensions de la règle de \emph{Leibniz} sur les intégrales (appelée également théorème de transport de \emph{Reynolds} \citep{flanders1973differentiation}), nous pouvons exprimer la dérivée temporelle de $\msWv$, en un point défini en $(l, j, \mbf{m})$, par :
\begin{eqnarray}
\nonumber
\frac{\mathrm{d}(\WT{\psi^{(l)}}{I_t})(j,\mbf{m})}{\mathrm{d}t} & = &\frac{\mathrm{d}\ip{{I_t}}{\psi^{(l)}_{j,\mbf{m}}}}{\mathrm{d}t}\\
& = & \iint\limits_{\bR^2} \frac{\partial I(x,y,t)}{\partial t} \psi^{(l)}_{j,\mbf{m}}(x,y)\, \mathrm{d}x\mathrm{d}y.
\label{equ.dmswv1}
\end{eqnarray}

En s'inspirant de méthodologie de la composition de la matrice d'interaction de la méthode photométrique \citep{collewet2011photometric}, nous pouvons écrire :
\begin{equation}
\frac{\mathrm{d}\ip{{I_t}}{\psi^{(l)}_{j,\mbf{m}}}}{\mathrm{d}t}= -\iint\limits_{\bR^2} \transp{\big(\nabla I_t(x,y)\big)}\mLp(x,y) \tc(t) \psi^{(l)}_{j,\mbf{m}}(x,y)\, \mathrm{d}x\mathrm{d}y.
\label{equ.dmswv2}
\end{equation}
où $\mLp(x,y)$ est la matrice d'interaction associée à un point 2D dans l'image \citep{chaumette92}. 

Dans un soucis de simplification de l'écriture, nous écrivons désormais :
\begin{equation}
 {I}^{(i)}_t = \transp{\big(\nabla I_t(x,y)\big)}\mLp(x,y)\mbf{e}_i,
 \label{equ.dmswv3}
\end{equation}
où $i \in \{1, \ldots, 6 \}$ et $\mbf {e}_i \in \bR^6$ représente le $i-eme$ vecteur unitaire canonique (à noter que $\mLp(x, y) \mbf{e}_i$ est la $i-eme$ colonne de la matrice $\mLp(x, y)$). En d'autres termes, ${I}^{(i)}_t$ mesure l'effet du gradient de l'image avec le mouvement généré par le $i-\'e me$ degré de liberté de la caméra. Grâce à (\ref{equ.dmswv3}), la matrice d'interaction  $\mathbf{L}_\texttt{sw}$ associée à $\msWv$ peut s'écrire comme suit :

\begin{equation}
\mathbf{L}_\texttt{sw}(t)  = - \begin{bmatrix} (\WT{\psi^{(l_1)}}{{I}^{(1)}_t})(j_1,\mbf{m}_1) & \cdots & (\WT{\psi^{(l_1)}}{{I}^{(6)}_t})(j_1,\mbf{m}_1)\\\vdots&\ddots&\vdots\\(\WT{\psi^{(l_N)}}{{I}^{(1)}_t})(j_N,\mbf{m}_N) & \cdots & (\WT{\psi^{(l_N)}}{{I}^{(6)}_t})(j_N,\mbf{m}_N)\end{bmatrix} \in \bR^{N\times 6}.
\label{equ.interactionmatrixwv1} 
\end{equation}

A noter que chaque colonne de $\mathbf{L}_\texttt{sw}$ représente la transformation en ondelettes d'une image ${I}^{(i)}_t $ par rapport à un système d'ondelettes $(\psi^{(l)})_{l \leq L}$ et à la position $(l_n, j_n, \mbf{m}_n)_ {n \leq N}$.\\

La matrice d'interaction $\mathbf{L}_\texttt{sw}$ est construite en utilisant les coefficients d'ondelettes non-sous-échantillonnés, pour utiliser moins de coefficients d'ondelettes dans le vecteur d'informations visuelles, et donc moins de transformations d'ondelettes. Pour ce faire nous avons travaillé sur une nouvelle forme de matrice d'interaction plus générique qui utilise uniquement deux transformées. 

Par conséquent, supposons que les fonctions génératrices $\psi^{(l)}$ sont compactes spatialement, il en résulte alors que les vitesses  $\left(\frac{dx}{dt},\frac{dy}{dt}\right)$ sont constantes (équation du flux optique) sur  $\psi^{(l)}$ \citep{bernard1999wavelets}. Cette supposition permet d'approximer la matrice d'interaction d'un point de l'image $\mLp(x, y)$ par une matrice $\mLpw(j,\mbf{m})$, qui dépend uniquement du centre du support des ondelettes $\psi^{(l)}_{j,\mbf{m}}$. Nous pouvons écrire alors 
\begin{equation}
\label{eq:Lw}
\mLpw(j,\mbf{m})  = \mLp(2^{-j}m_1,2^{-j}m_2).
\end{equation}

En s'appuyant sur \eqref{eq:Lw}, une approximation $\tilde{\mathbf{L}}_\texttt{sw}(t) \approx \mathbf{L}_\texttt{sw}(t)$ de la matrice d'interaction, fondée sur les ondelettes à un instant $t$, nécessitant que le calcul de deux transformées en ondelettes discrètes, est donnée par :
\begin{equation}
\tilde{\mathbf{L}}_\texttt{sw}(t)  = -\begin{bmatrix}\left((\WT{\psi^{(l_1)}}{\frac{\partial {I_t}}{\partial x}})(j_1,\mbf{m}_1),(\WT{\psi^{(l_1)}}{\frac{\partial {I_t}}{\partial y}})(j_1,\mbf{m}_1)\right)\mLpw(j_1,\mbf{m}_1) \\
 					\vdots \\
 					\left((\WT{\psi^{(l_N)}}{\frac{\partial {I_t}}{\partial x}})(j_N,\mbf{m}_N),(\WT{\psi^{(l_N)}}{\frac{\partial {I_t}}{\partial y}})(j_N,\mbf{m}_N)\right)\mLpw(j_N,\mbf{m}_N) \end{bmatrix} \in \bR^{N\times 6}.
\label{equ.interactionmatrixwv2} 
\end{equation}
%
\subsection{Loi de commande : ondelettes (deuxième approche)}
%
La formulation du contrôleur est similaire à celle des ondelettes ou encore de la méthode photométrique. L'erreur visuelle à éliminer par le contrôleur est définie par :
\begin{equation}
 \mbf{e}_\texttt{w}(t) = \msWv(t)  - \msWv^*. 
\label{equ.errorwv}
\end{equation}

L'erreur $\mbf{e}_\texttt{w}$ entre la position courante et celle désirée est annulée au fur à mesure pour converger vers zéro par la loi de commande suivante :
\begin{equation}
\tc(t) = -\lambda \Big(\mbf{H}(t) + \mu \operatorname{diag}\big(\mbf{H}(t)\big)\Big)^{-1}\transp{\Big(\mathbf{L}_\texttt{sw}(t)\Big)} \mbf{e}_\texttt{w}(t).
\label{equ.controller.levenberg}
\end{equation}
%
\subsection{Validations}
\begin{figure}[!h]
  \centering
  \includegraphics[width=0.8\columnwidth, height = 4.5cm]{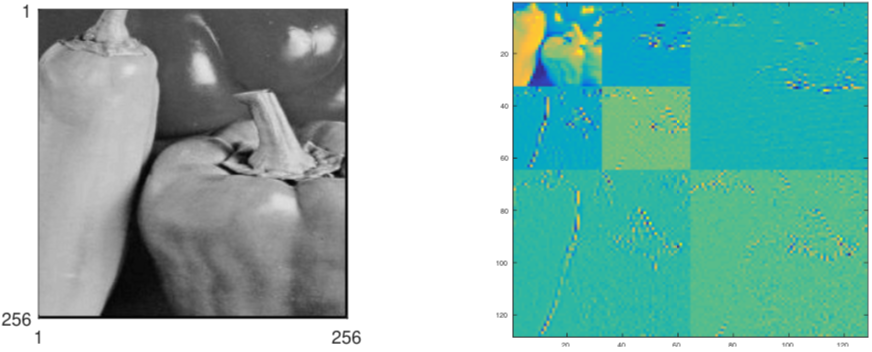}
  \caption{Coefficients d'ondelettes sous-échantillonnés $\mbf s_\texttt{sw}$ :  à gauche, l'image initiale et à droite, les coefficients d'ondelettes non-sous-échantillonnés ($\mbf s_\texttt{sw}$) obtenus.}
  \label{fig.sw_e}
\end{figure}

\begin{figure}[!h]
  \centering
  \includegraphics[width=\columnwidth]{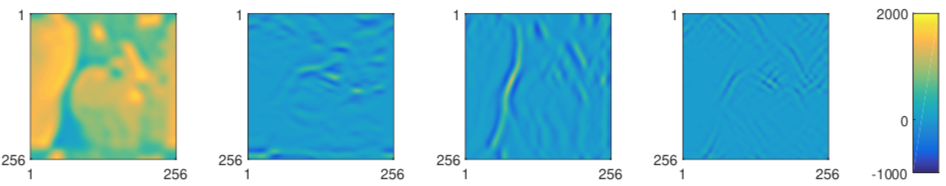}
  \caption{Coefficients d'ondelettes sous-échantillonnés.}
  \label{fig.sw_ne}
\end{figure}

%
Les deux méthodes d'asservissement visuel, discutées ci-dessus, ont été validées en simulation et dans des conditions expérimentales (sur un robot cartésien industriel et dans des conditions cliniques). Les deux méthodes diffèrent par le fait que dans un cas, les coefficients d'ondelettes modélisés sont échantillonnés ou sous-échantillonnés et en tenant compte de la contrainte du flux optique ou non dans l'estimation du mouvement de la caméra. La Fig.~\ref{fig.sw_e} donne un aperçu visuel de coefficients d'ondelettes pris en compte dans le calcul de la matrice d'interaction $\mathbf{L}_\texttt{sw}$ (coefficients non-sous-échantillonnés), quant à la Fig~\ref{fig.sw_ne} ceux utilisés dans la matrice d'interaction $\tilde{\mathbf{L}}_\texttt{sw}$ (coefficients sous-échantillonnés). 

%
\subsubsection{Validation numérique}

\begin{figure}[!h]
	\centering
	\subfigure[]{\includegraphics[width=0.30\columnwidth]{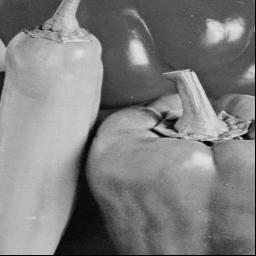}}
        \subfigure[]{ \includegraphics[width=0.30\columnwidth]{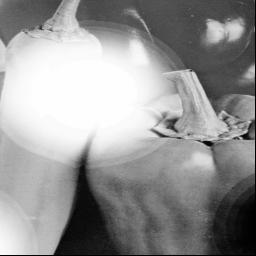}}
	\subfigure[]{\includegraphics[width=0.30\columnwidth]{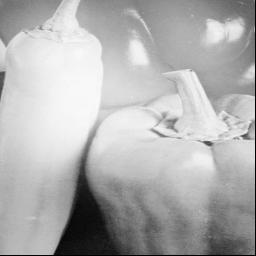}}
	\caption{(a) position désirée prise dans conditions nominales, (b) avec des occultations partielles, et (c) avec des variations d'éclairage.}
	\label{fig.simu_w}
\end{figure}

Afin d'analyser les performances des lois de commande présentées dans la section précédente, nous avons réalisé plusieurs simulations sous différentes conditions de travail. Chaque loi de commande a été testée dans des conditions nominales (Fig.~\ref{fig.simu_w}(a)) ou défavorables : occultations partielles (Fig.~\ref{fig.simu_w}(b)) et variations d'illumination (Fig.~\ref{fig.simu_w}(c)). \\

Les méthodes fondées sur les ondelettes ont été également comparées à la méthode photométrique~\citep{collewet2011photometric} considérée comme une référence en asservissement visuel direct. Dans les conditions nominales, les trois méthodes fonctionnement parfaitement avec une erreur moyenne de positionnement de l'ordre de $10^{-4}$mm. Néanmoins, lorsque les conditions sont défavorables, les ondelettes se montrent plus précises et plus robustes que la photométrie. Ceci est notamment est dû à la capacité de la transformée d'ondelettes à rejeter les informations visuelles bruitées grâces aux filtres passe-haut et passe-bas, qui interviennent dans la décomposition de l'image dans la base des ondelettes. En d'autres termes, la représentation parcimonieuse du signal-image dans une nouvelle base permet de rejeter automatiquement les informations visuelles, dont la valeur nulle ou proche de zéro (environ 80\% ou 90\% des images utilisées dans ce travail). Pour davantage de détails sur ces travaux, le lecteur est invité à consulter nos travaux dans~\citep{tamadazteIcra2016b, tamadazteIros2016b, tamadazteIjrr2018, tamadazteBiorob2018, OurakTMECH2019}. 
%
\subsubsection{Validation expérimentale}
\begin{figure}[!h]
  \centering
  \includegraphics[width=.7\columnwidth]{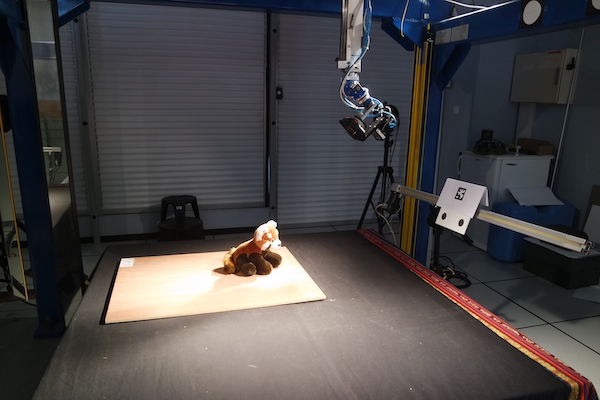}
  \caption{Vue du système expérimental avec une scène 3D.}
  \label{fig.setup_irisa}
\end{figure}

Pour la validation expérimentale, nous avons utilisé un robot cartésien à 6 degrés de liberté de l'équipe \emph{Rainbow} (IRISA, Rennes) sur lequel est montée une caméra CCD (Fig.~\ref{fig.setup_irisa}). La validation a été réalisée, à l'image de celle numérique, sous plusieurs conditions de travail favorables et non-favorables. Nos méthodes ont également été comparées à l'approche photométrique. Les résultats expérimentaux confirment ceux obtenus en simulation, notamment en termes de précision et de robustesse vis-à-vis des perturbations extérieures (occultations, changement d'éclairage), erreurs d'étalonnage de la caméra, utilisation d'une scène 3D sans reconstruction de la profondeur $Z$, etc.   

\begin{figure}[!h]
  \centering
  \includegraphics[width=\columnwidth]{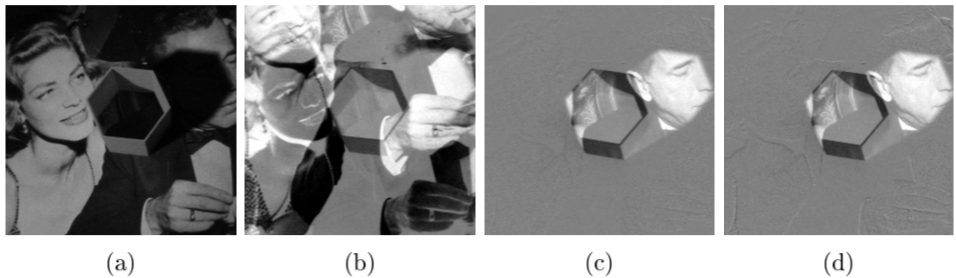}
  \caption{Tâche de positionnement par asservissement visuel par ondelettes :  (a) image désirée $\mbf I^*$, (b) image de différence initiale ($\mbf I - \mbf I^*$), (c) image de différence finale  ($\mbf I - \mbf I^*$) dans le cas des ondelettes non-sous-échantillonnées, et (d) l'image de différence finale ($\mbf I - \mbf I^*$) dans le cas d'ondelettes sous-échantillonnées.}
  \label{fig.seq_w_exp_occ}
\end{figure}

La Fig.~\ref{fig.seq_w_exp_occ} illustre la réalisation d'une tâche de positionnement à 6 degrés de liberté sous occultations partielles. Il est à noter que malgré l'occultation d'une partie (environ~$\frac{1}{6}$) de l'image durant la tâche de positionnement, les deux contrôleurs utilisant les coefficients sous-échantillonnés et non-sous-échantillonnés gardent une convergence précise vers la position désirée, comme le montre la différence entre les images finale et désirée (Fig.~\ref{fig.seq_w_exp_occ}(c) et Fig.~\ref{fig.seq_w_exp_occ}(d)).

\begin{figure}[!h]
  \centering
  \includegraphics[width=.8\columnwidth]{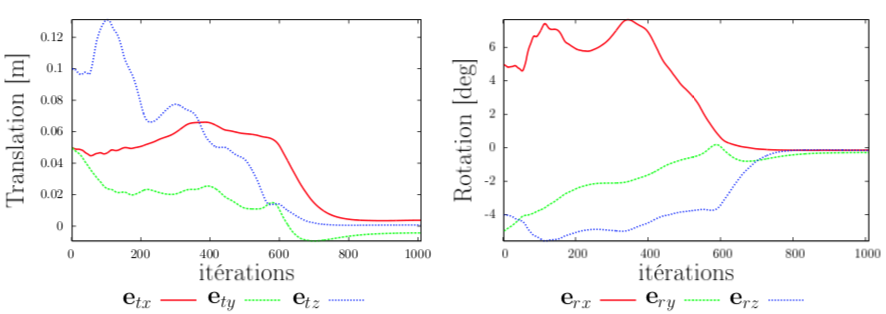}
  \caption{Erreurs de positionnement en fonction des itérations dans le cas des ondelettes non-sous-échantillonnées : (gauche) en translation et (droite) en rotation.}
  \label{fig.seq_w_exp_occ_er}
\end{figure}

Par ailleurs, la Fig.~\ref{fig.seq_w_exp_occ_er} montre la régulation vers zéro du vecteur erreur $ \mbf{e}_\texttt{w}(t)$ pendant la tâche de positionnement (cas des ondelettes non-sous-échantillonnées). Comme attendu, la décroissance n'est pas tout à fait exponentielle, ceci est dû notamment au choix de l'optimisation de type \emph{Levenberg-Marquardt} dans la loi de commande. Les valeurs numériques de la précision à convergence sont (dans les conditions nominales) de l'ordre de 0.15~mm (moyenne sur les 3 translations) et de 0.16$^\circ$ (moyenne sur les 3 rotations).  \\

Dans~\citep{tamadazteIjrr2018}, il a été démontré que la loi de commande utilisant des coefficients d'ondelettes sous-échantillonnés montre un comportement et une précision meilleure que celle avec les coefficients non-sous-échantillonnés. Ceci peut s'expliquer par le fait que la forme sous-échantillonnée des coefficients permet de ne modéliser qu'un nombre limité d'informations visuelles dans la matrice d'interaction, les plus pertinentes, c'est-à-dire celles qui se trouvent le plus loin de zéro dans la base de décomposition en ondelettes.  Les ondelettes sous-échantillonnées agissent comme un filtre dans l'espace parcimonieux. C'est probablement là que réside une des contributions les plus intéressantes de ce travail. Par ailleurs, ces mêmes méthodes ont montré une meilleure précision et robustesse par rapport à la méthode photométrique, dans des conditions de travail défavorables. 

%
\section{Transformée en shearlets}
\subsection{Shearlets :  les bases}
%
La transformée en shearlets apparue en 2005~\citep{LabateShearlets}, appelée également \emph{shearlets}, peut être considérée comme une extension naturelle des ondelettes. A l'inverse des ondelettes, les shearlets concernent uniquement les signaux 2D ou 3D, mais ne sont pas adaptées aux signaux 1D. La décomposition en shearlets utilise des opérateurs anisotropes pour mieux prendre en compte les informations visuelles anisotropes et singularités (autres que les verticales, horizontales et diagonales) d'une image, là où les ondelettes montrent des limites. Ces caractéristiques anisotropes et singulières sont des informations pertinentes (points d'intérêts, contours, etc.)  de l'image qui ne doivent pas être omises lors d'une décomposition. \\

La décomposition en shearlets est réalisée grâce à trois opérateurs anisotropes : dilatation, translation et découpage (shearing, en anglais) appliqués sur des fonctions génératrices carrées intégrables et sommables définies dans $L^2(\bR^2)$.  L'opérateur géométrique général (englobant les 3 opérateurs suscités) $\DO{\mbf {M}}$ est donné par :
\begin{equation} 
\DO{\mbf{M}}f(x)=  \abs{\operatorname{det}\mbf{M}}^{1/2}f(\mbf{M}x), ~ \text{avec}~~\mbf{M} \in \bR^{2\times2},
\end{equation}
avec $\abs{\operatorname{det}\mbf{M}}$  la valeur absolue du déterminant de la matrice $\mbf{M}$. Choisir $\mbf{M}$ comme matrice de dilatation anisotrope $\mbf{A}$, où $\tilde{\mbf{A}}$ représente les opérateurs de dilatation $\DO{\mbf{A}}$ et $\DO{\tilde{\mbf{A}}}$, avec
\begin{equation}
\mathbf{A} = \begin{pmatrix}
2 & 0\\ 
0 & \sqrt{2} 
\end{pmatrix} \quad \text{et} \quad \tilde{\mathbf{A}} = \begin{pmatrix}
 \sqrt{2}& 0 \\ 
0 & 2 
\end{pmatrix}.
\end{equation}

En définissant $\mbf{M}$ comme matrice de découpage $\mbf{S}_k$ permet d'exprimer l'opérateur de découpage associé $\DO{\mbf{S}_k}$. $\mbf{S}_k$ s'écrit :
\begin{equation}
\mbf{S}_k = \begin{pmatrix}
1 & k \\
0 & 1
\end{pmatrix} ~~~ \text{où} ~k \in \bZ.
\end{equation}

\begin{figure}[!h]
  \centering
  \includegraphics[width=.6\columnwidth]{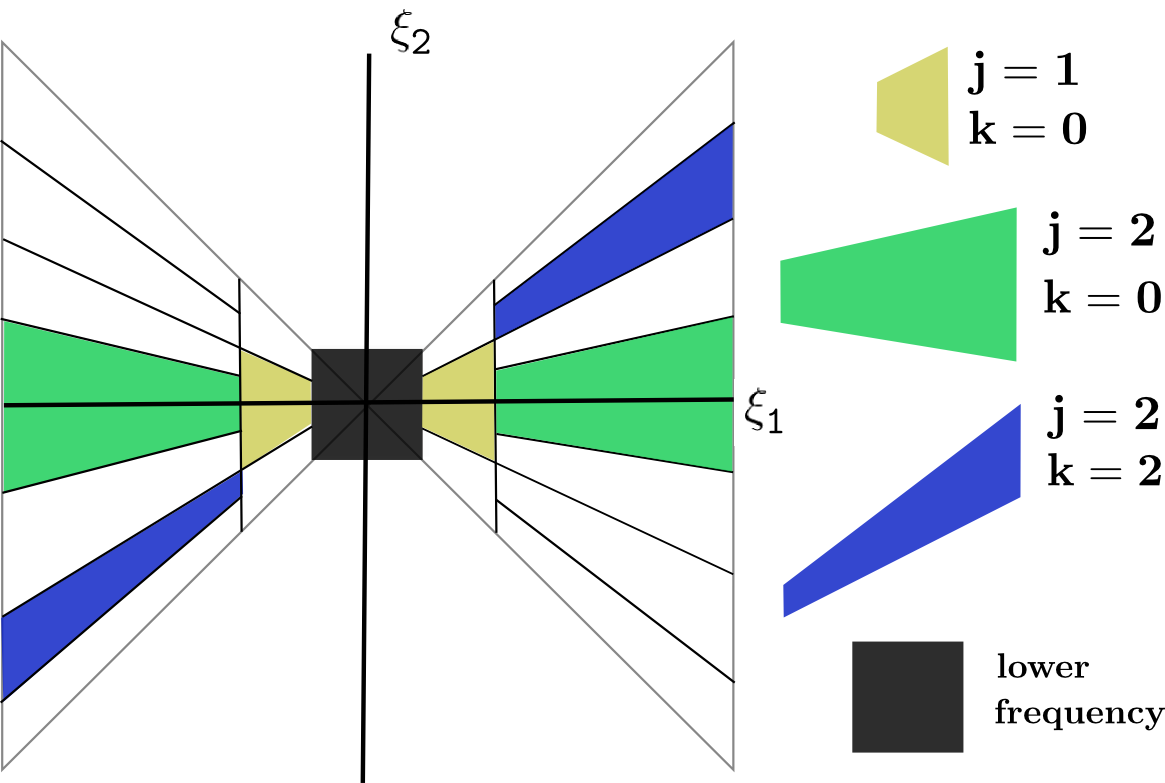}
  \caption{Découpage du cône de fréquences horizontales induit par le système des shearlets.}
  \label{fig.cone}
\end{figure}

Nous pouvons définir à présent le système des shearlets. Ainsi, pour une fonction  bi-dimensionnelle de mise à l'échelle $\phi^{(1)} \in L^2 (\bR^2)$ et les fonctions génératrices des shearlets $\psi^{(1)}, \psi^{(2)} \in L^2 (\bR^2)$, alors un système de shearlets appelé \emph{cone-adapted} (Fig.~\ref{fig.cone}) peut être défini (de la même manière que pour les ondelettes \eqref{equ.waveletsystem}), comme suit :
\begin{equation}
\label{equ.shearletsystem}
\begin{split}
&\left\{\phi^{(1)}_\mbf{m}  = \TO{\mbf{m}}\phi^{(1)}: \mbf{m}\in \bZ^2\right\} \; \cup\\
&\left\{\psi^{(1)}_{j,k,\mbf{m}} = \DO{\mbf{A}}^j\DO{\mbf{S}_k}\TO{\mbf{m}}\psi^{(1)}: j\in \bN_0, \abs{k}< \left\lceil 2^{\frac{j}{2}} \right\rceil, \mbf{m} \in \bZ^2 \right\} \; \cup\\
&\left\{\psi^{(2)}_{j,k,\mbf{m}} = \DO{\tilde{\mbf{A}}}^j\DO{\transp{\mbf{S}_k}}\TO{\mbf{m}}\psi^{(2)}: j\in \bN_0, \abs{k}< \left\lceil 2^{\frac{j}{2}} \right\rceil, \mbf{m} \in \bZ^2 \right\}.
\end{split}
\end{equation}

\begin{figure}[!h]
  \centering
  \includegraphics[width=\columnwidth]{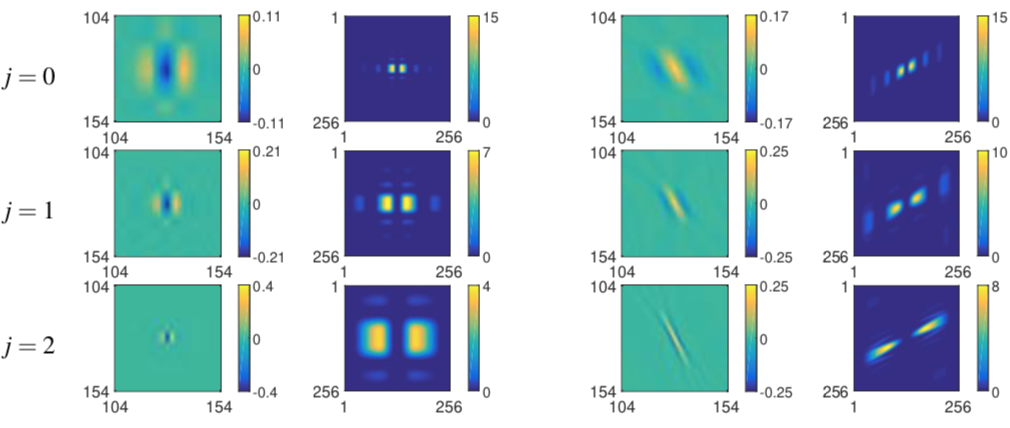}
  \caption{Différence fondamentale entre les ondelettes (à gauche), et les shearlets (à droite) lors de la décomposition multi-échelle de l'information image.}
  \label{fig.wav_vs_she}
\end{figure}

Sur la Fig.~\ref{fig.wav_vs_she}, il est montré l'apport des shearlets par rapport aux ondelettes grâce aux opérateurs anisotropes dont celui du découpage. Ainsi, toutes informations anisotropes et singularités sont bien encodées grâce aux shearlets, qui sont également un moyen de représentation parcimonieuse efficace.

Comme dans le cas des ondelettes, la décomposition en shearlets $\ST{\psi}{f}$ d'une fonction bi-dimensionnelle définie carrée intégrable $f\in L^2 (\bR^2)$ par rapport à une fonction génératrice $\psi \in L^2 (\bR^2)$, qui est encodée par les cônes de fréquences horizontales, est donnée à une échelle $j\in \bZ$, avec un découpage $\abs{k}<\left \lceil 2^{\frac{j}{2}} \right \rceil$, et une translation $\mbf{m} \in \bZ^2$, par les produits internes suivants :
\begin{equation}
(\ST{\psi}{f})(j,k,\mbf{m})= \langle f , \psi_{j,k,m} \rangle = \iint\limits_{\bR^2} f(x,y) \DO{\mbf{A}}^{j}\DO{\mbf{S}_k}\TO{\mbf{m}}\psi(x,y)  \mathrm{d}x\mathrm{d}y.
\label{equ.shearlettransform}
\end{equation}

Il existe trois différentes implémentations de la transformée en shearlets : 
\begin{enumerate}
\item \emph{Fast Finite Shearlet Transform - FFST} \citep{Huser2012FastFS}, qui calcule des coefficients des shearlets non-sous-échantillonnés grâce à des shearlets à bandes limitées,
\item \emph{Shearlab} \citep{Shearlab}, qui calcule des coefficients des shearlets sous-échantillonnés grâce à des fonctions génératrices à supports compacts,
\item \emph{ShearLab 3D} \citep{ShearLab3D}, qui permet une décomposition sous-échantillonnée des images 3D.
\end{enumerate}

Dans notre investigation des shearlets dans une boucle de commande par asservissement visuel direct, nous nous sommes inspirés des deux premières implémentations suscitées. 
%
\subsection{Matrice d'interaction analytique : shearlets}
%
Dans cette section, nous allons décrire la méthodologie suivie pour définir deux matrices d'interaction analytiques reliant les mouvements d'une caméra à la variation temporelle des coefficients des shearlets (sous-échantillonnés et non-sous-échantillonnés), de manière analogique aux travaux sur les ondelettes. 
\begin{figure}[!h]
  \centering
  \includegraphics[width=0.8\columnwidth]{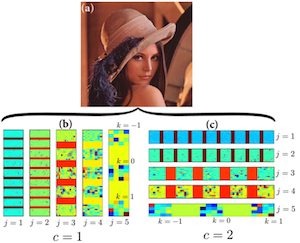}
  \caption{Coefficients des shearlets sous-échantillonnés : (a) l'image initiale, (b) les coefficients verticaux, et (c) ceux horizontaux, tous calculés à $j = [1, 2, 3, 4, 5]$.}
  \label{fig.s_sh}
\end{figure}

Pour ce faire, considérons un ensemble fini de fonctions génératrices $(\psi^{(l)})_{l\leq L}\subset L^2(\bR^2)$ défini dans $L \in \bN$. Le vecteur d'informations visuelles $\msSh(t)$ utilisé dans la boucle de commande est un ensemble de coefficients défini dans $N \in \bN$ par le quadruplet $(l_n, j_n, k_n, \mbf{m}_n)_ {n \leq N} \subset \{1, \ldots, L \} \times \bN \times \bZ \times \bZ ^2$, où $l_n$ est une fonction génératrice, $j_n$ l'échelle, $k_n$ l'opérateur de découpage, et $\mbf{m}_n$ la translation. Ainsi, $\msSh(t)$ (un exemple visuel de $\msSh(t)$  peut être vu sur Fig.~\ref{fig.s_sh} ) peut s'écrire 
 \begin{equation}
\msSh(t) = \transp{\left((\ST{\psi^{(l_1)}}{I_t})(j_1,k_1,\mbf{m}_1), \ldots,\, (\ST{\psi^{(l_N)}}{I_t})(j_N,k_N,\mbf{m}_N)\right)}.
\label{equ.featurevectorsh}
\end{equation}

Si l'évaluation des fréquences est horizontale, alors $\psi^{(l)}_{j,k,\mbf{m}}$ vaut 
\begin{equation}
\psi^{(l)}_{j,k,\mbf{m}} = \DO{\mbf{A}}^j\DO{\mbf{S}_k}\TO{\mbf{m}}\psi^{(l)}.
\end{equation}

Par contre, si cette évaluation est verticale, alors $\psi^{(l)}_{j,k,\mbf{m}}$ vaut 
\begin{equation}
\psi^{(l)}_{j,k,\mbf{m}} = \DO{\tilde{\mbf{A}}}^j\DO{\transp{\mbf{S}_k}}\TO{\mbf{m}}\psi^{(l)}.
\end{equation}

En s'inspirant de la méthodologie mise en \oe uvre dans les ondelettes (les détails de la méthode peuvent être trouvés dans~\citep{tamadazteIjrr2018}), nous pouvons calculer analytiquement la matrice d'interaction $\mLsSh$ reliant le tenseur vitesse $\tc = (v_x, v_y, v_z, \omega_x, \omega_y, \omega_z)$ de la caméra à la variation temporelle des coefficients des shearlets $\msSh$. Elle est donnée par :
\begin{equation}
\mLsSh(t)  = - \begin{bmatrix} (\ST{\psi^{(l_1)}}{{I}^{(1)}_t})(j_1,k_1,\mbf{m}_1) & \cdots & (\ST{\psi^{(l_1)}}{{I}^{(6)}_t})(j_1,k_1,\mbf{m}_1)\\\vdots&\ddots&\vdots\\(\ST{\psi^{(l_N)}}{{I}^{(1)}_t})(j_N,k_N,\mbf{m}_N) & \cdots & (\ST{\psi^{(l_N)}}{{I}^{(6)}_t})(j_N,k_N,\mbf{m}_N)\end{bmatrix} \in \bR^{N\times 6}.
\label{equ.interactionmatrixsh1} 
\end{equation}

Contrairement à la méthode des ondelettes, la matrice $\mLsSh$ est déterminée par rapport au centre du support des shearlets, qui dépend non seulement du paramètre $j$ (ce qui est le cas des ondelettes) mais aussi du paramètre de découpage $k$. En faisant, l'hypothèse que le centre du support des shearlets est situé autour de l'origine du cône des fréquences, alors nous pouvons écrire, pour évaluation verticale, la relation suivante :
\begin{equation}
\label{equ.Ls1}
\mLpsh(j,k,\mbf{m})  = \mLp\Big(2^{-j}(m_1-km_2),2^{-j/2}m_2\Big),
\end{equation}
quant à l'évaluation horizontale, nous pouvons écrire :
\begin{equation}
\label{equ.Ls2}
\mLpsh(j,k,\mbf{m})  = \mLp\Big(2^{-j/2}m_1,2^{-j}(m_2-km_1)\Big).
\end{equation}

Après quelques manipulations mathématiques, l'approximation de la matrice $\mLsSh(t)$ s'écrit sous la forme suivante :
 \begin{equation}
\mLsShTilde(t)  = -\begin{bmatrix}\left((\ST{\psi^{(l_1)}}{\frac{\partial {I_t}}{\partial x}})(j_1,k_1,\mbf{m}_1),(\ST{\psi^{(l_1)}}{\frac{\partial {I_t}}{\partial y}})(j_1,\mbf{m}_1)\right)\mLpsh(j_1,k_1,\mbf{m}_1) \\
 					\vdots \\
 					\left((\ST{\psi^{(l_N)}}{\frac{\partial {I_t}}{\partial x}})(j_N,k_N,\mbf{m}_N),(\ST{\psi^{(l_N)}}{\frac{\partial {I_t}}{\partial y}})(j_N,k_N,\mbf{m}_N)\right)\mLpsh(j_N,\mbf{m}_N) \end{bmatrix} \in \bR^{N\times 6}.
\label{equ.interactionmatrixsh2} 
\end{equation}

Maintenant que les matrices d'interaction associées aux shearlets sont définies, il est possible d'écrire les lois de commande qui en découlent.
%
\subsection{Loi de commande : shearlets}\label{subsec.shearlets}
La formulation du contrôleur est similaire à celle des ondelettes ou encore de la méthode photométrique. L'erreur visuelle à éliminer par le contrôleur est définie par :
\begin{equation}
 \mbf{e}_\text{sh}(t) = \msSh(t) - \msSh^*.
\label{equ.errorsh}
\end{equation}

Cette erreur entre la position courante et celle désirée est réduite jusqu'à ce qu'elle s'annule par :
\begin{equation}
	\tc(t) = -\lambda \Big(\mbf{H}(t) + \mu \operatorname{diag}\big(\mbf{H}(t)\big)\Big)^{-1}\transp{\Big(\mLsSh(t)\Big)} \mbf{e}_\text{sh}(t).
\label{equ.wshcontrol.levenberg}
\end{equation}
%
\subsection{Validation}
%
%
\begin{figure}[!h]
  \centering
  \includegraphics[width=\columnwidth]{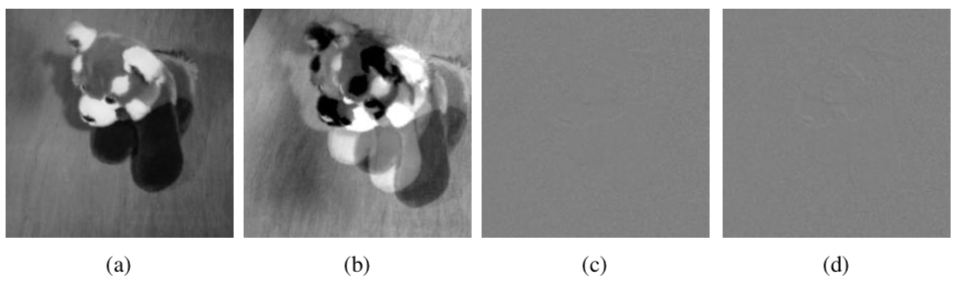}
  \caption{Tâche de positionnement par asservissement visuel fondé sur les shearlets.  (a) image désirée $ \mbf I^*$, (b) image de différence ($\mbf I - \mbf I^*$) initiale, (c) l'image de différence ($\mbf I - \mbf I^*$) finale dans le cas des shearlets non-sous-échantillonnées, et (d) l'image de différence ($\mbf I - \mbf I^*$) finale dans le cas des shearlets sous-échantillonnées.}
  \label{fig.seq_exp_sh_nom}
\end{figure}

Cette nouvelle loi de commande a été validée en simulation et sur une plateforme robotique de la Fig.~\ref{fig.setup_irisa}. Les scénarios de validation restent inchangés : dans des conditions optimales et dégradées (présence de perturbations extérieures ou d'erreurs d'étalonnage). La Fig.~\ref{fig.seq_exp_sh_nom} montre un exemple de la réalisation d'une tâche de positionnement à 6 degrés de liberté sur une scène 3D. Il est à noter que malgré que nous ne connaissions pas la profondeur $Z$ en chaque point de la scène (la profondeur $Z$ utilisée est estimée en un seul point de l'image, et ceci à la position désirée), le contrôleur converge vers la position désirée sans grande difficulté. La différence d'images ($\mbf I - \mbf I^*$) entre les images finale et désirée montre que le contrôleur est précis. Les erreurs de positionnement à convergence sont du même ordre que celles obtenues dans le cas des ondelettes.   

\begin{figure}[!h]
  \centering
  \includegraphics[width=.8\columnwidth]{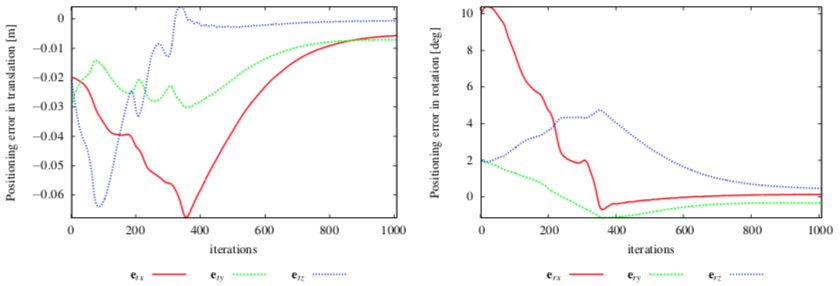}
  \caption{Erreurs de positionnement en fonction des itérations dans le cas des shearlets non-sous-échantillonnées : à gauche les translations et à droite les rotations.}
  \label{fig.seq_exp_sh_nom_err}
\end{figure}

En ce qui concerne, le comportement des méthodes fondées sur les shearlets, la Fig.~\ref{fig.seq_exp_sh_nom_err} donne un aperçu de la régulation des erreurs cartésiennes vers zéro. Le fait que la formalisme (optimisation de type \emph{Levenberg-Marquardt}) de la loi de commande est le même que dans le cas des ondelettes, le comportement l'est également dans le cas des shearlets. 
%
\section{Applications médicales}
%
Comme mentionné au début de ce chapitre, les méthodes décrites ici ont plusieurs visées cliniques. Même si les premières validations expérimentales ont été réalisées dans des conditions qui ne sont pas cliniques, ceci notamment en utilisant des robots industriels (bras manipulateurs ou robots parallèles) et des caméras conventionnelles, elles ont été, au moins partiellement, validées dans des conditions précliniques. Plus précisément, il s'agit de réaliser automatiquement des biopsies optiques répétitives dont le système d'imagerie est la tomographie par cohérence optique (OCT) ou la microscopie confocale. Il a été également question de réaliser des diagnostics échographiques assistés par un robot manipulateur sur lequel est montée une sonde échographique. Ci-dessous quelques tâches robotiques réalisées grâces aux méthodes discutées précédemment. 
%
\subsection{Examen échographique}
%
\begin{figure}[!h]
  \centering
  \includegraphics[width=.7\columnwidth]{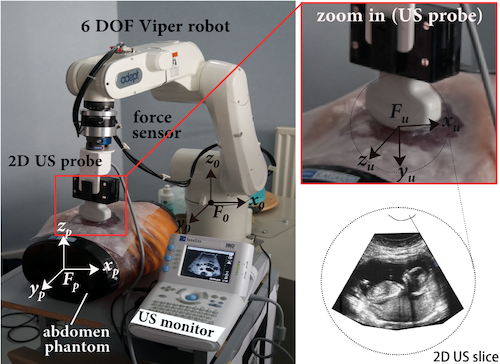}
  \caption{Vue de la plateforme robotique de l'équipe \emph{Rainbow} (IRISA, Rennes) utilisée pour la validation expérimentale.}
  \label{fig.setup_echo}
\end{figure}

Une des thématiques de la robotique médicale est la réalisation d'examens échographiques assistés par un système robotique. Il s'agit notamment de pallier les problèmes d'inégalités géographiques (télémédecine), de s'assurer d'un examen échographique le plus précis même par des radiologues avec peu d'expérience, et de compenser certains mouvements indésirables lors de l'examen (mouvements du patient, mouvements physiologiques, etc.). Plusieurs travaux ont vu le jour, notamment sur l'utilisation d'un couple robot-sonde contrôlé par une commande référencée capteur pour l'assistance aux gestes échographiques~\citep{Abolmaesumi2002, Mebarki10a, Nadeau13a,tamadazteIcra2016c}. C'est dans ce contexte que nous avons choisi de valider notre méthode des shearlets (les autres méthodes comme les ondelettes sont en cours). La Fig.~\ref{fig.setup_echo} montre la plateforme expérimentale de l'équipe \emph{Rainbow} (IRISA, Rennes) utilisée à cette occasion pour nos validations.

\begin{figure}[!h]
  \centering
  \includegraphics[width=\columnwidth]{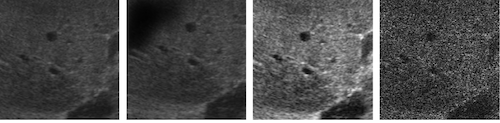}
  \caption{Exemples d'images échographiques réalisées sur un abdomen artificiel : (a) image "standard", (b) image avec des occultations partielles, (c) image avec des saturations, et (d) image avec un bruit de type "tavelure".}
  \label{fig.image_echo}
\end{figure}

Les images échographiques sont caractérisées par un rapport signal/bruit défavorable, l'absence de formes géométriques faciles à suivre visuellement (comme cela peut être le cas en imagerie conventionnelle), présence d'un bruit de type "tavelure" (speckle, en anglais), ... (Fig.~\ref{fig.image_echo}). En effet, grâce à la décomposition temps-fréquence des images échographiques, à l'aide de la transformée en shearlets, il est possible de sélectionner, de manière intuitive, uniquement les informations visuelles pertinentes (grâce à la représentation parcimonieuse) et de rejeter celles qui ne le sont pas. Ces informations visuelles sont ensuite utilisées dans la modélisation de la matrice d'interaction associée et de la loi de commande qui en découle. 

\begin{figure}[!h]
  \centering
  \includegraphics[width=\columnwidth, height = 3.7cm]{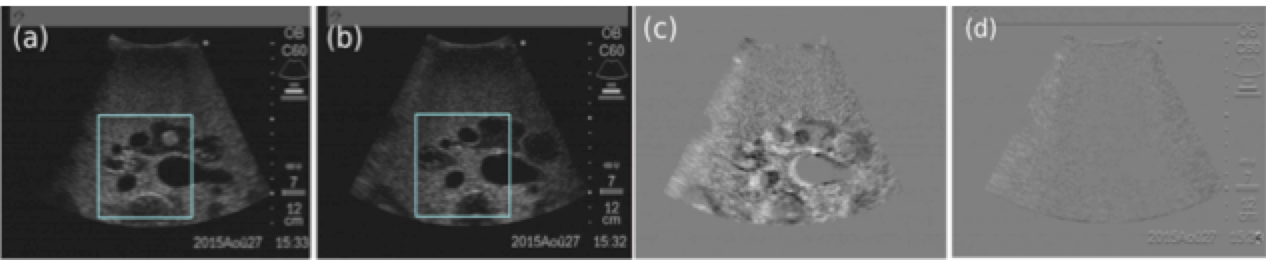}
  \caption{Séquence d'images acquises pendant la tâche de positionnement : (a) image initiale $\mbf I$ , (b) image désirée $\mbf I^*$, (c) image de différence $\mbf I - \mbf I^*$ initiale, et (d) image de différence $\mbf I - \mbf I^*$ finale.}
  \label{fig.seq_echo}
\end{figure}

La tâche de positionnement réalisée consiste à définir une position désirée dans laquelle l'examen échographique est optimal et ensuite de la reproduire automatiquement à partir d'une position arbitraire. La Fig~\ref{fig.seq_echo} montre un exemple de tâche de positionnement réalisée à l'aide d'un asservissement visuel, dont les informations visuelles modélisées sont les coefficients des shearlets sous-échantillonnés. Comme le montre cette figure, il est possible d'effectuer un positionnement précis de la sonde vis-à-vis d'un organe anatomique en utilisant uniquement les coefficients des shearlets comme informations visuelles et ceci sans algorithme de détection ou de suivi visuel de formes géométriques dans les images échographiques. Pour davantage de détails sur ces travaux, le lecteur est invité à consulter~\citep{tamadazteIcra2016c}.

\begin{figure}[!h]
  \centering
  \includegraphics[width=.8\columnwidth]{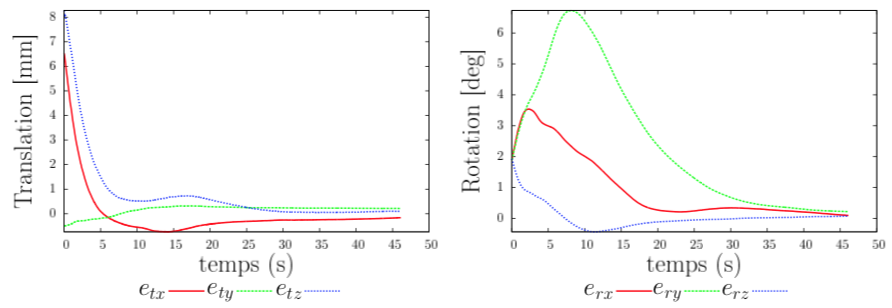}
  \caption{Graphiques montrant la régulation à zéro des erreurs cartésiennes dans l'exemple de  l'asservissement visuel échographique : à gauche, les translations et à droite, les rotations.}
  \label{fig.seq_echo_err}
\end{figure}

\begin{figure}[!h]
  \centering
  \includegraphics[width=.7\columnwidth, height = 4cm]{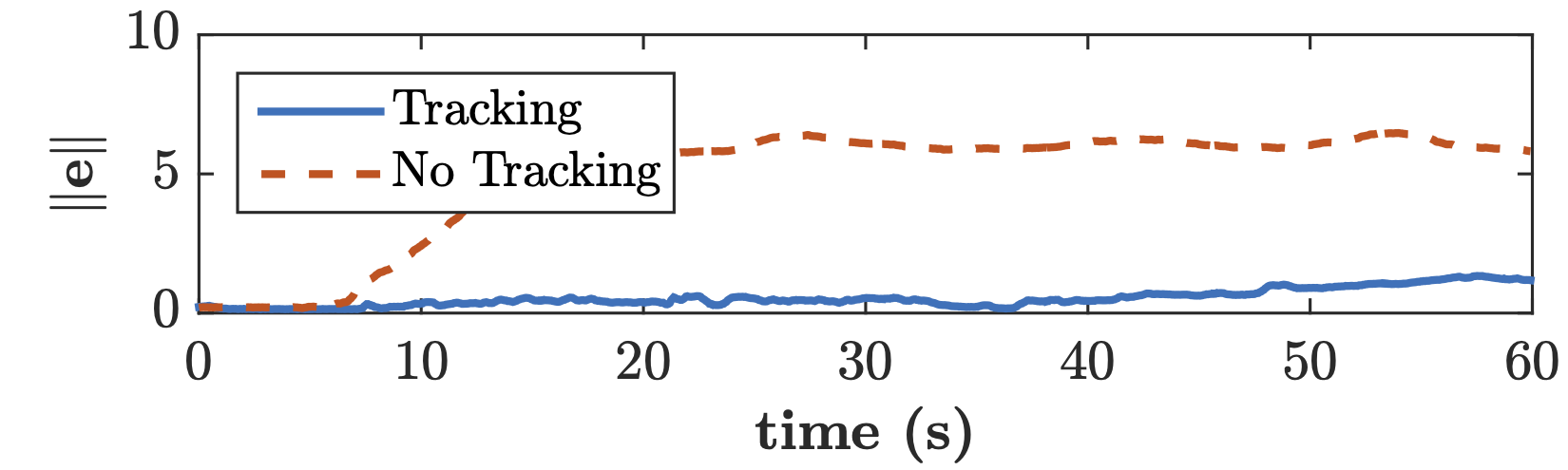}
  \caption{Exemple de la compensation de mouvements physiologiques par asservissement visuel, courbe en pointillés sans compensation, courbe en trait bleu avec compensation.}
  \label{fig.exp_echo_comp}
\end{figure}

La méthode avec les shearlets a été également testée pendant une tâche de compensation de mouvements physiologiques (des mouvements simulant la respiration ont été ajoutés pendant la tâche). Comme le montre le graphique de la Fig.~\ref{fig.exp_echo_comp}, la loi de commande par asservissement visuel permet de compenser ces mouvements de manière efficace. 

\subsection{Biopsie optique}
%
L'autre application médicale visée est la réalisation de biopsies optiques automatiques et répétitives. Il s'agit de reproduire avec précision l'acquisition d'une biopsie optique sur un tissu suspect, à un instant $t$, et ensuite de revenir reprendre la même biopsie à l'instant $t+\Delta t$ pour les comparer. Pour rappel, la biopsie optique est un examen clinique réalisé généralement \emph{in situ} à l'aide d'un système d'imagerie médicale caractérisé par une résolution micrométrique et par une capacité à visualiser le tissu en profondeur (sur plusieurs couches). \\

\begin{figure}[!h]
  \centering
  \includegraphics[width=.8\columnwidth]{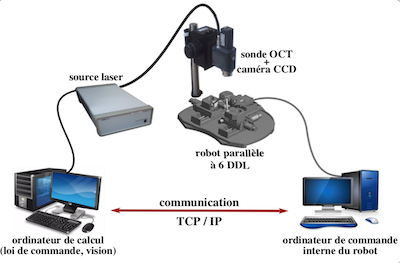}
  \caption{Station microrobotique utilisée pour la validation des méthodes d'asservissement visuel OCT.}
  \label{fig.setup_oct}
\end{figure}

Le type de biopsie optique utilisée ici est l'OCT. Pour ce faire, nous avons développé une plateforme microrobotique constituée d'un robot parallèle à 6 degrés de liberté à très haute résolution, sur lequel est monté un système OCT dans une configuration \emph{eye-to-hand} (Fig.~\ref{fig.setup_oct}). Nous avons réalisé la validation de la méthode des ondelettes \citep{tamadazteIcra2016a, tamadazteIros2016c, OurakTMECH2019}, ainsi que celle des shearlets. 

%
\subsubsection{Méthode des ondelettes}
%
Nous avons proposé de réaliser le positionnement automatique d'échantillons biologiques (un "\emph{insecte}") à l'aide du contrôleur fondé sur les coefficients des ondelettes \citep{tamadazteIros2016c}. Pour rappel, l'OCT permet l'acquisition de coupes 2D verticales successives (images OCT) dans un échantillon étudié. Aucun lien géométrique n'existe entre une coupe OCT et sa voisine. De fait, il n'y a pas de moyen de relier les mouvements hors-plan (rotations $R_x$, $R_z$ et translation $y$ si la coupe OCT est acquise suivant l'axe $x$) de la sonde OCT aux variations des informations visuelles 2D. Autrement dit, lorsque les informations visuelles sont bi-dimensionnelles, la tâche de positionnement à effectuer est restreinte à 3 degrés de liberté. Cependant, grâce à la caméra CCD attachée (co-axiale) à la sonde OCT, il est possible de contrôler les 3 degrés de liberté restants. 
 
\begin{figure}[!h]
  \centering
  \includegraphics[width=.6\columnwidth]{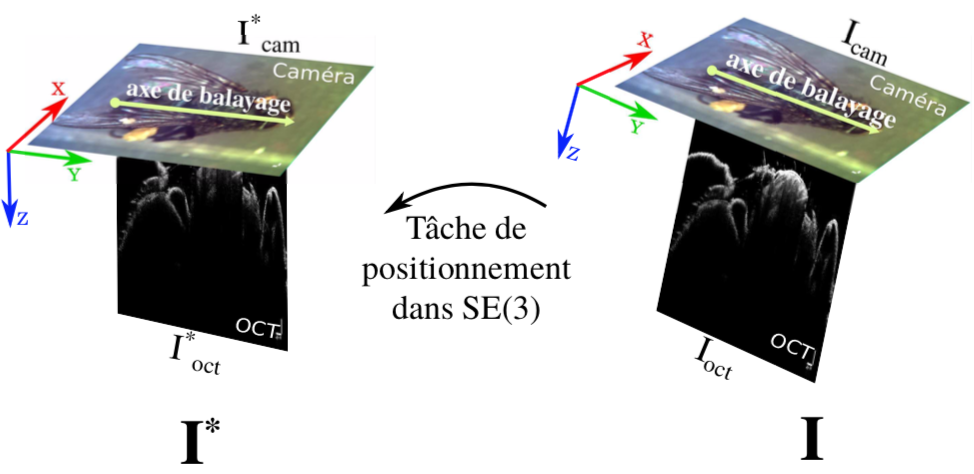}
  \caption{Tâche de positionnement dans $SE(3)$ réalisée par la loi de commande fondée sur les ondelettes.}
  \label{fig.task_se_3_oct}
\end{figure}

Pour ce faire, nous avons mis en \oe uvre une loi de commande hybride dite "partitionnée" (\ref{eq.v.part}) dont la matrice d'interaction (6 degrés de liberté) modélise à la fois les informations visuelles de l'OCT et celles de la caméra CCD. Comme le montre la Fig.~\ref{fig.task_se_3_oct}, l'image de la caméra CCD est perpendiculaire à la coupe OCT permettant de formaliser une tâche de positionnement définie dans $SE(3)$. Néanmoins, même si les informations visuelles viennent de deux capteurs différents, la loi de commande reste la même que l'asservissement visuel par ondelettes.

\begin{equation}
\tc = \mbf S \tc_{\texttt{cam}} + (\mathbb{I} - \mbf S)\tc_{\texttt{oct}},
\label{eq.v.part}
\end{equation} 
 où $\mbf S= diag (1, 0, 0, 0, 1, 1)$ une matrice de sélection et $ \mathbb{I}$ une matrice identité $6\times6$. 
 
 \begin{figure}[!h]
  \centering
  \includegraphics[width=\columnwidth, height = 8.5cm]{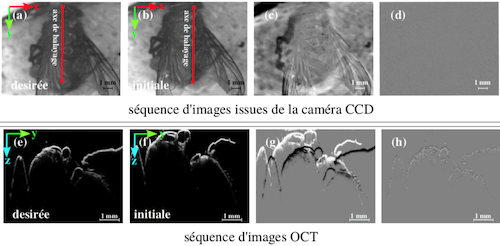}
  \caption{Séquence d'images de l'OCT et de la caméra CCD montrant la réalisation d'une biopsie optique automatique répétitive dans $SE(3)$.}
  \label{fig.seq_se3_oct}
\end{figure}

La Fig.~\ref{fig.seq_se3_oct} montre quelques images prises pendant le processus de positionnement de la sonde OCT sur une position désirée définie sur un échantillon biologique. Nous pouvons constater que malgré l'absence (sur les biopsies optiques) de texture, de contours bien définis ou encore de formes géométriques, ... la loi de commande converge sans grande difficulté vers la position où la première biopsie optique est prise. Sans surprise également, le contrôleur converge également pour les 3 degrés de liberté contrôlés à l'aide de la caméra CCD. D'autres résultats et de tests de validation peuvent être consultés dans~\citep{tamadazteIros2016c}.
%
\subsubsection{Méthode des shearlets}
%
La méthode des shearlets a également été validée dans un contexte d'acquisition de biopsies optiques, ceci sur la même plateforme expérimentale. L'échantillon biologique est, cette fois-ci, un bout de "\emph{crevette}". L'objectif est de pouvoir contrôler la position de l'acquisition d'une biopsie optique en 6 degrés de liberté et de la reproduire automatique par asservissement visuel après quelques jours, par exemple. Dans ce travail, nous avons souhaité utiliser uniquement les informations visuelles issues des images OCT pour contrôler l'ensemble des degrés de liberté de la plateforme de positionnement. 

\begin{figure}[!h]
  \centering
  \includegraphics[width=.5\columnwidth]{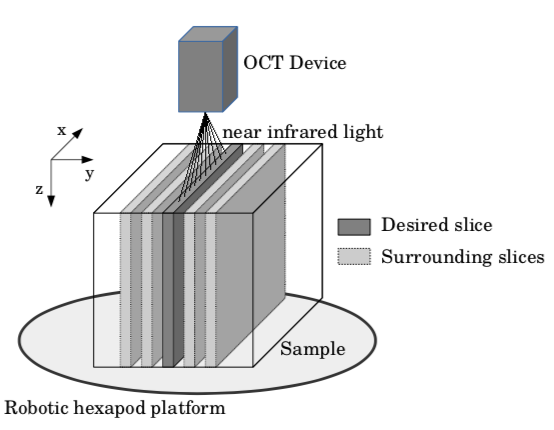}
  \caption{Méthode de calcul de la matrice d'interaction numérique à 6 ddl.}
  \label{fig.setup_oct_3d}
\end{figure}

Pour ce faire, nous avons modifié la loi de commande présentée dans la section~\ref{subsec.shearlets} en remplaçant la matrice d'interaction analytique par une forme numérique estimée hors-ligne~\citep{tamadazteBiorob2018}. Il s'agit de faire des petits déplacements successifs et indépendants de la sonde OCT suivant les 3 translations $\Delta T_x$, $\Delta T_y$ et $\Delta T_z$ et les 3 rotations $\Delta R_x$, $\Delta R_y$ et $ \Delta R_z$, ceci autour de la position désirée. Il faut ensuite effectuer une différentiation numérique sur les coefficients des shearlets $\mbf{s}_\texttt{sh}$ calculés à chaque petit déplacement. Ainsi, la nouvelle matrice d'interaction numérique est obtenue (à une résolution donnée $j$) par :
\begin{equation}
\mbf{L}_\texttt{nsh} = \Bigg( \frac{\Delta ^j\mbf{s}_{\Delta x}}{\Delta x}, \frac{\Delta ^j\mbf{s}_{\Delta y}}{\Delta y}, \frac{\Delta ^j\mbf{s}_{\Delta z}}{\Delta z}, \frac{\Delta ^j\mbf{s}_{\Delta R_x}}{\Delta R_x}, \frac{\Delta ^j\mbf{s}_{\Delta R_y}}{\Delta R_y}, 
\frac{\Delta ^j\mbf{s}_{\Delta R_z}}{\Delta R_z}\Bigg).
\label{eq.mat_num}
\end{equation}

Cette méthode présente l'avantage de fonctionner sur l'ensemble des degrés de liberté, cependant le domaine de convergence est très limité par rapport à une méthode analytique. 

\begin{figure}[!h]
  \centering
  \includegraphics[width=\columnwidth, height = 4.5cm]{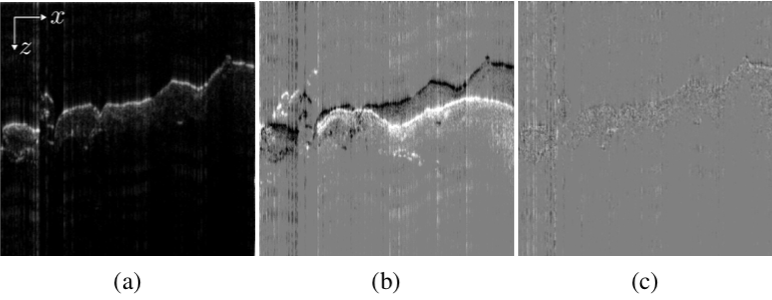}
  \caption{Exemple montrant la réalisation d'un positionnement à 6 ddl en utilisant uniquement les informations visuelles issues des images.}
  \label{fig.seq_oct_sh}
\end{figure}

La Fig.~\ref{fig.seq_oct_sh} illustre un exemple d'un repositionnement automatique de la sonde OCT à l'endroit où une biopsie optique a été acquise au préalable. Nous pouvons noter que malgré la mauvaise qualité des images OCT, le contrôleur converge tout de même à la position désirée. 
 
%
\section{Bilan}
%
Dans ce chapitre, nous avons discuté des méthodes d'asservissement visuel direct, qui se veulent plus robustes et plus précises que les méthodes traditionnelles. En effet, il s'agit d'utiliser les représentations parcimonieuses des images dans les domaines fréquentiel ou spatio-fréquentiel. Ces représentations s’obtiennent par des décompositions multi-échelles bien établies en traitement du signal/image, en l'occurrence la transformée de Fourier, d'ondelettes ou encore de shearlets. Le "c\oe ur" d'une commande par asservissement visuel est la définition de la matrice d'interaction analytique, qui modélise les mouvements de l'imageur en fonction des variations temporelles des informations visuelles considérées dans la tâche. A l'instar des méthodes d'asservissement visuel photométrique, nous avons proposé d'utiliser l'information spectrale (méthodes de Fourier), les coefficients d'ondelettes (multi-échelle ou non), ou encore les coefficients des shearlets dans l'expression de lois de commande à 6 degrés de liberté. Certaines méthodes développées sont détaillées dans ce chapitre, mais d'autres ne le sont pas, mais peuvent être consultées dans nos travaux~\citep{tamadazteTase2016, tamadazteTim2016, tamadazteIros2016a, tamadazteIcra2016c}. \\

Pour chacune des méthodes, nous avons toujours eu le souci de réaliser des validations en simulation et dans des conditions expérimentales. L'objectif final de ce travail sur l'asservissement visuel est les applications médicales. Nous avons eu l'occasion de voir certaines applications comme l'assistance d'un examen échographique, et le contrôle d'acquisition de biopsies optiques répétitives. Ce travail de validation clinique se poursuit à travers d'autres applications comme la navigation intracorporelle, l'acquisition comprimée d'images médicales, etc.  \\

A titre de bilan, nous avons mis en \oe uvre plusieurs lois de commande, que nous résumons ci-après :  
\begin{itemize}
\item asservissement visuel spectral à 6 degrés de liberté dont les informations visuelles considérées sont les représentations en \emph{log-polaire} des fréquences dans le domaine de Fourier,
\item asservissement visuel direct 3D à 4 degrés de liberté ($x$, $y$, $z$ et $R_x$) qui utilisent la pose d'une caméra reconstruite directement en utilisant l'information fréquentielle dans le domaine de Fourier sans étape de reconstruction,
\item asservissement visuel à 6 degrés de liberté qui utilise les coefficients des ondelettes. Nous avons mis en \oe uvre trois matrices d'interaction analytiques : 
	\begin{itemize}
	\item matrice d'interaction multi-échelle modélisant des coefficients d'ondelettes sont multi-échelle ;
	\item matrice d'interaction modélisant des coefficients d'ondelettes sous-échantillonnés ;
	\item matrice d'interaction modélisant des coefficients d'ondelettes non-sous-échantillonnés.
	\end{itemize}
\item asservissement visuel à 6 degrés de liberté utilisant les coefficients des shearlets dans l'expression de la matrice d'interaction associée. Deux versions de cette méthode sont proposées : 
	\begin{itemize}
	\item matrice d'interaction dont les informations visuelles modélisées sont des coefficients des shearlets sous-échantillonnés ;
	\item matrice d'interaction modélisant des coefficients de shearlets non-sous-échantillonnés.\\
	\end{itemize}
\end{itemize}

Les méthodes d'asservissement visuel direct sont caractérisées par une très bonne précision avec l'inconvénient de posséder un domaine de convergence plus restreint par rapport aux méthodes géométriques, par exemple. Ces caractéristiques, nous les retrouvons également dans nos méthodes avec l'avantage de rejeter intuitivement les informations visuelles bruitées grâce à la représentation parcimonieuse. \\

Ces travaux sur l'asservissement visuel direct ont donné lieu aux publications listées (liste non exhaustive) ci-dessous.  
\footnotesize
\subsubsection{Liste des publications scientifiques [depuis 2012]}
\begin{enumerate}
%
%
\item [Ji] \underline{M. Ourak}, \textbf{B. Tamadazte}, Olivier Lehmann, and N. Andreff, \emph{Direct Visual Servoing using Wavelet Coefficients}, IEEE Trans. on Mechatronics, doi :10.1109/TMECH.2019.2898509, 2019. 
\item [Ji] B. Rosa, B. Dahroug, \textbf{B. Tamadazte}, K. Rabenorosoa, P. Rougeot, N. Andreff, and P. Renaud, \emph{Online Robust Endomicroscopy Video Mosaicking using Robot Prior}, IEEE Robotics and Automation Letters, 2018, 3(4), 4163-4170.
\item [Ji] \underline{L.-A. Duflot}, R. Reisenhofer, \textbf{B. Tamadazte}, A. Krupa, and  N. Andreff, \emph{Wavelet and Shearlet-Based Image Representations for Visual Servoing}, The Int. J. of Robotics Research, 2018, 0278364918769739.
\item [Ji] \underline{N. Marturi}, \textbf{B. Tamadazte}, S. Demb\'el\'e, and N. Piat, \emph{Visual Servoing-Based Depth Estimation Technique for Manipulation inside SEM}, IEEE Trans. on Instrumentation and Measurement, DOI: 10.1109/ TIM.2016.2556898, 2016.
\item [Ji] \underline{N. Marturi}, \textbf{B. Tamadazte}, S. Demb\'el\'e, and N. Piat, \emph{Image-Guided Nanopositioning Scheme for SEM}, IEEE Trans. on Automation Science and Engineering, DOI: 10.1109/TASE.2016.2580660, 2016.
\item [ ----- ]
\item [CL] \underline{M. Ourak}, \textbf{B. Tamadazte}, N. Andreff, and E. Marchand, \emph{Visual Servoing-based Numerical Registration of Multimodal Images}, selected paper from ICINCO conference 2013, 10.1007/978-3-642-38085-3 12, LNCS Springer, 2015.
\item [ ----- ]
\item [Ci] B. Rosa, K. Rabenorosoa, \textbf{B. Tamadazte}, P. Rougeot, P. Renaud, and N. Andreff, \emph{Building robust confocal endomicroscopy mosaics despite image losses}, Hamlyn Symposium on Medical Robotics, London, United Kingdom, 2018.
\item [Ci] \underline{L.-A. Duflot}, \textbf{B. Tamadazte}, A. Krupa, and N. Andreff, \emph{Wavelet-based visual servoing using OCT images}, IEEE BioRob, Netherlands, 2018, 621-626.
\item [Ci] \underline{M. Ourak}, \textbf{B. Tamadazte}, G. J. Laurent, and N. Andreff, \emph{Geometric Calibration of an OCT Imaging System}, IEEE ICRA, Brisbane, 2018, 3993-3999.
\item [Ci] \underline{\underline{Y. Baran}}, K. Rabenorosoa, G. J. Laurent, P. Rougeot, N. Andreff, and \textbf{B. Tamadazte}, \emph{Preliminary Results on OCT-based Position Control of a Concentric Tube Robot}, IEEE/RSJ IROS, Vancouver, BC, 2017, pp. 3000-3005.
\item [Ci] \underline{L.-A. Duflot}, \textbf{B. Tamadazte}, A. Krupa, and N. Andreff, \emph{Shearlet Transform: a Good Candidate for Compressed Sensing in Optical Coherence Tomography}, IEEE EMBC, USA, DOI: 10.1109/EMBC.2016.7590733, 2016.
\item [Ci] \underline{L.-A. Duflot}, \textbf{B. Tamadazte}, A. Krupa, and N. Andreff, \emph{Shearlet-based vs. Photometric-based Visual Servoing for Robot-assisted Medical Applications}, IEEE/RSJ IROS, South Korea, DOI: 10.1109/ IROS.2016. 7759603, 2016.
\item [Ci]\underline{M. Ourak}, \textbf{B. Tamadazte}, and N. Andreff, \emph{Partitioned Camera-OCT based 6 DOF Visual Servoing for Automatic Repetitive Optical Biopsies}, IEEE/RSJ IROS, South Korea, 2016.
\item [Ci] V. Guelpa, G. J. Laurent, \textbf{B. Tamadazte}, P. Sandoz, N. Piat, and C. Clévy, \emph{Single Frequency-based Visual Servoing for Microrobotics Applications}, IEEE/RSJ IROS, South Korea, DOI: 10.1109/IROS.2016.7759629, 2016.
\item [Ci] \underline{M. Ourak}, \underline{\underline{A. De Simone}}, \textbf{B. Tamadazte}, G. J. Laurent, A. Menciassi, and N. Andreff, \emph{Repetitive and Automated Optical Biopsies using Optical Coherence Tomography}, IEEE ICRA, Sweden, pp. 4186-4191, 2016.
\item [Ci] \underline{M. Ourak}, \textbf{B. Tamadazte}, and N. Andreff, \emph{Wavelet-based 6 DOF Visual Servoing}, IEEE ICRA, Sweden, pp. 3414-3419, 2016.
\item [Ci] \underline{L.-A. Duflot}, A. Krupa, \textbf{B. Tamadazte}, and N. Andreff, \emph{Towards Ultrasound-based Visual Servoing using Shearlet Coefficients}, IEEE ICRA, Sweden, pp. 3420-3425, 2016.
\item [Ci]  \underline{M. Ourak}, \textbf{B. Tamadazte}, N. Andreff, and E. Marchand, \emph{From Multimodal Image Registration to Visual Servoing using Mutual Information with Simplex Optimization}, ICINCO, France, pp. 44-51, 2015.
\item [ ----- ]
\item [Wi]  \underline{M. Ourak}, \textbf{B. Tamadazte}, and N. Andreff, \emph{Registration using Wavelet Transform in Spectral Domain}, Surgitica, Chambéry, France, 2014. 
\item [Wi]  \textbf{B. Tamadazte}, \emph{Microrobotics and Optical Coherence Tomography}, Workshop "Medical Imaging Robotics" at IEEE/RSJ IROS, Vancouver, 2017 (\textbf{50 participants}).
\end{enumerate}
\normalsize

\vspace{.5cm}

\noindent
\underline{P. Nom} : doctorant(e)\\
\underline{\underline{P. Nom}} : stagiaire\\

\noindent
Ji : journal international avec comité de lecture\\
CL :  chapitre de livre\\
Ci : conférence internationale avec actes et comité de lecture\\
Wi : workshop international avec actes et comité de lecture

%% file: fichiers/cs_v3.tex
%
\chapter{Acquisition comprimée : \emph{résultats préliminaires}} \label{chap.cs}

\minitoc

\emph{Ce chapitre concerne une thématique récente dans mes activités de recherche, l'acquisition comprimée en imagerie médicale. Il sera donc question de discuter de mes premières contributions scientifiques en acquisition comprimée et les résultats qui en découlent. Cette thématique constitue une continuité scientifique naturelle des travaux menés sur les méthodes de décomposition multi-échelle (ondelettes, shearlets), de la parcimonie, de l'asservissement visuel, de la commande de laser, et de la biopsie optique présentés dans les chapitres précédents. \\
Nous allons discuter des méthodes d'acquisition comprimée combinant la décomposition en shearlets et des techniques de balayage continus adaptées aux contraintes cinématiques du système d'acquisition (exemple : miroirs galvanométriques). L'algorithme d'acquisition comprimée résultant a été directement implémenté sur un système d'imagerie médicale, en l'occurrence la tomographie à cohérence optique (OCT). Nous allons également décrire les premiers travaux sur les problèmes inverses et de complétion de tenseurs à rang faible rencontrés en statistiques et dans plusieurs autres domaines d'ingénierie, comme l'acquisition comprimée de données tri-dimensionnelles.}

%
\section{Acquisition comprimée}
%
\subsection{Définitions}
%
L'acquisition comprimée (en anglais, \emph{compressed sensing}) est une discipline, qui a émergé récemment et qui peut être considérée comme une extension naturelle des travaux sur l'analyse/décomposition des signaux ou images par des approches de type ondelettes. Un problème d'acquisition comprimée peut être vu, sous le spectre des mathématiques, comme la recherche d'une solution la plus parcimonieuse d'un système linéaire sous-déterminé. 

En effet, le théorème de \emph{Nyquist-Shannon} stipule, que l'échantillonnage d'un signal/image devrait être échantillonné à une fréquence $f_n$ supérieure ou égale à deux fois la fréquence maximale $f_m$ présente dans celui-ci, autrement dit $f_n \geq 2 f_m$. Si cette condition n'est pas respectée, la reconstruction du signal/image peut \^etre sujette \`a une perte d'information et de distorsions. Ce théorème est utilisé dans les systèmes d'acquisition numérique comme les caméras. A noter que respecter cette contrainte peut s'avérer très complexe pour certains systèmes de mesure ou d'acquisition numérique. Ceci est notamment vrai lorsque la taille des données à acquérir est élevée et que les fréquences qui caractérisent le signal/image sont élevées, ce qui impose une fréquence d'échantillonnage trop grande pour \^etre mise en pratique.  

\begin{figure}[!h]
  \centering
  \includegraphics[width=0.8\columnwidth]{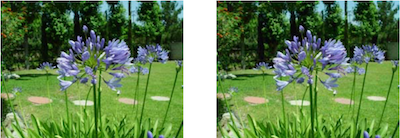}
  \caption{Compression JPEG :  à gauche, une image initiale brute (15MB), et à droite, l'image compressée JPEG (150KB).}
  \label{fig.jpeg}
\end{figure}

Candès~\citep{candes2006stable}, Donoho~\citep{donoho2006compressed} et leurs collègues se sont posés la question de "comment dépasser le théorème de \emph{Nyquist-Shannon} et de pouvoir reconstruire un signal/image avec beaucoup moins de mesures ?". L'idée est alors de représenter le signal/image dans une autre base où une grande partie de celui-ci est nul ou proche de zéro, c'est-à-dire parcimonieux (en anglais, \emph{sparse}). A titre d'exemple, la décomposition en ondelettes d'une image montre que, plus de 90\% des coefficients sont nuls ou proches de zéro. Cela veut dire, que seulement une petite partie de l'information contenue dans l'image "suffit" à remonter à (reconstruire) l'ensemble de l'information spatiale de l'image. Le meilleur exemple, de cette capacité à représenter une image avec si peu d'informations est, sans aucun doute, la compression JPEG ou JPEG2000 (Fig.~\ref{fig.jpeg}). 

\begin{figure}[!h]
  \centering
  \includegraphics[width=0.5\columnwidth]{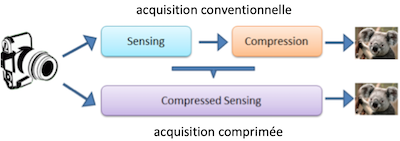}
  \caption{Différence entre la compression d'images et l'acquisition comprimée.}
  \label{fig.cs}
\end{figure}

Par conséquent, la question qui se pose alors est la suivante : "à quoi sert d'acquérir 100\% des informations (dont 90\% de données redondantes comme les zones homogènes) pour finalement en garder que 10\% après compression ?". Ne serait-ce pas plus judicieux de directement acquérir ces 10\% de mesures pertinentes pour reconstruire entièrement l'image. C'est le principe même de l'acquisition comprimée, comme son nom l'indique (Fig.~\ref{fig.cs}). 

L'acquisition comprimée peut résoudre un grand nombre de problèmes en imagerie numérique standard et médicale comme :
\begin{itemize}
\item la conception de capteurs distribués et intelligents à faible coût de fabrication. A titre d'exemple, concevoir une caméra à un pixel dotée d'un système de balayage permettant d'acquérir des images à très haute résolution grâce à l'acquisition comprimée ;
\item s'affranchir des problèmes de transmission des grands volumes de données avant de les compresser (la compression se fait directement à l'acquisition) ;
\item améliorer le temps d'acquisition des systèmes d'imageries 3D (capteurs sismiques, échographie, IRM, tomographie, etc.) ;
\item diminuer les risques de santé (patients et cliniciens) lors d'examens radiologiques en limitant le temps d'exposition aux rayons ;
\item augmenter la fréquence d'acquisition grâce à l'acquisition comprimée permet supprimer certains artéfacts lorsque la scène capturée est en mouvement (fréquence d'acquisition $> >$ à la dynamique de la scène).  
\end{itemize}
%
\subsection{Théorie et concepts mathématiques sous-jacents}
%
Les mathématiques, qui découlent du concept d'acquisition comprimée, peuvent être vues comme des problèmes de manipulation de matrices aléatoires (respectivement tenseurs) et de techniques d'optimisation avancées. Mathématiquement, le problème d'acquisition comprimée peut se définir comme suit. Soit $\mbf x$ un ensemble de données (un vecteur, une matrice ou un tenseur) défini dans l'espace Euclidien $\mathbb E$, qui admet une représentation parcimonieuse dans une base de décomposition des données $\Upsilon$ (ondelettes, shearlets, curvelets, etc.) :
\begin{align}
	\mbf x & = \sum_{j=1}^p \ \mbf c_j \Upsilon _j
\end{align}
où, $\mbf c\in \mathbb R^p$ est un vecteur parcimonieux (en anglais, $k$-sparse) qui signifie qu'il contient $k$ valeurs non nulles. Les mesures sont obtenues par un simple système d'équations linéaires sous la forme suivante 
\begin{align}
	\mbf y_i & = \big \langle \mbf m_i,  \mbf x \big \rangle
\end{align}
avec $\big \langle \mbf m_i,\cdot \big \rangle$ qui est un produit scalaire défini dans $\mathbb E$ pour toute valeur de $i=1,\ldots,n$. Le problème linéaire peut s'écrire alors :
\begin{align}
\mbf y & = \mbf A \mbf c
\label{linsys}
\end{align}
où $\mbf A \in \mathbb R^{n\times p}$ et les colonnes de $\mbf A$ sont obtenues par   
\begin{align*}
	\mbf A_{i,:} & =\transp{\mbf  m_i} \Upsilon
\end{align*}

Le problème d'acquisition comprimée peut être alors défini celui de reconstruire le vecteur $\mbf c$ avec le moins de mesures possibles. Par conséquent, le nombre de mesures $n$ devient très petit par rapport \`a la dimension du vecteur à reconstruire $p$, c'est-à-dire (n$\ll$p). La force des méthodes d'acquisition comprimée est de proposer une th\'eorie permettant de proposer des matrices $\mbf A$, poss\'edant les propri\'et\'es suffisantes pour reconstruire $x$ \`a partir du plus petit nombre possible (ou presque) $n$ (fonction de $k$) de valeurs observ\'ees. 

\begin{figure}[!h]
  \centering
  \includegraphics[width=0.6\columnwidth]{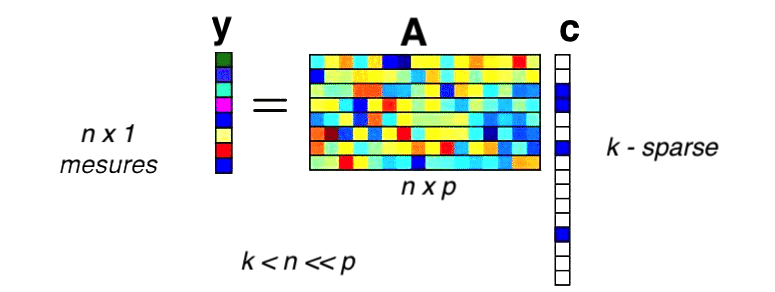}
  \caption{Représentation du problème d'acquisition comprimée.}
  \label{fig.cs_scheme}
\end{figure}

Une premi\`ere approche pour aborder ce problème (\ref{linsys}) est de trouver la solution la plus parcimonieuse possible.  A ce titre, si pour toute valeur de $\mbf c$ dans $\mathbb R^p$ et en notant $\| \mbf c\|_{l_0}$ la "norme $l_0$" de $\mbf c$, c'est \`a dire le nombre de composantes non-nulles de de $\mbf c$ (qui n'est en v\'erit\'e pas une norme), alors la reconstruction \`a partir de l'acquisition comprimée peut être obtenue par la solution de 
\begin{equation}
\label{l0}
\min_{\mbf c\in \mathbb R^p} \| \mbf c\|_{l_0} \hspace{.3cm} {\rm  sous~la~contrainte  } \hspace{.3cm} \mbf{A} \mbf{c}= \mbf{y}. 
\end{equation}

L'unicit\'e de la solution est garantie si la matrice $\mbf{A}$ satisfait la condition suffisante 
\begin{align}
    0 < & \Vert \mathbf{A}\mbf{c} \Vert_2
\end{align}
pour tout vecteur $\mbf{c}$ possédant au plus $k$ composantes non-nulles o\`u $k$ est la sparsité du signal d'origine \`a reconstruire. Un gros problème derri\`ere la minimisation de la "norme $l_0$" est que ce problème d'optimisation n'est pas convexe et est m\^eme $NP$-difficile. Pour ces raisons, Candès et Romberg~\citep{candes2006robust} proposent de simplement remplacer le problème de minimisation de la "norme" $l_0$ par celui de la norme $l_1$ (qui est bien une vraie norme, cette fois), afin de passer \`a une complexité polynomiale. Le probl\`eme obtenu est fait solvable par programmation linéaire. Ainsi, (\ref{l0}) devient 
\begin{equation}
\label{l1}
\min_{\mbf c\in \mathbb R^p} \| \mbf c\|_{l_1} \hspace{.3cm} {\rm  sous~la~contrainte  } \hspace{.3cm} \mbf{A} \mbf{c}= \mbf{y}. 
\end{equation}

Il existe une condition suffisante sur le choix de $\mbf A$ qui garantit que la solution du probl\`eme de minimiser la norme $l_1$ donne la m\^eme solution que pour celui de minimiser la norme $l_0$. Cette condition, introduite par~\citep{candes2006robust}, s’appelle la \emph{quasi-isométrie restreinte} (en anglais, restricted quasi-isometry property), qui est formalisée comme suit :
\begin{equation}
(1-\delta) \left \|  \mbf c \right \|_{2} \leq  \left \|  \mbf A \mbf c \right \|_{2} \leq (1+ \delta) \left \|  \mbf c \right \|_{2}.
\label{eq.rip}
\end{equation}
pour tout vecteur $\mbf c$ $k$-sparse. 

V\'erifier qu'une matrice satisfait la condition de quasi-isom\'etrie restreinte est en fait une probl\`eme lui m\^eme NP difficile. Heureusement, il est facile de montrer qu'une matrice al\'eatoire \`a composantes ind\'ependantes sous-gaussiennes satisfait cette condition \cite{vershynin}. 

De multiples extentions de cette approche pour le cas de la sparsit\'e par groupe, o\`u la sparsit\'e spectrale, ont d\'emontr\'e des propri\'et\'es remarquables de reconstruction. Ainsi, dans le cas de la sparsit\'e spectrale, la norme nucléaire~\citep{recht2010guaranteed} est souvent employ\'ee. 

Dans le cas o\`u les mesures sont entach\'ees d'erreur, une m\'ethode de moindres carr\'es p\'enalis\'ee est souvent employ\'ee et l'estimateur correspondant s'appelle de LASSO~\citep{tibshirani1996regression, chretien2010alternating}. 
 
%
%
Comme mentionnée ci-dessus, une des conditions nécessaires à la résolution d'un problème d'acquisition comprimée est la capacité du signal/image à être parcimonieux dans une base donnée. 

Définition : On dit qu'un vecteur $\mbf c$ est $k$-sparse (ou parcimonieux) lorsque $\left \|  \mbf c \right \|_{0} \leq k$ et en notant $\Sigma_k$ l'ensemble de tous les vecteurs $k$-sparse de $\bR^n$, alors 
\begin{equation}
\nonumber
\Sigma_k  = \{k \in \bR^N~: \left \|  \mbf c \right \|_{0} \leq k \}
\end{equation}

\begin{figure}[!h]
  \centering
  \includegraphics[width=0.7\columnwidth]{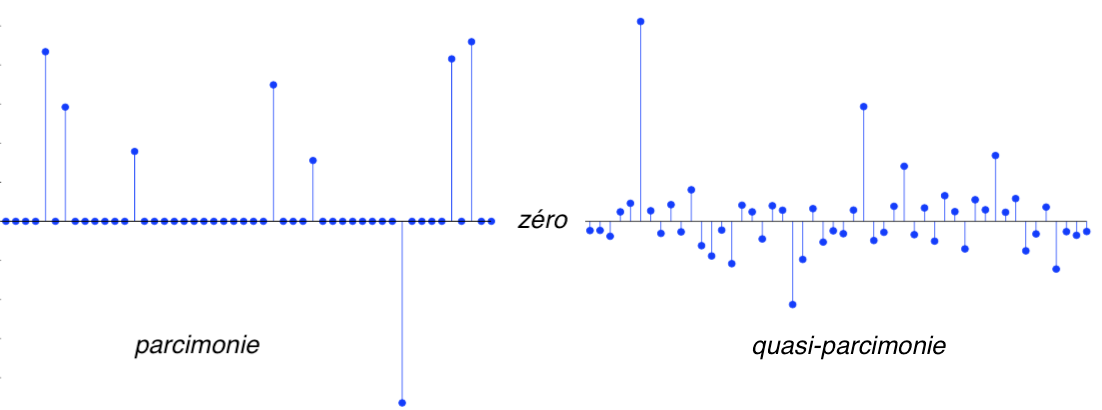}
  \caption{La parcimonie d'un signal 1D.}
  \label{fig.sparse}
\end{figure}

Une autre condition suffisante pour que la majorit\'e des probl\`emes de reconstruction $k$-sparse puisse \^etre r\'esolus par la minimisation de la norme $l_1$, est celle d'avoir une faible \emph{cohérence}\citep{candes2007sparsity, chretien2014sparse}. Nous pouvons dire que des mesures sont incohérentes par rapport au dictionnaire dans lequel le signal est sparse, si leurs représentations dans la base de mesure et de projection (base parcimonieuse) sont peu corrélées.

Définition : Soit deux bases, la base de mesure $\Psi \in \bR^n$ et la base de projection $\Upsilon \in \bR^n$ où le signal est parcimonieux. La cohérence entre les données entre les deux bases est définie par :
\begin{equation}
\nonumber
\mu \big( \Upsilon, \Psi  \big) = \sqrt{n}.\max_{1 \leq k,j \leq n}\left | \big \langle \Upsilon_k, \Psi_j   \rangle \right |
 \end{equation}
 où $\mu$ désigne la cohérence. 
%
Un syst\`eme de mesure sera dit incoh\'erent par rapport au dictionnaire si $\mu$ est petit, de l'ordre de $\log(p)^{-1}$.
%
\subsection{Imagerie médicale et acquisition comprimée}
%
L'imagerie médicale, en particulier le scanner, l'échographie 2D (mais surtout 3D), l'IRM ou encore la tomographie à cohérence optique (OCT) représentent des candidats très intéressants pour les applications des méthodes d'acquisition comprimée. D'une part, ces systèmes d'imagerie posent le problème de temps d'acquisition de fait d'une acquisition 3D (balayage spatial), de grandes quantités de données à transmettre, à (pré)traiter, à stocker, etc. D'autre part, les images médicales sont généralement caractérisées par des grandes zones homogènes, souvent "noires" qui font qu'elles sont naturellement parcimonieuses une fois projetées dans une base de décomposition. Par conséquent, elles présentent des taux de compression très élevés par rapport aux images conventionnelles. D'autres caractéristiques font des images médicales un "excellent" domaine d'application de l'acquisition comprimée. Par exemple, l'IRM, la tomographie à cohérence optique, ... acquièrent les mesures dans le domaine spectral, autrement dit directement dans une base où le signal image est directement parcimonieux. Finalement, contrairement aux caméras CCD/CMOS, la plupart des systèmes d'imagerie médicale opèrent de manière séquentielle lors de l'acquisition des mesures. C'est-à-dire ils acquièrent une donnée à 1-dimension pour chaque position $(x, y)$ du mécanisme de balayage (bobines magnétiques de l'IRM, miroirs galvanométriques d'un OCT, etc.). 

\begin{figure}[!h]
  \centering
  \includegraphics[width=0.8\columnwidth]{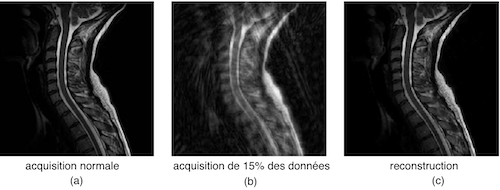}
  \caption{Exemple de résultats d'un algorithme d'acquisition comprimée sur des images IRM.}
  \label{fig.irm}
\end{figure}

La valeur ajoutée clinique de l'acquisition comprimée est bien établie. Il s'agit, par exemple, de réduire l'exposition du patient et du personnel médical aux rayons lors d'un examen radiologique, en réduisant considérablement le temps d'acquisition. Un autre exemple de l'intérêt de l'acquisition comprimée en IRM (Fig.~\ref{fig.irm}), c'est celui de pouvoir faire un examen dans de meilleures conditions ; le patient n'a plus autant besoin de retenir sa respiration ou d'éviter de faire des mouvements lors de l'acquisition\footnote{Siemens Heathineers, propose depuis peu un IRM rapide qui intègre des méthodes d'acquisition comprimée.}. 

Dans les travaux décrits dans ce chapitre, nous nous intéressons à l'acquisition comprimée dans le cas de la tomographie à cohérence optique, ceci pour plusieurs raisons. D'abord, il s'agit un besoin clinique récurrent dans trois de nos projets de recherche sur le diagnostic des maladies neurodégénératives (ANR-NEMRO), sur la chirurgie de l'oreille moyenne (ANR-$\mu$RoCS) et sur le diagnostic de cancers dans le tube digestif (INSERM-ROBOT). Aussi, l'OCT est une technique d'imagerie à fort potentiel médical (caractérisation micrométrique des tissus), mais qui souffre de problème de temps d'acquisition, notamment lorsqu'il s'agit de réaliser \emph{in situ} des biopsies optiques tri-dimensionnelles. 
%
\section{Acquisition comprimée et biopsie optique}
Avant d'aborder les méthodes développées dans le cadre de nos travaux sur l'acquisition comprimée en tomographie (mais qui peuvent s'appliquer à d'autres types d'imagerie), il faut comprendre le fonctionnement d'un tel type de système d'imagerie. 
\subsection{Fonctionnement d'un OCT}
Le principe de fonctionnement d'un système d'imagerie OCT est très proche de celui d'un appareil échographique par ultrasons. Si des ultrasons sont utilisés dans le cas d'échographies, dans le cas de l'OCT, de la lumière (à faible cohérence temporelle) dans l'infrarouge proche (entre 600~nm et 1300~nm) est utilisée. Le faisceau lumineux est absorbé par les tissus biologiques sur environ un millimètre de profondeur. Le temps de parcours du rayon lumineux renseigne sur la position des différentes couches du tissu et sur ses propriétés optiques~\citep{fujimoto2000optical}. La vitesse de la lumière ne permet pas d'étudier le délai entre l'émission du signal et la réception de son écho, comme c'est le cas pour les ultrasons en échographie. C'est grâce à une technique d'interférométrie à faible cohérence de type \emph{Michelson}, qu'une image peut être reconstruite à partir des rayons lumineux envoyés sur l'échantillon.

\begin{figure}[!h]
  \centering
  \includegraphics[width=0.65\columnwidth]{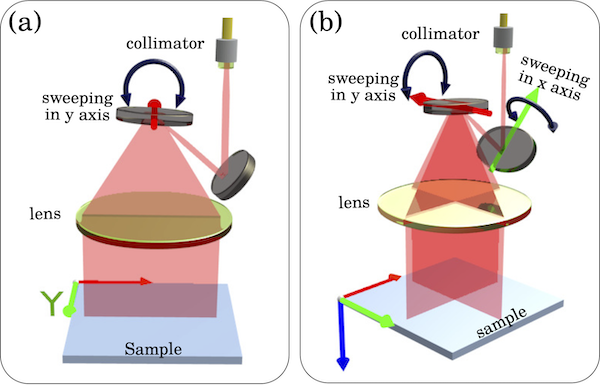}
  \caption{Illustration du fonctionnement du dispositif du système de balayage OCT \citep{tamadazteIros2018a}.}
  \label{fig.opti_path}
\end{figure}

L'OCT offre trois modes de fonctionnement : A-scan (carotte optique 1D), B-scan (section verticale 2D) et C-scan (volume). C'est surtout sa résolution spatiale (environ 5~$\mu$m/pixel) et sa capacité à voir au-delà du visible (profondeur de 1 à 3~mm) qui font de la tomographie à cohérence optique un moyen de caractérisation des tissus très intéressant. 

Les acquisitions en A-scan et B-scan ne posent pas beaucoup de problèmes en clinique. Néanmoins, l'acquisition des volumes OCT peut prendre plusieurs minutes pour plus de 10GB de données, ce qui limite le mode C-scan en usage clinique. Comme le montre la Fig.~\ref{fig.opti_path}, l'OCT utilise deux miroirs galvanométriques orthogonaux. Les deux miroirs contrôlés indépendamment ou ensemble permettent de "viser" n'importe quel point de l'échantillon étudié.  Aussi, les deux miroirs (comme dans le cas des bobines magnétiques d'un IRM), le balayage s'effectue à l'aide d'un système mécatronique (robot) et non électronique comme c'est le cas des caméras CCD/CMOS, par exemple. Par conséquent, les méthodes d'acquisition comprimée sur lesquelles nous travaillons intègrent la "composante mécatronique/robotique" dans le formalisme du problème. 
%
\subsection{Quelle base de représentation parcimonieuse choisir ?}
Avant de s'intéresser à l'acquisition comprimée en ligne sur une système d'acquisition physique, en l'occurrence l'OCT, nous avons étudié dans~\citep{tamadazteEmbc2016c}, l'apport des différentes bases de décomposition (ondelettes, curvelets, shearlets, ...) dans le problème d'acquisition comprimée. Pour ce faire, nous avons implémenté plusieurs méthodes de décomposition associées à une méthode de reconstruction du signal image en utilisation l'algorithme \emph{Fast Iterative Soft-Thresholding} (FISTA). L'algorithme FISTA \citep{Fista2009} est largement utilisé dans divers problèmes d'optimisation, pour sa simplicité et son temps de convergence acceptable en pratique. Par ailleurs, nous avons proposé une mise à jour de l'algorithme FISTA (que nous appelons $\mu$-FISTA) pour mieux l'adapter à notre problème de reconstruction (cette mise à jour est consultable dans~\citep{tamadazteEmbc2016c}.

\begin{figure}[!h]
  \centering
  \includegraphics[width=\columnwidth]{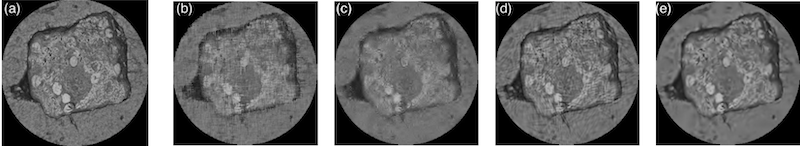}
  \caption{Exemple de reconstruction d'images OCT. (a) : image originale, (b) : reconstruction par interpolation bilinéaire, (c) : $\mu$FISTA + ondelettes, (d) : $\mu$FISTA + curvelets, et (e)  : $\mu$FISTA + shearlets.}
  \label{fig.seq_cs_fista}
\end{figure}

Pour la validation numérique, nous avons choisi trois transformées : ondelettes, curvelets et shearlets. Nous avons comparé la qualité des reconstructions réalisées en utilisant les différentes bases de décomposition suscitées et l'algorithme d'optimisation $\mu$FISTA. Les images reconstruites par acquisition comprimée ont également été comparées à une simple reconstruction par interpolation bilinéaire. La Fig.~\ref{fig.seq_cs_fista} montre un exemple de reconstruction d'un échantillon de "kiwi" en sous-échantillonnant aléatoirement l'image originale, c'est-à-dire uniquement 30\% des mesures sont utilisées pour la reconstruction.

Chacun peut constater que la reconstruction par l'interpolation bilinéaire est de mauvaise qualité quand les reconstructions par la méthode $\mu$FISTA sont de bien meilleure qualité. A noter également que les décompositions en curvelets et en shearlets surpassent largement celle des ondelettes. Nous avons évoqué, dans le chapitre précédent, que malgré l'intérêt que présente les ondelettes en décomposition multi-échelle d'une image, elles présentent néanmoins l'inconvénient de laisser passer les informations visuelles anisotropes contenues dans l'image, là où les curvelets (développées par Candès~\citep{candes2000curvelets} dans ces travaux sur l'acquisition comprimée) et les shearlets décomposent bien mieux ces informations. 

\begin{figure}[!h]
  \centering
  \includegraphics[width=.5\columnwidth]{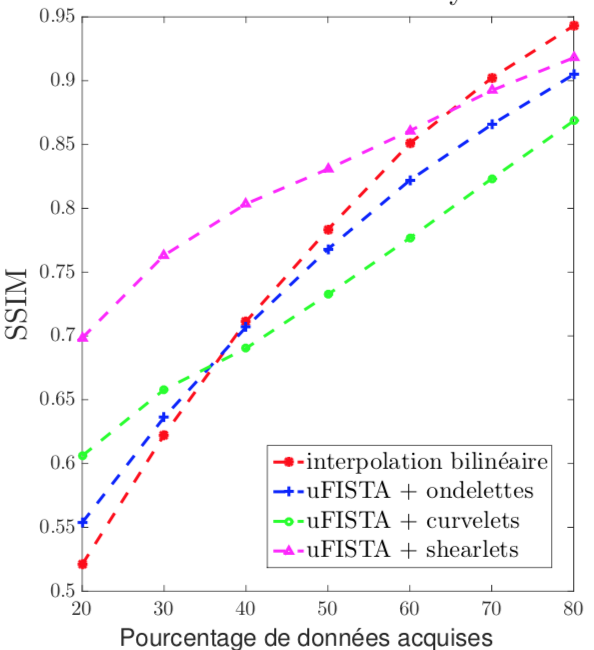}
  \caption{Analyse du score de similarité des différentes méthodes en fonction du taux d'échantillonnage (nombre de mesures utilisées).}
  \label{fig.ssim_fista}
\end{figure}

Autres renseignements à tirer de cette comparaison : les shearlets sont plus efficaces que les curvelets et les ondelettes lorsqu'il s'agit de reconstruire l'image avec de moins en moins de mesures. La Fig.~\ref{fig.ssim_fista} montre les scores de similarité SSIM (\emph{Structural SIMilarity}), généralement utilisé en traitement d'images, obtenus pour différents taux d'échantillonnage (de 20\% à 80\%).  

%
\subsection{Acquisition comprimée sur un système physique}
%
La plupart des méthodes d'acquisition comprimée existantes dans la littérature utilisent un sous-échantillonnage pseudo-aléatoire ou aléatoire incohérent de données dans la base parcimonieuse~\citep{candes2006stable, candes2008restricted}. Ces techniques de sous-échantillonnage, aussi appelées \emph{masques} ou \emph{matrices de mesures}, sont souvent peu pertinentes car peu adaptées à des systèmes d'acquisition physiques~\citep{bigot2016analysis}. Par exemple, dans le cas des systèmes d'imagerie à balayage mécanique (IRM, OCT, etc.), faire une acquisition aléatoire peut trop solliciter le dispositif de balayage et surtout consommer davantage de temps qu'une acquisition complète (100\% des échantillons). 

\begin{figure}[!h]
  \centering
  \includegraphics[width=.8\columnwidth, height = 4cm]{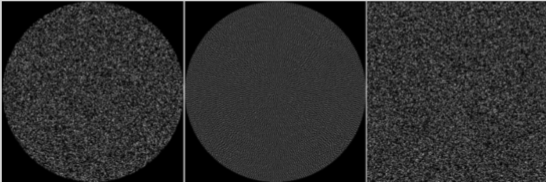}
  \caption{Exemples de trajectoires de balayage (taux de sous-échantillonnage de 20\%). De gauche à droite : spirale, rosace ou Lissajous.}
  \label{fig.trajs}
\end{figure}

Dans cette section, nous allons discuter une méthode d'acquisition comprimée directement implémentée sur système d'imagerie OCT physique. Parmi les originalités de ce travail, la mise en \oe uvre de trajectoires continues pour réaliser un sous-échantillonnage paramétrable de l'espace d'acquisition : taux de recouvrement, nombre de mesures, longueur de chemin d'acquisition, champ de vision, etc.  Les trajectoires étudiées sont de type rosace, spirale et Lissajous qui tiennent compte des limites imposées par la cinématique des miroirs galvanométriques de l'OCT. 

\begin{figure}[!h]
  \centering
  \includegraphics[width=.7\columnwidth]{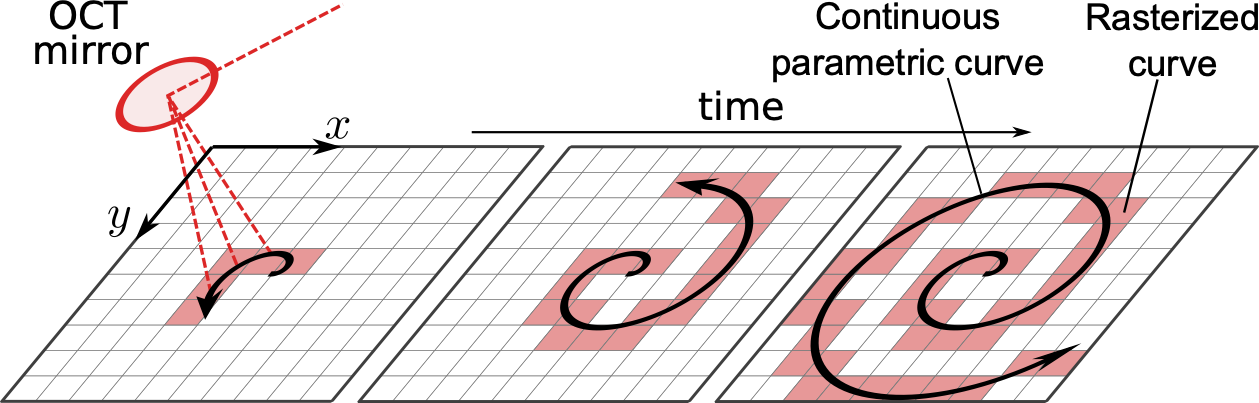}
  \caption{Illustration du fonctionnement de l'algorithme de rastérisation développé.}
  \label{fig.raster}
\end{figure}

Nous avons également développé un algorithme de rastérisation pour garantir une cohérence entre les coordonnées (en pixels) des éléments de la matrice de mesure (trajectoire de balayage) prédéfinie au préalable et les positions métriques ciblées sur l'échantillon à caractériser par biopsie optique (Fig.~\ref{fig.raster}). Dans ce travail, nous avons proposé deux versions de l'algorithme de reconstruction : 2D (shearlets 2D) et 3D (shearlets 3D).  

%
\subsection{Validation}
%
Les méthodes d'acquisition comprimée proposées ont d'abord été validées directement sur un système physique OCT dit Spectral-Domain\footnote{Telesto-II 1325nm commercialisé par Thorlabs (www.thorlabs.us).} . Les premiers tests de validation consistent à reconstruire des images OCT de 512$\times$512~pixels, tandis que les seconds visent à reconstruire des volumes OCT de 281$\times$281$\times$199~pixels. Les deux validations (2D et 3D) sont réalisées grâces à deux versions de l'algorithme d'acquisition comprimée utilisant pour l'un, une décomposition des données par des shearlets 2D et pour l'autre les shearlets 3D. 

\subsubsection{Images OCT : shearlets 2D}
\begin{figure}[!h]
  \centering
  \includegraphics[width=.9\columnwidth, height = 3.7cm]{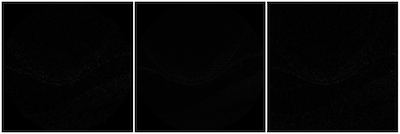}
    \includegraphics[width=.9\columnwidth,  height = 3.7cm]{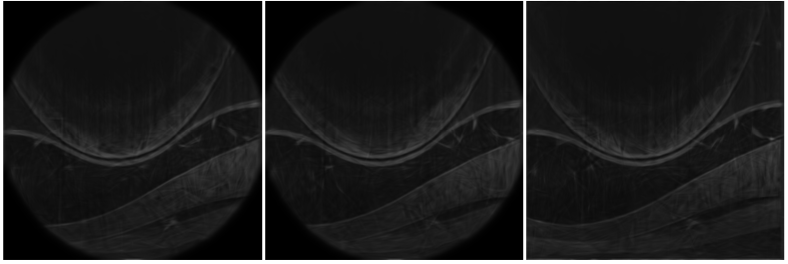}
  \caption{Exemple de reconstruction avec seulement 10\% de mesures. De gauche à droite : spirale, rosace, Lissajous. }
  \label{fig.cs_eye_10_pct}
\end{figure}

Nous avons commencé par la validation numérique, en utilisant des images OCT préenregistrées, et expérimentale en implémentant la méthode d'acquisition comprimée directement sur un système physique d'imagerie OCT. Plusieurs scénarios sont mis en \oe uvre pour en juger les performances de la méthode. D'abord, plusieurs types de biopsies optiques 2D (rétine d'un \oe il de poisson, un grain de raisin, etc.) sont utilisés. Ensuite pour chacun des échantillons, nous avons testé les trajectoires continues de balayage (Fig.~\ref{fig.trajs}) avec des taux de sous-échantillonnage, allant de 10\% à 80\%, pour chaque image OCT. 

La Fig.~\ref{fig.cs_eye_10_pct} illustre un exemple de reconstruction d'une biopsie optique 2D (rétine) à partir de 10\% de mesures seulement en utilisant les trajectoires de balayage développées. Chacun peut noter que l'algorithme est capable de reconstruire une biopsie optique de bonne qualité à partir d'une image "quasi-noire". 
%
\subsubsection{Volumes OCT : shearlets 3D}
%
Pour rappel, c'est l'acquisition de biopsies 3D qui posent généralement problème en usage clinique. Par conséquent, nous nous sommes davantage intéressés à l'acquisition comprimée 3D, donc à la version 3D de notre méthode. Il s'agit de l'une des contributions majeures de ce travail : réaliser une réelle acquisition comprimée 3D directement sur le système physique avec des stratégies de balayage associées à l'algorithme de rastérisation tout en respectant les contraintes cinématiques des miroirs galvanométriques. 

\begin{figure}[!h]
\centering
\subfigure[ ]{
\includegraphics[width=0.25\columnwidth]{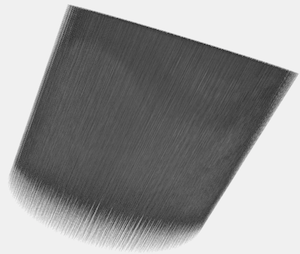}}
\subfigure[ ]{
\includegraphics[width=0.25\columnwidth]{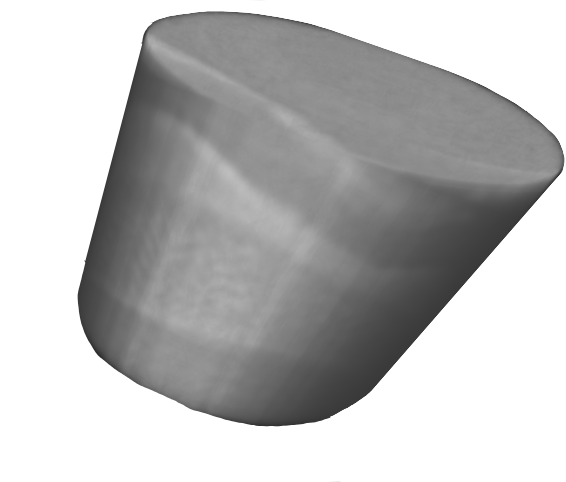}}
\quad
\subfigure[ ]{
\includegraphics[width=0.21\columnwidth]{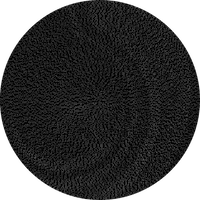}}
\subfigure[ ]{
\includegraphics[width=0.21\columnwidth]{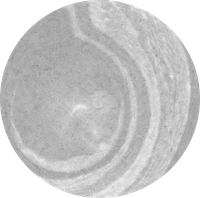}}
\caption{Reconstruction 3D d'une rétine de poisson [30\% de mesures] : (a) mesures 3D acquises, (b) volume 3D reconstruit, (c) visualisation d'une coupe 2D dans masque 3D, et (d) visualisation d'une coupe dans le volume reconstruit.}
\label{fig.cs_3d_1}
\end{figure}

\begin{figure}[!h]
\centering
\subfigure[ ]{
\includegraphics[width=0.25\columnwidth]{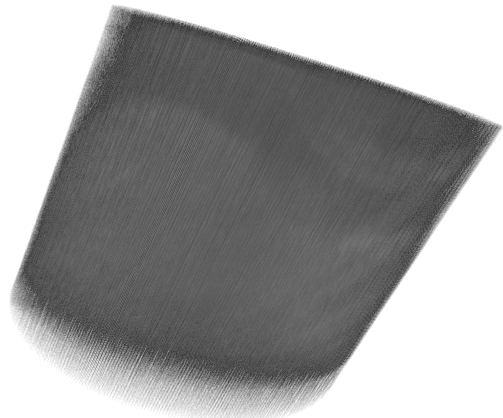}}
\subfigure[ ]{
\includegraphics[width=0.25\columnwidth]{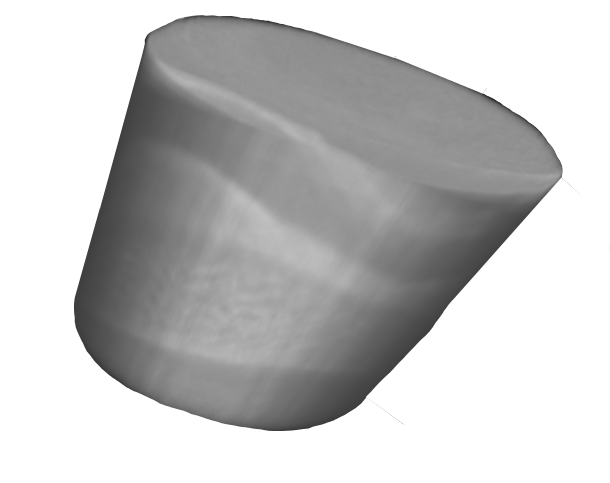}}
\quad
\subfigure[ ]{
\includegraphics[width=0.21\columnwidth]{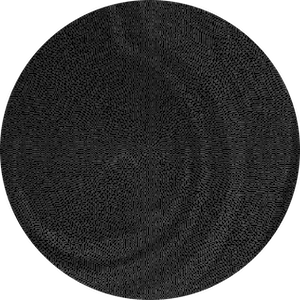}}
\subfigure[ ]{
\includegraphics[width=0.21\columnwidth]{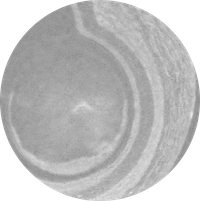}}
\caption{Reconstruction 3D d'une rétine de poisson [50\% de mesures] : (a) mesures 3D acquises, (b) volume 3D reconstruit, (c) visualisation d'une coupe 2D dans masque 3D, et (d) visualisation d'une coupe dans le volume reconstruit.}
\label{fig.cs_3d_2}
\end{figure}

La Fig.~\ref{fig.cs_3d_1} montre un exemple de reconstruction 3D (Fig.~\ref{fig.cs_3d_1}(b)) sur la rétine d'un poisson en utilisant 30\% de mesures acquises grâce à une trajectoire Lissajous (Fig.~\ref{fig.cs_3d_1}(a)). Une coupe 2D a été réalisée dans le volume pour une meilleure visualisation de la reconstruction (Fig.~\ref{fig.cs_3d_1}(d)). La même opération a été reproduite en utilisant cette fois-ci 50\% de mesures. Comme le montre la Fig.~\ref{fig.cs_3d_2}, la qualité de reconstruction est visuellement très similaire à celle à 30\% de mesures. Par conséquent, 30\% (de mesures voire moins) suffisent largement pour obtenir des biopsies 3D exploitables en clinique d'où tout l'intérêt de ces méthodes.

Le tableau~\ref{tab.simu.fisheye} résume les valeurs numériques obtenues de chaque test en utilisant l'image OCT d'une rétine d'un \oe il de poisson. En plus du score SSIM, nous avons également calculé le \emph{peak signal noise ratio} (PSNR) pour évaluer qualitativement les résultats obtenus. Généralement, si les valeurs du PSNR se situent entre 30db et 50db (pour les images codées 8 bits), alors l'image reconstruite est considérée comme fidèle à l'originale. Ainsi, en analysant le tableau~\ref{tab.simu.fisheye}, nous pouvons constater que pour seulement 10\% de mesures, nous obtenons un PSNR de près de 40db, ce qui est très intéressant. Autrement dit, il suffit d'acquérir 10\% de données, et nous sommes capables de reconstruire une image OCT de très bonne qualité. La même conclusion peut être tirée en analysant le SSIM. Finalement, les trois trajectoires de balayage étudiées permettent d'obtenir des résultats quasi-similaires, même si la courbe de Lissajous permet un meilleur recouvrement de l'échantillon avec le même nombre de points d'acquisition.   

\begin{table}[!h] 
		\begin{center}
			\caption{Évaluation de la qualité de la reconstruction en fonction de différents taux de sous-échantillonnage et de trajectoires de balayage.}
			\label{tab.simu.fisheye} 
			\begin{tabular}{c|c|c|c}
			\hline 
				taux de sous-échantillonnage & type de trajectoire de balayage & PSNR (db) & SSIM \\
				\hline
				& rosace   & 38.944 &  0.834\\
				10\%  & spirale   & 39.059 &  0.833 \\ 
				& Lissajous & 37.825 &  0.859 \\  \hline
				& rosace   & 39.902 &  0.849 \\
				20\% & spirale    & 39.961 &  0.847 \\ 
				& Lissajous & 38.911 &  0.882 \\  \hline
				& rosace   & 41.048 &  0.868 \\
				30\% & spirale    & 41.129 &  0.870 \\ 
				& Lissajous & 40.103 &  0.910 \\  \hline
				& rosace   & 43.361 &  0.898 \\
				50\% & spirale    & 43.767 &  0.903 \\ 
				& Lissajous & 42.006 &  0.945 \\  \hline
				& rosace   & 45.921 &  0.917 \\
				70\% & spirale    & 45.893 &  0.917 \\ 
				& Lissajous & 44.202 &  0.969 \\  \hline
			\end{tabular}
		\end{center}
\end{table}
	
\begin{figure}[!h]
  \centering
  \includegraphics[width=.6\columnwidth, height = 5cm]{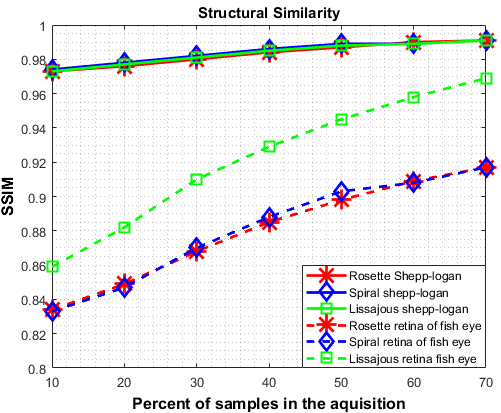}
  \caption{Volume 3D: représentation des valeurs SSIM en fonction du taux de sous-échantillonnage utilisé.}
  \label{fig.ssim1}
\end{figure}

La Fig.~\ref{fig.ssim1} montre les valeurs du score de similarité (SSIM) obtenues pour chaque taux de sous-échantillonnage entre 10\% et 70\%. Nous pouvons constater que la valeur du SSIM est supérieure à 0.84 avec seulement 10\% de mesures, voire même 0.97 en fonction de la structure de l'échantillon visualisé qui donne une meilleure décomposition (données très parcimonieuses) dans le domaine des shearlets.

\subsection{Complétion de tenseurs à faible rang}
%
Un travail fondamental a également été mené sur la théorie de la complétion de tenseurs à faible rang, aussi appelés tenseurs creux \citep{kolda2009tensor}. Au cours des deux dernières décennies, plusieurs travaux relatifs au développement d'algorithmes de remplissage (\emph{complétion}) de matrices de grandes dimensions à partir d'un petit sous-ensemble de ses données~\citep{Candes2010tao}, ont vu le jour. De nombreuses applications en statistiques, en imagerie et en ingénierie, ou encore sur les sites d'achat en ligne ou des plateformes de streaming comme \emph{Netflix} ont poussé un fort développement des méthodes de complétion de données de faible rang.  En imagerie, ces méthodes rejoignent les problèmes inverses et d'acquisition comprimée sous les contraintes de parcimonie. C'est dans ce contexte, que nous avons travaillé sur l'extension de certains concepts mathématiques bien établis dans le cas des matrices, vers le cas des tenseurs de dimension $n_1 \times n_2 \times n_3$ représentant des biopsies optiques 3D. Cependant, les problèmes de complétion de tenseurs à grandes dimensions sont des problèmes $NP$-difficile. 

\begin{figure}[!h]
  \centering
  \includegraphics[width=.8\columnwidth]{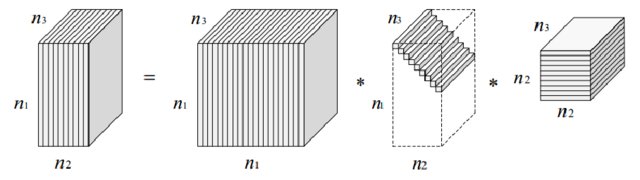}
  \caption{Décomposition $SVD$ d'un tenseur de dimension $n_1 \times n_2 \times n_3$.}
  \label{fig.t_svd}
\end{figure}

Dans~\citep{tamadazteInvProb2018}, nous avons travaillé sur une méthode de reconstruction "implémentable" de tenseurs (biopsies optiques 3D) à partir de données parcimonieuses. Dans ce travail, il s'agissait d'utiliser la décomposition en valeurs singulières $t$-SVD d'un tenseur (Fig.\ref{fig.t_svd}). Plusieurs méthodes de décomposition en $t$-SVD existent dans la littérature. Néanmoins, la plus adaptée pour résoudre un problème "réel" de complétion de tenseurs est celle proposée dans~\citep{anandkumar2014tensor}. 
%
\subsection{Validation}
%
Pour ce faire, nous nous sommes inspirés des travaux~\citep{yuan2014tensor} sur les matrices creuses pour l'étendre au cas de tenseurs. Dans cette méthode, nous avons également travaillé pour la résolution du problème inverse (reconstruction) en utilisant une optimisation en norme nucléaire $l_n$. 

\begin{figure}[!h]
  \centering
  \includegraphics[width=.7\columnwidth]{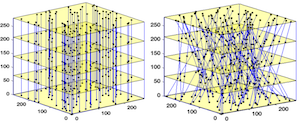}
  \caption{Tenseurs (masques) de mesures : à gauche, un masque vertical, à droite, un masque en biais.}
  \label{fig.masks_3d}
\end{figure}

\begin{figure}[!h]
  \centering
  \includegraphics[width=\columnwidth]{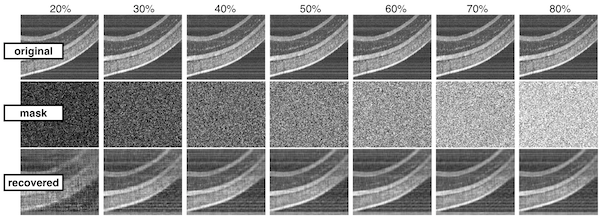}
  \caption{Exemples d'images OCT tirées d'un volume OCT reconstruit avec la méthode $t$-SVD avec l'optimisation en norme $l_n$.}
  \label{fig.cs_comp_1}
\end{figure}

Cette analyse a été appliquée au cas d'imagerie OCT 3D sur des données préenregistrées (l'implémentation actuelle sous \textsc{MatLab} ne permet pas son usage sur un système physique pour des raisons de compatibilité). Les tenseurs de mesures (masques) ont été conçus de manière à s'adapter à la fois à tout type d'imagerie médicale (OCT, IRM, scanner) où les "lignes de tir" (laser, rayon X, etc.) peuvent être verticales mais également de biais par rapport à la cible (Fig.~\ref{fig.masks_3d}). Un exemple de reconstruction est montré sur la Fig.~\ref{fig.cs_comp_1}. Sur cette figure, plusieurs taux de sous-échantillonnage ont été utilisés pour réaliser la reconstruction du volume OCT. 

%
%
%

\section{Bilan}
	
Les travaux sur l'acquisition comprimée ont démarré il y a deux ans. Ils constituent une continuité logique des travaux sur le suivi de chemin (chapitre~\ref{chap.laser}), ainsi que ceux sur la parcimonie et l'asservissement visuel direct fondés sur les ondelettes et les shearlets (chapitre~\ref{chap.wavelet}). Nous avons proposé une méthode d'acquisition comprimée de biopsies optiques implémentée directement sur un système physique, qui constitue une réelle contribution scientifique. Dans cette méthode, nous utilisons une décomposition en shearlets associée à des trajectoires continues de balayage (d'acquisition), qui prennent en compte les caractéristiques physiques du dispositif de balayage (miroirs galvanométriques). Un algorithme de rastérisation permet de faire le lien entre les coordonnées des pixels de la trajectoire de balayage définie dans l'espace image et les coordonnées métriques de l'échantillon cible. Les résultats obtenus sont très satisfaisants. En effet, l'algorithme d'acquisition comprimée permet de reconstruire une biopsie optique 2D ou 3D avec seulement 10\% de mesures. 

Par ailleurs, nous avons entamé un travail sur les problèmes inverses qui découlent de la manipulation de tenseurs à faible rang. Ces problèmes inverses, nous les retrouvons dans les schémas d'acquisition comprimée de volumes de données (ici des biopsies optiques 3D). Pour ce faire, nous avons travaillé sur une méthode de décomposition en valeurs singulières que nous avons étendue cas des tenseurs. \\

L'état actuel de ces travaux ont donné lieu aux publications ci-dessous. 
\footnotesize
\subsubsection{Liste des publications scientifiques [depuis 2012]}
\begin{enumerate}
%
%
\item [Ji] \underline{B. Hayda}r, A.V. Kudryavtsev, A. Bartoli, S. Chrétien, N. Andreff, and \textbf{B. Tamadazte}, \emph{3-D OCT compressed sensing using the shearlet transform under continuous trajectories}, IEEE Trans. on Biomedical Engineering,  \todo{(under review)}. 
\item [Ji] M. I. Assoweh, S. Chrétien, and \textbf{B. Tamadazte}, \emph{Tensor Compelation using t-SVD and Nuclear Norm Minimization}, Inverse Problems and Imaging,  \todo{(under review)}.
\item [Ji] M. I. Assoweh, S. Chrétien, and \textbf{B. Tamadazte}, \emph{Low tubal rank tensor recovery using the Burer-Monteiro factorisation approach. Application to Optical Coherence Tomography}, Inverse Problems and Imaging, \todo{(under review)}.
\item [ ----- ]
\item [Ci] \underline{L. A. Duflo}t, \textbf{B. Tamadazte}, A. Krupa, and N. Andreff, \emph{Shearlet Transform: a Good Candidate for Compressed Sensing in Optical Coherence Tomography}, IEEE EMBC, USA, DOI: 10.1109/EMBC.2016.7590733, 2016.
\item [ ----- ]
\item [Wi]  B. Haydar, B. Tamadazte, N. Andreff, and A. Bartoli, 3D OCT image compression using shearlet transform, Workshop "Medical Imaging Robotics" at Workshop "Medical Imaging Robotics" at IEEE/RSJ IROS, Vancouver, 2017.
\end{enumerate}

\normalsize

\vspace{.5cm}

\noindent
\underline{P. Nom} : doctorant(e)\\

\noindent
Ji : journal international avec comité de lecture\\
Ci : conférence internationale avec actes et comité de lecture\\
Wi : workshop international avec actes et comité de lecture

%% file: fichiers/conclusion_v3.tex

%
\chapter{Conclusion et projet de recherche} \label{chap.conclusion}
\minitoc
%
%
\section{Conclusion}
%
Ce document a résumé les différentes contributions scientifiques réalisées depuis mon recrutement au CNRS, en 2012. Trois thématiques de recherche, liées scientifiquement et chronologiquement, ont été abordées : la vision géométrique (à deux ou trois vues) dans un contexte de chirurgie laser, l'asservissement visuel direct fondé sur l'utilisation des représentations parcimonieuses, et enfin l'acquisition comprimée appliquée à la réalisation \emph{in situ} de biopsies optiques 2D et 3D. Pour chacune de ces thématiques, plusieurs solutions méthodologiques ont été apportées et discutées dans ce manuscrit. 

\subsubsection{Contrôle d'un laser}
En ce qui concerne la chirurgie laser, nous avons développé plusieurs schémas de commande en boucle fermée pour contrôler les déplacements d'un faisceau laser (respectivement d'un spot laser dans le plan-image). Le cadre applicatif (microchirurgie laser des cordes vocales) impose plusieurs contraintes supplémentaires : simplicité d'utilisation et ergonomie, précision, fréquence de commande, stabilité, robustesse aux erreurs d'étalonnage, ... qui sont primordiales dans le contexte clinique. Nous avons, d'abord, "revisité" la formulation de la contrainte trifocale sous une nouvelle forme simplifiée et plus adaptée (un double produit vectoriel) à une utilisation dans une boucle de commande d'asservissement visuel. L'autre originalité de ce travail réside dans l'utilisation de la source laser comme la troisième vue (virtuelle) dans la construction de la contrainte trifocale. Cette loi de commande a montré des performances très intéressantes en termes de simplicité de mise en \oe uvre (une vingtaine de lignes de code suffise pour l'implémentation), de précision, de fréquence de commande, et de stabilité. Une autre loi de commande a été également proposée en s'inspirant des méthodes de suivi de chemin utilisées en robotique mobile. Ce contrôleur permet le contrôle d'un spot laser dans le plan-image, en suivant une courbe quelconque non paramétrée définie à main levée. En d'autres termes, le clinicien dessine une courbe de résection/ablation sur une interface tactile (exemple, tablette) et le spot laser suit le chemin à très haute cadence (plusieurs centaines de Hertz), et à vitesse constante (réglable en fonction des besoins cliniques). Bien que cette technique se soit avérée efficace pour les courbes définies dans l'image (2D), elle présente néanmoins des limites pour les trajectoires 3D. Nous avons, ainsi, proposé un nouveau schéma de commande en combinant les avantages de la méthode trifocale et de ceux de l’approche du suivi de chemin. Par conséquent, la nouvelle approche offre un suivi de chemin à fort potentiel (précision, rapidité, robustesse, etc.) à la fois dans le repère image mais aussi dans le repère monde. 

Ces différentes méthodes ont été validées suivant plusieurs scénarios, allant de la simulation aux tests sur banc d'essais, jusqu'à la validation préclinique réalisée au laboratoire d'anatomie (sur cadavre) par un chirurgien ORL. Ces résultats font partie d'un projet de transfert via la création d'une spin-off sur la chirurgie laser des cordes vocales, \emph{AmaRob} (www.amarob.com).

\subsubsection{Asservissement visuel direct}
L'autre thématique développée dans ce manuscrit est l'asservissement visuel direct (sans extraction d'informations visuelles proprement dit) dont le signal d'entrée de la boucle de commande est une représentation parcimonieuse du signal-image. Une représentation parcimonieuse d'une image est obtenue grâce une transformée de type ondelettes, curvelets, shearlets, etc. Les avantages, parmi d'autres, d'une décomposition espace-fréquence et parcimonieuse est la capacité à filtrer "intuitivement" les informations visuelles. Les contributions scientifiques résident principalement dans l'établissement du rapport entre la dérivation temporelle des coefficients d'ondelettes ou de shearlets (différentes variantes ont été développées) et les mouvements de la caméra. Pour ce faire, nous avons formulé plusieurs matrices d'interaction analytiques et les lois de commande par asservissement visuel, qui en découlent. Par ailleurs, des méthodes utilisant uniquement l'information fréquentielle dans la dérivation de la matrice d'interaction ont été développées, bien qu'elles ne soient pas discutées dans ce document. 

Les méthodes développées dans cette thématique ont été validées en utilisant différents types d'imagerie : caméra conventionnelle, échographie, et tomographie à cohérence optique. Les deux derniers imageurs sont généralement caractérisés par un rapport signal sur bruit défavorable. L'utilisation des informations visuelles pertinentes, obtenues par la représentation parcimonieuse du signal-image, permet aux lois de commande proposées d'être robustes malgré la présence du bruit (par exemple, du type speckle), là où les autres méthodes directes, comme la photométrie, échouent. Ceci a été démontré grâce aux différentes séries de tests de validation, qui consistaient à réaliser automatiquement des tâches de positionnement dans $SE(3)$. Les tests ont été, à la fois, réalisés sur des robots de taille conventionnelle (robot cartésien à 6DDL et un bras anthropomorphique) ainsi que sur des plateformes microrobotiques, notamment un robot à 6DDL à structure parallèle (un hexapode). Les résultats obtenus ont démontré des performances intéressantes en termes de précision et de robustesse, même si le domaine de convergence reste limité par rapport aux méthodes d'asservissement visuel géométrique (points, droites, moments, etc.). 

La visée clinique de ces travaux est la réalisation de biopsies optiques répétitives et automatiques. Plus précisément, il s'agit d'être capable de retrouver une position de référence, où une biopsie optique a été acquise au préalable, pour un second examen/vérification réalisé quelques jours plus tard. Ici les images médicales (OCT ou échographie) sont utilisées à la fois pour la caractérisation optique des tissus/cellules visualisés mais, également, comme capteurs pour le contrôle des mouvements du robot. 

%
\subsubsection{Acquisition comprimée} 
Dans la continuité des activités sur la représentation parcimonieuse (asservissement visuel), nous avons travaillé sur la mise en \oe uvre de nouvelles approches d'acquisition comprimée en imagerie médicale. Pour rappel, un algorithme d'acquisition comprimée permet de reconstruire un signal (respectivement image) à partir d'un nombre très limité de mesures, environ 10\% pour les meilleures méthodes. 
 
La littérature fait généralement référence à des méthodes d'acquisition comprimée réalisées sur des images préenregistrées, lesquelles sont sous-échantillonnées à l'aide d'un masque (matrice des mesures) aléatoire ou pseudo-aléatoire. Néanmoins, lorsqu'il s'agit d'appliquer ces méthodes sur un système physique, le problème devient plus complexe, notamment à cause des techniques de sous-échantillonnage (balayage sur un système physique) aléatoires ou pseudo-aléatoires. Pour ce faire, nous avons proposé toute une méthodologie permettant une acquisition comprimée de biopsies optiques 2D et 3D (données OCT) directement sur un système physique. L'algorithme consiste à utiliser une décomposition multi-échelle (transformée en shearlets) et des techniques de rastérisation et de reconstruction. Dans ce travail, il ne s'agissait pas d'effectuer un balayage aléatoire de l'échantillon, ce qui ne respecte pas les contraintes physiques liées au système de balayage (miroirs galvanométriques), mais des trajectoires continues paramétrables (rosace, spirale, Lissajous). Nous avons réussi à reconstruire des images ou des volumes de bonne qualité (au sens des scores de similarité utilisés en traitement d'images) avec seulement 10\% de mesures. 

Par ailleurs, nous avons entamé un travail sur les problèmes inverses, qui découlent de la manipulation de volumes (tenseurs) à faible rang. Ces problèmes, nous les retrouvons habituellement dans les schémas d'acquisition comprimée de données tridimensionnelles de grandes tailles (ici des biopsies optiques 3D). Pour ce faire, nous avons travaillé sur une méthode de décomposition en valeurs singulières (SVD), bien établie dans le cas des matrices, que nous avons étendue au cas des tenseurs. Nous avons obtenu des résultats préliminaires originaux et encourageants pour une future implémentation sur un système physique.  

\subsubsection{Conception de démonstrateurs précliniques}
%
Finalement, ce document a décrit succinctement ma contribution aux travaux de conception mécatronique relatifs aux démonstrateurs précliniques des projets de recherche, que je coordonne ou auxquels je participe. J'ai le privilège d'être membre d'un département de recherche qui offre des conditions de travail exceptionnelles. Par conséquent, nous avons toujours eu le souci de faire évoluer les démonstrateurs au plus proche de la version clinique ou à défaut préclinique. C'est dans ce contexte, que j'ai participé à la réalisation de plusieurs démonstrateurs microrobotiques/robotiques utilisés pour la validation préclinique d'un grand nombre des contributions scientifiques décrites dans ce document. 
%
\section{Projet de recherche}
%
L'écriture de ce document est l'occasion de faire le bilan des travaux de recherche passés (et en cours), mais surtout d'établir les grandes lignes de mes futures investigations scientifiques sous forme d'un projet de recherche. Il ne s'agit pas de faire un changement de thématique(s), mais plutôt d'aller vers davantage de transversalité dans ma démarche scientifique. Cette transversalité sera à l'intersection de la robotique, des mathématiques appliquées (acquisition comprimée) et des sciences biologiques/neurosciences. 

Les trois grandes thématiques, discutées dans ce manuscrit, constituent un réel cheminement scientifique, entre la commande d'un faisceau laser en utilisant les données-images, l'asservissement visuel direct parcimonieux appliqué à la réalisation de biopsies optiques (ici l'imageur joue à la fois le rôle de capteur et celui du moyen d'examen clinique) et l'acquisition comprimée contrôlée, qui peut être considérée comme l'imbrication des méthodologies de commande de systèmes de balayage (par exemple, miroirs) et de la représentation parcimonieuse.

\begin{figure}[!h]
\centering
	 \includegraphics[width=.9\textwidth]{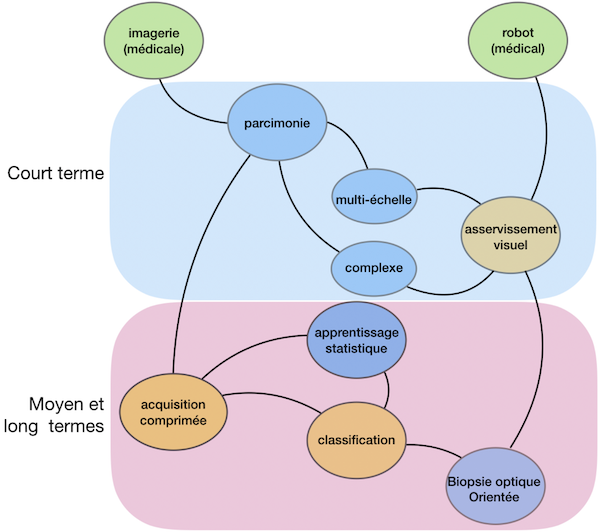}
	 \caption{Organisation des thématiques de mes travaux de recherches futurs.}
	 \label{fig.organisation_perspectives}
\end{figure}

Ce projet de recherche peut être décomposé comme suit : une thématique à moyen terme (asservissement visuel parcimonieux et acquisition comprimée), et une autre à moyen et long termes (apprentissage statistique, acquisition comprimée et commande référencée capteur de vision). La Fig.~\ref{fig.organisation_perspectives} donne un aperçu sur l'organisation de mes futurs travaux de recherche.  

\subsection{Asservissement visuel et acquisition comprimée : quel lien?}
%
Les perspectives, décrites dans cette section, peuvent être considérées comme la continuité (perspectives à court terme) des travaux décrits dans les chapitres~\ref{chap.wavelet} et \ref{chap.cs}.
\subsubsection{Asservissement visuel parcimonieux}
\begin{figure}[!h]
\centering
	 \includegraphics[width=.9\textwidth]{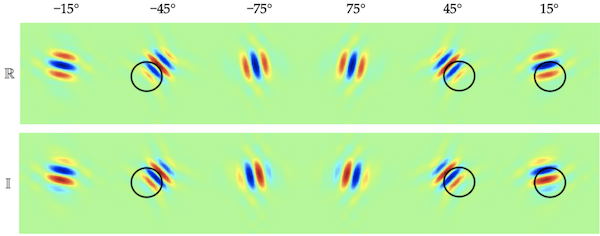}
	 \caption{Comparaison entre les parties réelle et imaginaire dans le cas des ondelettes appliquées à une image.}
	 \label{fig.wavelets_complex}
\end{figure}

%
Il est montré, dans le chapitre~\ref{chap.wavelet}, que les lois de commande par ondelettes ou shearlets sont globalement précises et robustes, même si le domaine de convergence est relativement restreint, comparées, par exemple, aux méthodes géométriques. Dans la conception des matrices d'interaction analytiques associées à ces méthodes, uniquement la partie réelle des coefficients est prise en compte. Il serait intéressant de considérer, également, la partie imaginaire de ces coefficients dans la modélisation, ce qui apporterait certainement davantage de précision, de robustesse, et par conséquent, augmenterait le domaine de convergence (uniquement une partie de l'espace temps-fréquence est utilisée). La Fig.~\ref{fig.wavelets_complex} montre l'apport de la partie imaginaire en terme de contenu visuel (parties entourées d'un cercle). 

L'autre volet, relatif à ce travail, est l'utilisation de l'ensemble du potentiel de la décomposition multi-échelle qu'offrent les transformées en ondelettes et en shearlets. Nous avons proposé une première tentative de dérivation d'une matrice d'interaction multi-échelle en utilisant les ondelettes. Il serait intéressant d'approfondir ce travail sur l'asservissement visuel par ondelettes et de l'étendre aux shearlets. L'intérêt d'une matrice d'interaction multi-échelle est d'avoir la possibilité de changer d'échelle pendant la tâche de positionnement. Par exemple, en fonction de la norme de l'erreur entre les positions courante et désirée, la loi de commande peut choisir une résolution basse (loin de la position désirée) ou haute (proche de la position désirée). Ceci permet d'augmenter sensiblement le domaine de convergence de ces approches tout en gardant une grande précision. 
%
\subsubsection{Asservissement visuel sans reconstruction}
%
Une autre perspective commune à l'asservissement visuel et à l'acquisition comprimée peut consister à utiliser les informations visuelles (coefficients parcimonieux) sans reconstruction proprement dite de l'image. Jusqu'à présent, les méthodes proposées utilisent les coefficients d'ondelettes ou de shearlets extraits d'images entières acquises par un capteur. En réalité, l'étape d'acquisition/reconstruction n'est pas nécessaire au fonctionnement de la commande. Ainsi, avec des capteurs adaptés, essentiellement en imagerie médicale, il est possible d'acquérir uniquement un certain nombre de points-images pour la conception de la matrice d'interaction associée. S'affranchir de la phase de reconstruction des images permet de réduire considérablement le temps de la boucle de commande, notamment dans le cas des capteurs d'images à faible fréquence d'acquisition. 

Il est vrai que cette approche peut être ambitieuse en termes de méthodologie et aussi du fait qu'il n'existe pas, à ma connaissance, de travaux relatifs dans la littérature. Cependant, les travaux sur la parcimonie (acquisition comprimée et asservissement visuel) ont permis de poser les premières bases et de lever un certain nombre de verrous scientifiques. 
%
\subsection{De l'histopathologie à la caractérisation \emph{in situ}}
%
L'histopathologie est l'observation microscopique (optique) de tissus prélevés par biopsie chirurgicale (par exemple, par ponction), pour permettre d'établir un diagnostic précis pour certaines pathologies non détectables par imagerie externe, notamment dans le cas de la classification de tumeurs cancéreuses. Cependant, l'histopathologie classique a deux limitations importantes : 1) l'efficacité de l'examen est limitée par la précision de l'excision, et 2) la caractérisation optique des tissus étant réalisée \textit{a posteriori}, il est nécessaire de s'assurer que l'échantillon est parfaitement prélevé et qu'il soit représentatif des tissus suspects à examiner. Si le diagnostic s'avère non concluant, ce qui peut atteindre un taux de 39\% pour certains actes, il est nécessaire de réaliser une seconde, voire davantage, de biopsies supplémentaires. Chaque prélèvement induit un risque clinique dû au retard dans le traitement de la pathologie éventuelle, voire de son aggravation en raison d'un risque accru de dispersion des cellules atteintes (contamination). 

\begin{figure}[!h]
\centering
	 \includegraphics[width=.8\textwidth]{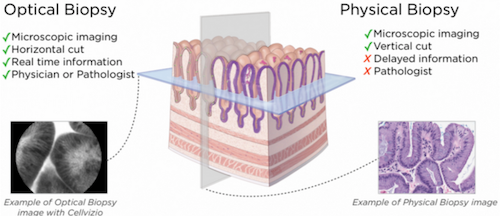}
	 \caption{Comparaison de la biopsie optique et physique. \scriptsize CC BY-NC-SA 3.0 Cellvizio.net \normalsize}
	 \label{fig.biopsies}
\end{figure}

\normalsize
La biopsie optique peut clairement pallier les limites de celle physique, et apporter plusieurs valeurs ajoutées cliniques (Fig.~\ref{fig.biopsies}). En revanche, dans la pratique actuelle, le clinicien ne s'appuie que sur son "expertise intrinsèque" pour établir un diagnostic partiel (différencier un tissu pathologique d'un autre sain), mais ne dispose pas de suffisamment d'informations pour réaliser simultanément le diagnostic et la classification des tissus à l'échelle cellulaire.  

\begin{figure}[!h]
\centering
	 \includegraphics[width=.85\textwidth]{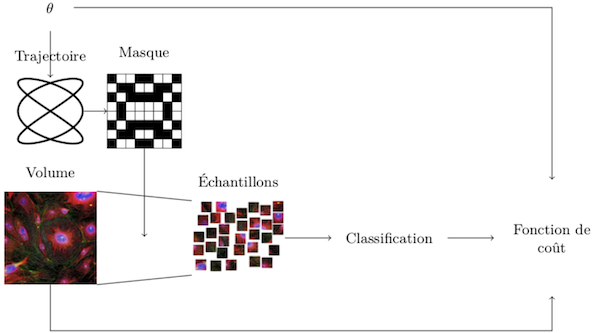}
	\caption{Classification directe des tissus à partir de la représentation parcimonieuse.
	Se passer de l'étape de reconstruction permet de diminuer le temps de traitement.}
	\label{fig.classification}
\end{figure}

Pour ce faire, il serait intéressant d'associer, pendant la tâche d'acquisition, des méthodes d'apprentissage statistique afin d'évaluer cliniquement (en ligne) les données acquises en prédisant spatialement la probabilité de présence d'indicateurs pathologiques sur le tissu examiné. En effet, effectuer une analyse \textit{in situ} permettrait d'augmenter la certitude du diagnostic, d'être capable de réaliser une classification de la pathologie, et de diminuer le risque de faux-négatifs. L'acquisition des données, ainsi que la classification de celles-ci, peuvent être réalisées sans avoir recours à la reconstruction de la biopsie optique, ou à minima, à ne reconstruire que lorsque les indicateurs pathologiques sont positifs (Fig.~\ref{fig.classification}). L'avantage de cette approche réside notamment dans la capacité à traiter des zones de tissus plus grandes et plus rapidement. 
 
%
\subsubsection{Approche globale}
Pour répondre à ce double challenge scientifique et clinique, plusieurs approches peuvent être explorées, telles que la caractérisation à partir d'une seule ligne de tir, ou à partir d'échantillonnages bi- ou tridimensionnels partiels (de géométries variables), en utilisant des trajectoires de balayage définies au préalable. Parmi les méthodes statistiques à étudier pour répondre aux problématiques d'acquisition parcimonieuse et de classification, il serait opportun de s'intéresser aux modèles d'apprentissage de l'erreur résiduelle~\citep{lee_deep_2017, quan_compressed_2018} ou aux  réseaux génératifs antagonistes (GANs)~\citep{yang_dagan_2018}. A titre d'exemple, dans le cas d'une exploration endomicroscopique d'un tissu (organe, cavité, etc.), le clinicien a l'habitude de scanner de manière "globale" le site suspecté à la recherche d’un éventuel changement de texture ou d'aspect dans les biopsies optiques successivement acquises. Cette pratique peut s'avérer complexe et inefficace. D'abord le clinicien n'a aucune vision directe du site à examiner, et par conséquent, aucune information sur les positions préalablement explorées. Autrement dit, il existe un risque d'effectuer plusieurs passages sur la même zone, tout en oubliant plusieurs autres. D'un autre côté, pour certains examens, le clinicien est contraint de réaliser le diagnostic dans un temps imparti, la recherche exhaustive d'indicateurs pathologiques peut alors poser un problème de temps de réalisation. Avec les méthodes, que nous visons à développer, nous ouvrons clairement la voie de ce qu'on pourrait appeler "\emph{la biopsie optique orientée}". En effet, la sonde endomicroscopique montée sur un robot, peut être contrôlée de manière à scanner suivant une trajectoire définie (pouvant être mise à jour en ligne) par la position des indicateurs pathologiques issus de la classification. 
%
\subsubsection{Approche multi-échelle (1)}
\begin{figure}[!h]
\centering
	 \includegraphics[width=\textwidth]{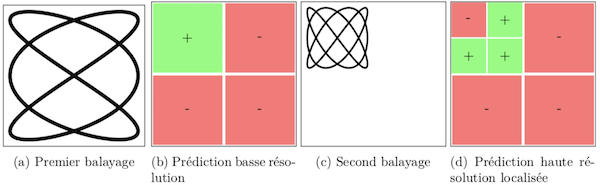}
	\caption{Balayage multi-échelle. Une première prédiction basse résolution permet de détecter les zones d'intérêt sur lesquelles effectuer une acquisition plus détaillée.}
	\label{fig.multiscale_classification}
\end{figure}

%
En opposition à la méthode globale discutée ci-dessus, j'envisage d'étudier des approches multi-échelle inspirées des décompositions à plusieurs niveaux existantes en imagerie, à l'exemple de la transformée en ondelettes. Ces approches consistent généralement à effectuer une première analyse de l'information à basse résolution (grande échelle), puis à réaliser une ou plusieurs analyses plus fines, à haute résolution (petite échelle) uniquement dans les zones pertinentes. Dans le cas de la classification en ligne, je propose d'utiliser cette approche pour la recherche des indicateurs pathologiques dans les données-images. La technique se prête particulièrement bien à ce type de problèmes, car il est effectivement possible de choisir des trajectoires (résolution, forme géométrique, etc.) d'investigation des tissus selon les échantillons acquis à l'instant $t$. Un premier balayage rapide offrira un aperçu de la zone d'intérêt. Le choix des zones à analyser plus finement pourrait alors se faire selon plusieurs critères (Fig.~\ref{fig.multiscale_classification}). Les algorithmes de classification développés dans l'approche globale serviront à prédire la probabilité spatiale de la présence d'indicateurs pathologiques dans la zone scannée. Les zones ayant la plus forte probabilité de contenir ces indicateurs pourront alors être sélectionnées pour un balayage plus fin, pour mieux confirmer le diagnostic et en quantifier le degré ou l'étendue. Finalement, à l'aide de cette méthode, il est possible de créer une sorte de carte multi-échelle du tissu, dont la résolution indique la probabilité de présence ou non d'indicateurs pathologiques. 

Par ailleurs, les positions spatiales des indicateurs pathologiques définissent un nouveau moyen de création d'une "\emph{parcimonie pathologique}" sans avoir recours aux méthodes de décomposition de type ondelettes, shearlets, etc. A partir de cette nouvelle parcimonie, il est tout à fait envisageable d'appliquer les méthodes d'acquisition comprimée décrites dans le chapitre~\ref{chap.cs}, par exemple, celles de la complétion de matrices ou de tenseurs à faible rang, pour construire des biopsies optiques contenant les informations sur la pathologie. 

Par ailleurs, l'approche multi-échelle est, certainement, beaucoup plus adaptée à une future intégration sur un système endomicroscopique, voire même d'imagerie de manière générale. Elle permet de réduire considérablement le temps d'acquisition et de reconstruction des données, sans faire appel aux méthodes traditionnelles de représentations parcimonieuses, qui sont généralement "gourmandes" en temps de calcul (transformées directe et inverse).  
%
\subsubsection{Approche multi-échelle (2)}
\begin{figure}[!h]
\centering
	 \includegraphics[width=\textwidth]{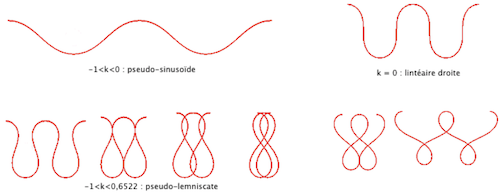}
	\caption{Exemple d'une courbe élastique à utiliser pour l'investigation multi-échelle et adaptative \small (www.mathcurve.com). \normalsize}
	\label{fig.elastic}
\end{figure}

%
L'aspect multi-échelle peut être également envisagé d'une autre manière, c'est-à-dire plus compact. Il s'agit non pas de décomposer l'espace de recherche d'indicateurs pathologiques en plusieurs sous-espaces, comme évoqué ci-dessus, mais de faire une investigation globale et adaptative "\emph{à la volée}". Pour ce faire, il est nécessaire de disposer d'une procédure de balayage, simple à mettre en \oe uvre et qui se prête à ce type de problématique. Il existe une famille de courbes géométriques paramétrées dont le changement d'un seul coefficient permet de modifier efficacement l'échelle et la forme de la courbe elle-même. Les courbes élastiques, par exemple l'\emph{elastica} (Fig.~\ref{fig.elastic}), dont les équations paramétriques sont données par :
\begin{eqnarray}
\nonumber
x & = & a \sqrt{k + \cos (t)} \\
y & = & \frac{a}{2} \int_{0}^{t} \frac{\cos u}{\sqrt{k + \cos u}} du ~~~\texttt{avec}~t= \theta-\frac{\pi}{2}  
\label{eq.elsatic}
\end{eqnarray}
peut être un premier cas d'étude. 

Le paramètre le plus intéressant dans les équations (\ref{eq.elsatic}), est $k$. Il permet d'adapter l'échelle et la forme de la courbe élastique, comme le montre la Fig.~\ref{fig.elastic} pour différentes valeurs. Par conséquent, la variation de la valeur de $k$ peut être reliée directement à la classification des indicateurs pathologiques acquis en ligne. Lorsque l'indicateur pathologique est négatif (probabilité faible), la valeur de $k$ peut être comprise dans l'intervalle [-1, 0] pour une recherche à basse résolution (Fig.~\ref{fig.elastic}~(haut)). A l'inverse, si la probabilité que l'indicateur soit positif est haute, la valeur de $k$ peut être continuellement adaptée en ligne pour avoir une recherche plus fine (haute résolution), comme le montre la Fig.~\ref{fig.elastic}~(bas) pour un $k$ = [-1, 0.6522]. 
 
Cette approche peut s'appliquer également aux méthodes d'acquisition comprimée de manière générale indépendamment des problématiques de classification statistique. 
%
\subsubsection{Approche hybride : acquisition comprimée, classification et asservissement visuel}
%
A plus long terme, l'objectif est de développer des approches méthodologiques associant l'acquisition comprimée, la classification et la commande référencée capteur de vision (images endoscopiques) dans le même paradigme. Ce travail peut s'inscrire dans la continuité des travaux sur la commande (chapitre~\ref{chap.wavelet}) dont le signal d'entrée sont des données-images (coefficients d'ondelettes multi-échelle ou non, coefficients des shearlets, etc.) pour le positionnement automatique d'une sonde endomicroscopique à une position désirée (localisation d'une biopsie optique). La fonction tâche est une "simple" régression de l'erreur cartésienne entre une position courante et une autre désirée. Chacun peut noter qu'il existe un dénominateur commun entre l'acquisition comprimée, la classification des données parcimonieuses et les lois de commande d'asservissement visuel par ondelettes ou shearlets. Par conséquent, l'imbrication des différentes tâches devient presque "naturelle" et plusieurs intérêts et challenges scientifiques peuvent en découler. Pour ce faire, la nouvelle fonction tâche, qui associe les trois objectifs suscités, peut être définie sous la forme suivante :
\begin{equation}
\nonumber
\tau_{totale} = \tau_{acquisition} + \tau_{classification} + \tau_{commande}
\end{equation}

\begin{figure}[!h]
\centering
	 \includegraphics[width=.9\textwidth]{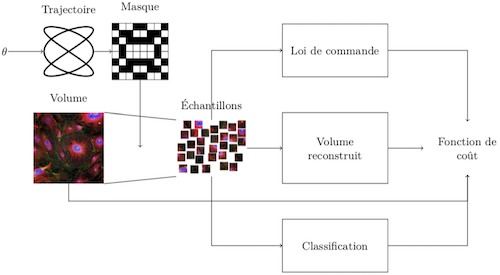}
	\caption{Schématisation d'une approche multi-tâche : acquisition comprimée, classification et asservissement visuel.}
	\label{fig.cmd_class_cs}
\end{figure}

%
%
Le fonctionnement recherché est le suivant : contrôler la structure robotique, sur laquelle est montée une sonde endomicroscopique pour retrouver la localisation d'une biopsie optique prise lors d'un examen initial. Par exemple, les déplacements de la sonde seront contrôlés par la loi de commande (tâche principale), ainsi que par la tâche relative à la recherche d'indicateurs pathologiques (tâche secondaire~1). Si ces derniers sont positifs, alors il faut procéder à la reconstruction de la biopsie optique par les méthodes d'acquisition comprimée (tâche secondaire~2). La priorité des tâches peut varier en fonction d'un ou plusieurs paramètres de pondération ou de poids définis au préalable, par exemple par apprentissage. 
%
\subsubsection{Au-delà des problématiques scientifiques}
%
Au-delà du fait que les méthodes décrites, ci-dessus, permettront de rendre la pratique du diagnostic en ligne, à l'aide de biopsies optiques, réalisable et efficace, elles ouvrent la voie à plusieurs autres applications cliniques. A titre d'exemple, les approches de classifications, qui seront développées, peuvent être utilisées, indépendamment des aspects et des contraintes d'acquisition, c'est-à-dire directement sur des grandes bases de données (données histopathologiques, imagerie, etc.) disponibles dans les centres médicaux (publics ou privés). Ceci sera réalisé dans le strict respect  de l'éthique et des lois qui régissent les données personnelles et médicales. Des cadres législatifs, notamment dans celui du programme "\emph{Hôpital du Futur}", seront, certainement, mis en place dans les prochaines années pour rendre possible l'utilisation de ces données indispensables à la recherche. 

Parmi les applications cliniques visées se trouvent le diagnostic des maladies neurodégénératives (Alzheimer, Parkinson) à un stage précoce. Dans le chapitre~\ref{chap.clinique}, nous avons abordé l'idée de faire la caractérisation \textit{in situ} de l'épithélium olfactif à la recherche de la présence des protéines \emph{Bêta-Amyloïdes} à l'aide de biopsies optiques (travail en cours). De la même manière, il est possible d'utiliser efficacement les méthodes d'apprentissage statistique de classification entrainées sur de grandes bases de données (exemple, IRM fonctionnel) disponibles dans les services concernés. L'idée est de créer des "benchmarks", qui seront mis à la disposition de la communauté, comme c'est la pratique dans d'autres disciplines de la robotique, de l'intelligence artificielle, de la vision par ordinateur, etc. J'ai la conviction que collecter ces données (dans le cadre d'un protocole), les standardiser, et les mettre à disposition des scientifiques, est déjà une avancée significative en soi. La mise à disposition, pour la communauté, des méthodes d'extraction d'indicateurs pathologiques sur des biopsies optiques pour certaines maladies, comme l'Alzheimer (un grand défi de santé publique), serait un pas de "géant".